\pdfoutput=1

\documentclass[11pt]{article}
\usepackage{acl} 

\AtBeginDocument{
  \hypersetup{
    colorlinks=true,
    linkcolor=[RGB]{205,92,92},   
  }
}

\usepackage{times,latexsym}
\usepackage{url}
\usepackage{tabularx}
\usepackage[T1]{fontenc}
\usepackage{microtype}
\usepackage{graphicx} 
\usepackage{inconsolata}
\usepackage{subcaption}
\usepackage{soul}
\usepackage{float}

\usepackage{longtable, booktabs, array}
\usepackage{makecell}
\usepackage[utf8]{inputenc}
\usepackage{colortbl}
\renewcommand{\arraystretch}{1.2} 
\usepackage{tcolorbox}   
\usepackage{xcolor}      
\usepackage{mathtools}
\usepackage{amsmath}   
\usepackage{amssymb}   
\usepackage[mathscr]{euscript} 
\usepackage{geometry}  
\usepackage{etoolbox}  

\usepackage{amsfonts}
\usepackage{booktabs} 
\usepackage{multirow} 

\definecolor{C1}{RGB}{240, 248, 255}
\definecolor{C2}{RGB}{255, 248, 252}
\definecolor{C3}{RGB}{240, 255, 240}

\newcolumntype{Y}{>{\centering\arraybackslash}X}
\newcolumntype{L}{>{\raggedright\arraybackslash}X}
\definecolor{darkred}{RGB}{200, 0, 0}
\definecolor{darkyellow}{RGB}{250, 153, 0}
\definecolor{darkgreen}{RGB}{0, 180, 0}
\definecolor{lise}{RGB}{128,0,32}
\definecolor{deepblue}{RGB}{68,85,102}
\definecolor{graygreen}{RGB}{85,107,47}
\definecolor{tomato}{RGB}{255,99,71}
\definecolor{india}{RGB}{205,92,92}

\definecolor{saddlebrown}{RGB}{139, 69, 19}

\definecolor{forestgreen}{RGB}{34, 139, 34}
\definecolor{maroon}{RGB}{217,38,38}

\definecolor{navy}{RGB}{164, 101, 138}

\definecolor{moccasin}{RGB}{230, 159, 0}

\definecolor{darkred}{RGB}{200, 0, 0}
\definecolor{darkyellow}{RGB}{250, 153, 0}
\definecolor{darkgreen}{RGB}{0, 180, 0}

\usepackage{hyperref}

\usepackage{times}
\usepackage{latexsym}
\usepackage[T1]{fontenc}
\usepackage[utf8]{inputenc}
\usepackage{microtype}
\usepackage{inconsolata}
\usepackage{graphicx}

\title{How Chain-of-Thought Works? Tracing Information Flow from Decoding, Projection, and Activation}

\author{
{\bf Hao Yang}$^{1}$, 
{\bf Qinghua Zhao}$^{2,\dagger}$,
{\bf Lei Li}$^{3}$, 
{\bf Lingyi Meng}$^{4}$, 
{\bf Mengda Yu}$^{5}$
\\[0.5em]
$^{1}$State Key Laboratory for Novel Software Technology, Nanjing University, Suzhou, China \\
$^{2}$School of Artificial Intelligence and Big Data, Hefei University, Hefei, China \\
$^{3}$School of Artificial Intelligence, Beijing Institute of Technology, Beijing, China \\
$^{4}$School of Computing and Information, University of Pittsburgh, Pittsburgh, PA 15260, USA \\
$^{5}$Center for Biostatistics, The Ohio State University Wexner Medical Center, Columbus, OH, USA
}

\begin{document}
\maketitle
{\let\thefootnote\relax\footnotetext{$^\dagger$ Corresponding author.}}
\begin{abstract}
Chain-of-Thought (CoT) prompting significantly enhances model reasoning, yet its internal mechanisms remain poorly understood. We analyze CoT’s operational principles by reversely tracing information flow across decoding, projection, and activation phases. Our quantitative analysis suggests that CoT may serve as a decoding space pruner, leveraging answer templates to guide output generation, with higher template adherence strongly correlating with improved performance. Furthermore, we surprisingly find that CoT modulates neuron engagement in a task-dependent manner: reducing neuron activation in open-domain tasks, yet increasing it in closed-domain scenarios.  These findings offer a novel mechanistic interpretability framework and critical insights for enabling targeted CoT interventions to design more efficient and robust prompts. We released our code and data at \url{https://github.com/How-Young-X/cot}. 
\end{abstract}

\section{Introduction}

Large language models (LLMs) excel in diverse tasks but falter in multi-step reasoning. Chain-of-Thought (CoT) prompting, introduced by \citet{wei2022chain}, mitigates this by guiding models through step-by-step reasoning, markedly improving performance in arithmetic, commonsense, and symbolic reasoning tasks \citep{wang2023towards, jin2024impact, prabhakar2024deciphering, nayab2025concisethoughtsimpactoutput}. Despite its effectiveness, CoT’s underlying mechanisms remain poorly understood.

Prior studies propose hypotheses for CoT’s efficacy. For instance, \citet{madaan2023makes} argue that CoT reduces task complexity, making tasks more manageable for models. Similarly, \citet{madaan2022text} propose that models imitate answer templates in CoT prompts, and \citet{schaeffer2023invalid} suggest that prompt features unrelated to logical reasoning drive performance gains. Although these insights are intuitively compelling, they lack direct experimental support linking models' internal states to observed outcomes.

This work advances our understanding of CoT’s mechanisms by adopting a mechanistic interpretability  approach \citep{dumas2024how, 10.5555/3692070.3693122}. 
Mechanistic interpretability aims to ``open'' the model's ``black box'', investigating how its internal ``\textit{parts}'' (computational units) and ``\textit{wiring}'' (connection units) enable information to be sequentially processed and flow. It is crucial for  CoT that significantly alters model behavior. Consistent with prior work \citep{chia2023contrastive, madaan2023makes, jin2024impact}, we focus on vanilla CoT \citep{wei2022chain}, as its internal mechanisms remain underexplored.

To investigate CoT’s mechanisms, we select six models (ranging from 3B to 70B) and nine datasets spanning three core reasoning types: arithmetic, commonsense, and symbolic reasoning, covering a wide range of task complexities, input formats, and answer spaces (e.g., open-domain GSM8K and closed-domain AQuA). 
Our multi-faceted analysis yields several key mechanistic insights: 
  

\begin{itemize}
    \setlength{\itemsep}{-1.3mm}
    \item[\color{india}$\circ$] We propose a novel mechanism by which CoT may constrain the decoding space by leveraging answer templates.  While previous work has hinted at the importance of prompt structure, we \textit{quantify} this by demonstrating a strong correlation between reasoning structure adherence and performance.
    \item[\color{saddlebrown}$\diamond$] Our analysis demonstrates that CoT yields concentrated probability distributions, potentially minimizing prediction uncertainty, \textit{providing new evidence} for CoT's role in enhancing model confidence.
    \item[\color{moccasin}$\star$] We identify a task-dependent modulation of neuron engagement by CoT, which may vary across datasets: reducing activation in open-domain tasks, while {unexpectedly increasing} it in closed-domain tasks, a phenomenon \textit{not previously reported}.
\end{itemize}
\vspace{-2mm}
These findings provide a more nuanced understanding of CoT's  mechanisms and offer a novel mechanistic interpretability  and critical insights for designing more efficient and robust prompts.

\section{Related Work}
This section reviews mechanistic interpretability for LLM analysis and CoT prompting with its influencing factors. We then position our work, which applies mechanistic interpretability to investigate CoT's internal mechanisms.

\paragraph{Mechanistic Interpretability}
Mechanistic interpretability seeks to reverse-engineer neural network computations by mapping inputs through internal states to outputs, rather than relying solely on input–output correlations \citep{nanda2023progress}.
Research has examined specific components: transformer \textit{feed-forward networks} (FFNs) as key-value memories linking textual patterns to output distributions \citep{geva-etal-2021-transformer}; neuron properties, such as universal neurons \citep{gurnee2024universal}, reasoning neurons in FFNs \citep{rai-yao-2024-investigation}, and activation sparsity \citep{voita-etal-2024-neurons}.

Other approaches analyze information flow and \textit{representation} through intermediate layers. Techniques include patching representations to localize computations \citep{fierro2025multilinguallanguagemodelsremember}, tracking embeddings across spaces \citep{wendler-etal-2024-llamas}, and extending the logit lens on vocabulary embeddings \citep{cancedda-2024-spectral}. 
Recent work applies causal interventions such as activation patching to measure faithfulness of natural language explanations \citep{yeo2024faithfulnaturallanguageexplanations}, while others use mechanistic analysis to reveal phase transitions and specialized attention pathways in step-by-step reasoning \citep{dutta2024thinkstepbystepmechanisticunderstanding}. Together, these methods provide tools to probe LLM internals and reasoning mechanisms. 
 
\begin{table*}[!htb]
\scriptsize
\setlength{\tabcolsep}{2pt} 
\captionsetup[table]{skip=3pt}  
\renewcommand{\arraystretch}{0.85} 
\begin{tabular}{p{1.3cm} p{2.2cm} p{12cm}}

\toprule
\textbf{Task} & \textbf{\makecell[l]{Dataset \\ (Answer Space)}} & \textbf{Example} \\
\midrule

\rowcolor{C1}
& \makecell[l]{GSM8K / SVAMP \\ (Open,   Numerical)} & 
\textbf{Question}: If there are 3 cars in the parking lot and 2 more cars arrive, how many cars are in the parking? \newline
\textbf{Answer}: There are originally 3 cars. 2 more cars arrive. \textcolor{darkred}{$3 + 2 = 5$}. So the answer is \textcolor{darkred}{5}. \\
\cmidrule(lr){2-3}

\rowcolor{C1}
\textbf{Arithmetic} & \makecell[l]{AQuA \\(Closed, Options)} & 
\textbf{Question}: A person is traveling at 20 km/hr and reached his destiny in 2.5 hr then find the distance? \newline
\textbf{Choices}: (a) 53 km (b) 55 km (c) 52 km (d) 60 km (e) 50 km \newline
\textbf{Answer}: The distance that the person traveled would have been $20 \, \text{km/hr} \times 2.5 \, \text{hrs} = 50 \, \text{km}$. So the answer is \textcolor{darkred}{(e)}. \\

\midrule

\rowcolor{C2}
& \makecell[l]{Bamboogle \\ (Open, Text)} & 
\textbf{Question}: Who lived longer, Theodor Haecker or Harry Vaughan Watkins? \newline
\textbf{Answer}: Theodor Haecker was 65 years old when he died. Harry Vaughan Watkins was 69 years old when he died. So the answer is \textcolor{darkred}{Harry Vaughan Watkins}. \\
\cmidrule(lr){2-3}

\rowcolor{C2}
\textbf{Common sense}& \makecell[l]{StrategyQA / Sports \\ (Binary, Yes/No) }& 
\textbf{Question}: Do hamsters provide food for any animals? \newline
\textbf{Answer}: Hamsters are prey animals. Prey are food for predators. Thus, hamsters provide food for some animals. So the answer is \textcolor{darkred}{Yes}. \\
\cmidrule(lr){2-3}

\rowcolor{C2}
 & \makecell[l]{Date \\ (Formatted)} & 
\textbf{Question}: The concert was scheduled to be on 06/01/1943, but was delayed by one day to today. What is the date 10 days ago in MM/DD/YYYY? \newline
\textbf{Answer}: One day after 06/01/1943 is 06/02/1943, so today is 06/02/1943. 10 days before today is 05/23/1943. So the answer is \textcolor{darkred}{05/23/1943}. \\

\midrule

\rowcolor{C3}
& \makecell[l]{Coin Flip \\(Binary, Yes/No)} & 
\textbf{Question}: A coin is heads up. Maybelle flips the coin. Shalonda does not flip the coin. Is the coin still heads up? \newline
\textbf{Answer}: The coin was flipped by Maybelle. So the coin was flipped 1 time, which is an odd number. The coin started heads up, so after an odd number of flips, it will be tails up. So the answer is \textcolor{darkred}{No}. \\
\cmidrule(lr){2-3}

\rowcolor{C3}
\textbf{Symbolic}& \makecell[l]{Last Letters\\ (Open, Text) }& 
\textbf{Question}: Take the last letters of each words in ``Lace\underline{y} Nor\underline{a} Debr\underline{a} Ashleig\underline{h}'' and concatenate them. \newline
\textbf{Answer}: The answer is \textcolor{darkred}{yaah}. \\
\bottomrule
\end{tabular}
\caption{Part Dataset \textit{Examples} (Full examples is in Appendix \ref{sec:dataset_examples}).} 
\label{tab:part_examples} 
\end{table*}

 
\paragraph{Chain-of-Thought}
CoT prompting improves reasoning by generating intermediate steps \citep{wei2022chain}, proving effective across diverse tasks \citep{tanneru2024hardness} and inspiring many extensions  \citep{li2023mot, bi2024program, chenboosting, kgqgipm24, agenteaicde25, agentqgcikm24, hydra26}.

Factors shaping CoT effectiveness suggest format matters more than content or logical validity: models still succeed with irrelevant \citep{webson-pavlick-2022-prompt} or invalid steps \citep{wang2023towards, schaeffer2023invalid}, or without keywords, if reasoning structure is preserved \citep{li2025llms}. Counterfactual studies further show that consistent formats, not symbols or grammar, drive CoT gains \citep{madaan2023makes}. 

Other influences include rationale length longer boosts, shorter harms \citep{jin2024impact}; imitation of style and instruction rather than factual reasoning \citep{gudibande2023false}; effects on feature attribution robustness \citep{wu2023analyzing}; benefits from diverse exemplar reasoning skills \citep{ye2023complementary}; and task-specific roles of probability, memorization, and noisy reasoning \citep{prabhakar2024deciphering}. This gap motivates our work.

\section{Experimental Setup}

Following \citep{wei2022chain, chia2023contrastive, madaan2023makes, jin2024impact}, we selected nine datasets spanning three core reasoning task types: arithmetic, commonsense, and symbolic reasoning, where CoT demonstrates substantial performance gains. Examples are shown in Table \ref{tab:part_examples}.

\paragraph{Arithmetic reasoning tasks.} These tasks require to solve mathematical problems through multi-step calculations. We used three  widely adopted datasets: GSM8K \citep{cobbe2021training}, SVAMP \citep{patel-etal-2021-nlp}, and AQuA \citep{ling2017program}.

\paragraph{Commonsense reasoning tasks.} These tasks involve answering questions based on commonsense knowledge. We employed four  commonly used datasets: Bamboogle \citep{press-etal-2023-measuring}, StrategyQA \citep{geva2021strategyqa}, Date, and Sports \citep{srivastava2023beyond}.

\paragraph{Symbolic reasoning tasks.} These tasks involve processing symbolic sequences using logical rules. We considered two  available  datasets: the Coin Flip  \citep{wei2022chain} and the Last Letters Concatenation \citep{kojima2022large}.

\subsection{Models \& Parameters Settings}  \label{sec:models}

Consistent with prior work \citep{chia2023contrastive, madaan2023makes, jin2024impact}, we focus on vanilla CoT \citep{wei2022chain}, as its internal mechanisms remain underexplored.
We used 4-shot prompts derived from \citep{wei2022chain, kojima2022large, wang2023towards}. We evaluated six pretrained models from diverse families and scales: LLaMA3.1 (8B, 70B) \citep{grattafiori2024llama3herdmodels}, Gemma2 (2B, 9B, 27B) \citep{gemmateam2024gemma2improvingopen}, and LLaMA3.2-3B \citep{llama3_2_connect_2024}. Greedy decoding with a 300-token limit ensured deterministic output. Each dataset’s test set was evaluated, with performance measured by accuracy, extracted via regular expressions. CoT prompt exemplars are shown in Table \ref{tab:part_examples}, with full prompts in Appendix \ref{sec:prompts}. This study compared LLM behaviors under CoT and standard prompts. Due to space constraints, representative results are presented here, with comprehensive results in Appendix \ref{appendix:keyword_imitation} $\sim$ \ref{appendix:activation_difference}.

\begin{figure*}[!ht]
\centering
\captionsetup[sub]{skip=-1mm}  
    \begin{subfigure}{0.32\linewidth}
        \includegraphics[width=\linewidth]{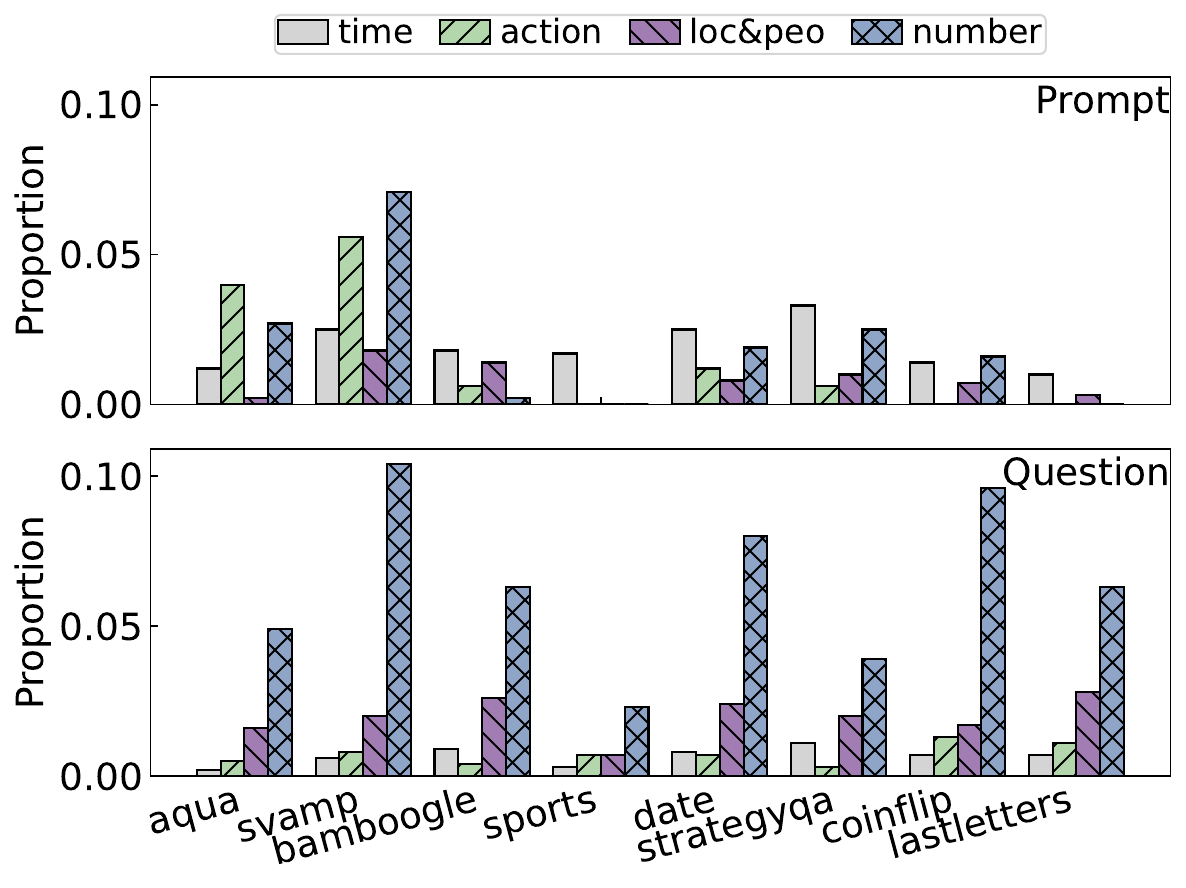}
        \caption{\textit{GSM8K}}
        \label{fig:imatition_gsm8k}
    \end{subfigure}
    \begin{subfigure}{0.32\linewidth}
        \includegraphics[width=\linewidth]{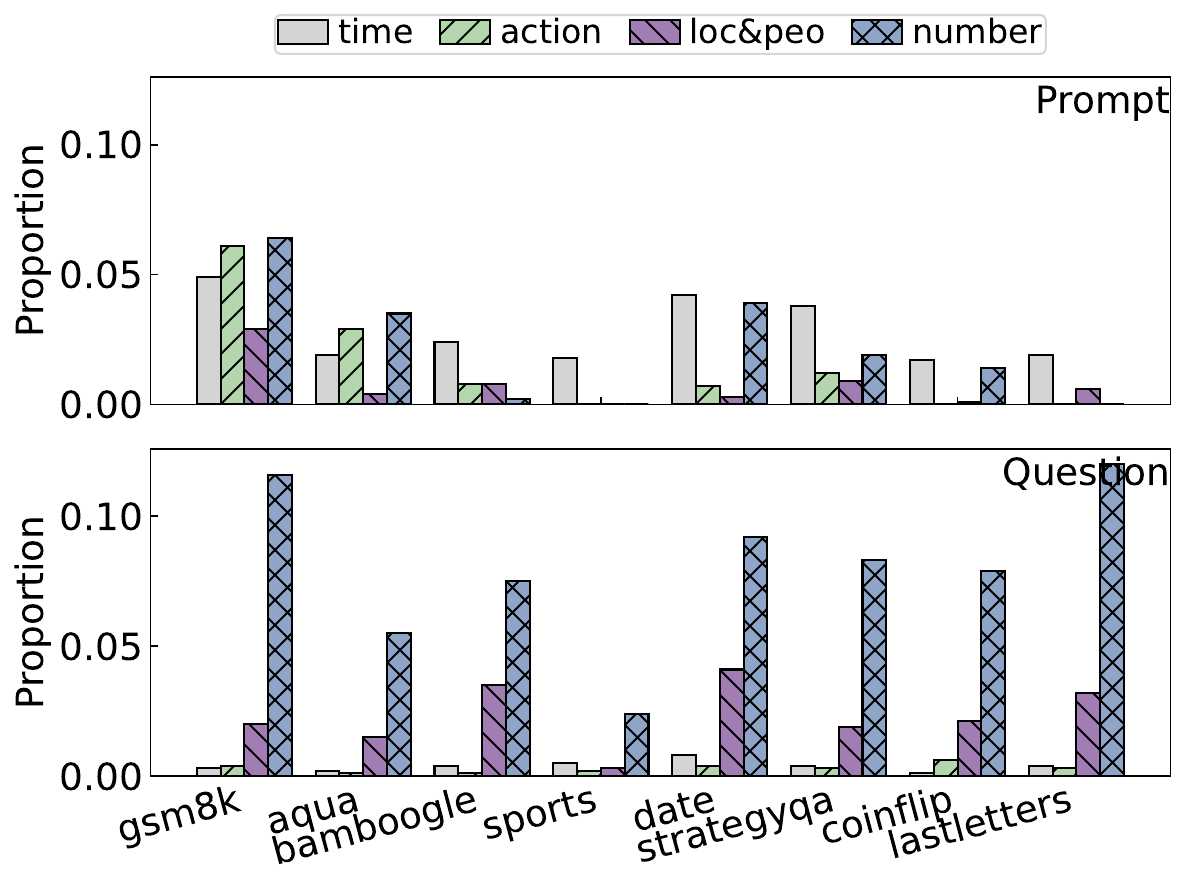}
        
        \caption{\textit{SVAMP}}
        \label{fig:imatition_svamp}
    \end{subfigure}
    \begin{subfigure}{0.32\linewidth}
        \includegraphics[width=\linewidth]{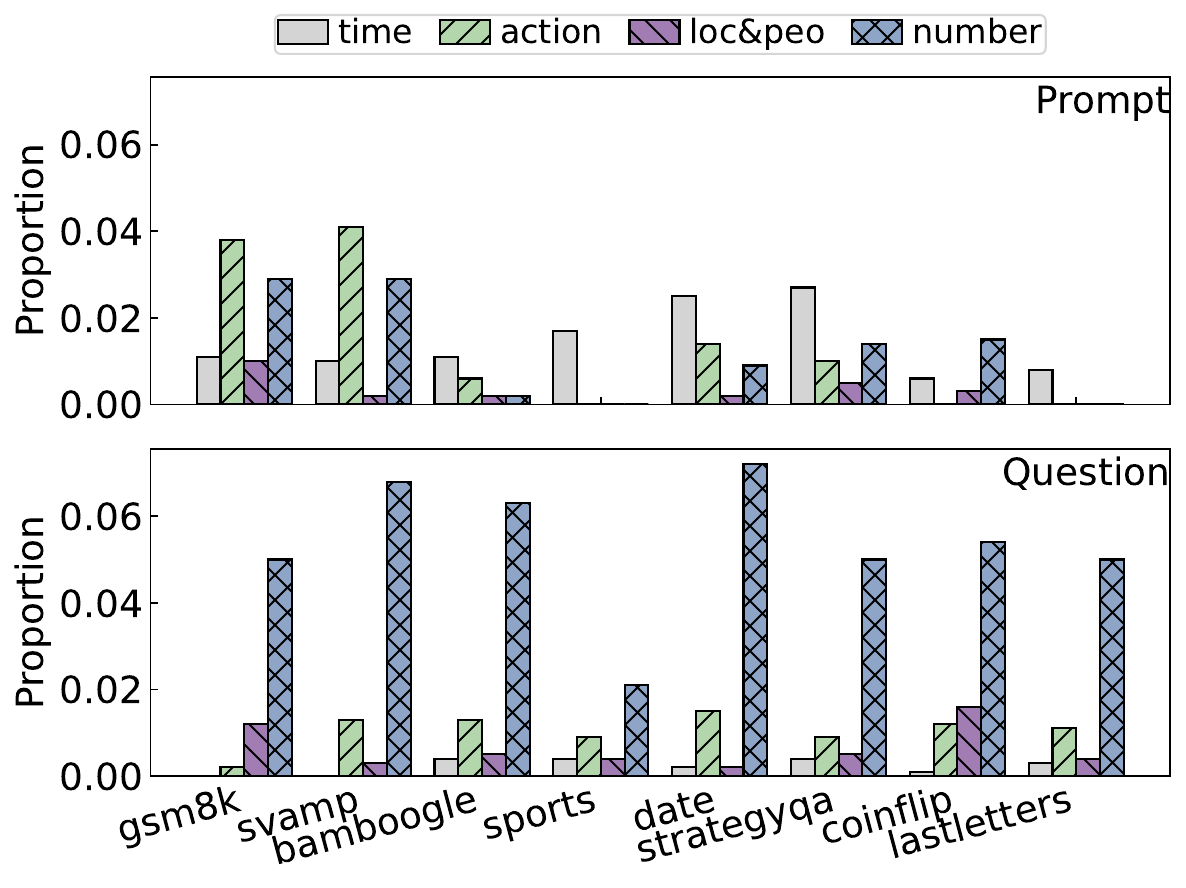}
        \caption{\textit{AQUA}}
        \label{fig:imatition_aqua}
    \end{subfigure}
    \begin{subfigure}{0.32\linewidth}
        \includegraphics[width=\linewidth]{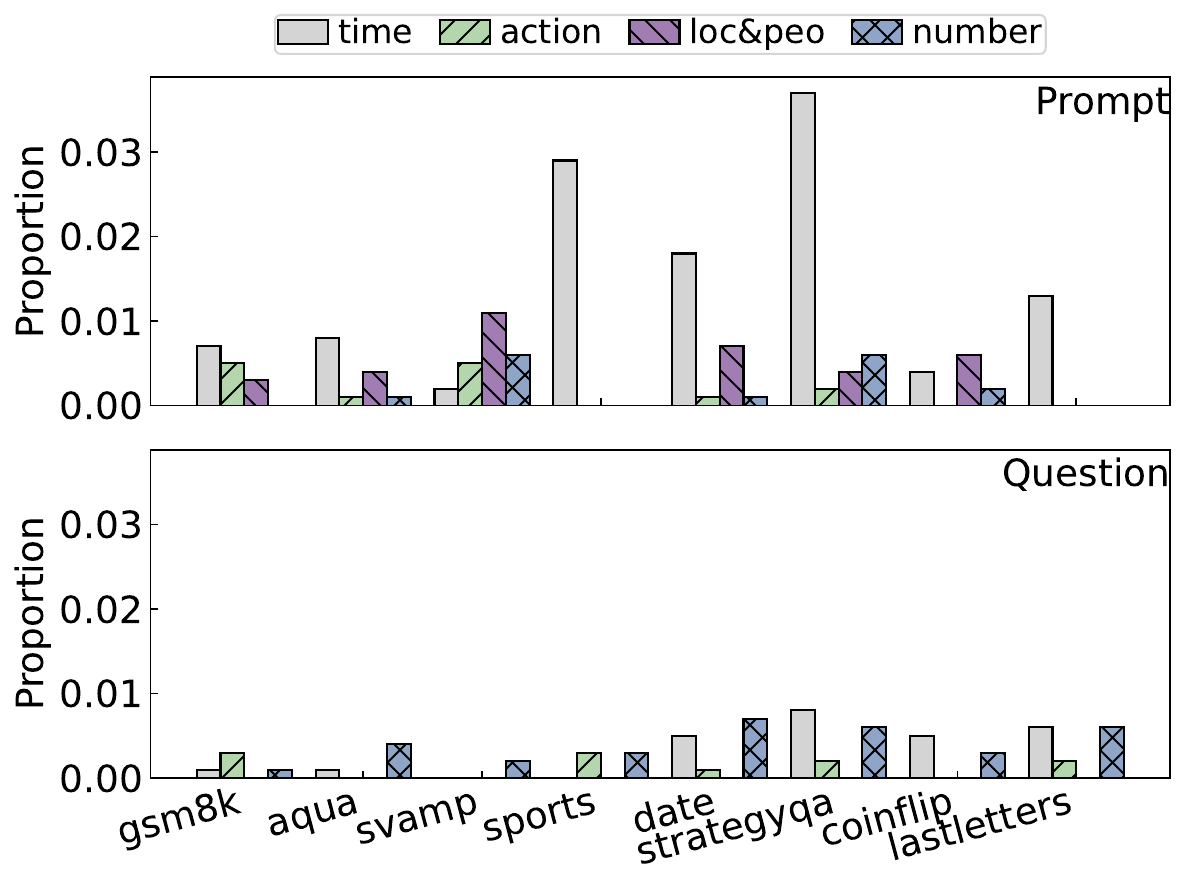}
        \caption{\textit{Bamboogle}}
        \label{fig:imatition_bamboogle}
    \end{subfigure}
    \begin{subfigure}{0.32\linewidth}
        \includegraphics[width=\linewidth]{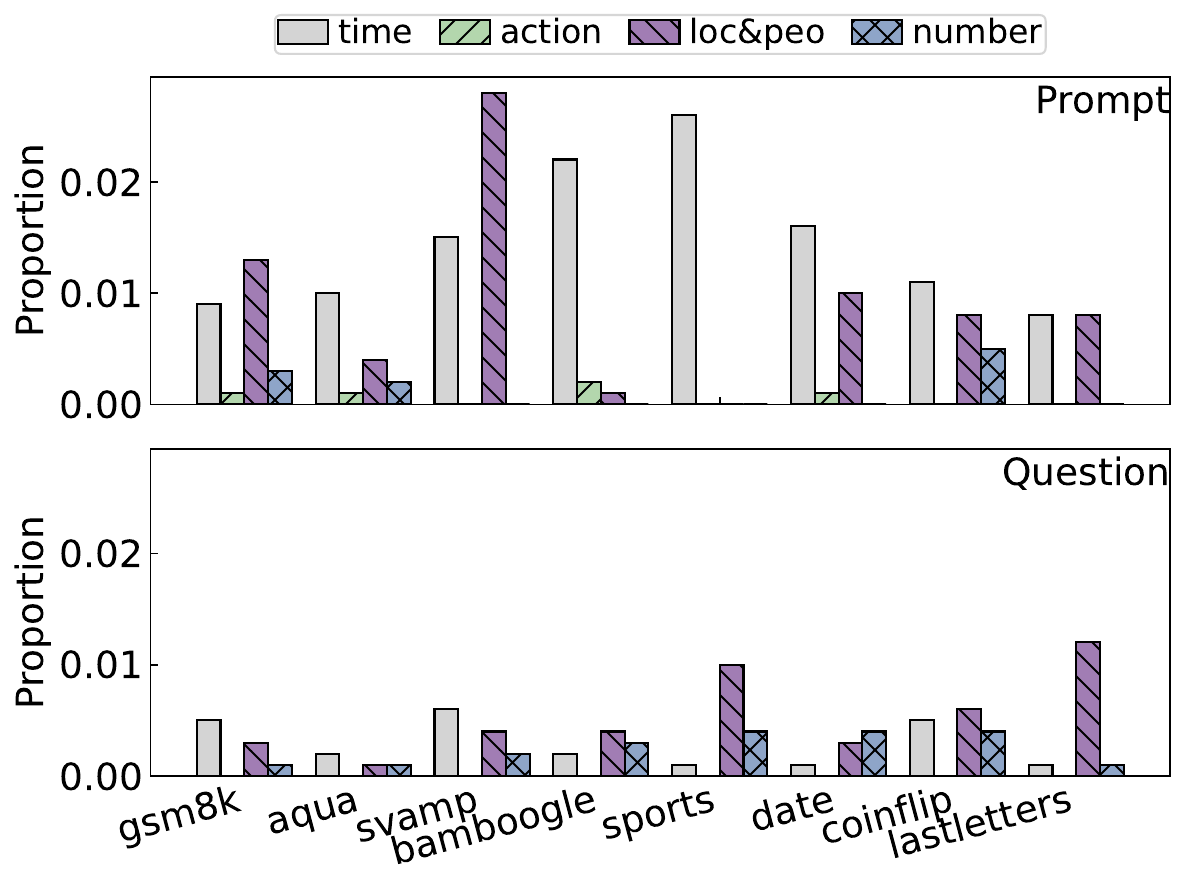}
        \caption{\textit{StrategyQA}}
        \label{fig:imatition_strategyqa}
    \end{subfigure}
    \begin{subfigure}{0.32\linewidth}
        \includegraphics[width=\linewidth]{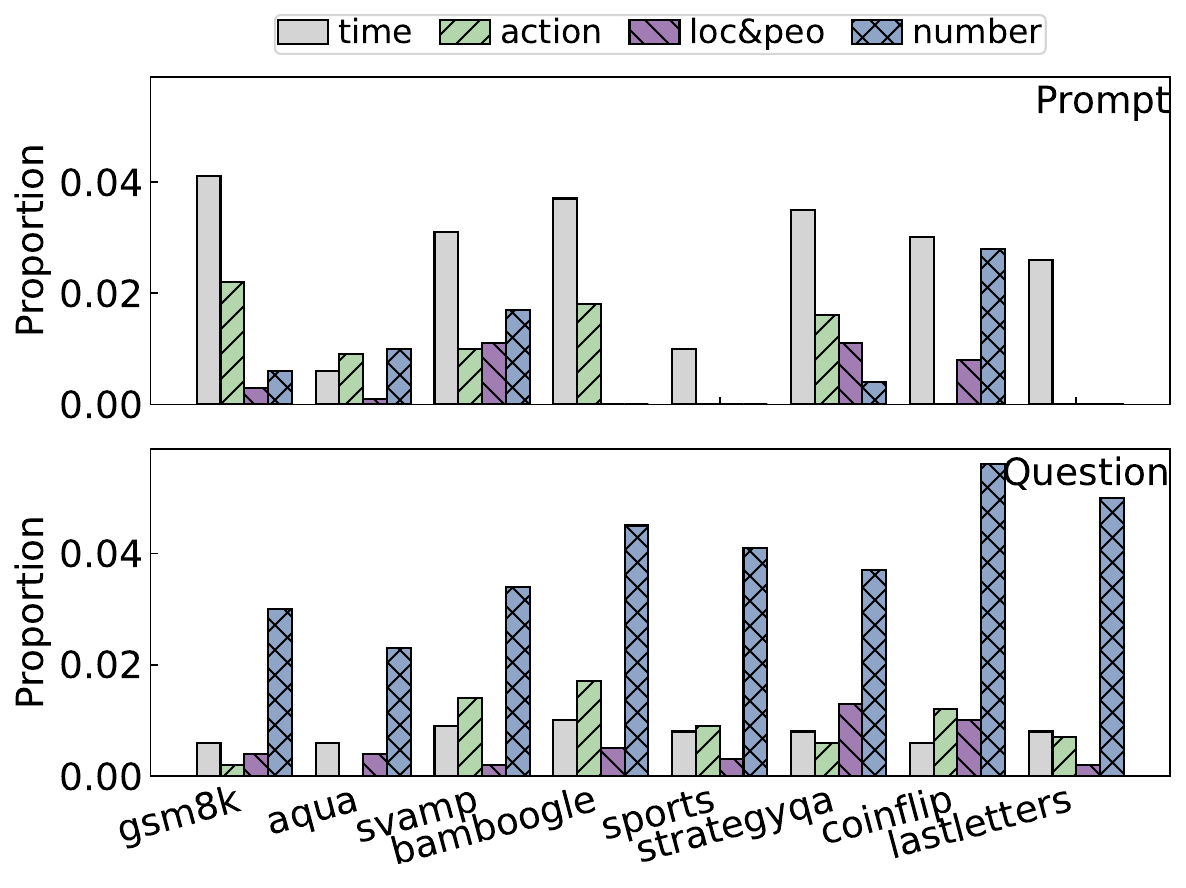}
        \caption{\textit{Date}}
        \label{fig:imatition_date}
    \end{subfigure}
    \begin{subfigure}{0.32\linewidth}
        \includegraphics[width=\linewidth]{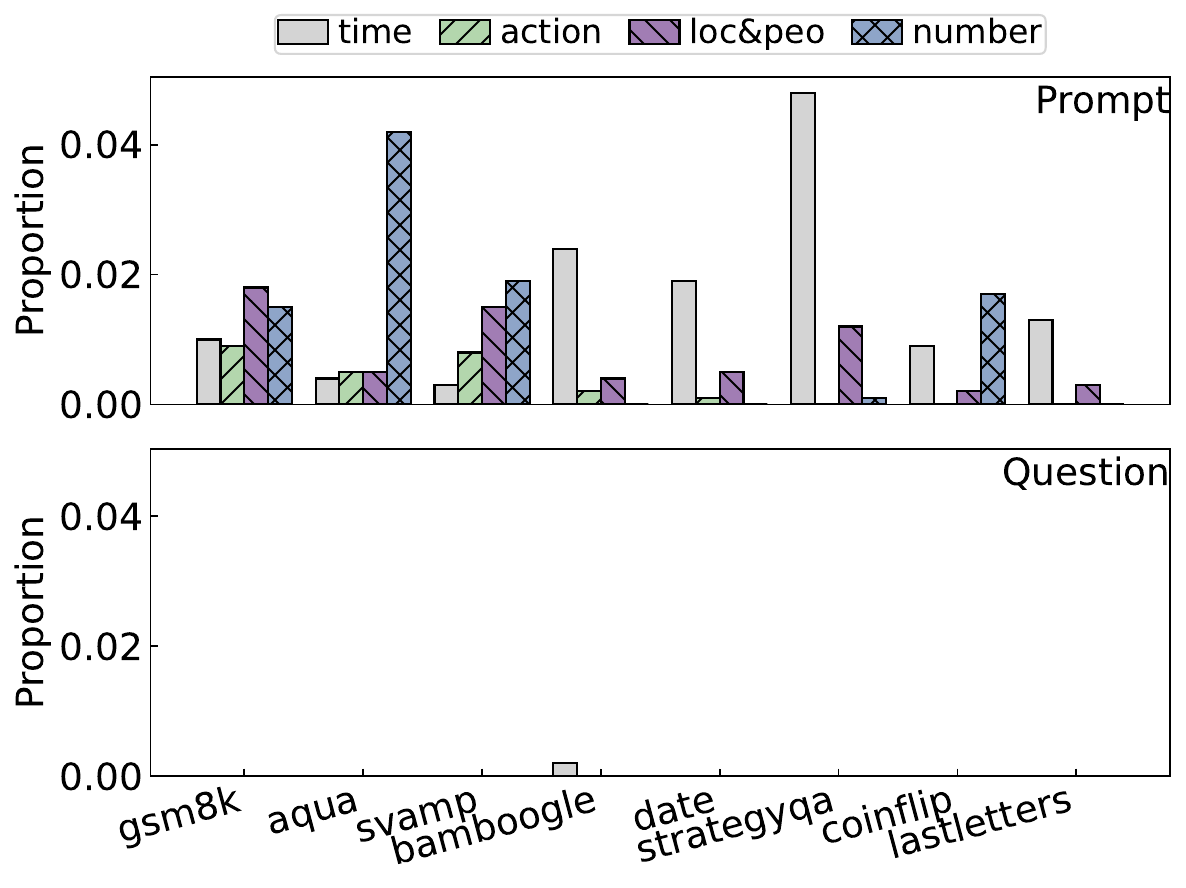}
        \caption{\textit{Sports}}
        \label{fig:imatition_sports}
    \end{subfigure}
    \begin{subfigure}{0.32\linewidth}
        \includegraphics[width=\linewidth]{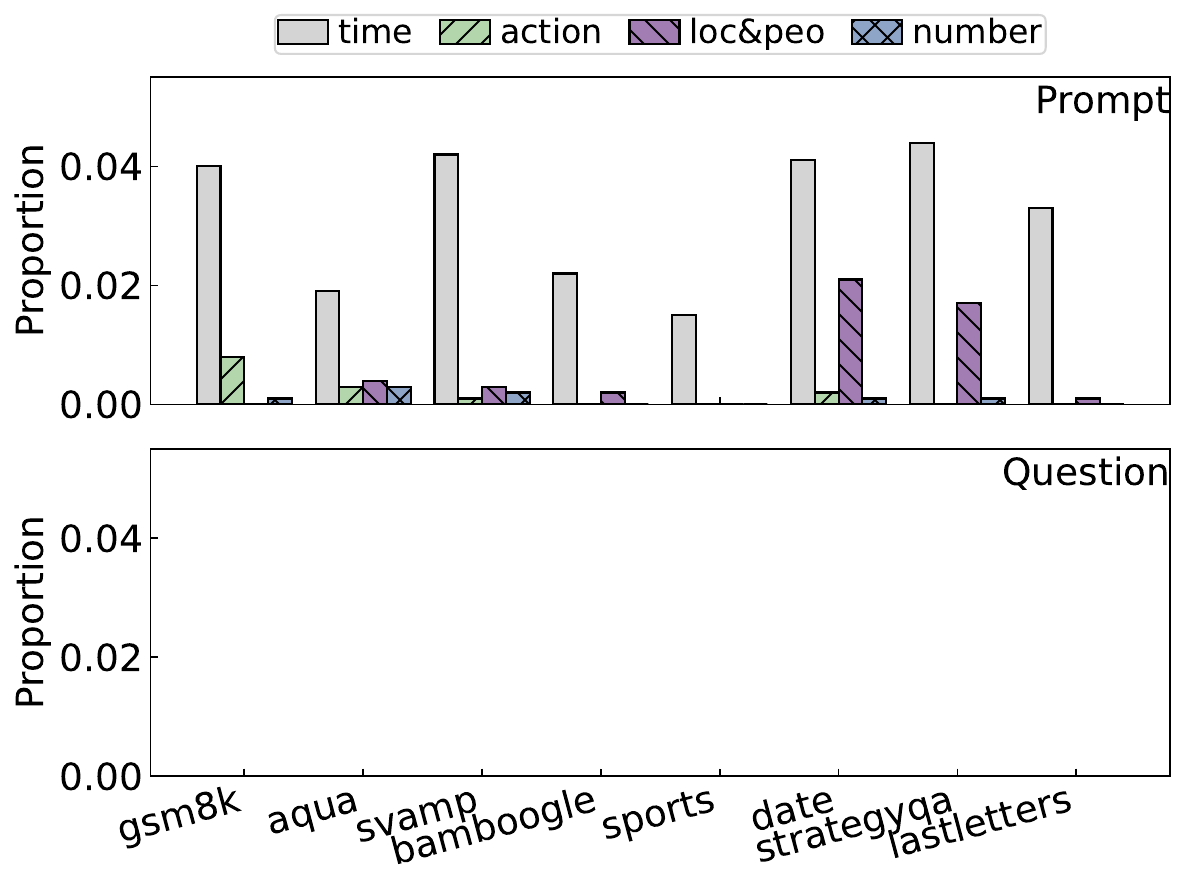}
        \caption{\textit{Coin Flip}}
        \label{fig:imatition_coinflip}
    \end{subfigure}
    \begin{subfigure}{0.32\linewidth}
        \includegraphics[width=\linewidth]{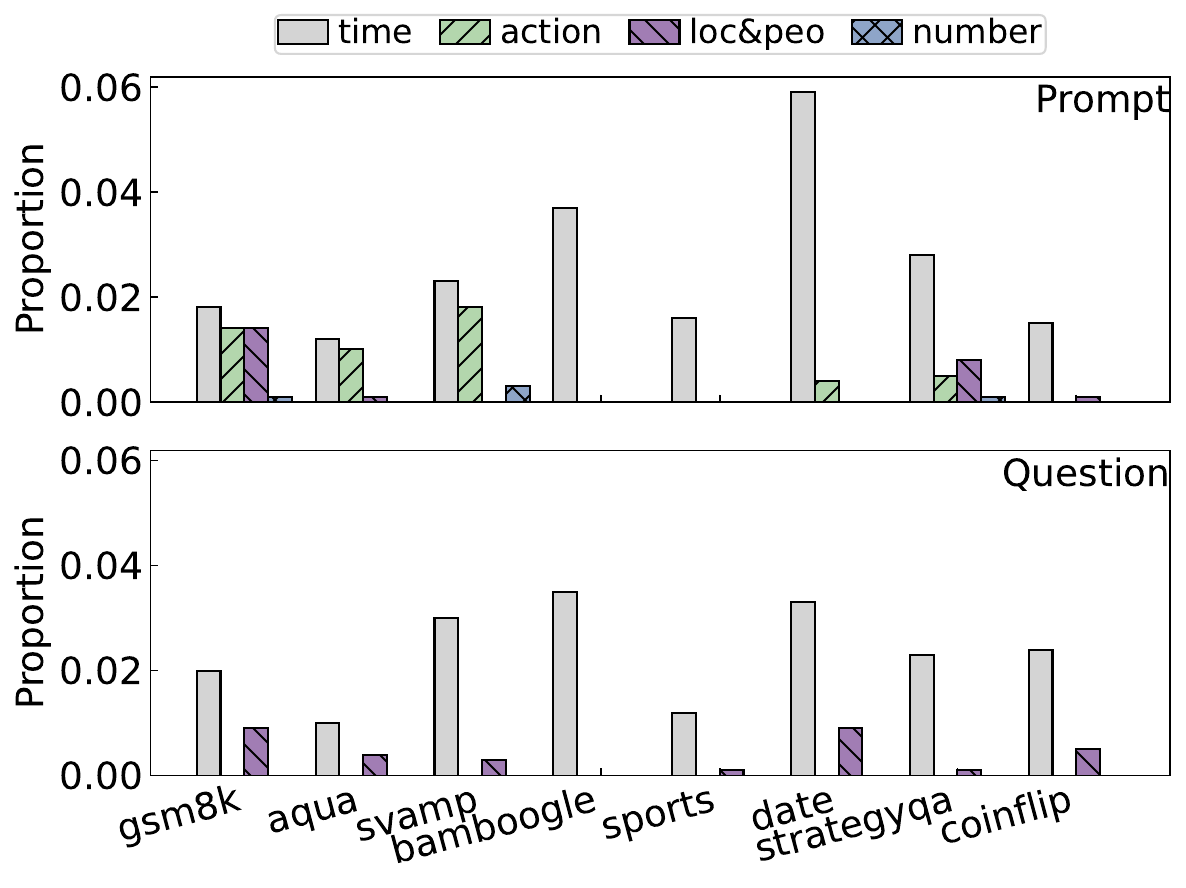}
        \caption{\textit{Last Letters Concatenation}}
        \label{fig:imatition_lastletter}
    \end{subfigure}
\caption{\textcolor{darkred}{Keyword imitation} results (Gemma2-27B is reported).}
\label{fig:transfer_test_27b}
\end{figure*}

\section{Methodology}
CoT prompts elicit step-by-step generation, so our analysis starts with decoding (generated tokens), tracing backward through projection (probability projections) to activation (FFN neurons). By analyzing these interconnected layers, we aim to provide a comprehensive understanding of CoT.

\subsection{\textcolor{darkred}{Decoding}} \label{sec:decoding}
CoT prompting reshapes model outputs by guiding the generation of intermediate reasoning steps before the final answer. We hypothesize that this guidance narrows the decoding space, yielding more structured, task-relevant outputs.  To investigate this, we analyzed the characteristics of the generated text from two perspectives: the {imitation of specific keywords} present in prompts and questions, and the {adherence to a answer structure} defining CoT reasoning.

\paragraph{\textcolor{darkred}{Keyword Imitation Analysis} }
Prior work \citep{madaan2022text, madaan2023makes, gudibande2023false} suggest that CoT improves reasoning by prompting models to mimic prompt formats, but the degree of imitation remains unquantified. To bridge this gap, we \textit{first} sought to  measure imitation by identifying specific keywords and comparing their presence in model outputs against the input (prompts and questions). To this end, we introduce the concept of ``\textit{test points}'', which are keywords reflecting  key reasoning aspects as observed in the generated CoT steps.

\textbf{Test Points.} We classified test points into four types based on observed patterns in input imitation: 1) \textit{time} (e.g., ``before'', ``therefore), indicating temporal order, logical sequence, or causality; 
2) \textit{action} (e.g., ``add'', ``increase''), representing operations; 
3) location and people (\textit{loc \& peo}) (e.g., ``there'', ``someone''); 
and 4) \textit{numbers} (e.g., ``1'', ``two'').
Keywords were extracted by analyzing prompts and questions across all test datasets. Using the \textit{Spacy} library, we performed part-of-speech tagging to identify candidate words (e.g., verbs for actions, adverbs for time), followed by manual verification for accurate categorization.  The keyword list is shown in Table \ref{tab:test_points}. Imitation is measured as the proportion of test points in the model’s generated text that match those in the prompt or question. Additionally, to assess the generalizability of CoT prompts and models’ adaptability to diverse reasoning structures, we conducted cross-dataset prompt transfer experiments. CoT prompts designed for dataset $X$ were applied to dataset $Y$, and we evaluated changes in generated content characteristics.

\noindent \textbf{Analysis.} Figure \ref{fig:transfer_test_27b} illustrates the average proportion of generated tokens that correspond to the four defined test points for both prompts and questions across different datasets.  Our analysis of keyword imitation reveals two primary patterns, shedding light on how CoT influences the structure and content of generated reasoning steps.
First, we observe \textbf{varied imitation patterns by source}: Models tend to imitate ``time'' and ``action'' keywords more frequently from the CoT prompts, suggesting that prompts effectively convey sequential and operational structuring cues for the reasoning process. In contrast, ``number''  keywords are often imitated more from the input questions, particularly in arithmetic datasets like GSM8K, SVAMP, AQUA, and Date (i.e., Figure \ref{fig:imatition_gsm8k} $\sim$ \ref{fig:imatition_aqua}, and \ref{fig:imatition_date}). This indicates that while CoT prompts provide the structural framework, models primarily extract specific content and entities directly from the problem statement to populate this structure.

\vspace{-1.3mm}
Second, we find \textbf{varied imitation patterns across datasets}: Imitation of question test points is notably \textit{lower} for tasks requiring significant external commonsense knowledge, such as Bamboogle, StrategyQA, and Sports (Figure \ref{fig:imatition_bamboogle}, \ref{fig:imatition_strategyqa}, and \ref{fig:imatition_sports}). This is likely because the necessary reasoning information is not contained within the input question, compelling models to rely more heavily on their internal knowledge. Similarly, in Coin Flip (Figure \ref{fig:imatition_coinflip}), the reasoning involves concepts not explicit in the question, reducing direct keyword imitation from the input. Conversely, the Last Letters Concatenation (Figure \ref{fig:imatition_lastletter}) exhibits balanced imitation, reflecting the high relevance of both prompt and question keywords to its structured symbolic task.

These findings demonstrate that  CoT prompting effectively steers models to produce text that conforms to a specified reasoning structure by selectively imitating structural keywords (e.g., ``time'', ``action'') from the prompt, while incorporating task-relevant content (e.g., ``number'') by mimicking keywords from the input question. This differential keyword imitation underpins CoT's ability to impose order on model outputs. We posit that such structural guidance narrows the decoding space, enhancing the focus and accuracy of token predictions, as further explored in the subsequent analysis of probability projections.

\vspace*{-1mm}
\paragraph{\textcolor{darkred}{Reasoning Structure Adherence Analysis} }
Building upon the keyword imitation findings, we further analyze the model's adherence to a higher-level, abstract reasoning structure commonly observed in effective CoT generations. Our observation that ``time'', ``action'', ``loc\&peo'', and ``number'' frequently fulfill specific syntactic roles within the reasoning process motivates the formalization of a CoT Reasoning Structure:
\vspace*{-2mm}
\begin{equation}
\label{eq:Cot_template}
    \texttt{Reasoning Structure: } \mathcal{E}_p  \xrightarrow{\mathcal{O}} \mathcal{E}_g + \mathcal{S}_l ,
\end{equation}
where $\mathcal{E}_p$ represents input entities (often corresponding to ``number'' or ``loc\&peo''), $\mathcal{O}$ represents reasoning operations or predicates  (i.e., ``action''), $\mathcal{E}_g$ signifies derived intermediate entities, and $\mathcal{S}_l$ is the final answer statement (i.e.,  ``the answer is...''). For example, in a step like ``3 + 2 = 5. So the answer is 5.'', ``3'' and ``2'' are  $\mathcal{E}_p$,  ``+'' is  $\mathcal{O}$, ``5'' is $\mathcal{E}_g$, and ``So the answer is 5.'' is $\mathcal{S}_l$. 
By quantifying the extent to which generated samples adhere to this structure (measured as ``Imitation Count'' based on keyword patterns and their sequence), we aim to assess the model's ability to capture the structural properties of CoT reasoning. We hypothesize that this ability to generate structurally coherent reasoning steps is essential for  multi-step problem-solving, as it provides a clear path towards the correct answer, thereby helping the model navigate the problem space more effectively.

\begin{figure}[!t]
\centering
\captionsetup[sub]{skip=-0.8mm}  
\setlength{\abovecaptionskip}{-1mm}
    \begin{subfigure}{0.49\linewidth}
        \includegraphics[width=\linewidth, height=0.8\linewidth]{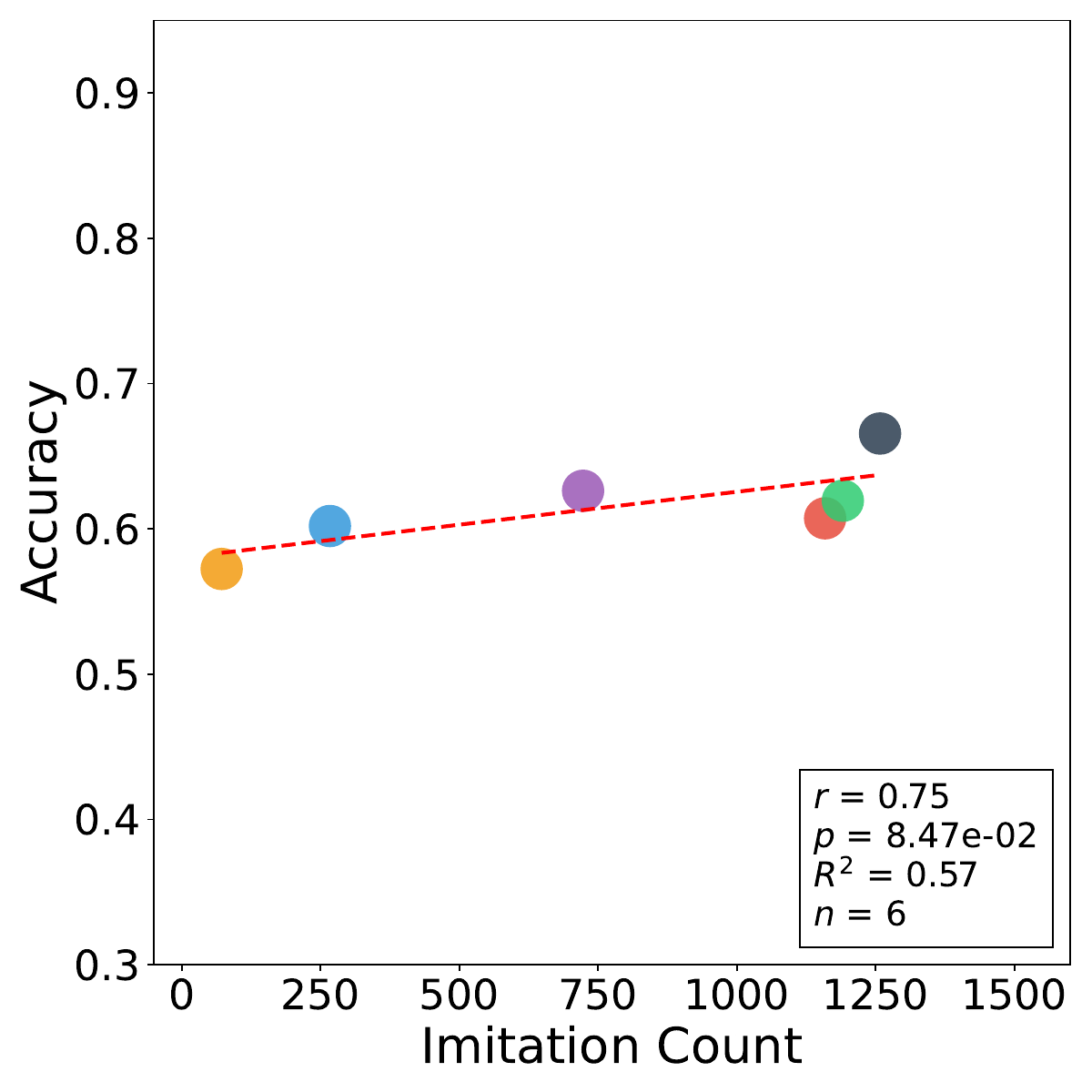}
        \caption{{LLaMA3.2-3B}} 
        \label{fig:question1_a2_a}
    \end{subfigure}
    \begin{subfigure}{0.49\linewidth}
        \includegraphics[width=\linewidth, height=0.8\linewidth]{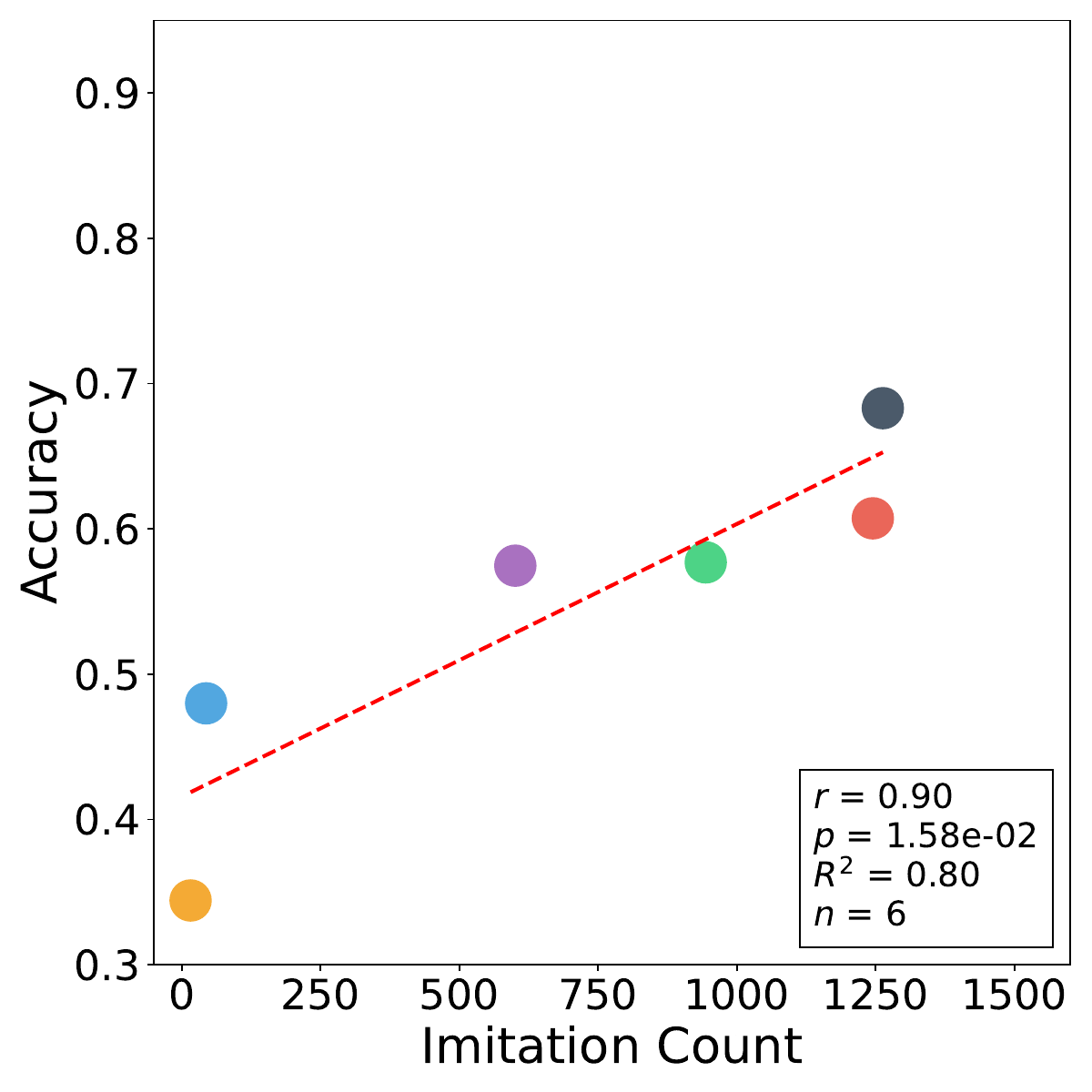}
        \caption{{Gemma2-9B}}
    \end{subfigure}
    \begin{subfigure}{0.49\linewidth}
        \includegraphics[width=\linewidth, height=0.9\linewidth]{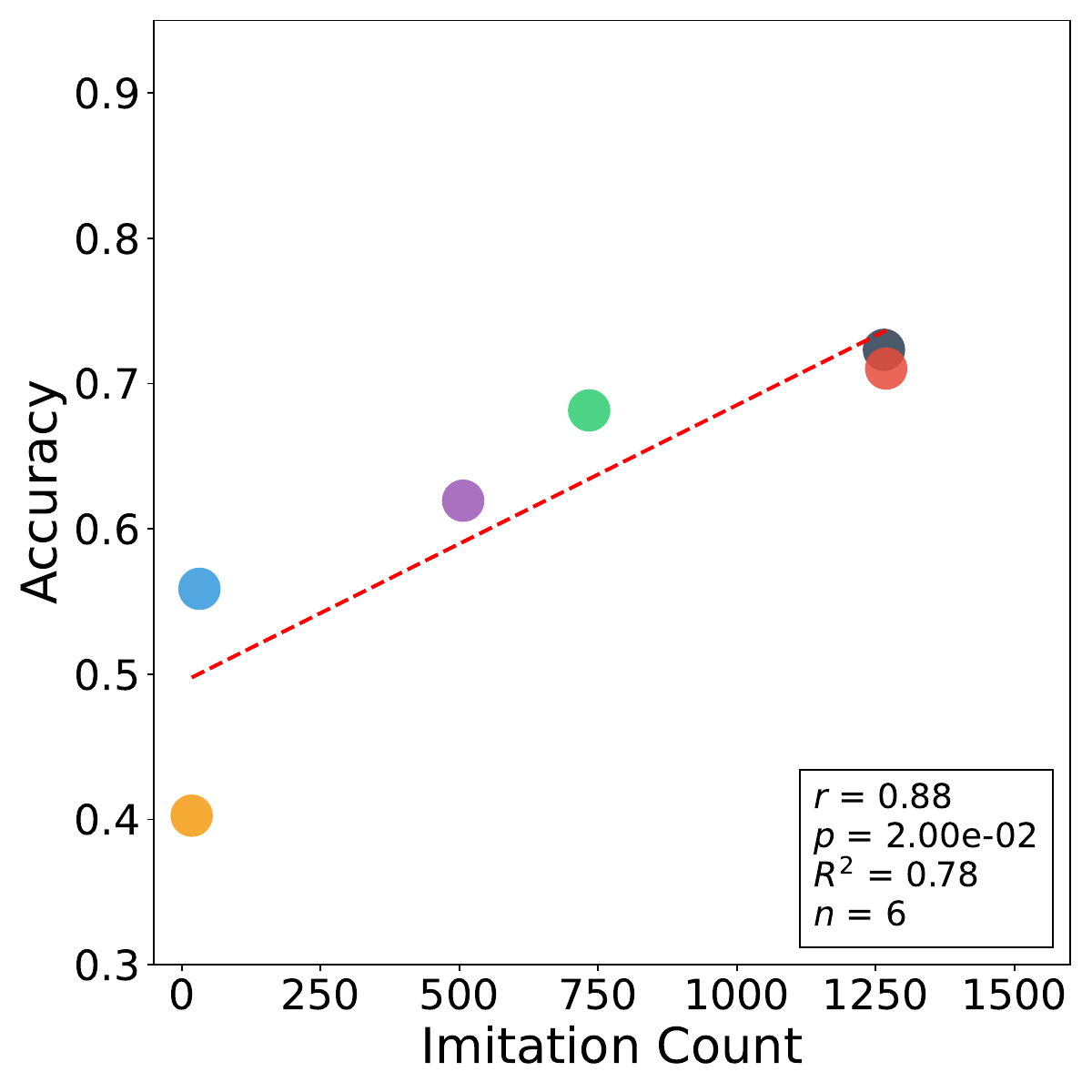}
        \caption{{Gemma2-27B}}
    \end{subfigure}
  \begin{subfigure}{0.49\linewidth}
        \includegraphics[width=\linewidth, height=0.9\linewidth]{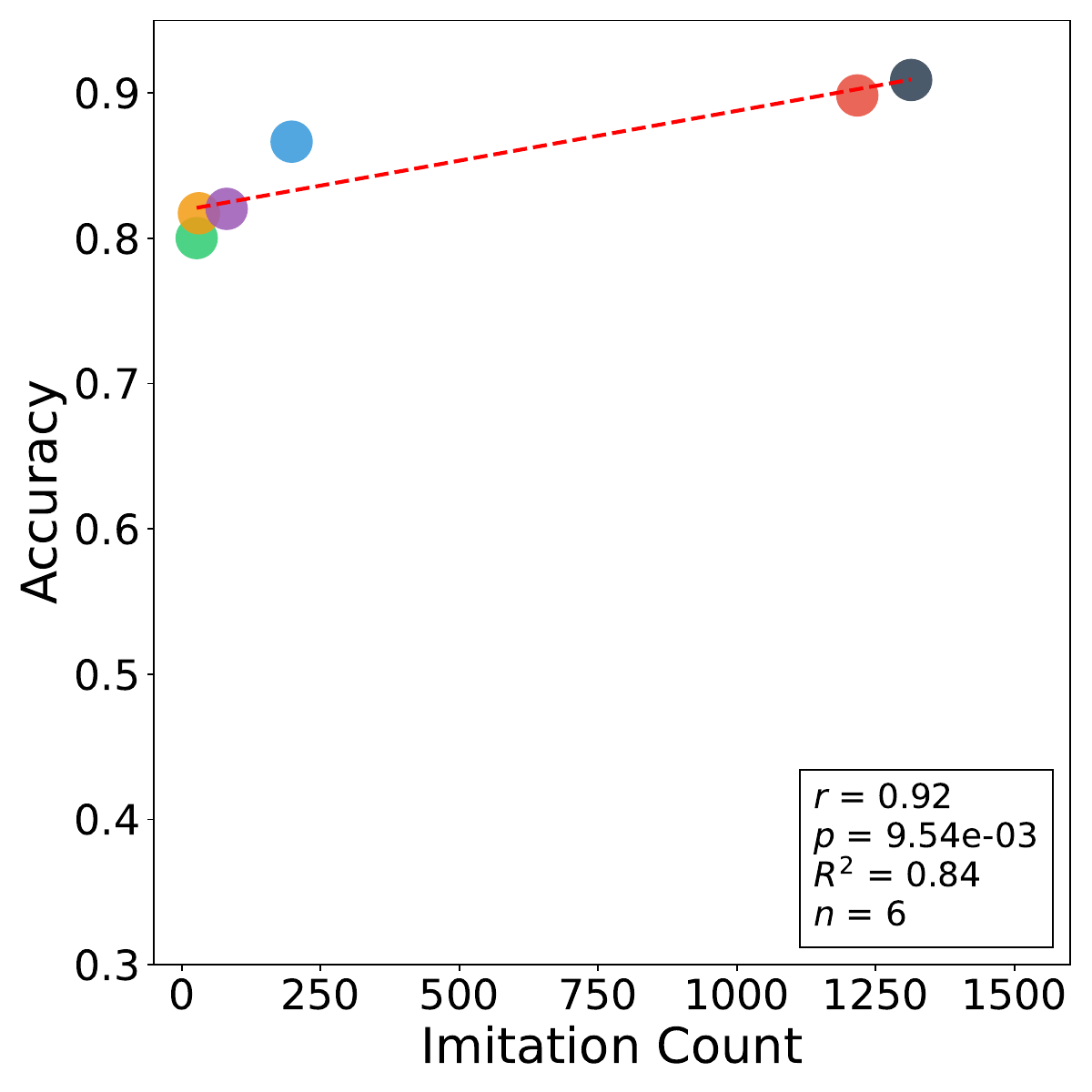}
        \caption{{LLaMA3.1-70B}}
    \end{subfigure}
      \begin{subfigure}{\linewidth}
      \centering
        \includegraphics[width=0.78\linewidth, height=0.1\linewidth]{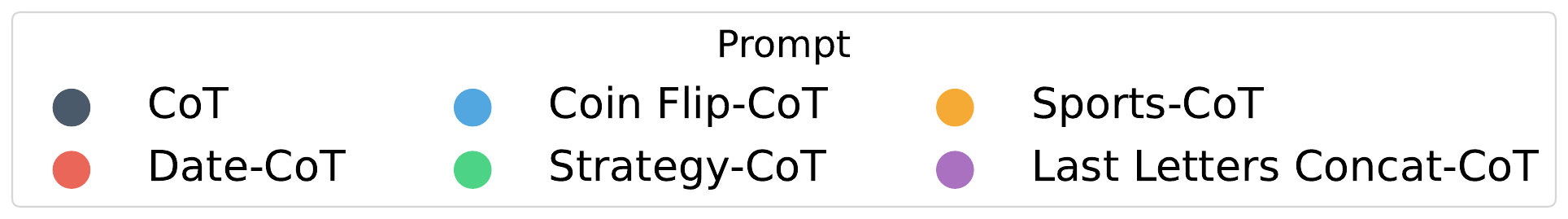}
    \end{subfigure}
    \vspace{1mm}
\caption{\textcolor{darkred}{Correlation} between reasoning structure adherence and task accuracy on the GSM8K dataset.
Pearson correlation coefficients ($r$), $p$-values, and $R^2$ scores
are reported for each model.}
\label{fig:correlation}
\end{figure}

\textbf{Analysis.}
We examine the relationship between adherence to the reasoning structure
(``Imitation Count'') and task accuracy on the GSM8K dataset using the original
and transferred CoT prompts (Figure~\ref{fig:correlation}).
We quantify this relationship using Pearson correlation
coefficients, $p$-values, and $R^2$ scores.

We observe a strong positive correlation between Imitation Count and
Accuracy. Across models, Pearson correlations range from $r=0.75$ to
$r=0.92$, with four out of five models achieving statistical significance
($p<0.05$); the remaining model exhibits a similar trend ($r=0.75$) but does not
reach statistical significance, likely due to its smaller model capacity.
As adherence to the CoT reasoning structure increases, GSM8K performance
consistently improves. High $R^2$ values ($0.57$--$0.84$) indicate that a substantial
portion of performance variance is explained by structural adherence alone.

Scatter plots further reveal differences between prompt types.
Prompts whose reasoning structures align with GSM8K (e.g., arithmetic or
sequential patterns) induce higher imitation counts and better performance,
whereas structurally misaligned prompts (e.g., Sports-CoT) lead to lower
adherence and accuracy.
These results suggest that CoT's effectiveness primarily stems from its ability
to guide the model toward generating a coherent reasoning structure, rather than
from the logical correctness of individual reasoning steps. This provides quantitative evidence that adherence to the
reasoning structure is a key factor underlying CoT's performance gains.

\begin{figure*}[!ht]
\centering
    \includegraphics[width=0.32\linewidth]{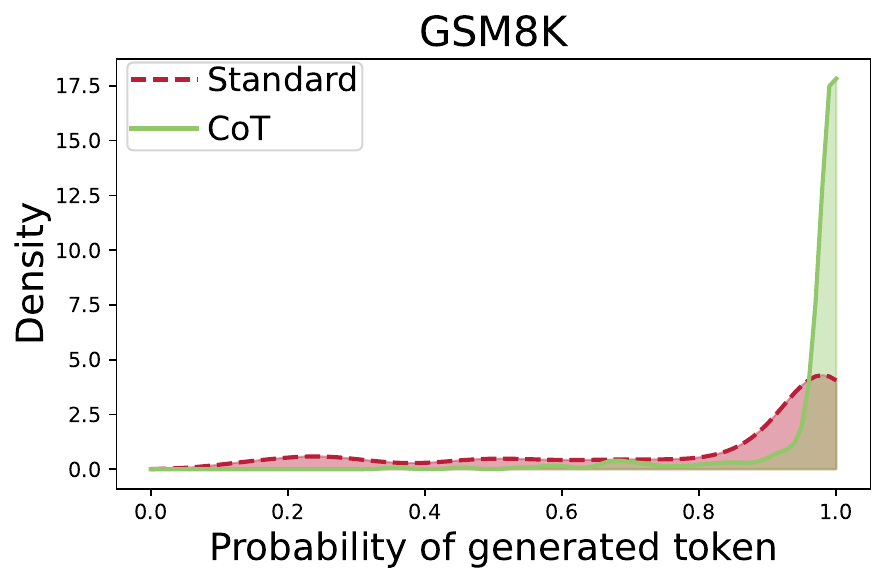}
    \includegraphics[width=0.32\linewidth]{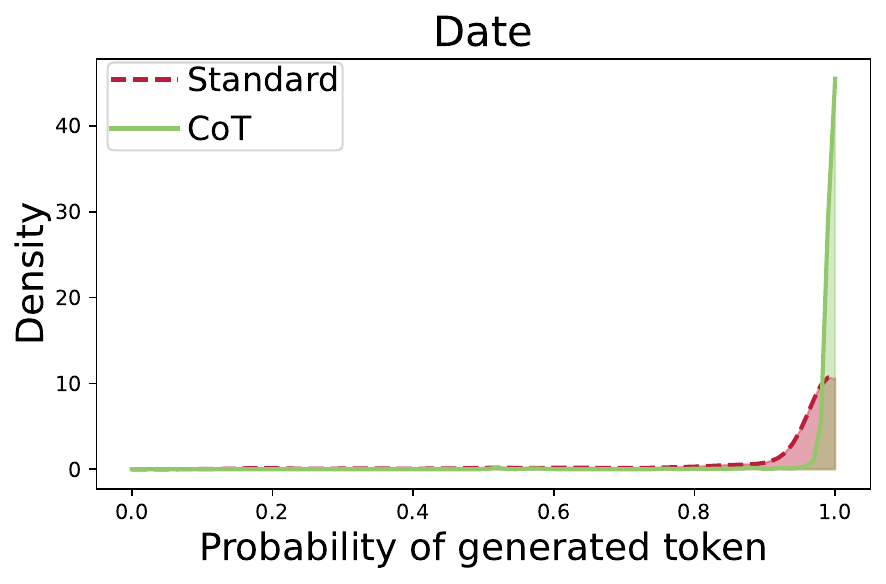}
     \includegraphics[width=0.32\linewidth]{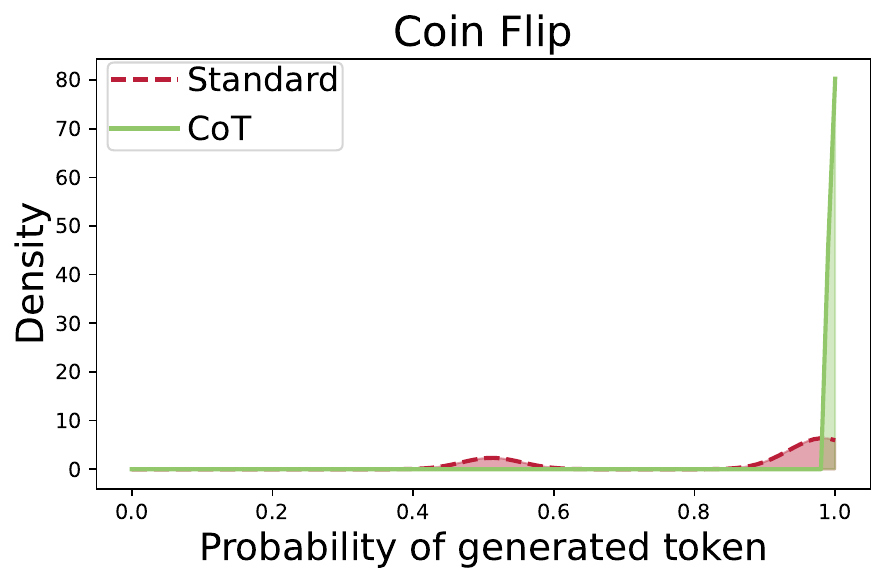}
    \vspace{-3.5mm}
    \caption{\textcolor{darkyellow}{Kernel density estimation} (Gemma2-9B is reported, see Appendix \ref{appendix:density} for more details).    }
    \label{fig:question2_horizontal_2}
\end{figure*}

\begin{figure*}[!ht]
\centering
\captionsetup[sub]{skip=-1mm}  
    \begin{subfigure}{0.32\linewidth}
        \includegraphics[width=\linewidth]{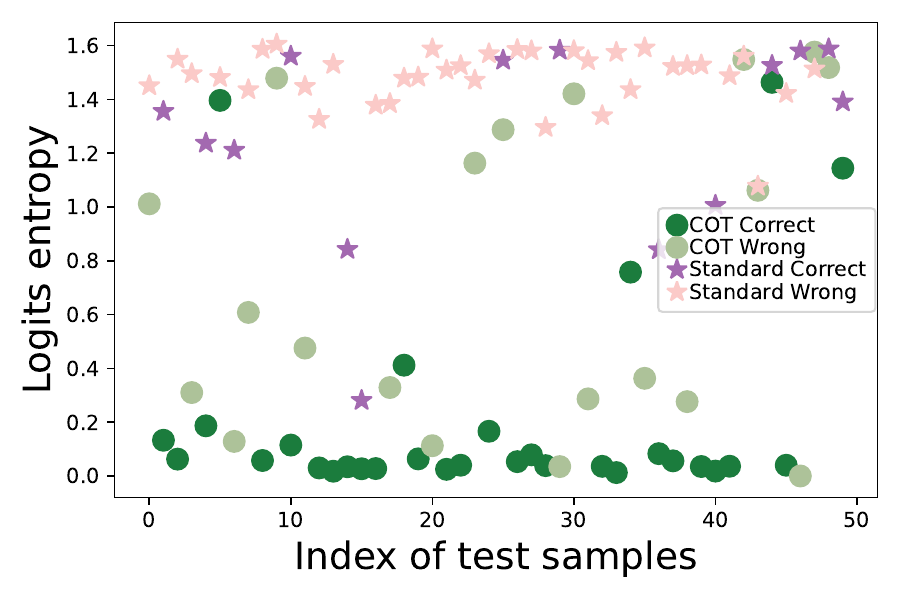}
        \caption{\textit{AQuA}}
    \end{subfigure}
    \begin{subfigure}{0.32\linewidth}
        \includegraphics[width=\linewidth]{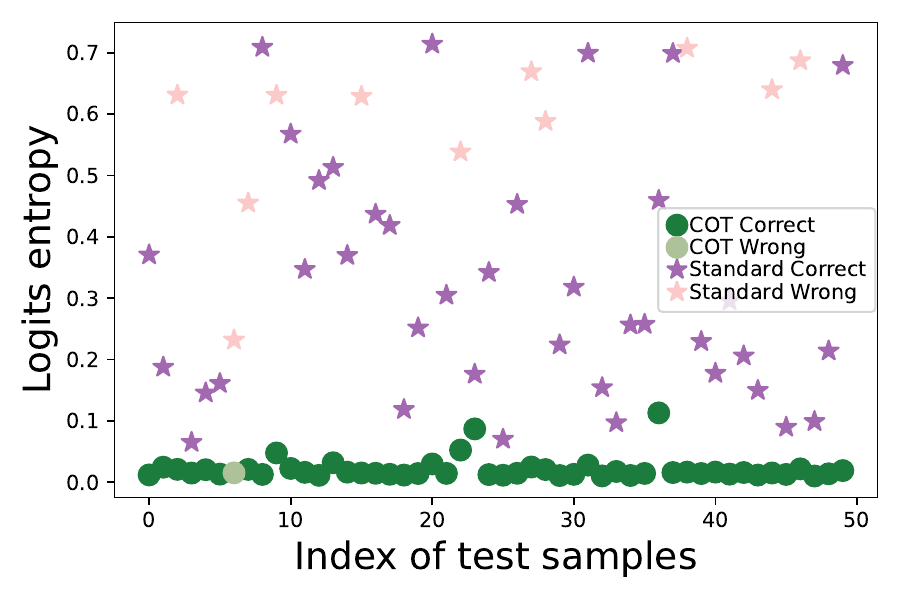}
        \caption{\textit{Sports}}
    \end{subfigure}
    \begin{subfigure}{0.32\linewidth}
        \includegraphics[width=\linewidth]{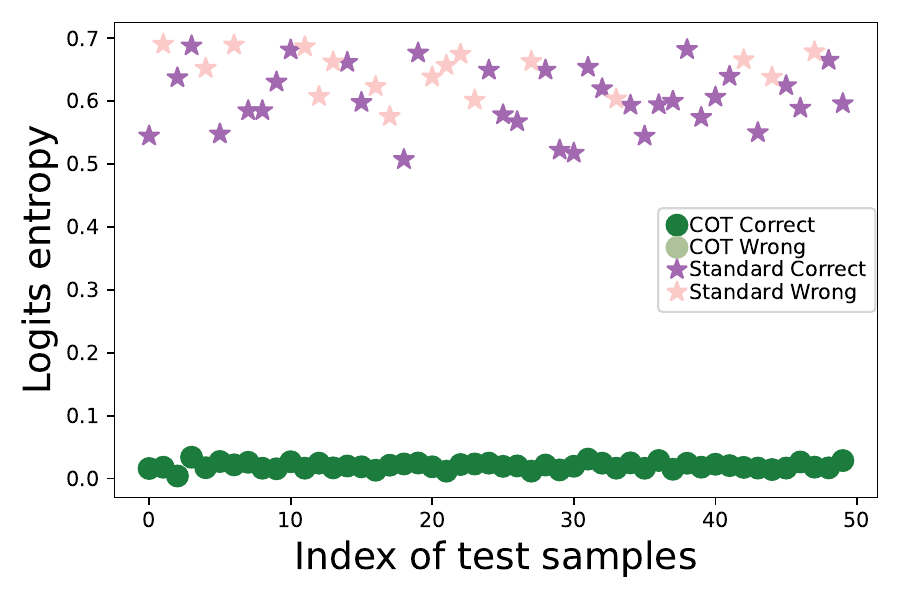}
        \caption{\textit{Coin Flip}}
    \end{subfigure}
\vspace{-3.5mm}
\caption{\textcolor{darkyellow}{Entropy} (Gemma2-27B is reported, see Appendix \ref{appendix:entropy} for more details)}
\label{fig:question2_vertical_entropy_27b}
\end{figure*}

\subsection{\textcolor{darkyellow}{Projection}}
To examine how CoT reshapes model behavior within the information flow framework, we analyzed the projection phase, where internal states are mapped to probability distributions over the vocabulary. We study this phase from two perspectives: the probability of generated sequences and the probability distribution of individual tokens.

\paragraph{\textcolor{darkyellow}{Probability of Generated Token \textit{Sequences}.}}
To evaluate the model's generation confidence, we analyzed the probability of generated token sequences, focusing on the common phrase ``answer is ...'' across all datasets and prompt types. We select this phrase as it marks the final decision point, revealing CoT’s impact on decision certainty. We model the probability sequence as $\mathcal{P} = [ p(\text{``answer''}), p(\text{``is''}), \ldots ]$ and compute kernel density estimates (KDE) to visualize the probability distribution’s \textit{density}. See Appendix \ref{appendix:density} for details about  KDE.

\paragraph{Analysis.} As shown in Figure~\ref{fig:question2_horizontal_2}, the generated probabilities of CoT  ($\mathcal{P}_{\texttt{CoT}}$) are \textit{consistently higher and more concentrated} than those of standard prompts ($\mathcal{P}_{\texttt{Standard}}$). This implies that with CoT, the model is assigning significant probability mass to a smaller and more specific set of next tokens at each step of generation. This shift in probability distribution indicates that CoT's structural guidance, provided through the intermediate steps (as discussed in Section \ref{sec:decoding}), effectively constrains the decoding space. By limiting the set of plausible next tokens and narrowing down the possible continuations, this guidance leads to reduced uncertainty and increased model conviction, facilitating more decisive generation of the concluding tokens.

\paragraph{\textcolor{darkyellow}{Probability Distribution of \textit{Individual} Tokens.}}
Beyond specific token probabilities, we analyzed the entire probability distribution across the vocabulary at each generation step. This allows us to understand  how the model's attention  was divided  across vocabulary. We used entropy, calculated as  $H(\mathcal{P}) = -\sum_{i=1}^{n} p_i \log(p_i)$ for a probability distribution   $H(\mathcal{P})$, to measure the uncertainty in the distribution, where lower entropy indicates a more concentrated distribution.
Additionally, to provide a controlled setting  allowing us to verify  the  probabilities  correspond to legitimate answer choices,  we strategically selected closed-domain datasets with fixed answer options, including AQuA (answer space: ``\textit{a, b, c, d, e}''), Sports, and Coin Flip (answer space: ``\textit{yes, no}'').


\paragraph{Analysis.}
Continuing our investigation into how CoT shapes information flow,
Figure~\ref{fig:question2_vertical_entropy_27b} shows token level entropy, reflecting the model’s predictive uncertainty.
We observe two key findings.
First, correct answers consistently exhibit lower entropy than incorrect ones,
aligning with prior work \citep{li2024rome}.
Second, CoT prompts yield substantially lower entropy than standard prompts,
indicating that CoT narrows the model’s predictive focus to more relevant tokens
and sharpens decision boundaries.
Overall, CoT fosters a more concentrated output distribution and reduced
uncertainty, suggesting that it fundamentally alters the generation process by
reducing ambiguity in token prediction, complementing our findings on structural
adherence in the Decoding phase.

\subsection{\textcolor{darkgreen}{Activation}}
Motivated by the findings of \citet{yi-etal-2024-badacts, rai-yao-2024-investigation, chen-etal-2024-learnable, voita-etal-2024-neurons}, which showed that different neurons are activated by different types of information, we analyzed the activated neurons in FFNs, which is formulated as:
\vspace{-2mm}
\[
\mathbf{h}^{(l)} = \mathbf{W}_{\text{down}}^{(l)}(\text{Act}(\mathbf{\tilde{h}}^{(l)} \mathbf{W}_{\text{up}}^{(l)}),
\]

where $\mathbf{\tilde{h}}^{(l)} \in \mathbb{R}^{d}$ is the hidden state output by attention module, serving as the input information flow to the FFN layer. $\mathbf{W}_{\text{up}}^{(l)} \in \mathbb{R}^{d \times d_1}$ is the upward projection weight, projecting $\mathbf{\tilde{h}}^{(l)}$ into higher-dimensional space (typically $d_1 \gg d$), enabling the FFN to represent a far greater number of features than its neuron count. $\text{Act}(\cdot)$ denotes the activation function (e.g., SwiGLU) and $\mathbf{W}_{\text{down}}^{(l)} \in \mathbb{R}^{d_1 \times d}$  maps the processed high-dimensional features back to the original space, reflecting the outflow of information.    $\text{Act}(\mathbf{\tilde{h}}^{(l)} \mathbf{W}_{\text{up}}^{(l)}) \in \mathbb{R}^{d_1}$ represents the  neurons. 
For GeLU or SwiGLU activation functions, a neuron is considered activated if its output is greater than zero, i.e., $\text{Act}(\mathbf{\tilde{h}}^{(l)} \mathbf{W}_{\text{up}}^{(l)}) > 0$. This definition aligns with prior work by \citet{geva-etal-2021-transformer, voita-etal-2024-neurons}

\paragraph{\textcolor{darkgreen}{Overall Neuron Activation.}}  At the generation step $t \in [1, \cdots, T]$,  the amount of activated neurons is calculated by:
\vspace{-3mm}
\[
A_t^{(l)} = \sum_{j=1}^{d_1} \mathbb{I}\left[\text{Act}(\mathbf{\tilde{h}}_{t}^{(l)},\mathbf{W}_{\text{up}}^{(l)})_j > 0\right],
\]
\vspace{-4mm}

where $\mathbb{I}[\cdot]$ is the indicator function and $\mathbf{\tilde{h}}_{t}^{(l)}$ represents the hidden state for token $x_t$ at layer $l$. The average Neuron Activation Count during generation is $\bar{\mathcal{A}} = \frac{1}{T} \sum_{t=1}^T \mathcal{A}_t$ and $\mathcal{A}_t = \sum_{l=1}^L A_t^{(l)}$.

\begin{figure}[!t]
\centering
\captionsetup[sub]{skip=-1mm}  
    \includegraphics[width=0.99\linewidth, height=0.4\linewidth]{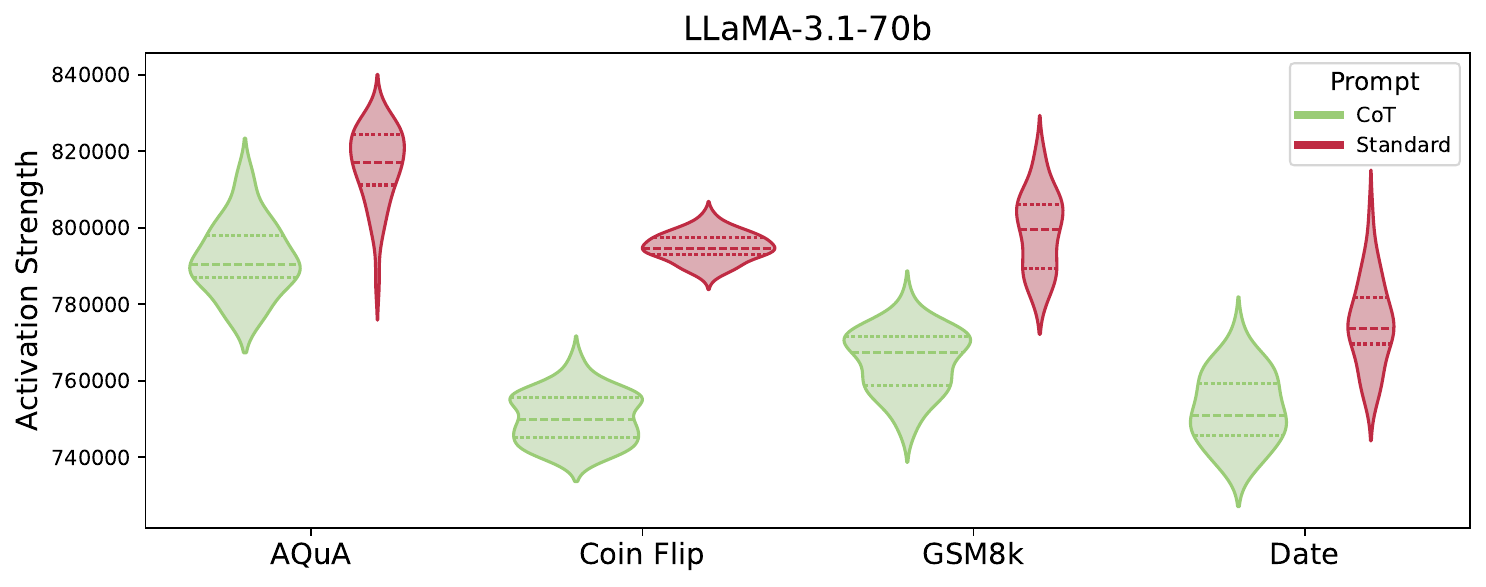}
    \vspace{-8mm}
    \caption{\textcolor{darkgreen}{Average Neuron Activation Count} (Full results are in Appendix~\ref{nurons}).
    }
    \label{fig:violin}
\end{figure}

\textbf{Analysis.}   
Figure \ref{fig:violin} presents the average Neuron Activation Count for CoT and standard prompts across various datasets for the LLaMA3.1-70B model. Across datasets, the distribution of  CoT generally exhibits a downward shift compared to standard prompts, resulting in a \textit{lower} overall number of activated neurons during the generation process. For instance, on the AQuA dataset, standard prompts engage $\sim820K$ neurons on average, whereas CoT prompts induce a lower average activation of  $\sim790K$ neurons. 

This observation suggests that CoT facilitates a more focused  processing regimen. Drawing upon our earlier findings, a potential mechanism is that CoT's structured decomposition of complex problems into sequential steps guides the model's processing attention. This focused guidance may allow FFNs to operate more selectively, activating a smaller, task-relevant subset of neurons at each step. This selective processing is hypothesized to be related to the reduced predictive uncertainty observed in the Projection phase.

Furthermore, we observed that the distribution for Coin Flip appears relatively ``short and wide'', indicating a more concentrated range of activation counts, while that for other datasets (e.g. GSM8K) is ``tall and narrow'', suggesting a tighter cluster around the mean. These differences likely reflect the varying activation patterns associated with different reasoning tasks. Tasks requiring discrete state tracking (e.g., ``head up'', ``even/odd'') might involve a different distribution of neural activity compared to tasks requiring continuaus processing of information (e.g, $25-9=16,16+3=19$). 

While these task-dependent variations in the distribution of overall activation counts are observed, the precise mechanistic implications of these specific distribution shapes and their relation to different reasoning strategies warrant further detailed investigation in future work.
While this analysis of overall activation provides insight into the general efficiency changes induced by CoT, a more nuanced and task-dependent picture of how CoT modulates neural activity emerges when examining activation differences at the layer level.

\begin{figure*}[!ht]
\centering
    \includegraphics[width=0.99\linewidth]{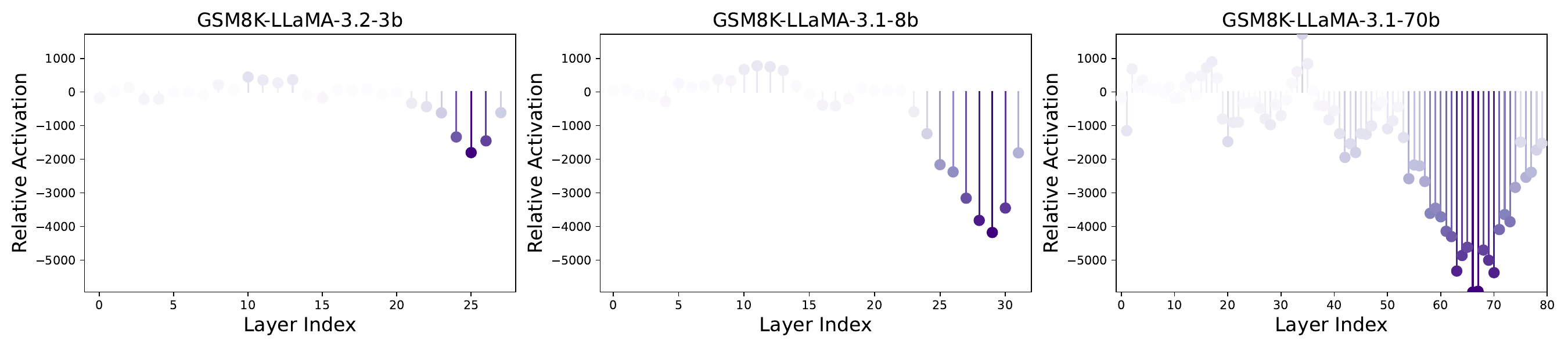} 
    \includegraphics[width=0.99\linewidth]{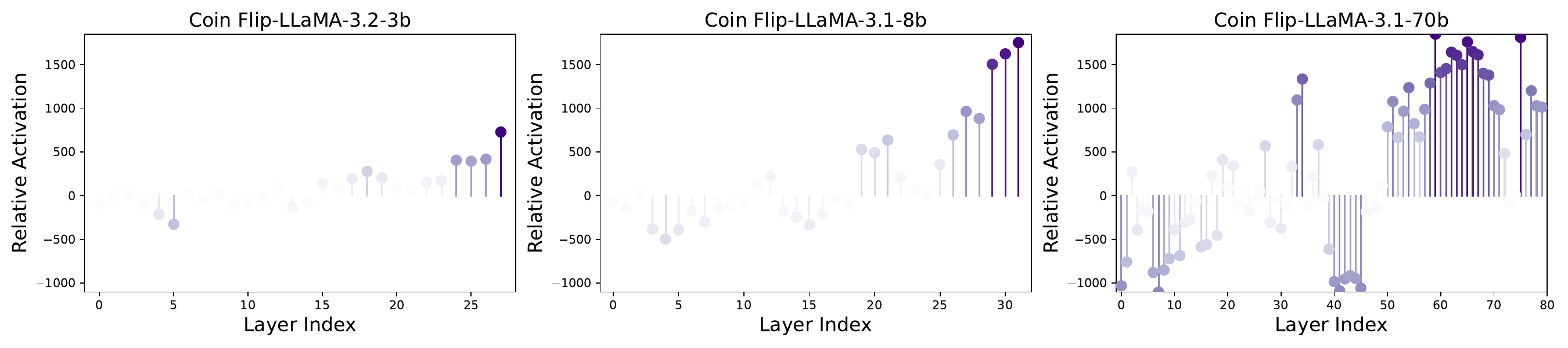}
\vspace{-5mm}
\caption{\textcolor{darkgreen}{Layer-wise activation differences} across layers for models (3B, 8B, 70B)  and datasets (GSM8K and Coin Flip). For full datasets results, please refer to Appendix \ref{appendix:activation_difference}.}
\label{fig:activation_difference}
\end{figure*}

\paragraph{\textcolor{darkgreen}{Layer-wise Activation Differences.}}
To understand precisely where in the model CoT influences neuron activation, we analyzed the difference in activated neuron count between CoT and Standard prompts at each layer. This layer-wise activation difference is defined as:
\vspace{-2mm}
\[
\mathcal{A}^{(l)} = \frac{1}{T} \sum_{t=1}^{T} \mathcal{A}^{(l)}{t}, \quad \Delta\mathcal{A}^{(l)} = \mathcal{A}_{\texttt{CoT}}^{(l)} - \mathcal{A}_{\texttt{Standard}}^{(l)}
\]

\paragraph{Analysis.}
Figure \ref{fig:activation_difference}  illustrates the evolution of this layer-wise activation difference ($\Delta\mathcal{A}^{(l)}$)  across model layers ($l$) for representative datasets, with more intense colors represent larger $| \Delta\mathcal{A}^{(l)}|$. This visualization reveals that:

 \textbf{Concentration in Later Layers.} The most pronounced differences  consistently occur in the final $1/3$ of the model's layers.  This implies that CoT primarily influences the later stages of processing within the model, which are typically associated with higher-level semantic processing, reasoning, and output generation. This observed concentration of CoT's impact in later layers may support  that CoT's guidance is most relevant during the structuring of final reasoning steps and the formulation of the answer, processes that are predominantly handled by these later layers.

 \textbf{Model Size Influence.}  Larger models (e.g., 70B) exhibit generally \textit{larger and more widespread layer-wise activation differences} compared to smaller models (e.g., 3B).   While the precise reasons are complex, these larger differences might reflect how CoT allows larger models to better utilize their greater representational capacity or interact differently with their hierarchical structures, potentially influencing a broader range of neuron functions or abstraction levels.

\textbf{Task-Specific Patterns.} 
A key finding  is the emergence of contrasting task-specific patterns in the layer-wise activation differences $\Delta\mathcal{A}^{(l)}$, particularly in later layers.  Specifically, for open-domain tasks like GSM8K and Bamboogle, we consistently observe $\Delta\mathcal{A}^{(l)} < 0$ in the final layers,  conversely, for closed-domain tasks such as Coin Flip, AQUA, and Sports, we find $\Delta\mathcal{A}^{(l)} > 0$ in the later layers.  These contrasting patterns reflect how CoT structures the model's processing in response to different task demands. 
We hypothesize that in open-domain tasks, which often require navigating a vast solution space, CoT's step-by-step guidance facilitates a more focused processing mode. By explicitly laying out a reasoning path, CoT could allow the model to selectively engage relevant features or knowledge, leading to the observed reduction in later-layer activation. 
Conversely, for closed-domain tasks, where the primary challenge is discriminating among limited options, CoT's guidance might encourage the model to thoroughly evaluate features associated with all plausible choices. 

This comprehensive consideration of options, potentially by activating relevant FFNs encoding those features, could explain the observed increase in later-layer activation.
These distinct modes of modulation suggest interpretive analogies for CoT's function at the neural level: akin to a ``pruner'' reducing activity in open-domain tasks, and an ``amplifier'' boosting relevant features within the predefined answer space for closed-domain task, revealing that CoT's influence on model processing 
\textit{maybe not uniform but dynamically tailored to the demands of the task}, providing novel insights into its operational principles.

\section{Conclusion}
In this work, we provided a comprehensive mechanistic interpretability analysis of CoT prompting by tracing information flow from decoding to activation. Our findings suggest that CoT may constrain the decoding space, reduces predictive uncertainty, and modulates neuron engagement in a task-dependent manner. 
These findings provide a deeper understanding of CoT's internal mechanisms and offer a novel mechanistic interpretability framework for analyzing and comparing prompting techniques for LLMs. 
Future research should investigate the interplay between task  properties (including difficulty and type)  in shaping activation patterns and information flow.

\section*{Limitations}
This study offers valuable mechanistic insights into the effects of CoT prompting on LLMs by analyzing information flow across different internal stages. However, it is crucial to contextualize our findings within the broader limitations inherent to the current state of LLM interpretability research and the fundamental opacity of these complex systems.

Modern LLMs, with their immense scale and intricate architectures, largely function as ``black boxes''. Our understanding of the precise roles of individual modules, the nature of high-dimensional representations, and how billions of parameters collectively give rise to complex behaviors like CoT-guided reasoning remains nascent. This fundamental lack of transparency makes it exceedingly difficult to establish definitive causal links between external inputs (like a CoT prompt) and specific internal computational processes or neural activities that drive observed outputs and performance gains. The problem of attribution, i.e.,  pinpointing why a model behaves in a certain way, is a significant challenge in this nascent field.

Consequently, research in LLM interpretability, including this work, is often necessarily fragmented. Studies tend to focus on isolated aspects of the model (e.g., specific layers, activation patterns, or particular types of inputs/outputs) or employ specific analytical techniques to probe individual phenomena. While each such piece contributes to the overall puzzle, assembling these disparate findings into a complete, unified, and fully causal mechanistic account of complex behaviors like CoT's influence is a significant undertaking that lies at the frontier of the field and is beyond the scope of any single study at this time.

Furthermore, a common challenge in current interpretability research, stemming from the difficulty in establishing clear causal chains within the black box, is that the relationship between experimental observations and derived conclusions is often heuristic or suggestive rather than strictly conclusive. We can observe strong correlations (e.g., between structural adherence and performance, or CoT and certain activation patterns) and propose plausible mechanistic hypotheses based on these correlations. However, definitively proving causality through direct intervention or formal verification within large-scale models remains a complex methodological hurdle for the field. Our findings, while empirically supported and offering valuable insights, should be interpreted as strong evidence supporting particular mechanistic hypotheses about CoT's operation within the constraints of current interpretability methods.

This work contributes significantly to empirically grounded investigations of CoT mechanisms under these prevailing conditions. A comprehensive and ultimately causal understanding of how CoT fundamentally alters LLM processing will require continued advancements across the entire field of mechanistic interpretability.

\section*{Acknowledgements}
This work was supported by the National Natural Science Foundation of China (Grant No.~62306138), the Jiangsu Natural Science Foundation (Grant No.~BK20230784), and the Innovation Program of the State Key Laboratory for Novel Software Technology at Nanjing University (Grant Nos.~ZZKT2024B15 and ZZKT2025B25). This work was also partially supported by the Hefei College Talent Research Fund Project (No.~24RC20) and the Scientific Research Project of the Anhui Provincial Education Department (No.~2025AHGXZK40379).
\bibliography{custom}

@inproceedings{
chenboosting,
title={Boosting of Thoughts: Trial-and-Error Problem Solving with Large Language Models},
author={Sijia Chen and Baochun Li and Di Niu},
booktitle={The Twelfth International Conference on Learning Representations},
year={2024},
url={https://openreview.net/forum?id=qBL04XXex6}
}

@inproceedings{bi2024program,
author = {Bi, Zhen and Zhang, Ningyu and Jiang, Yinuo and Deng, Shumin and Zheng, Guozhou and Chen, Huajun},
title = {When do program-of-thought works for reasoning?},
year = {2024},
isbn = {978-1-57735-887-9},
url = {https://doi.org/10.1609/aaai.v38i16.29721},
doi = {10.1609/aaai.v38i16.29721},
booktitle = {Proceedings of the Thirty-Eighth AAAI Conference on Artificial Intelligence and Thirty-Sixth Conference on Innovative Applications of Artificial Intelligence and Fourteenth Symposium on Educational Advances in Artificial Intelligence},
articleno = {1973},
numpages = {9},
}

@inproceedings{li2023mot,
    title = "{M}o{T}: Memory-of-Thought Enables {C}hat{GPT} to Self-Improve",
    author = "Li, Xiaonan  and
      Qiu, Xipeng",
    editor = "Bouamor, Houda  and
      Pino, Juan  and
      Bali, Kalika",
    booktitle = "Proceedings of the 2023 Conference on Empirical Methods in Natural Language Processing",
    month = dec,
    year = "2023",
    address = "Singapore",
    publisher = "Association for Computational Linguistics",
    url = "https://aclanthology.org/2023.emnlp-main.392/",
    doi = "10.18653/v1/2023.emnlp-main.392",
    pages = "6354--6374"
}

@article{tanneru2024hardness,
  title={On the Hardness of Faithful Chain-of-Thought Reasoning in Large Language Models},
  author={Tanneru, Sree Harsha and Ley, Dan and Agarwal, Chirag and Lakkaraju, Himabindu},
  journal={arXiv preprint arXiv:2406.10625},
  year={2024}
}

@inproceedings{jin2024impact,
    title = "The Impact of Reasoning Step Length on Large Language Models",
    author = "Jin, Mingyu  and
      Yu, Qinkai  and
      others",
    editor = "Ku, Lun-Wei  and
      Martins, Andre  and
      Srikumar, Vivek",
    booktitle = "Findings of the Association for Computational Linguistics: ACL 2024",
    month = aug,
    year = "2024",
    address = "Bangkok, Thailand",
    publisher = "Association for Computational Linguistics",
    url = "https://aclanthology.org/2024.findings-acl.108/",
    doi = "10.18653/v1/2024.findings-acl.108",
    pages = "1830--1842"
}

@inproceedings{madaan2023makes,
    title = "What Makes Chain-of-Thought Prompting Effective? A Counterfactual Study",
    author = "Madaan, Aman  and
      Hermann, Katherine  and
      Yazdanbakhsh, Amir",
    editor = "Bouamor, Houda  and
      Pino, Juan  and
      Bali, Kalika",
    booktitle = "Findings of the Association for Computational Linguistics: EMNLP 2023",
    month = dec,
    year = "2023",
    address = "Singapore",
    publisher = "Association for Computational Linguistics",
    url = "https://aclanthology.org/2023.findings-emnlp.101/",
    doi = "10.18653/v1/2023.findings-emnlp.101",
    pages = "1448--1535"
}

@article{chia2023contrastive,
  title={Contrastive chain-of-thought prompting},
  author={Chia, Yew Ken and Chen, Guizhen and others},
  journal={arXiv preprint arXiv:2311.09277},
  year={2023}
}

@inproceedings{wang2023towards,
    title = "Towards Understanding Chain-of-Thought Prompting: An Empirical Study of What Matters",
    author = "Wang, Boshi  and
      Min, Sewon  and
      Deng, Xiang  and
      Shen, Jiaming  and
      Wu, You  and
      Zettlemoyer, Luke  and
      Sun, Huan",
    editor = "Rogers, Anna  and
      Boyd-Graber, Jordan  and
      Okazaki, Naoaki",
    booktitle = "Proceedings of the 61st Annual Meeting of the Association for Computational Linguistics (Volume 1: Long Papers)",
    month = jul,
    year = "2023",
    address = "Toronto, Canada",
    publisher = "Association for Computational Linguistics",
    url = "https://aclanthology.org/2023.acl-long.153/",
    doi = "10.18653/v1/2023.acl-long.153",
    pages = "2717--2739"
}

@article{wu2023analyzing,
  title={Analyzing chain-of-thought prompting in large language models via gradient-based feature attributions},
  author={Wu, Skyler and Shen, Eric Meng and Badrinath, Charumathi and Ma, Jiaqi and Lakkaraju, Himabindu},
  journal={arXiv preprint arXiv:2307.13339},
  year={2023}
}

@inproceedings{ye2023complementary,
    title = "Complementary Explanations for Effective In-Context Learning",
    author = "Ye, Xi  and
      Iyer, Srinivasan  and
      Celikyilmaz, Asli  and
      Stoyanov, Veselin  and
      Durrett, Greg  and
      Pasunuru, Ramakanth",
    editor = "Rogers, Anna  and
      Boyd-Graber, Jordan  and
      Okazaki, Naoaki",
    booktitle = "Findings of the Association for Computational Linguistics: ACL 2023",
    month = jul,
    year = "2023",
    address = "Toronto, Canada",
    publisher = "Association for Computational Linguistics",
    url = "https://aclanthology.org/2023.findings-acl.273/",
    doi = "10.18653/v1/2023.findings-acl.273",
    pages = "4469--4484"
}

@inproceedings{schaeffer2023invalid,
  title = {Invalid Logic, Equivalent Gains: The Bizarreness of Reasoning in Language Model Prompting},
  author = {Schaeffer, Rylan and Pistunova, Kateryna and others},
  booktitle = {ICML 2023 Workshop on Knowledge and Logical Reasoning in the Era of Data-driven Learning},
  year = {2023},
  url = {https://icml.cc/virtual/2023/27230},
}

@inproceedings{prabhakar2024deciphering,
    title = "Deciphering the Factors Influencing the Efficacy of Chain-of-Thought: Probability, Memorization, and Noisy Reasoning",
    author = "Prabhakar, Akshara  and
      Griffiths, Thomas L.  and
      McCoy, R. Thomas",
    editor = "Al-Onaizan, Yaser  and
      Bansal, Mohit  and
      Chen, Yun-Nung",
    booktitle = "Findings of the Association for Computational Linguistics: EMNLP 2024",
    month = nov,
    year = "2024",
    address = "Miami, Florida, USA",
    publisher = "Association for Computational Linguistics",
    url = "https://aclanthology.org/2024.findings-emnlp.212/",
    doi = "10.18653/v1/2024.findings-emnlp.212",
    pages = "3710--3724"
}

@article{madaan2022text,
  title={Text and patterns: For effective chain of thought, it takes two to tango},
  author={Madaan, Aman and Yazdanbakhsh, Amir},
  journal={arXiv preprint arXiv:2209.07686},
  year={2022}
}

@inproceedings{
gudibande2023false,
title={The False Promise of Imitating Proprietary Language Models},
author={Arnav Gudibande and Eric Wallace and others},
booktitle={The Twelfth International Conference on Learning Representations},
year={2024},
url={https://openreview.net/forum?id=Kz3yckpCN5}
}

@article{cobbe2021training,
  title={Training verifiers to solve math word problems},
  author={Cobbe, Karl and Kosaraju, Vineet  and others},
  journal={arXiv preprint arXiv:2110.14168},
  year={2021}
}

@inproceedings{ling2017program,
    title = "Program Induction by Rationale Generation: Learning to Solve and Explain Algebraic Word Problems",
    author = "Ling, Wang  and
      Yogatama, Dani  and
      Dyer, Chris  and
      Blunsom, Phil",
    editor = "Barzilay, Regina  and
      Kan, Min-Yen",
    booktitle = "Proceedings of the 55th Annual Meeting of the Association for Computational Linguistics (Volume 1: Long Papers)",
    month = jul,
    year = "2017",
    address = "Vancouver, Canada",
    publisher = "Association for Computational Linguistics",
    url = "https://aclanthology.org/P17-1015/",
    doi = "10.18653/v1/P17-1015",
    pages = "158--167"
}

@inproceedings{wei2022chain,
author = {Wei, Jason and Wang, Xuezhi and Schuurmans, Dale and Bosma, Maarten and Ichter, Brian and Xia, Fei and Chi, Ed H. and Le, Quoc V. and Zhou, Denny},
title = {Chain-of-thought prompting elicits reasoning in large language models},
year = {2022},
isbn = {9781713871088},
publisher = {Curran Associates Inc.},
address = {Red Hook, NY, USA},
booktitle = {Proceedings of the 36th International Conference on Neural Information Processing Systems},
articleno = {1800},
numpages = {14},
location = {New Orleans, LA, USA},
url = "https://dl.acm.org/doi/10.5555/3600270.3602070",
series = {NIPS '22}
}

@inproceedings{patel-etal-2021-nlp,
    title = "Are {NLP} Models really able to Solve Simple Math Word Problems?",
    author = "Patel, Arkil  and
      Bhattamishra, Satwik  and
      Goyal, Navin",
    editor = "Toutanova, Kristina  and
      Rumshisky, Anna  and
      Zettlemoyer, Luke  and
      Hakkani-Tur, Dilek  and
      Beltagy, Iz  and
      Bethard, Steven  and
      Cotterell, Ryan  and
      Chakraborty, Tanmoy  and
      Zhou, Yichao",
    booktitle = "Proceedings of the 2021 Conference of the North American Chapter of the Association for Computational Linguistics: Human Language Technologies",
    month = jun,
    year = "2021",
    address = "Online",
    publisher = "Association for Computational Linguistics",
    url = "https://aclanthology.org/2021.naacl-main.168",
    doi = "10.18653/v1/2021.naacl-main.168",
    pages = "2080--2094",
}

@article{srivastava2023beyond,
  title={Beyond the Imitation Game: Quantifying and extrapolating the capabilities of language models},
  author={BIG-bench authors},
  journal={Transactions on Machine Learning Research},
  issn={2835-8856},
  year={2023},
  url={https://openreview.net/forum?id=uyTL5Bvosj},
  note={}
}

@misc{gemmateam2024gemma2improvingopen,
      title={Gemma 2: Improving Open Language Models at a Practical Size}, 
      author={Gemma Team},
      year={2024},
      eprint={2408.00118},
      archivePrefix={arXiv},
      primaryClass={cs.CL},
      url={https://arxiv.org/abs/2408.00118}, 
}

@inproceedings{press-etal-2023-measuring,
    title = "Measuring and Narrowing the Compositionality Gap in Language Models",
    author = "Press, Ofir  and
      Zhang, Muru  and
      others",
    editor = "Bouamor, Houda  and
      Pino, Juan  and
      Bali, Kalika",
    booktitle = "Findings of the Association for Computational Linguistics: EMNLP 2023",
    month = dec,
    year = "2023",
    address = "Singapore",
    publisher = "Association for Computational Linguistics",
    url = "https://aclanthology.org/2023.findings-emnlp.378",
    doi = "10.18653/v1/2023.findings-emnlp.378",
    pages = "5687--5711",
}

@article{geva2021strategyqa,
    title = "Did Aristotle Use a Laptop? A Question Answering Benchmark with Implicit Reasoning Strategies",
    author = "Geva, Mor  and
      Khashabi, Daniel  and
      Segal, Elad  and
      Khot, Tushar  and
      Roth, Dan  and
      Berant, Jonathan",
    editor = "Roark, Brian  and
      Nenkova, Ani",
    journal = "Transactions of the Association for Computational Linguistics",
    volume = "9",
    year = "2021",
    address = "Cambridge, MA",
    publisher = "MIT Press",
    url = "https://aclanthology.org/2021.tacl-1.21/",
    doi = "10.1162/tacl_a_00370",
    pages = "346--361"
}

@misc{grattafiori2024llama3herdmodels,
      title={The Llama 3 Herd of Models}, 
      author={Aaron Grattafiori and Abhimanyu Dubey etc.},
      year={2024},
      eprint={2407.21783},
      archivePrefix={arXiv},
      primaryClass={cs.AI},
      url={https://arxiv.org/abs/2407.21783}, 
}

@misc{llama3_2_connect_2024,
  author       = {Meta AI},
  title        = {LLaMA 3.2: Connect 2024 — Vision on the Edge for Mobile Devices},
  year         = {2024},
  url          = {https://ai.meta.com/blog/llama-3-2-connect-2024-vision-edge-mobile-devices/},
  note         = {Accessed: 2025-02-16}
}

@article{li2025llms,
  title={LLMs Can Easily Learn to Reason from Demonstrations Structure, not content, is what matters!},
  author={Li, Dacheng and Cao, Shiyi and Griggs, Tyler and Liu, Shu and Mo, Xiangxi and Patil, Shishir G and Zaharia, Matei and Gonzalez, Joseph E and Stoica, Ion},
  journal={arXiv preprint arXiv:2502.07374},
  year={2025}
}

@misc{nayab2025concisethoughtsimpactoutput,
      title={Concise Thoughts: Impact of Output Length on LLM Reasoning and Cost}, 
      author={Sania Nayab and Giulio Rossolini and others},
      year={2025},
      eprint={2407.19825},
      archivePrefix={arXiv},
      primaryClass={cs.CL},
      url={https://arxiv.org/abs/2407.19825}, 
}

@inproceedings{yi-etal-2024-badacts,
    title = "{B}ad{A}cts: A Universal Backdoor Defense in the Activation Space",
    author = "Yi, Biao  and
      Chen, Sishuo  and
     others",
    editor = "Ku, Lun-Wei  and
      Martins, Andre  and
      Srikumar, Vivek",
    booktitle = "Findings of the Association for Computational Linguistics: ACL 2024",
    month = aug,
    year = "2024",
    address = "Bangkok, Thailand",
    publisher = "Association for Computational Linguistics",
    url = "https://aclanthology.org/2024.findings-acl.317/",
    doi = "10.18653/v1/2024.findings-acl.317",
    pages = "5339--5352"
}

@inproceedings{chen-etal-2024-learnable,
    title = "Learnable Privacy Neurons Localization in Language Models",
    author = "Chen, Ruizhe  and
      Hu, Tianxiang  and
      Feng, Yang  and
      Liu, Zuozhu",
    editor = "Ku, Lun-Wei  and
      Martins, Andre  and
      Srikumar, Vivek",
    booktitle = "Proceedings of the 62nd Annual Meeting of the Association for Computational Linguistics (Volume 2: Short Papers)",
    month = aug,
    year = "2024",
    address = "Bangkok, Thailand",
    publisher = "Association for Computational Linguistics",
    url = "https://aclanthology.org/2024.acl-short.25/",
    doi = "10.18653/v1/2024.acl-short.25",
    pages = "256--264"
}

@inproceedings{10.5555/3692070.3693122,
author = {Lee, Andrew and Bai, Xiaoyan and Pres, Itamar and Wattenberg, Martin and Kummerfeld, Jonathan K. and Mihalcea, Rada},
title = {A mechanistic understanding of alignment algorithms: a case study on DPO and toxicity},
year = {2024},
publisher = {JMLR.org},
booktitle = {Proceedings of the 41st International Conference on Machine Learning},
articleno = {1052},
numpages = {18},
location = {Vienna, Austria},
url = "https://dl.acm.org/doi/10.5555/3692070.3693122",
series = {ICML'24}
}

@inproceedings{
dumas2024how,
title={How do Llamas process multilingual text? A latent exploration through activation patching},
author={Cl{\'e}ment Dumas and Veniamin Veselovsky and Giovanni Monea and Robert West and Chris Wendler},
booktitle={ICML 2024 Workshop on Mechanistic Interpretability},
year={2024},
url={https://openreview.net/forum?id=0ku2hIm4BS}
}

@inproceedings{
nanda2023progress,
title={Progress measures for grokking via mechanistic interpretability},
author={Neel Nanda and Lawrence Chan and Tom Lieberum and Jess Smith and Jacob Steinhardt},
booktitle={The Eleventh International Conference on Learning Representations },
year={2023},
url={https://openreview.net/forum?id=9XFSbDPmdW}
}

@article{
gurnee2024universal,
title={Universal Neurons in {GPT}2 Language Models},
author={Wes Gurnee and Theo Horsley and others},
journal={Transactions on Machine Learning Research},
issn={2835-8856},
year={2024},
url={https://openreview.net/forum?id=ZeI104QZ8I},
note={}
}

@misc{fierro2025multilinguallanguagemodelsremember,
      title={How Do Multilingual Language Models Remember Facts?},
      author={Constanza Fierro and Negar Foroutan and Desmond Elliott and Anders Søgaard},
      year={2025},
      eprint={2410.14387},
      archivePrefix={arXiv},
      primaryClass={cs.CL},
      url={https://arxiv.org/abs/2410.14387}, 
}

@inproceedings{wendler-etal-2024-llamas,
    title = "Do Llamas Work in {E}nglish? On the Latent Language of Multilingual Transformers",
    author = "Wendler, Chris  and
      Veselovsky, Veniamin  and
      Monea, Giovanni  and
      West, Robert",
    editor = "Ku, Lun-Wei  and
      Martins, Andre  and
      Srikumar, Vivek",
    booktitle = "Proceedings of the 62nd Annual Meeting of the Association for Computational Linguistics (Volume 1: Long Papers)",
    month = aug,
    year = "2024",
    address = "Bangkok, Thailand",
    publisher = "Association for Computational Linguistics",
    url = "https://aclanthology.org/2024.acl-long.820/",
    doi = "10.18653/v1/2024.acl-long.820",
    pages = "15366--15394",
}

@inproceedings{kojima2022large,
author = {Kojima, Takeshi and Gu, Shixiang Shane and Reid, Machel and Matsuo, Yutaka and Iwasawa, Yusuke},
title = {Large language models are zero-shot reasoners},
year = {2022},
isbn = {9781713871088},
publisher = {Curran Associates Inc.},
address = {Red Hook, NY, USA},
booktitle = {Proceedings of the 36th International Conference on Neural Information Processing Systems},
articleno = {1613},
numpages = {15},
location = {New Orleans, LA, USA},
url = "https://dl.acm.org/doi/10.5555/3600270.3601883",
series = {NIPS '22}
}

@inproceedings{geva-etal-2021-transformer,
    title = "Transformer Feed-Forward Layers Are Key-Value Memories",
    author = "Geva, Mor  and
      Schuster, Roei  and
      Berant, Jonathan  and
      Levy, Omer",
    editor = "Moens, Marie-Francine  and
      Huang, Xuanjing  and
      Specia, Lucia  and
      Yih, Scott Wen-tau",
    booktitle = "Proceedings of the 2021 Conference on Empirical Methods in Natural Language Processing",
    month = nov,
    year = "2021",
    address = "Online and Punta Cana, Dominican Republic",
    publisher = "Association for Computational Linguistics",
    url = "https://aclanthology.org/2021.emnlp-main.446/",
    doi = "10.18653/v1/2021.emnlp-main.446",
    pages = "5484--5495",
}

@inproceedings{rai-yao-2024-investigation,
    title = "An Investigation of Neuron Activation as a Unified Lens to Explain Chain-of-Thought Eliciting Arithmetic Reasoning of {LLM}s",
    author = "Rai, Daking  and
      Yao, Ziyu",
    editor = "Ku, Lun-Wei  and
      Martins, Andre  and
      Srikumar, Vivek",
    booktitle = "Proceedings of the 62nd Annual Meeting of the Association for Computational Linguistics (Volume 1: Long Papers)",
    month = aug,
    year = "2024",
    address = "Bangkok, Thailand",
    publisher = "Association for Computational Linguistics",
    url = "https://aclanthology.org/2024.acl-long.387/",
    doi = "10.18653/v1/2024.acl-long.387",
    pages = "7174--7193",
}

@inproceedings{voita-etal-2024-neurons,
    title = "Neurons in Large Language Models: Dead, N-gram, Positional",
    author = "Voita, Elena  and
      Ferrando, Javier  and
      Nalmpantis, Christoforos",
    editor = "Ku, Lun-Wei  and
      Martins, Andre  and
      Srikumar, Vivek",
    booktitle = "Findings of the Association for Computational Linguistics: ACL 2024",
    month = aug,
    year = "2024",
    address = "Bangkok, Thailand",
    publisher = "Association for Computational Linguistics",
    url = "https://aclanthology.org/2024.findings-acl.75/",
    doi = "10.18653/v1/2024.findings-acl.75",
    pages = "1288--1301",
}

@inproceedings{webson-pavlick-2022-prompt,
    title = "Do Prompt-Based Models Really Understand the Meaning of Their Prompts?",
    author = "Webson, Albert  and
      Pavlick, Ellie",
    editor = "Carpuat, Marine  and
      de Marneffe, Marie-Catherine  and
      Meza Ruiz, Ivan Vladimir",
    booktitle = "Proceedings of the 2022 Conference of the North American Chapter of the Association for Computational Linguistics: Human Language Technologies",
    month = jul,
    year = "2022",
    address = "Seattle, United States",
    publisher = "Association for Computational Linguistics",
    url = "https://aclanthology.org/2022.naacl-main.167/",
    doi = "10.18653/v1/2022.naacl-main.167",
    pages = "2300--2344"
}

@inproceedings{cancedda-2024-spectral,
    title = "Spectral Filters, Dark Signals, and Attention Sinks",
    author = "Cancedda, Nicola",
    editor = "Ku, Lun-Wei  and
      Martins, Andre  and
      Srikumar, Vivek",
    booktitle = "Proceedings of the 62nd Annual Meeting of the Association for Computational Linguistics (Volume 1: Long Papers)",
    month = aug,
    year = "2024",
    address = "Bangkok, Thailand",
    publisher = "Association for Computational Linguistics",
    url = "https://aclanthology.org/2024.acl-long.263/",
    doi = "10.18653/v1/2024.acl-long.263",
    pages = "4792--4808",
}

@article{li2024rome,
  title={ROME: Memorization Insights from Text, Logits and Representation},
  author={Li, Bo and Zhao, Qinghua and Wen, Lijie},
  journal={arXiv preprint arXiv:2403.00510},
  year={2024}
}

@misc{yeo2024faithfulnaturallanguageexplanations,
      title={Towards Faithful Natural Language Explanations: A Study Using Activation Patching in Large Language Models}, 
      author={Wei Jie Yeo and Ranjan Satapathy and Erik Cambria},
      year={2024},
      eprint={2410.14155},
      archivePrefix={arXiv},
      primaryClass={cs.CL},
      url={https://arxiv.org/abs/2410.14155}, 
}

@misc{dutta2024thinkstepbystepmechanisticunderstanding,
      title={How to think step-by-step: A mechanistic understanding of chain-of-thought reasoning}, 
      author={Subhabrata Dutta and Joykirat Singh and Soumen Chakrabarti and Tanmoy Chakraborty},
      year={2024},
      eprint={2402.18312},
      archivePrefix={arXiv},
      primaryClass={cs.CL},
      url={https://arxiv.org/abs/2402.18312}, 
}

@article{kgqgipm24,
  author       = {Runhao Zhao and
                  Jiuyang Tang and
                  Weixin Zeng and
                  Yunxiao Guo and
                  Xiang Zhao},
  title        = {Towards human-like questioning: Knowledge base question generation
                  with bias-corrected reinforcement learning from human feedback},
  journal      = {Inf. Process. Manag.},
  volume       = {62},
  number       = {3},
  pages        = {104044},
  year         = {2025},
  url          = {https://doi.org/10.1016/j.ipm.2024.104044},
  doi          = {10.1016/J.IPM.2024.104044},
  bibsource    = {dblp computer science bibliography, https://dblp.org}
}

@inproceedings{agenteaicde25,
  author       = {Runhao Zhao and
                  Weixin Zeng and
                  Jiuyang Tang and
                  Yawen Li and
                  Guanhua Ye and
                  Junping Du and
                  Xiang Zhao},
  title        = {Towards Unsupervised Entity Alignment for Highly Heterogeneous Knowledge
                  Graphs},
  booktitle    = {41st {IEEE} International Conference on Data Engineering, {ICDE} 2025,
                  Hong Kong, May 19-23, 2025},
  pages        = {3792--3806},
  publisher    = {{IEEE}},
  year         = {2025},
  url          = {https://doi.org/10.1109/ICDE65448.2025.00283},
  doi          = {10.1109/ICDE65448.2025.00283},
  bibsource    = {dblp computer science bibliography, https://dblp.org}
}

@inproceedings{agentqgcikm24,
  author       = {Runhao Zhao and
                  Jiuyang Tang and
                  Weixin Zeng and
                  Ziyang Chen and
                  Xiang Zhao},
  editor       = {Edoardo Serra and
                  Francesca Spezzano},
  title        = {Zero-shot Knowledge Graph Question Generation via Multi-agent LLMs
                  and Small Models Synthesis},
  booktitle    = {Proceedings of the 33rd {ACM} International Conference on Information
                  and Knowledge Management, {CIKM} 2024, Boise, ID, USA, October 21-25,
                  2024},
  pages        = {3341--3351},
  publisher    = {{ACM}},
  year         = {2024},
  url          = {https://doi.org/10.1145/3627673.3679805},
  doi          = {10.1145/3627673.3679805},
  bibsource    = {dblp computer science bibliography, https://dblp.org}
}

@article{hydra26,
  author       = {Runhao Zhao and
                  Weixin Zeng and
                  Wentao Zhang and
                  Xiang Zhao and
                  Jiuyang Tang and
                  Lei Chen},
  title        = {Towards Temporal Knowledge Graph Alignment in the Wild},
  journal      = {CoRR},
  volume       = {abs/2507.14475},
  year         = {2025},
  url          = {https://doi.org/10.48550/arXiv.2507.14475},
  doi          = {10.48550/ARXIV.2507.14475},
  eprinttype   = {arXiv},
  eprint       = {2507.14475},
  bibsource    = {dblp computer science bibliography, https://dblp.org}
}

\appendix
\onecolumn

\clearpage

\section{Examples of Tested Datasets} \label{sec:dataset_examples}

\begin{table*}[!ht]
\scriptsize
\setlength{\tabcolsep}{2pt} 
\renewcommand{\arraystretch}{0.9} 
\begin{longtable}{>{\raggedright\arraybackslash}p{1.3cm}>{\raggedright\arraybackslash}p{2.1cm}>{\raggedright\arraybackslash}p{12cm}}
\toprule
\textbf{Task} & \textbf{Dataset (Answer Space)} & \textbf{Example} \\
\midrule

\rowcolor{C1}
& GSM8K (Open, Numerical) & 
\textbf{Question}: If there are 3 cars in the parking lot and 2 more cars arrive, how many cars are in the parking? \newline
\textbf{Answer}: There are originally 3 cars. 2 more cars arrive. \textcolor{darkred}{$3 + 2 = 5$}. So the answer is \textcolor{darkred}{5}. \\
\cmidrule(lr){2-3}

\rowcolor{C1}
\textbf{Arithmetic}  & SVAMP (Open, Numerical) & 
\textbf{Question}: Jason had 20 lollipops. He gave Denny some lollipops. Now Jason has 12 lollipops. How many lollipops did Jason give to Denny? \newline
\textbf{Answer}: Jason started with 20 lollipops. Then he had 12 after giving some to Denny. So he gave Denny \textcolor{darkred}{$20 - 12 = 8$}. So the answer is \textcolor{darkred}{8}. \\
\cmidrule(lr){2-3}

\rowcolor{C1}
\textbf{Arithmetic} & AQuA (Closed, Options) & 
\textbf{Question}: A person is traveling at 20 km/hr and reached his destiny in 2.5 hr then find the distance? \newline
\textbf{Choices}: (a) 53 km (b) 55 km (c) 52 km (d) 60 km (e) 50 km \newline
\textbf{Answer}: The distance that the person traveled would have been $20 \, \text{km/hr} \times 2.5 \, \text{hrs} = 50 \, \text{km}$. So the answer is \textcolor{darkred}{(e)}. \\

\midrule

\rowcolor{C2}
& Bamboogle (Open, Text) & 
\textbf{Question}: Who lived longer, Theodor Haecker or Harry Vaughan Watkins? \newline
\textbf{Answer}: Theodor Haecker was 65 years old when he died. Harry Vaughan Watkins was 69 years old when he died. So the answer is \textcolor{darkred}{Harry Vaughan Watkins}. \\
\cmidrule(lr){2-3}

\rowcolor{C2}
\textbf{Common sense}& StrategyQA  (Binary, Yes/No) & 
\textbf{Question}: Do hamsters provide food for any animals? \newline
\textbf{Answer}: Hamsters are prey animals. Prey are food for predators. Thus, hamsters provide food for some animals. So the answer is \textcolor{darkred}{Yes}. \\

\rowcolor{C2}
& Sports (Binary, Yes/No) & 
\textbf{Question}: Is the following sentence plausible? ``Kyle Palmieri was called for slashing'' \newline
\textbf{Answer}: Kyle Palmieri is a hockey player. Being called for slashing is part of hockey. So the answer is \textcolor{darkred}{Yes}. \\
\cmidrule(lr){2-3}

\rowcolor{C2}
 & Date (Formatted) & 
\textbf{Question}: The concert was scheduled to be on 06/01/1943, but was delayed by one day to today. What is the date 10 days ago in MM/DD/YYYY? \newline
\textbf{Answer}: One day after 06/01/1943 is 06/02/1943, so today is 06/02/1943. 10 days before today is 05/23/1943. So the answer is \textcolor{darkred}{05/23/1943}. \\
\midrule

\rowcolor{C3}
& Coin Flip (Binary, Yes/No) & 
\textbf{Question}: A coin is heads up. Maybelle flips the coin. Shalonda does not flip the coin. Is the coin still heads up? \newline
\textbf{Answer}: The coin was flipped by Maybelle. So the coin was flipped 1 time, which is an odd number. The coin started heads up, so after an odd number of flips, it will be tails up. So the answer is \textcolor{darkred}{No}. \\
\cmidrule(lr){2-3}

\rowcolor{C3}
\textbf{Symbolic}& Last Letters Concatenation (Open, Text) & 
\textbf{Question}: Take the last letters of each words in ``Ti\underline{m} Candac\underline{e} Ceci\underline{l} Misae\underline{l}'' and concatenate them.  \newline
\textbf{Answer}: The last letter of ``Tim'' is ``\textcolor{darkred}{m}''. The last letter of ``Candace'' is ``\textcolor{darkred}{e}''. The last letter of ``Cecil'' is ``\textcolor{darkred}{l}''. The last letter of ``Misael'' is ``\textcolor{darkred}{l}''. Concatenating them is ``\textcolor{darkred}{mell}''. So the answer is \textcolor{darkred}{mell}. \\

\bottomrule
\end{longtable}
\caption{Full Dataset Examples} \label{tab:dataset_examples} 
\end{table*}

\section{Test Points}
\begin{table*}[!ht]
\begin{longtable}{@{}p{2cm}p{12cm}@{}}
\toprule
\textbf{Test points} & \textbf{Contained words} \\
\midrule
\endfirsthead
\multicolumn{2}{l}{\textit{Continued from previous page}}\\
\toprule
\textbf{Test points} & \textbf{Contained words} \\
\midrule
\endhead
\midrule
\multicolumn{2}{r}{\textit{Continued on next page}}\\
\endfoot
\bottomrule
\endlastfoot
time & originally, then, after, so, start, first, next, last, finally, 
      before, later, afterwards, subsequently, meanwhile, during, while, 
      when, once, as, since, because, due, hence, therefore, thus, 
      consequently, accordingly, result, resulting, resulted, initially,
      earlier, until, at the same time \\
\midrule
action & +, -, *, /, =, >, <, add, subtract, multiply, divide, average, 
        increase, decrease, equal, calculate, total, square, root, cube, prime \\
\midrule
loc\&peo & there, location, site, area, spot, venue, someone, somebody, anyone, 
          nobody, everyone, person, individual, participant, operator, handler, 
          's, his, her, their, its, he, she, they, it \\
\midrule
number & ``it was obtained by regular expression''  \\
\bottomrule
\end{longtable}
\caption{Test points and their associated words.}\label{tab:test_points}
\end{table*}

\clearpage
\section{Results on Tested Datasets}\label{appendix:results_on_tests}
\begin{table*}[htbp]
\centering
\caption{LLaMA3.1-8B}
\label{tab:performance_comparison}
\begin{tabular}{l|ccc|ccc}
\toprule
& \multicolumn{3}{c|}{Closed-domain} & \multicolumn{3}{c}{Open-domain} \\
\cmidrule(lr){2-4} \cmidrule(lr){5-7}
Metric & \multicolumn{1}{c}{AQuA} & \multicolumn{1}{c}{Sports} & \multicolumn{1}{c|}{Coin Flip} & \multicolumn{1}{c}{GSM8K} & \multicolumn{1}{c}{Date} & \multicolumn{1}{c}{Last Letter Concat} \\
\midrule
Standard Acc & 0.3110 & 0.7497 & 0.4580 & 0.1774 & 0.4417 & 0.0 \\
CoT Acc      & 0.4961 & 0.9395 & 1.000 & 0.7771 & 0.7100 & 0.4496 \\
Relative Improvement     & +59.49\%  & +25.30\%  & +118.34\%  & +338.03\%  & +67.4\%  & $+\infty$  \\
\bottomrule
\end{tabular}
\end{table*}

\clearpage
\section{Resutls of Keyword Imitation}\label{appendix:keyword_imitation}
\vspace{-3mm}
\begin{figure*}[!h]
\centering
\captionsetup[sub]{skip=-1mm}
    \begin{subfigure}{0.3\linewidth}
        \includegraphics[width=\linewidth]{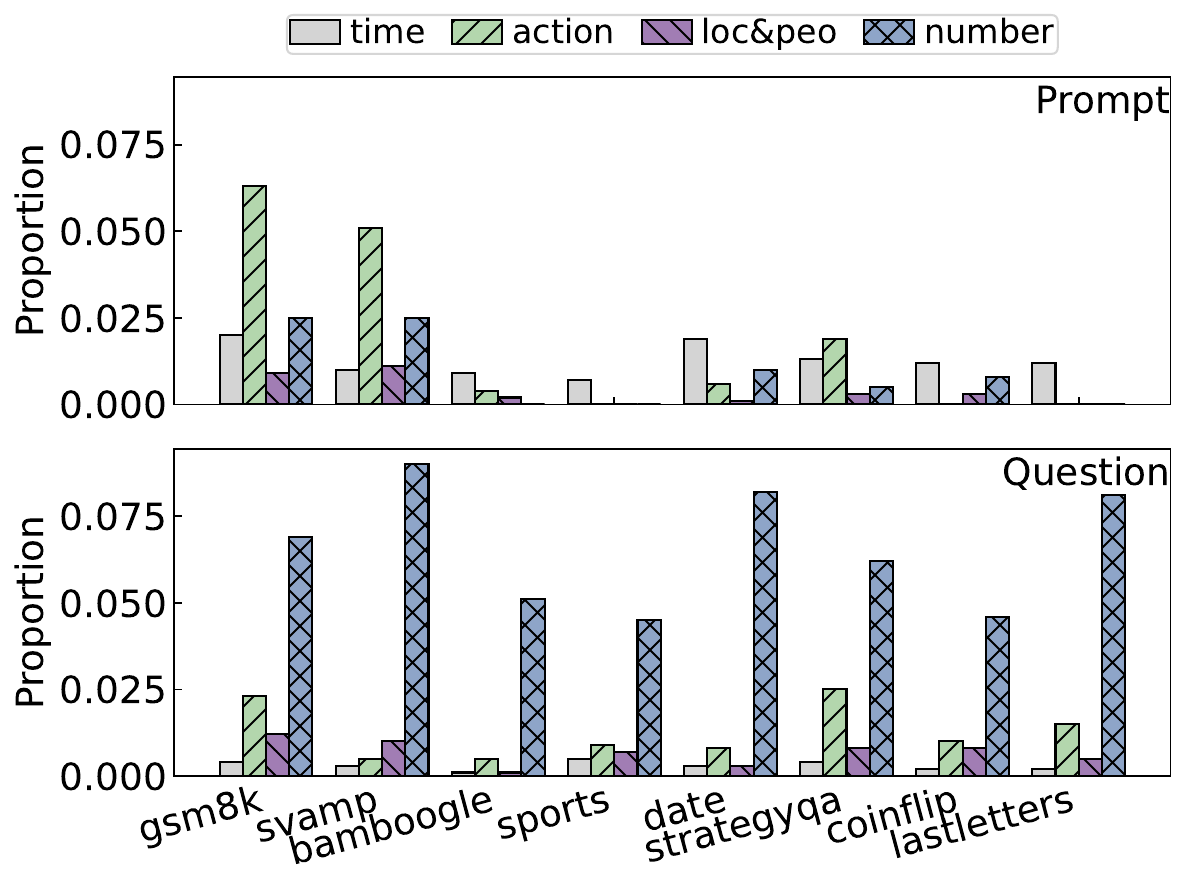}
        \caption{\textit{AQuA}}
        \label{fig:question1_a2_a}
    \end{subfigure}
    \begin{subfigure}{0.3\linewidth}
        \includegraphics[width=\linewidth]{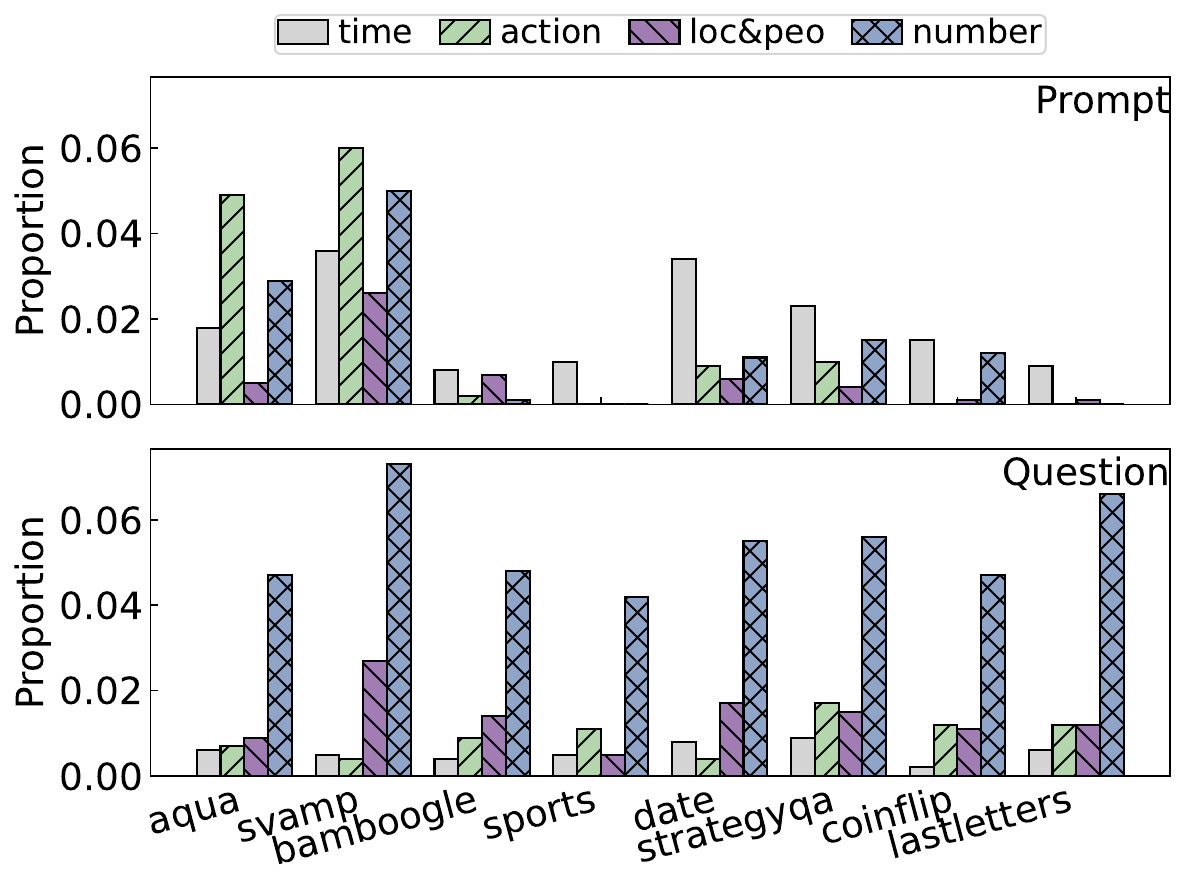}
        \caption{\textit{GSM8K}}
        \label{fig:question1_a2_b}
    \end{subfigure}
    \begin{subfigure}{0.3\linewidth}
        \includegraphics[width=\linewidth]{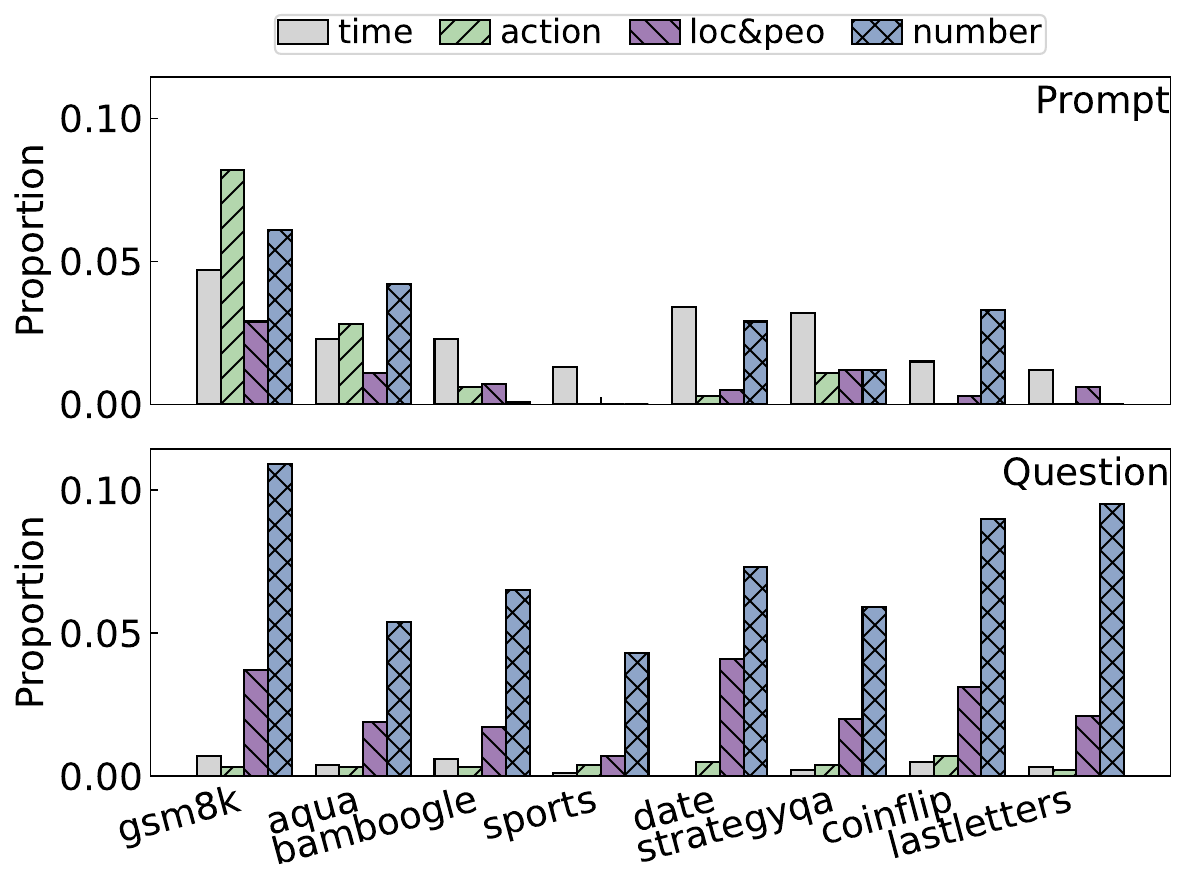}
        \caption{\textit{SVAMP}}
        \label{fig:question1_a2_c}
    \end{subfigure}

  \begin{subfigure}{0.3\linewidth}
        \includegraphics[width=\linewidth]{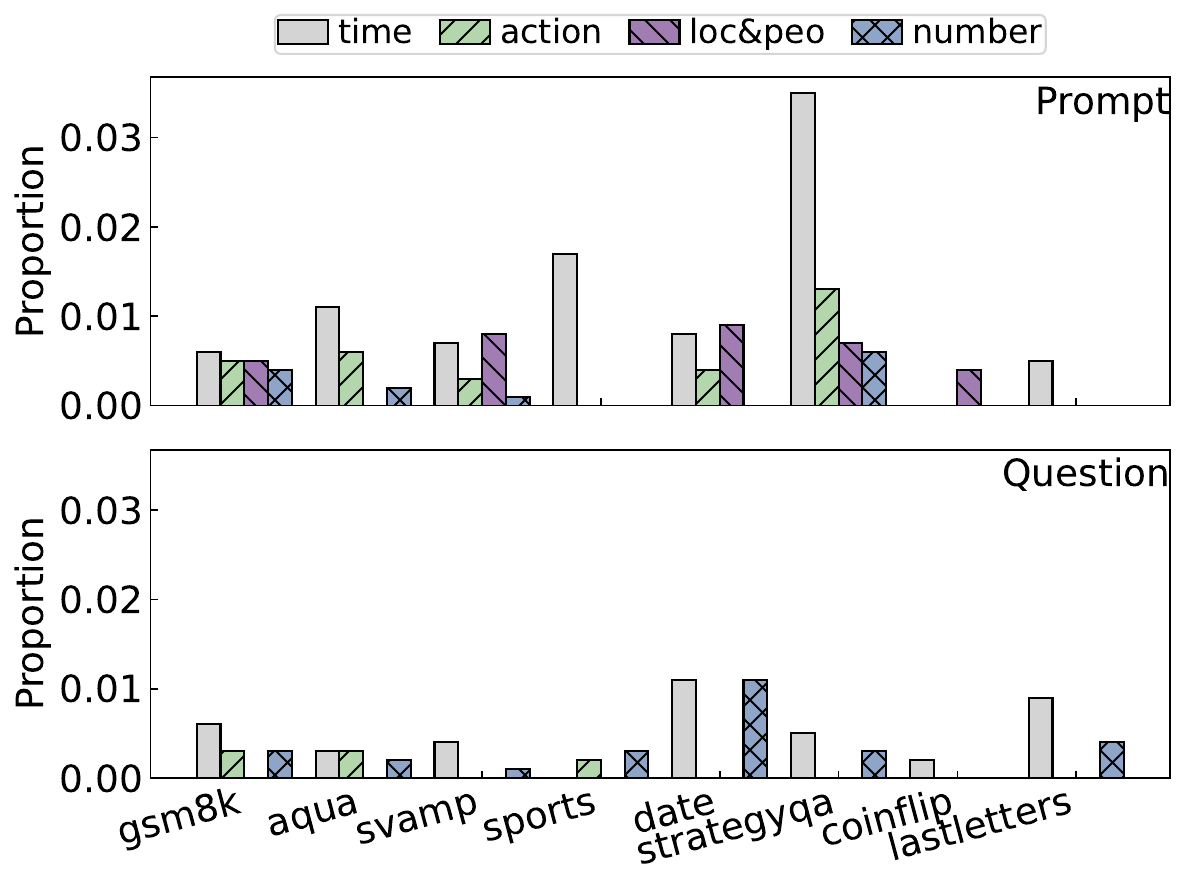}
        \caption{\textit{Bamboogle}}
        \label{fig:question1_a2_e}
    \end{subfigure}
  \begin{subfigure}{0.3\linewidth}
        \includegraphics[width=\linewidth]{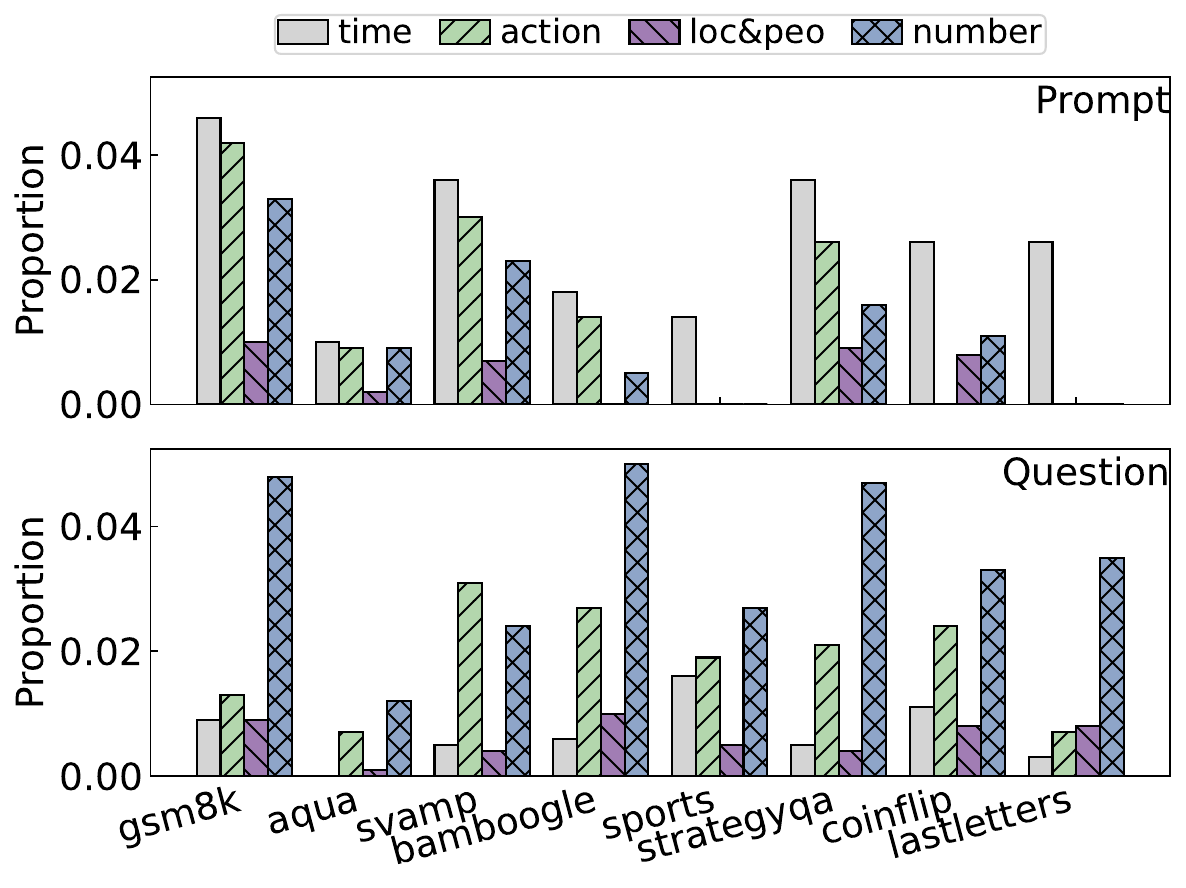}
        \caption{\textit{Date}}
        \label{fig:question1_a2_d}
    \end{subfigure}
    \begin{subfigure}{0.3\linewidth}
        \includegraphics[width=\linewidth]{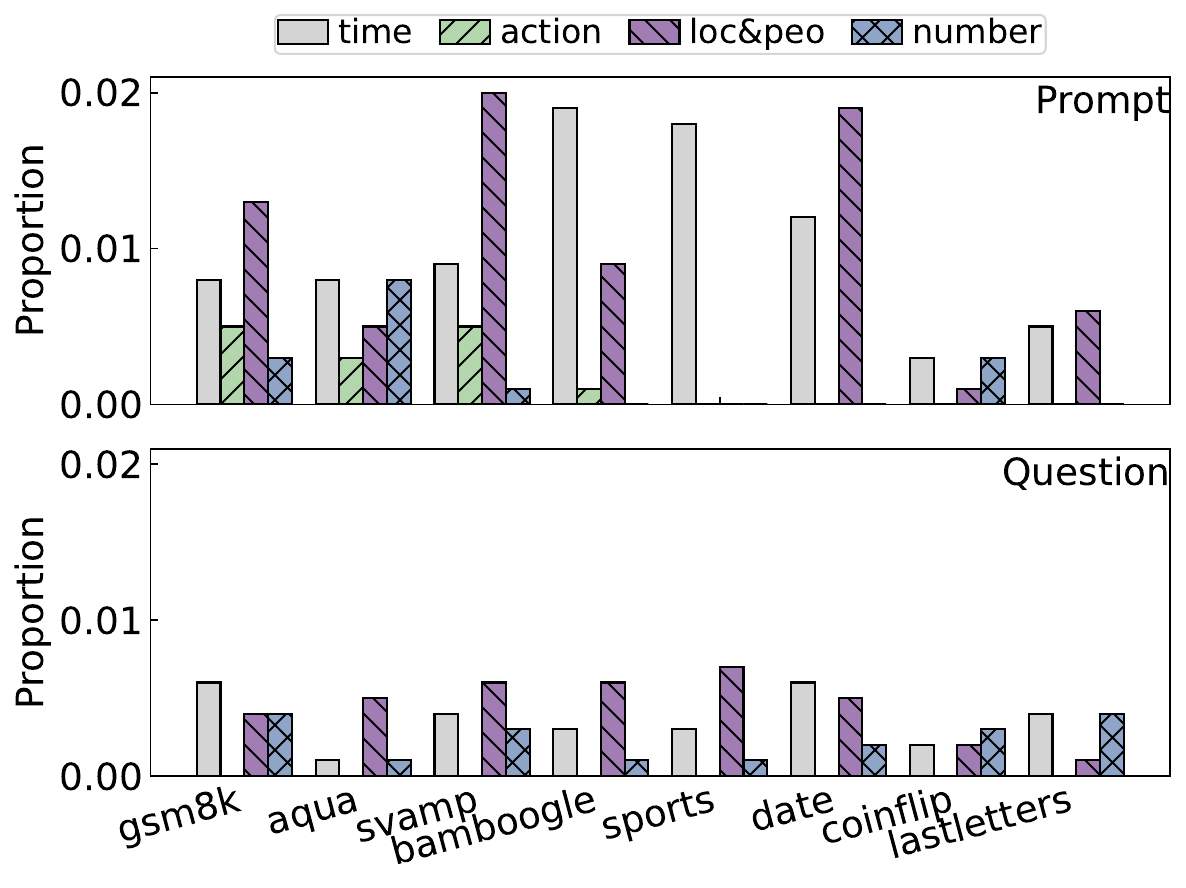}
        \caption{\textit{StrategyQA}}
    \end{subfigure}
    
     \begin{subfigure}{0.3\linewidth}
        \includegraphics[width=\linewidth]{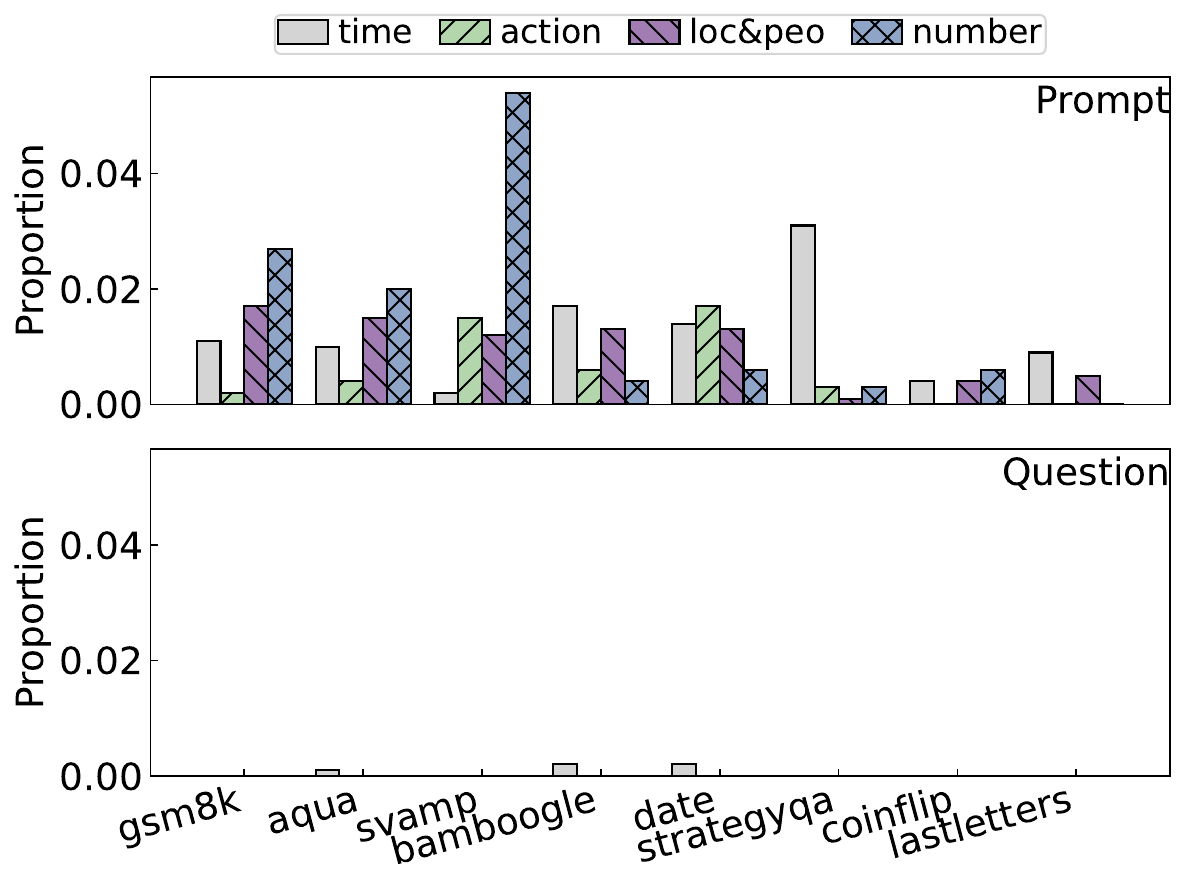}
        \caption{\textit{Sports}}
    \end{subfigure}
    \begin{subfigure}{0.3\linewidth}
        \includegraphics[width=\linewidth]{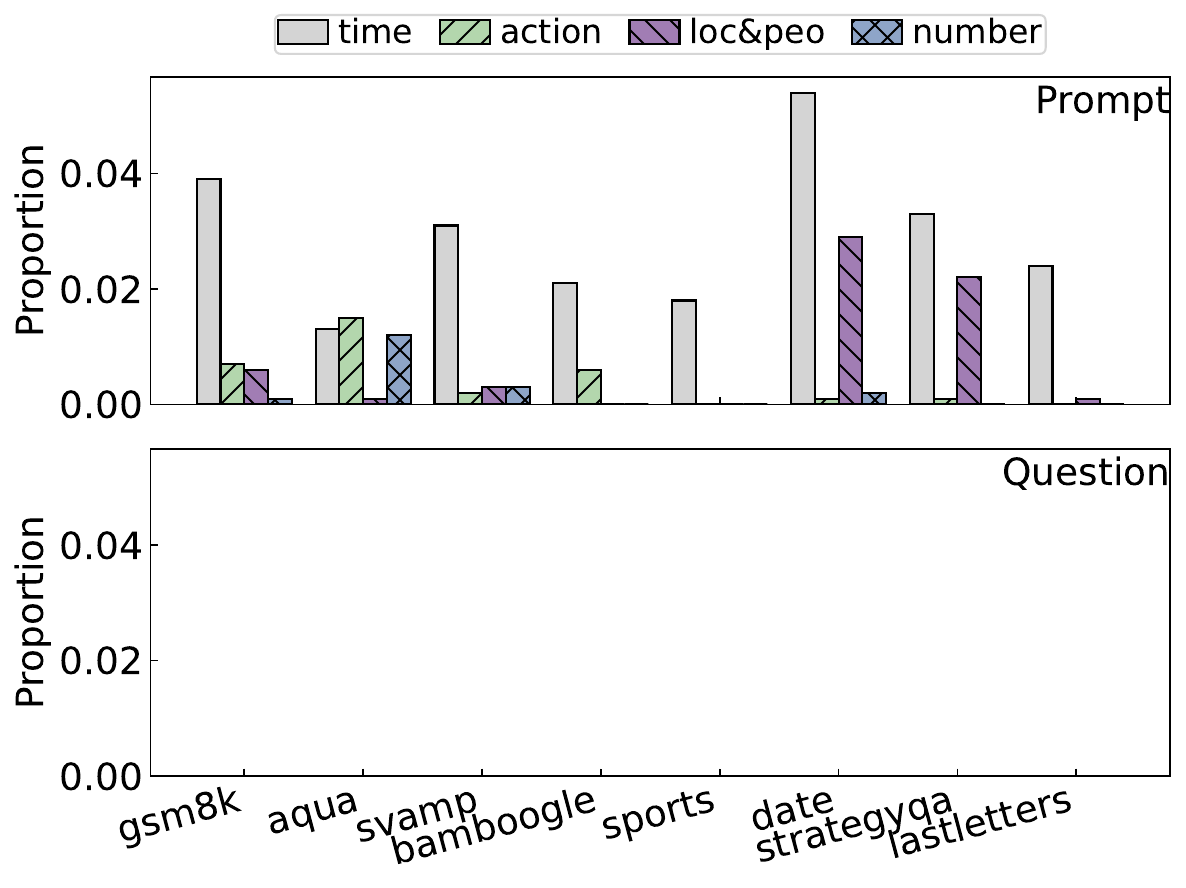}
        \caption{\textit{Coin Flip}}
    \end{subfigure}
    \begin{subfigure}{0.3\linewidth}
        \includegraphics[width=\linewidth]{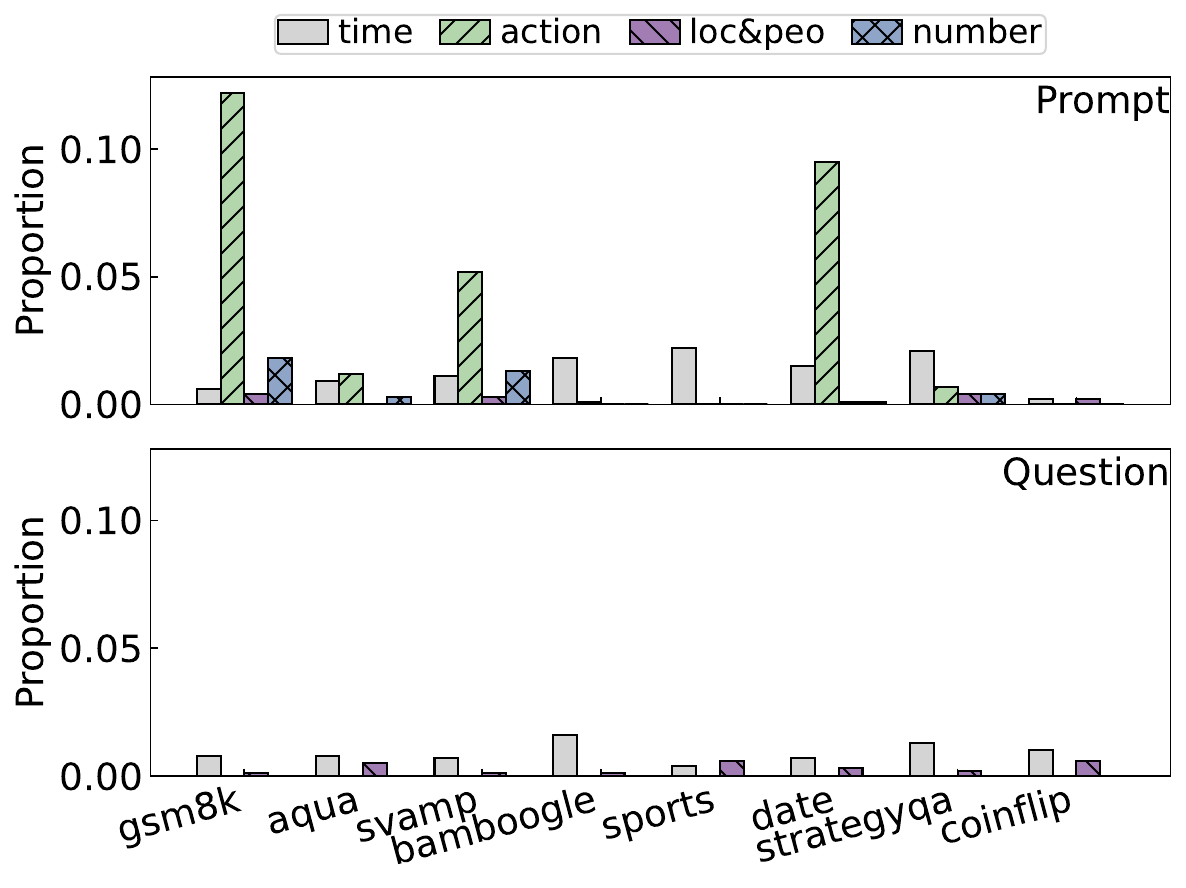}
        \caption{\textit{Last Letter}}
    \end{subfigure}
\vspace{-3mm}
\caption{Results of Gemma2-2b.}
\label{fig:transfer_test_2b}
\end{figure*}
\vspace{-3mm}
\begin{figure*}[!h]
\centering
\captionsetup[sub]{skip=-1mm}
    \begin{subfigure}{0.3\linewidth}
        \includegraphics[width=\linewidth]{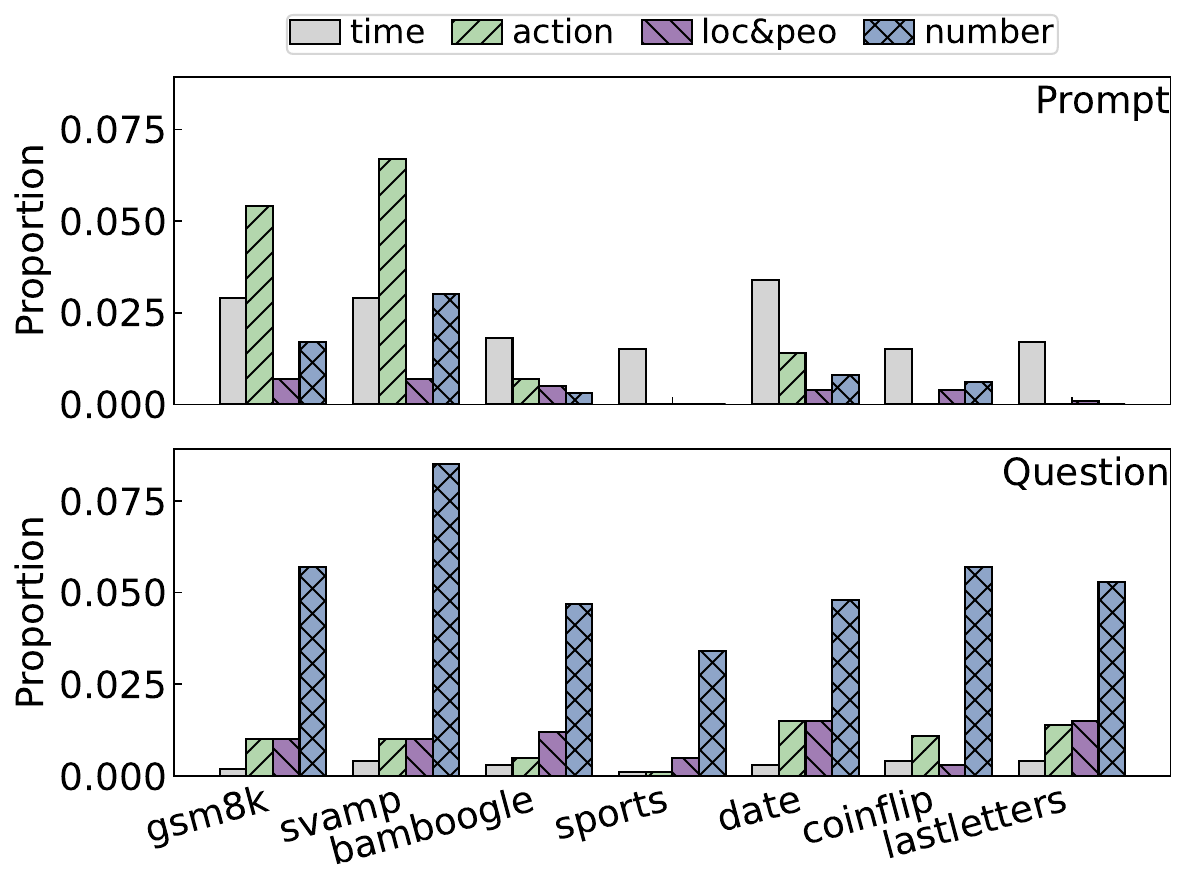}
        \caption{\textit{AQuA}}
        \label{fig:question1_a2_a}
    \end{subfigure}
    \begin{subfigure}{0.3\linewidth}
        \includegraphics[width=\linewidth]{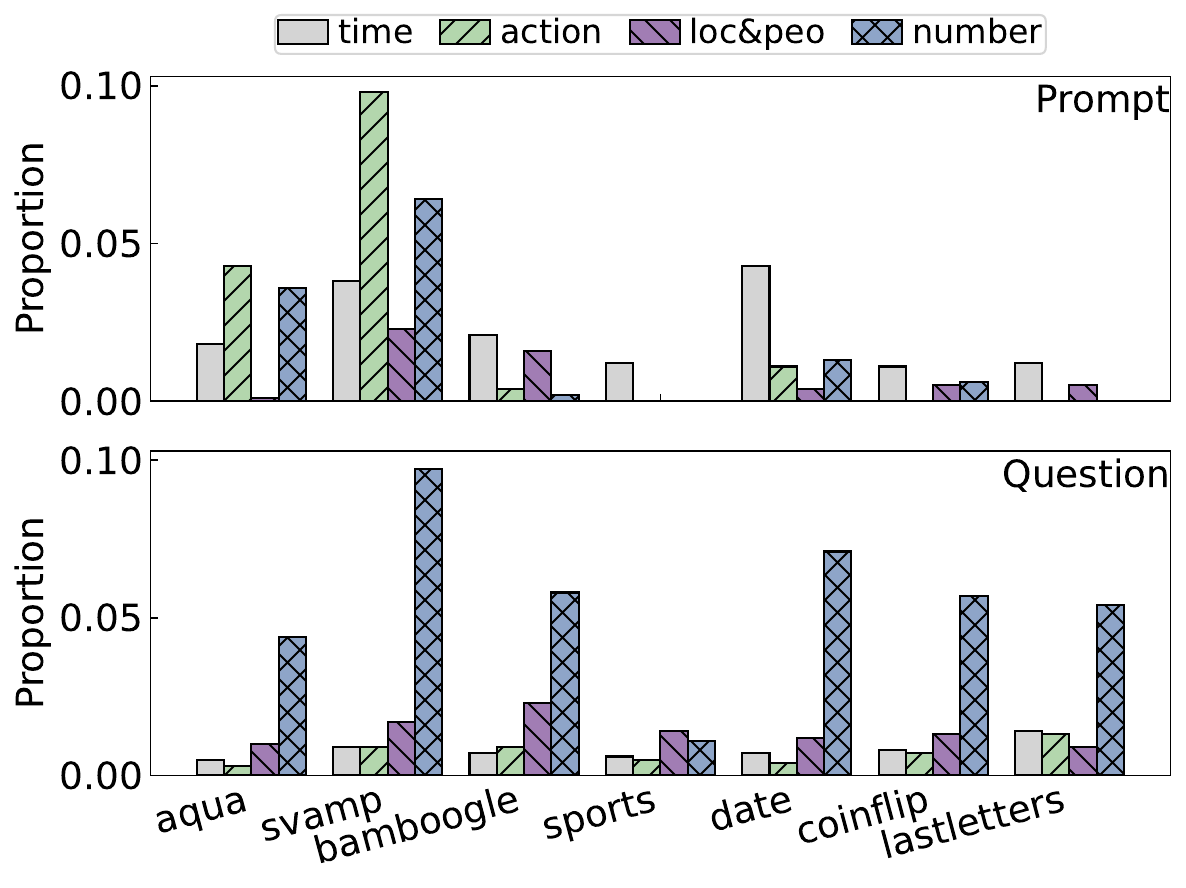}
        \caption{\textit{GSM8K}}
        \label{fig:question1_a2_b}
    \end{subfigure}
    \begin{subfigure}{0.3\linewidth}
        \includegraphics[width=\linewidth]{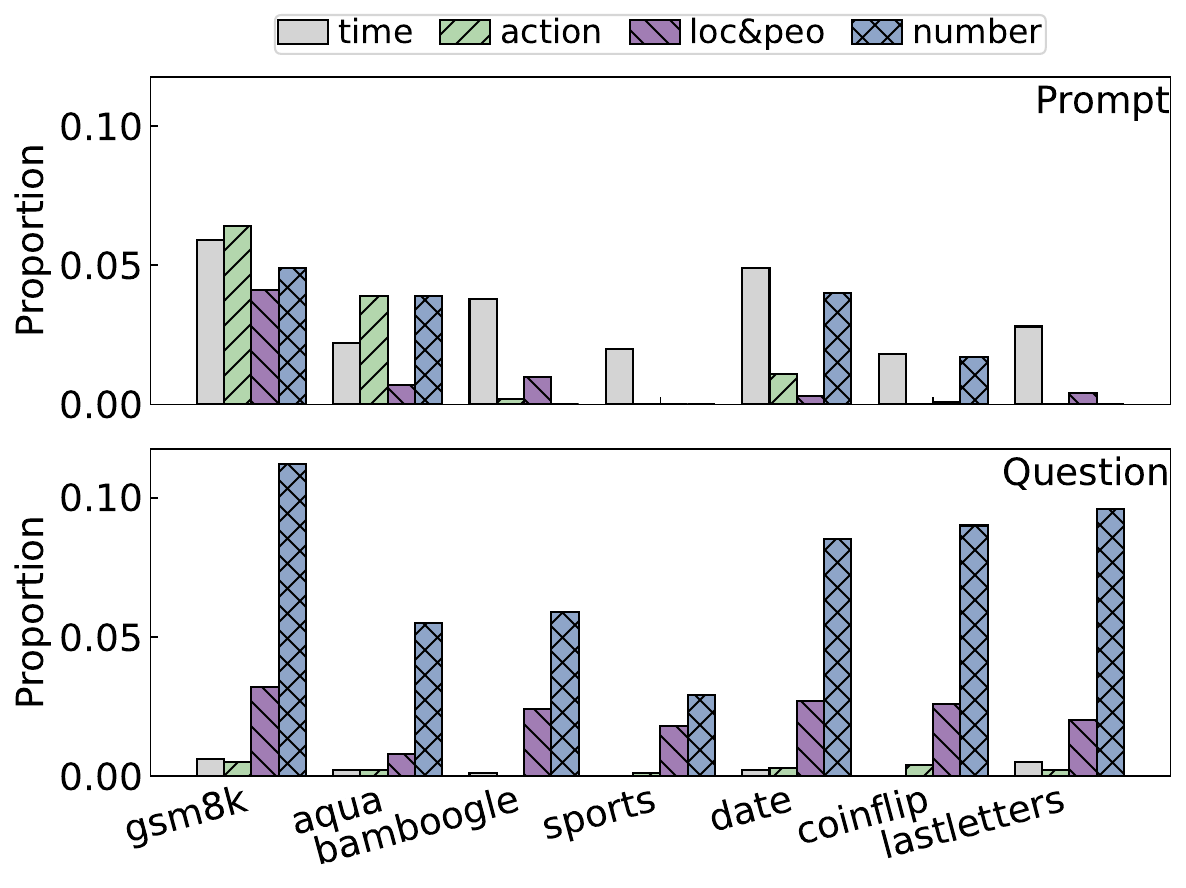}
        \caption{\textit{SVAMP}}
        \label{fig:question1_a2_c}
    \end{subfigure}

  \begin{subfigure}{0.3\linewidth}
        \includegraphics[width=\linewidth]{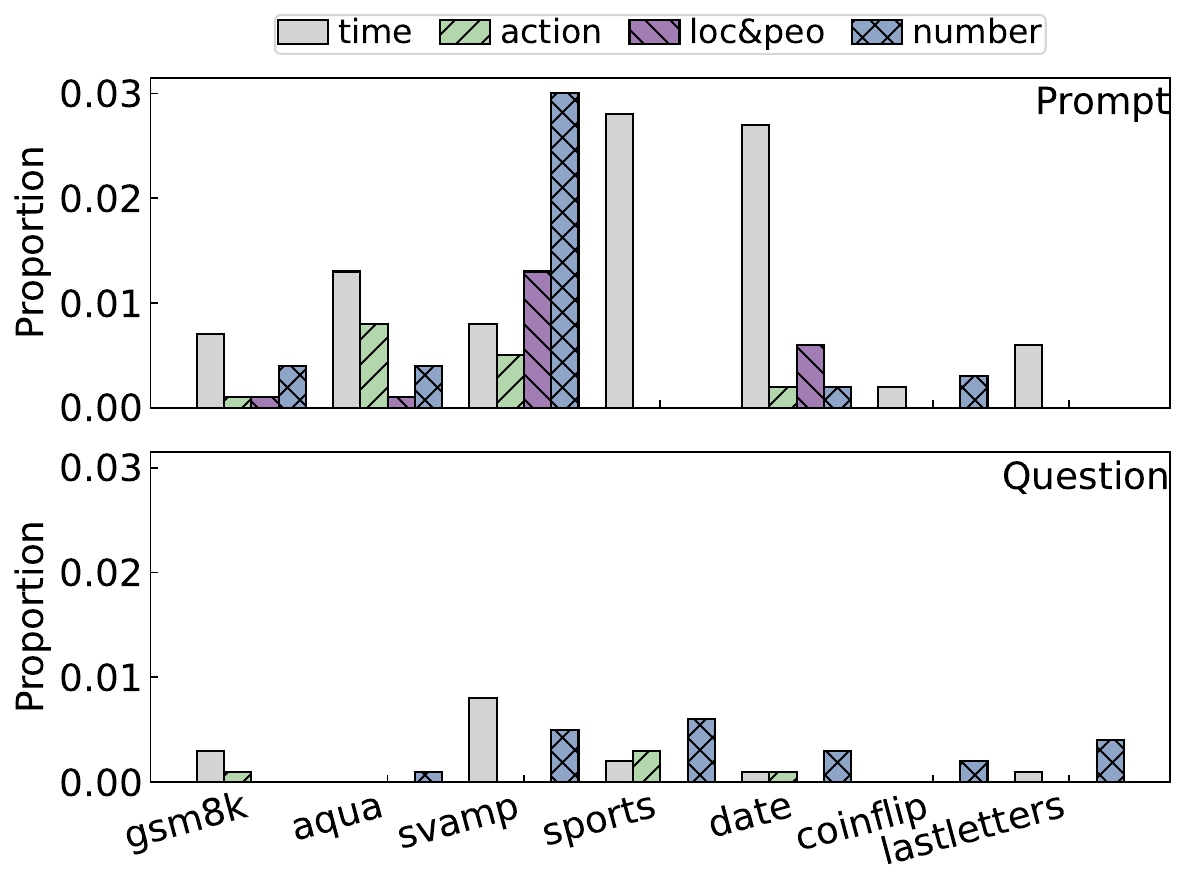}
        \caption{\textit{Bamboogle}}
        \label{fig:question1_a2_e}
    \end{subfigure}
  \begin{subfigure}{0.3\linewidth}
        \includegraphics[width=\linewidth]{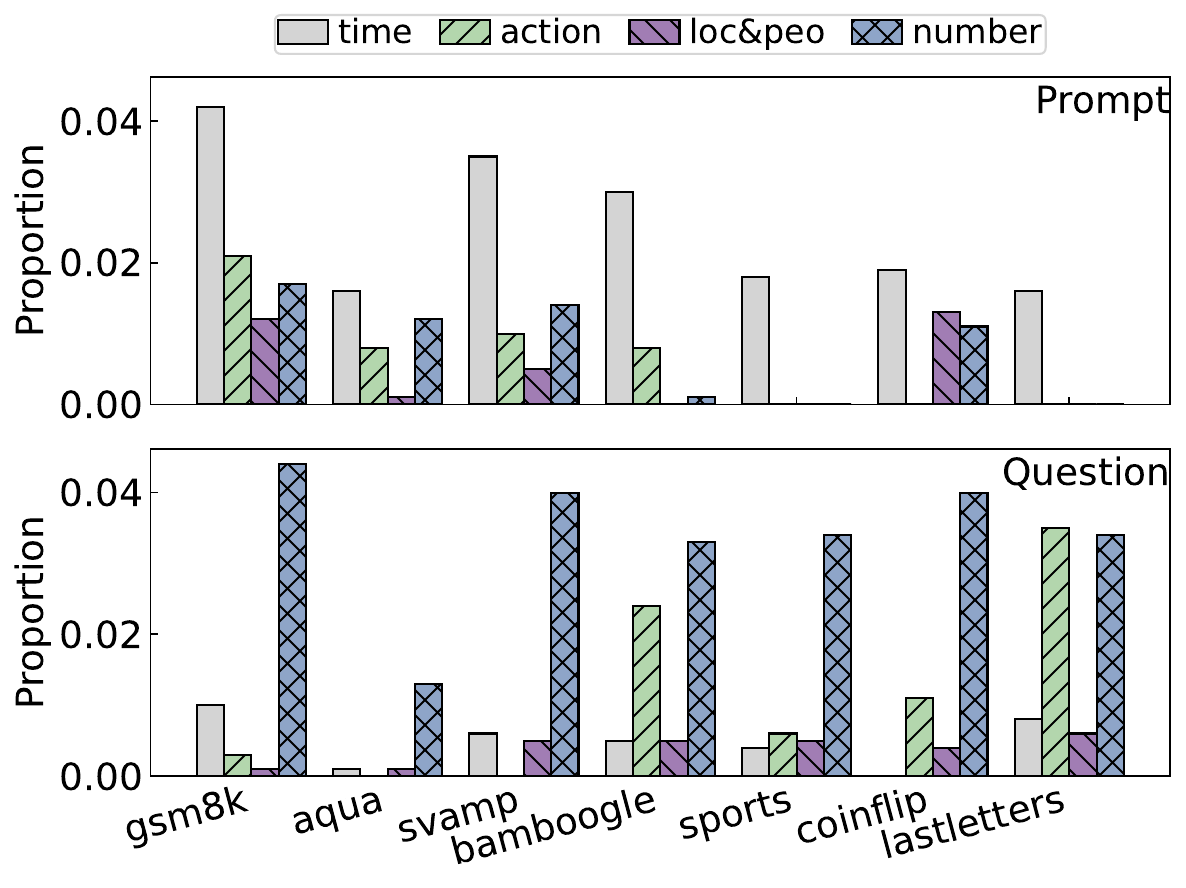}
        \caption{\textit{Date}}
        \label{fig:question1_a2_d}
    \end{subfigure}
    \begin{subfigure}{0.3\linewidth}
        \includegraphics[width=\linewidth]{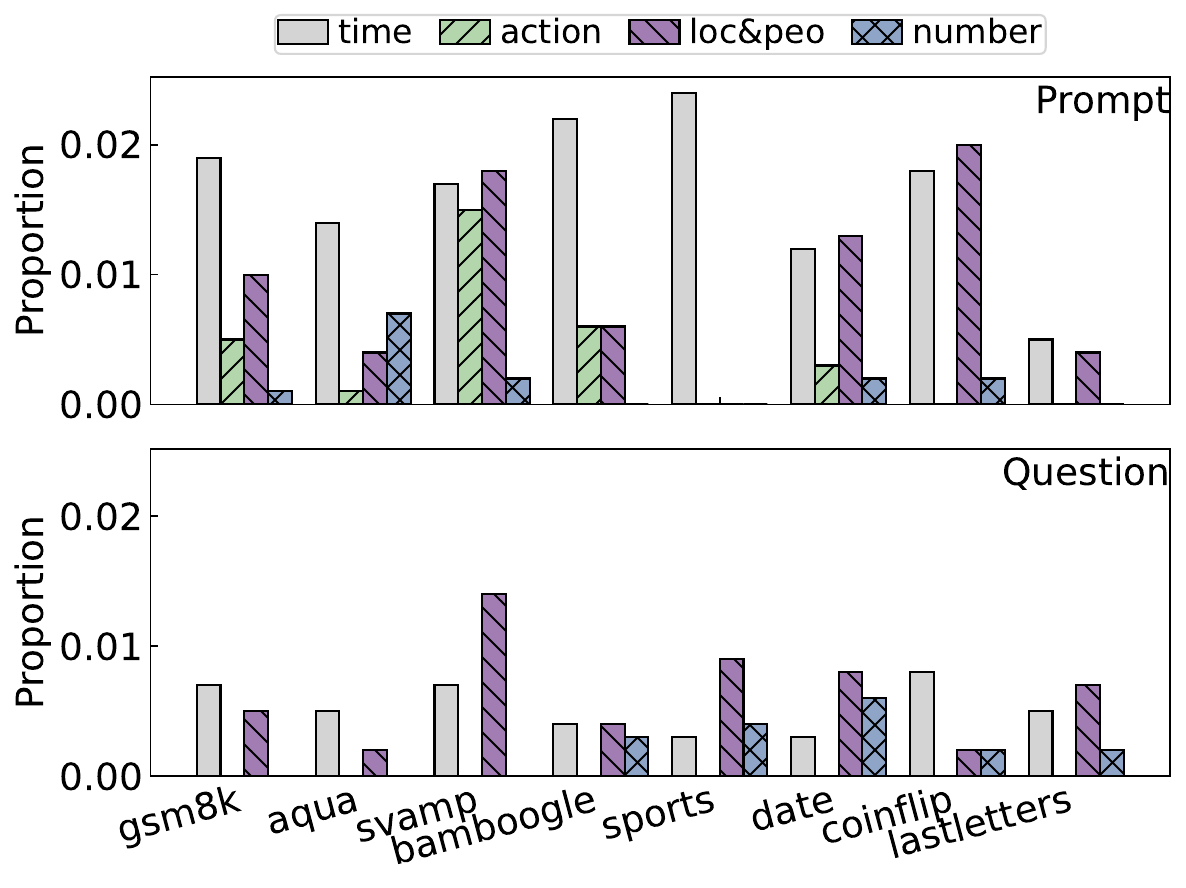}
        \caption{\textit{StrategyQA}}
    \end{subfigure}
    
     \begin{subfigure}{0.3\linewidth}
        \includegraphics[width=\linewidth]{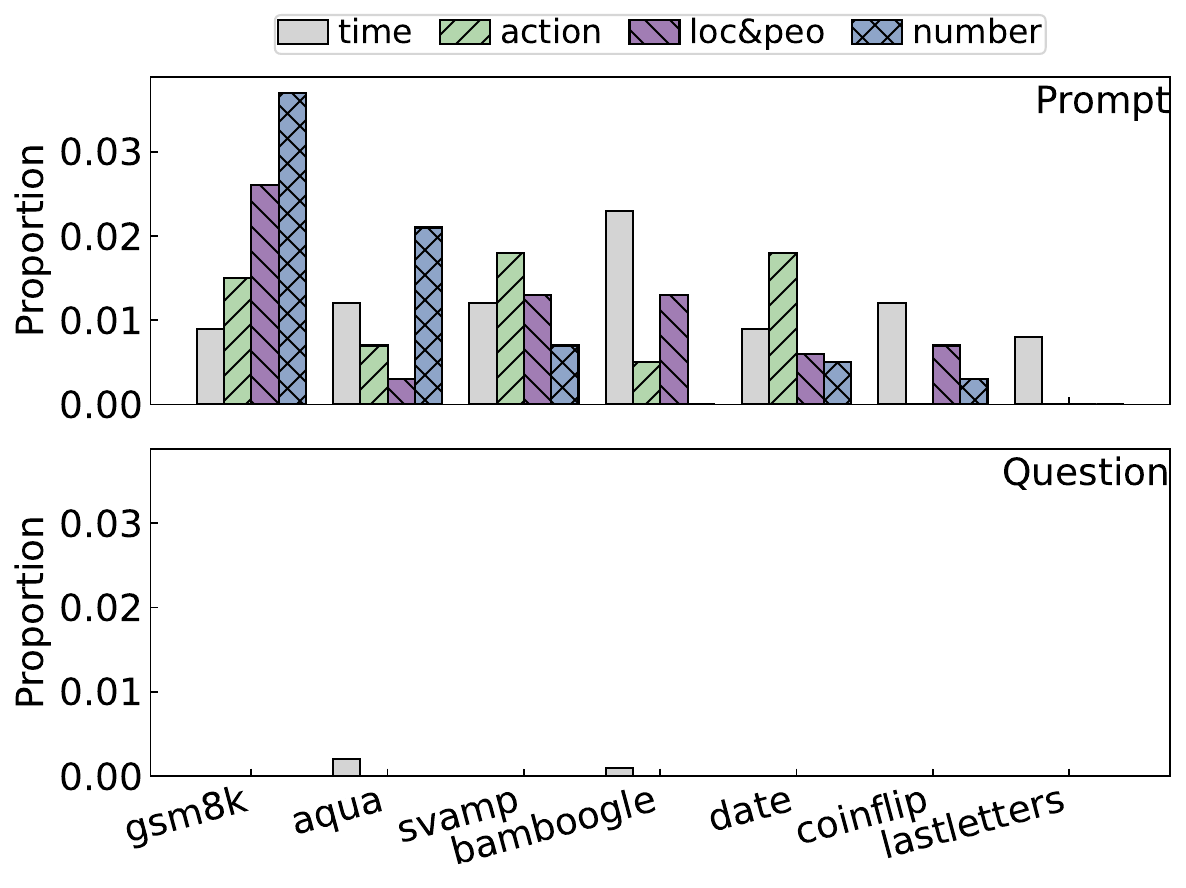}
        \caption{\textit{Sports}}
    \end{subfigure}
    \begin{subfigure}{0.3\linewidth}
        \includegraphics[width=\linewidth]{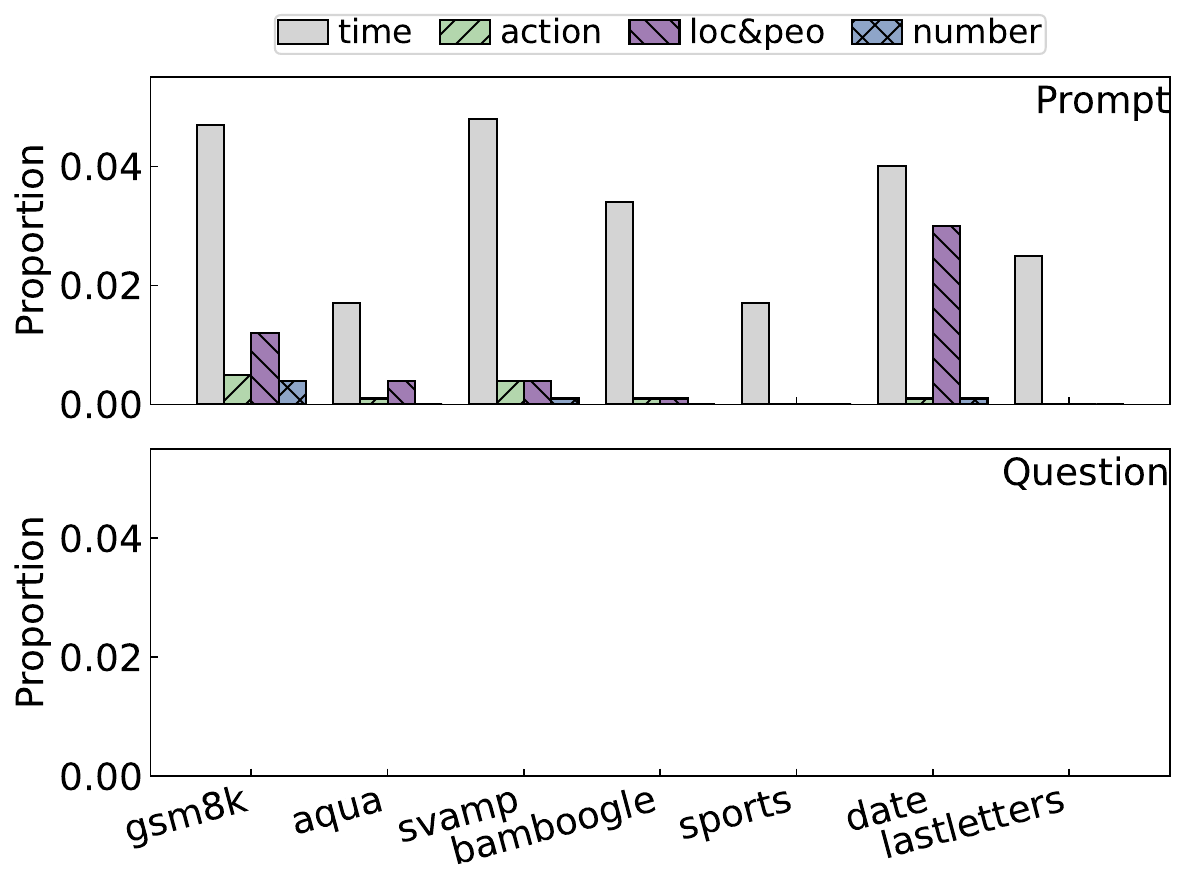}
        \caption{\textit{Coin Flip}}
    \end{subfigure}
    \begin{subfigure}{0.3\linewidth}
        \includegraphics[width=\linewidth]{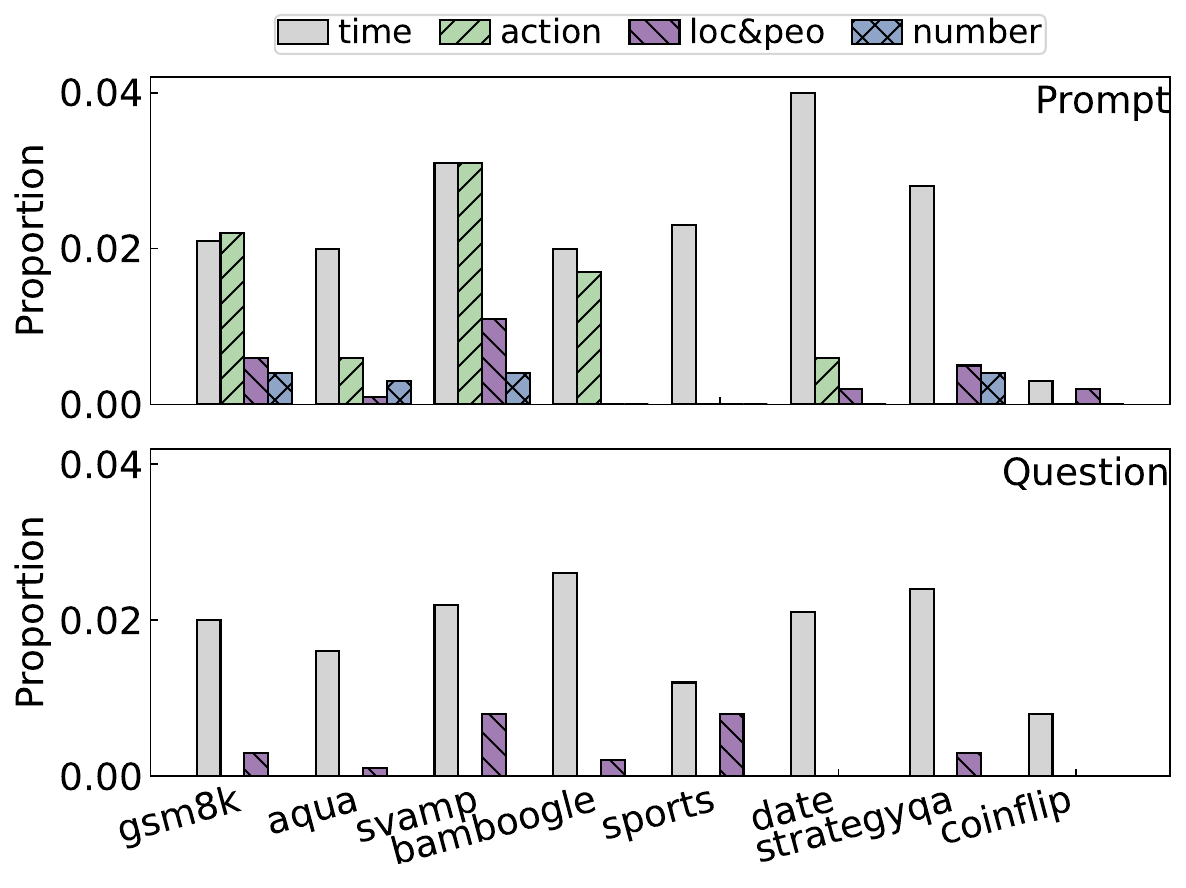}
        \caption{\textit{Last Letter}}
    \end{subfigure}
\vspace{-3mm}
\caption{Results of  Gemma2-9b.}
\label{fig:transfer_test_9b}
\end{figure*}
\vspace{-3mm}
\begin{figure*}[!h]
\centering
\captionsetup[sub]{skip=-1mm}
    \begin{subfigure}{0.3\linewidth}
        \includegraphics[width=\linewidth]{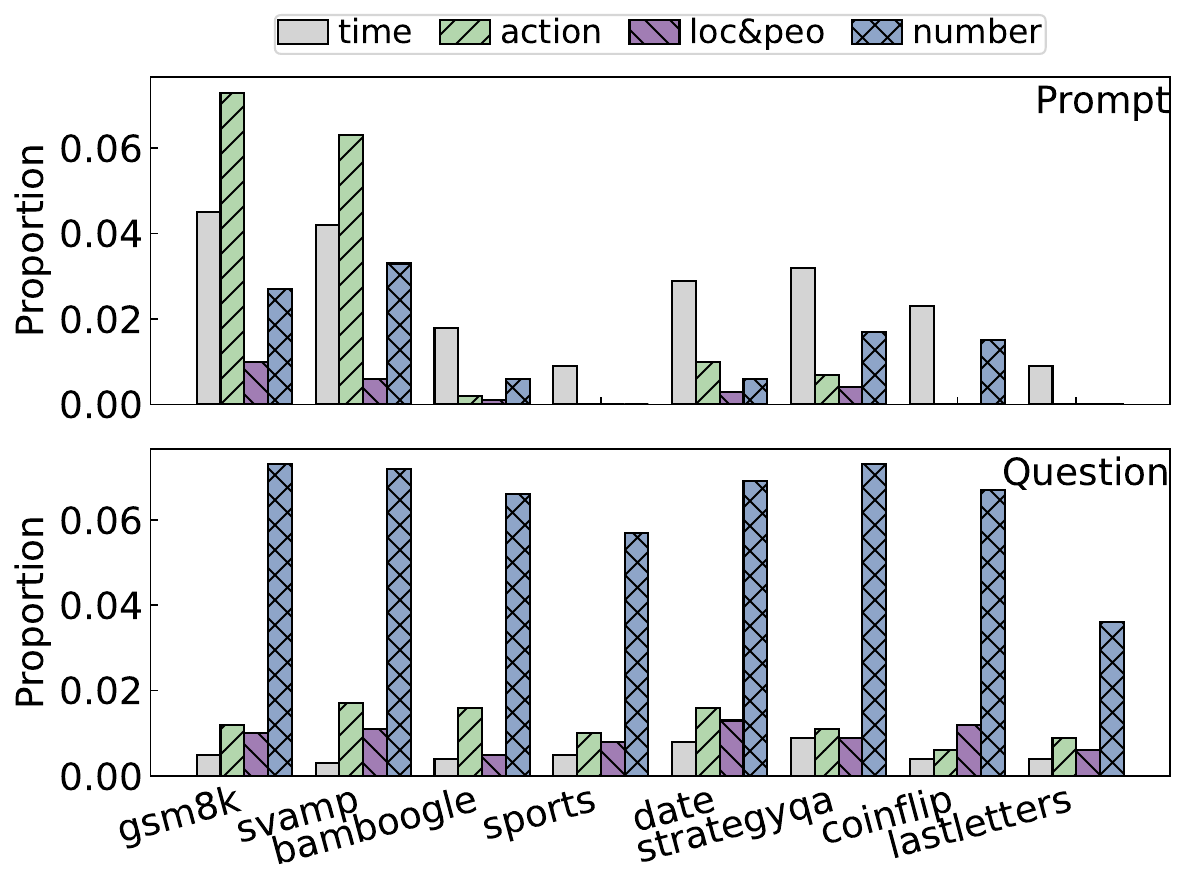}
        \caption{\textit{AQuA}}
        \label{fig:question1_a2_a}
    \end{subfigure}
    \begin{subfigure}{0.3\linewidth}
        \includegraphics[width=\linewidth]{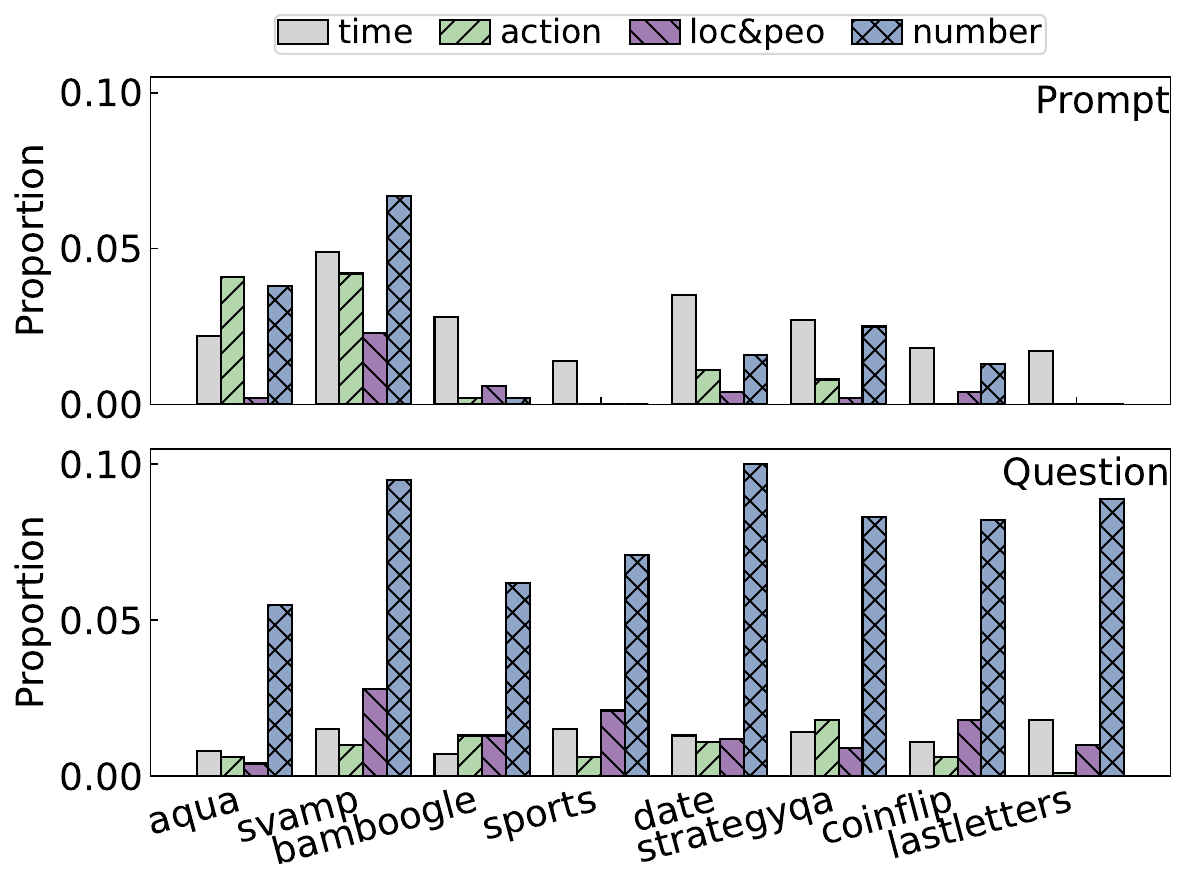}
        \caption{\textit{GSM8K}}
        \label{fig:question1_a2_b}
    \end{subfigure}
    \begin{subfigure}{0.3\linewidth}
        \includegraphics[width=\linewidth]{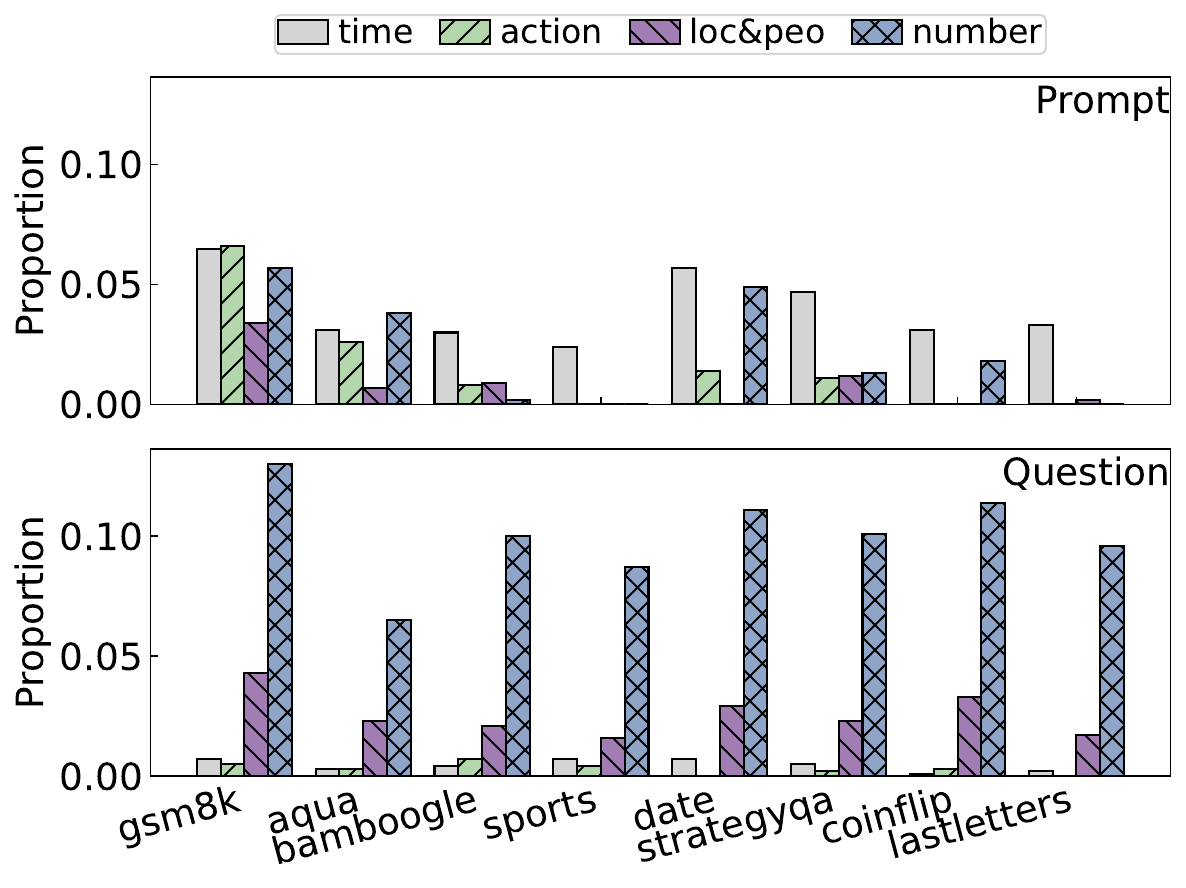}
        \caption{\textit{SVAMP}}
        \label{fig:question1_a2_c}
    \end{subfigure}

  \begin{subfigure}{0.3\linewidth}
        \includegraphics[width=\linewidth]{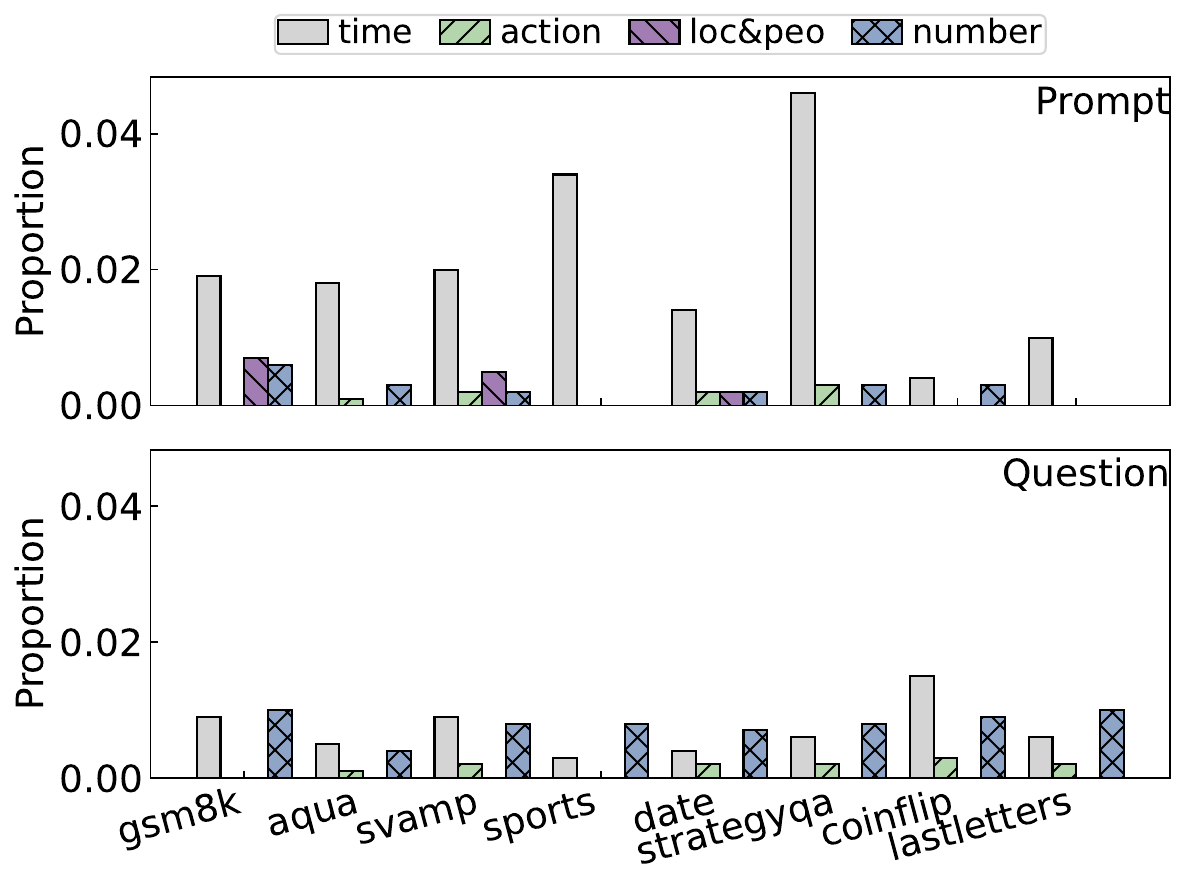}
        \caption{\textit{Bamboogle}}
        \label{fig:question1_a2_e}
    \end{subfigure}
  \begin{subfigure}{0.3\linewidth}
        \includegraphics[width=\linewidth]{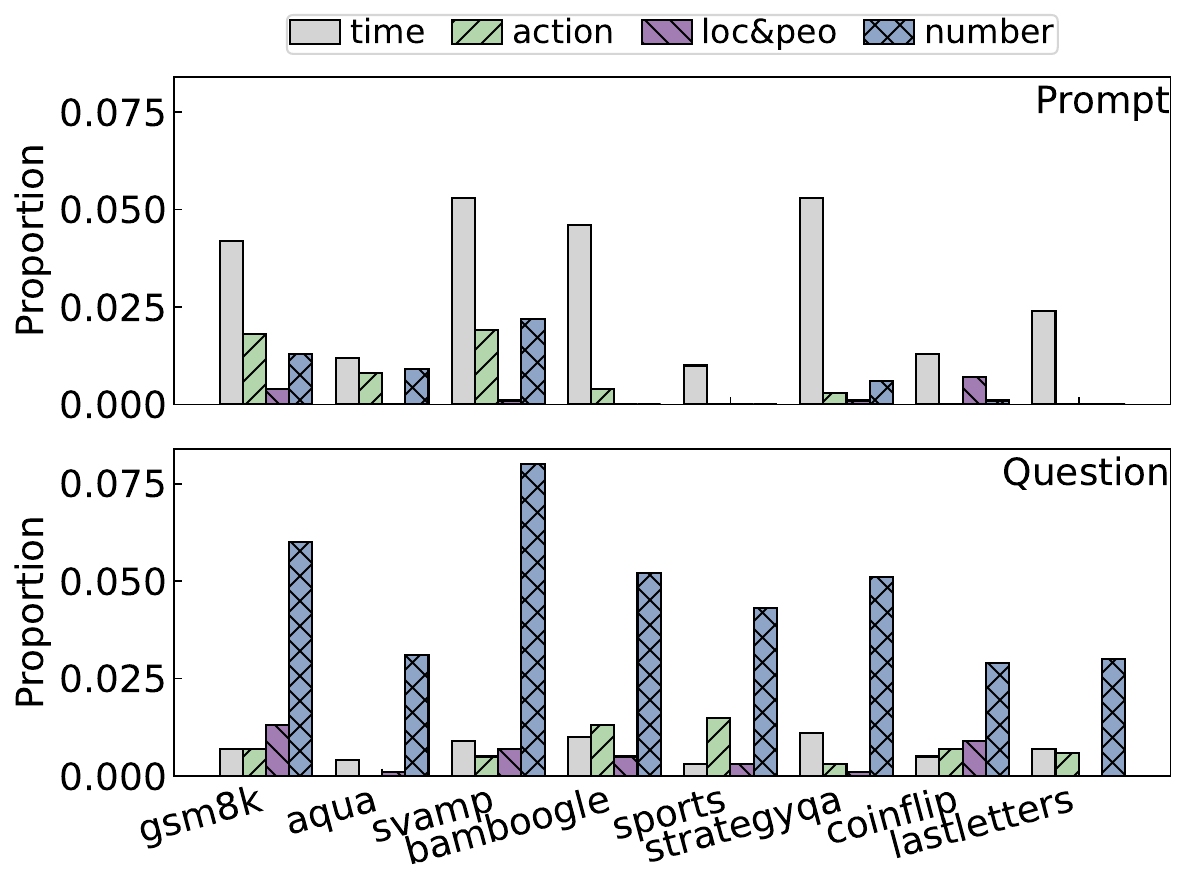}
        \caption{\textit{Date}}
        \label{fig:question1_a2_d}
    \end{subfigure}
    \begin{subfigure}{0.3\linewidth}
        \includegraphics[width=\linewidth]{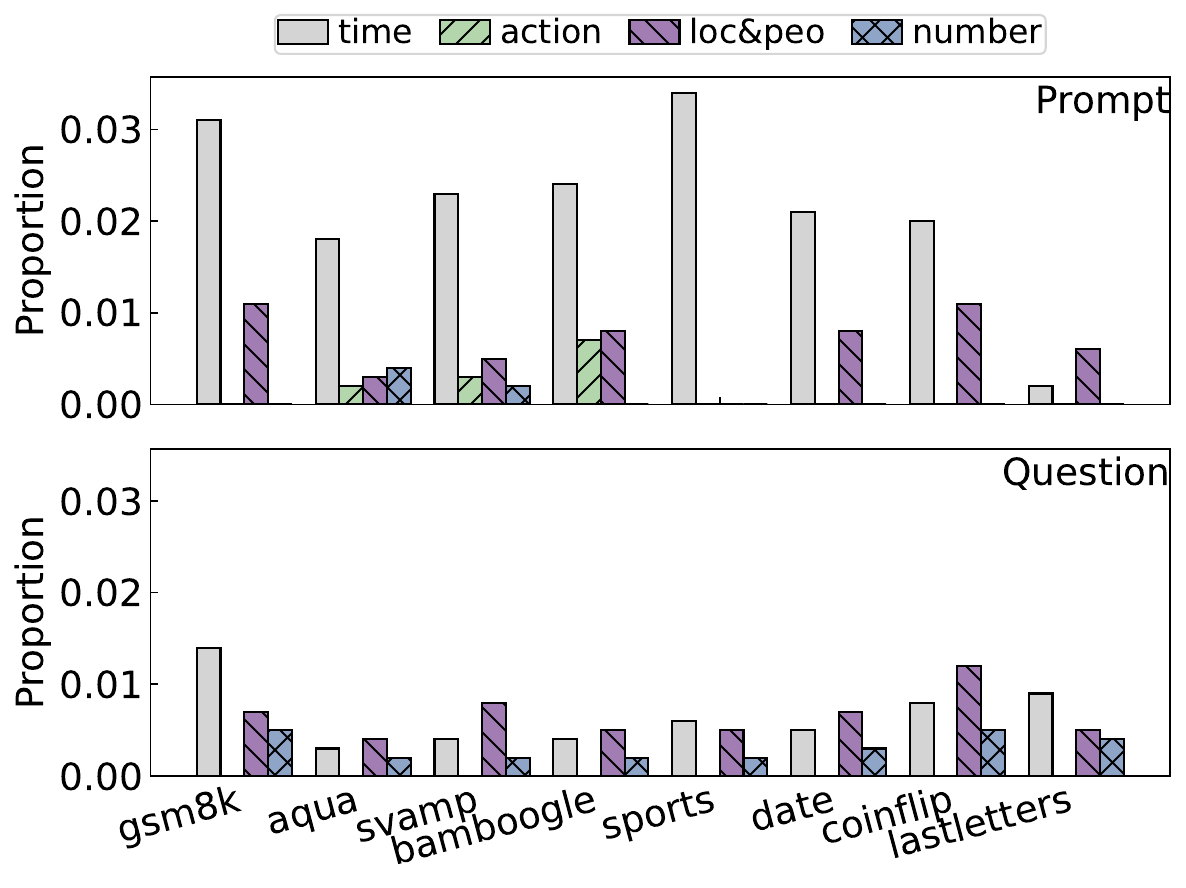}
        \caption{\textit{StrategyQA}}
    \end{subfigure}
    
     \begin{subfigure}{0.3\linewidth}
        \includegraphics[width=\linewidth]{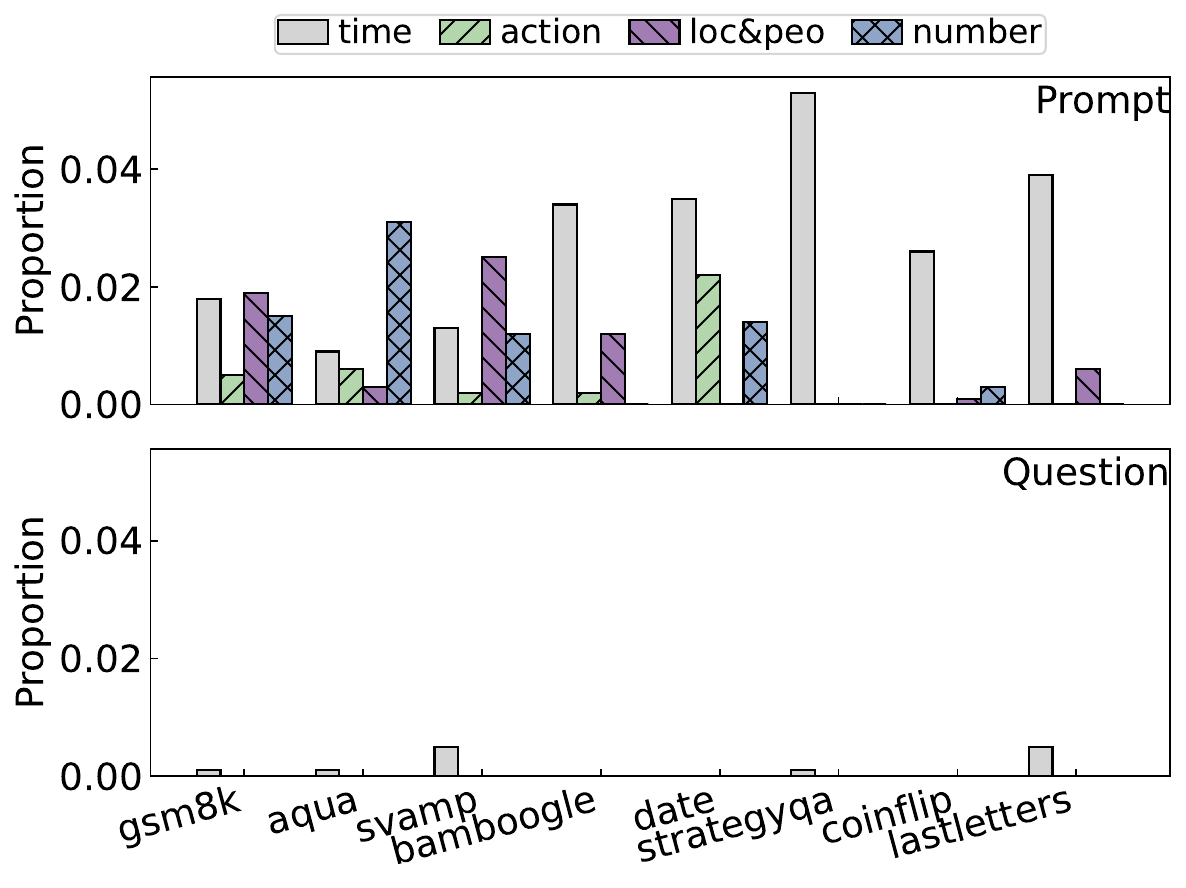}
        \caption{\textit{Sports}}
    \end{subfigure}
    \begin{subfigure}{0.3\linewidth}
        \includegraphics[width=\linewidth]{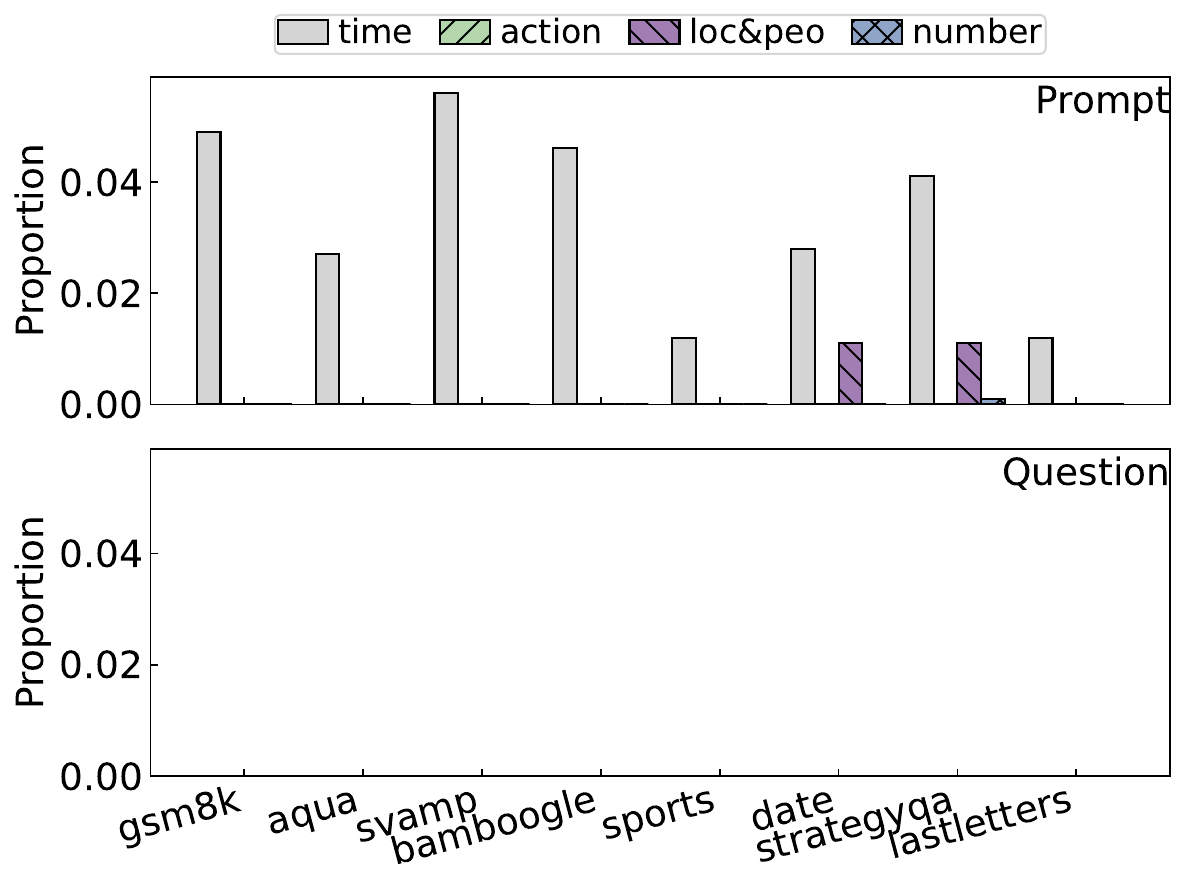}
        \caption{\textit{Coin Flip}}
    \end{subfigure}
    \begin{subfigure}{0.3\linewidth}
        \includegraphics[width=\linewidth]{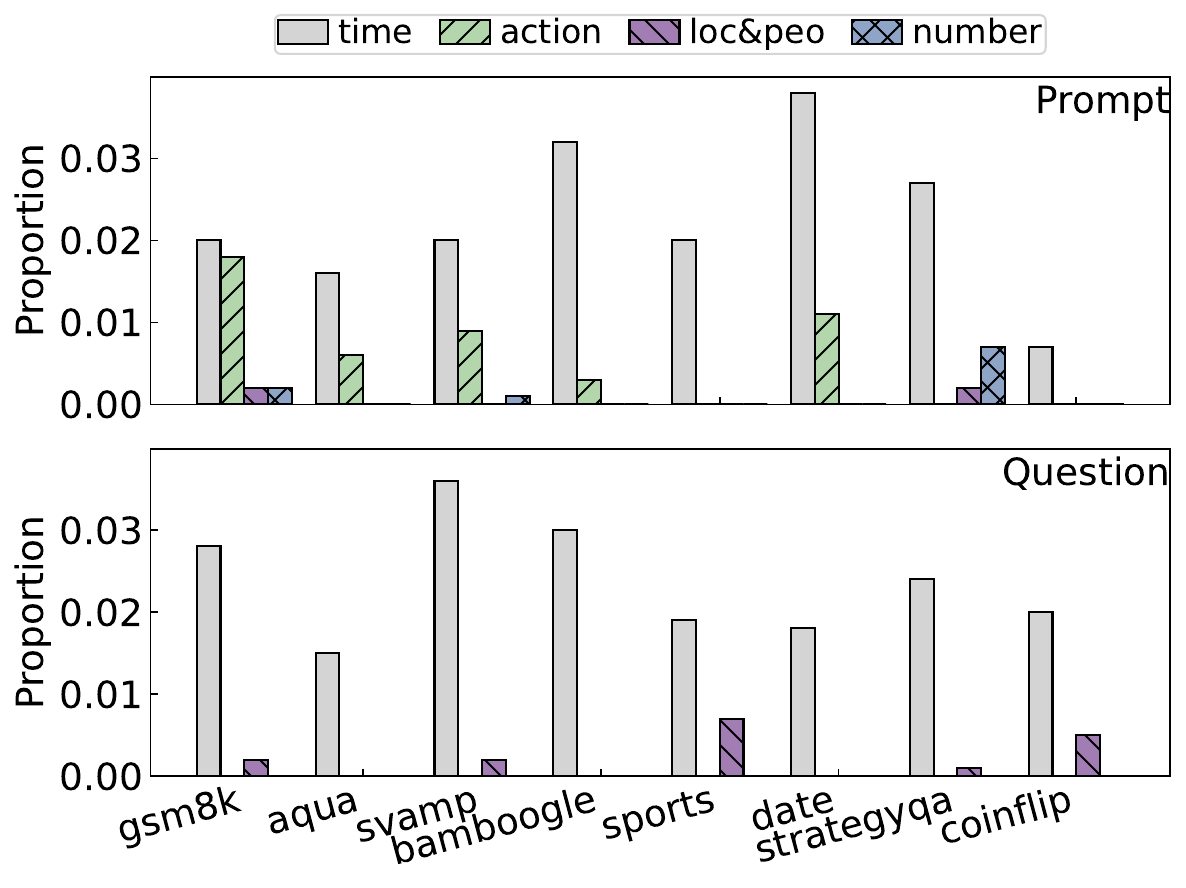}
        \caption{\textit{Last Letter}}
    \end{subfigure}
\vspace{-3mm}
\caption{Results of  LLaMA2-13b.}
\label{fig:transfer_test_13b}
\end{figure*}
\clearpage

\twocolumn
\section{Quantifying Structure Adherence}\label{sec:template_adherence}
Evaluating the fidelity of the model's generated content to the CoT reasoning format involved a three-stage assessment.
Initially, entities were extracted from the input prompts. The definition and extraction methodology varied by task domain: numbers for arithmetic tasks; names, locations, and temporal entities using SpaCy for commonsense benchmarks (e.g., Bamboogle, Sports), regular expressions for the Date dataset, names for symbolic reasoning like Coin Flip, and target words for the Last Letter Concatenation.
Subsequently, we determined the presence of intermediate reasoning steps. For most tasks, this involved identifying the generation of new entities not present in the initial prompt. However, for tasks less prone to explicit entity generation (e.g., Coin Flip, Last Letter), a verb-based heuristic was employed: counting occurrences of key process verbs such as ``flips, is, was, are, be, were''. A count exceeding four for these tasks was taken as an indicator of reasoning activity.
Finally, the completion of the reasoning process with a conclusive answer was verified by locating the explicit phrase ``the answer is'' at the terminal position of the generated text.
Content was deemed to exhibit CoT adherence if it satisfied all three aforementioned conditions.
\onecolumn


\clearpage

\twocolumn
\section{Probability Distribution of Individual Tokens}\label{appendix:probability_score}
\begin{figure*}[ht]
\centering
    \begin{subfigure}{0.45\linewidth}
        \includegraphics[width=\linewidth]
        {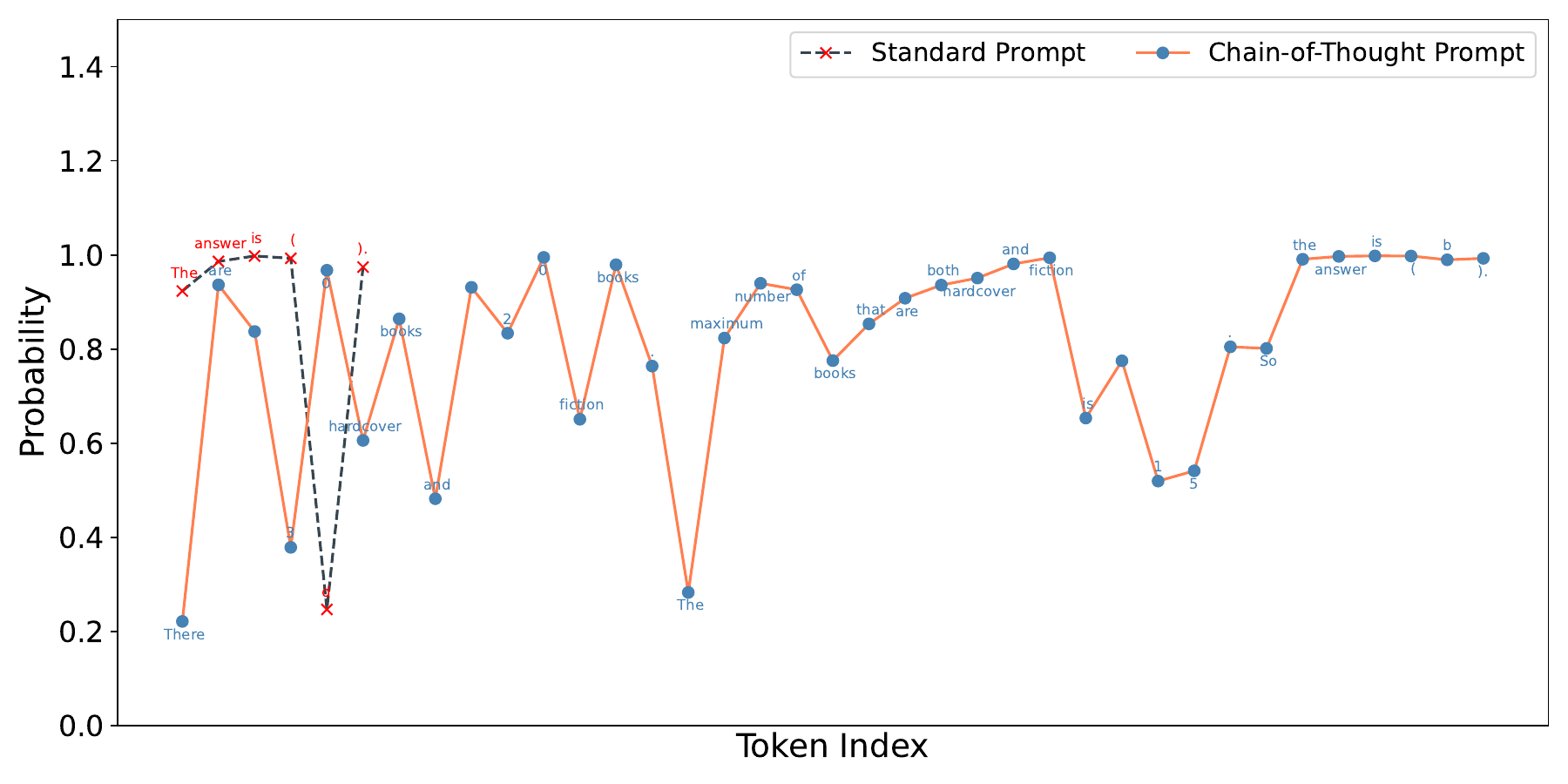}
        \caption{\textit{AQuA}}
    \end{subfigure}
    \begin{subfigure}{0.45\linewidth}
        \includegraphics[width=\linewidth]
        {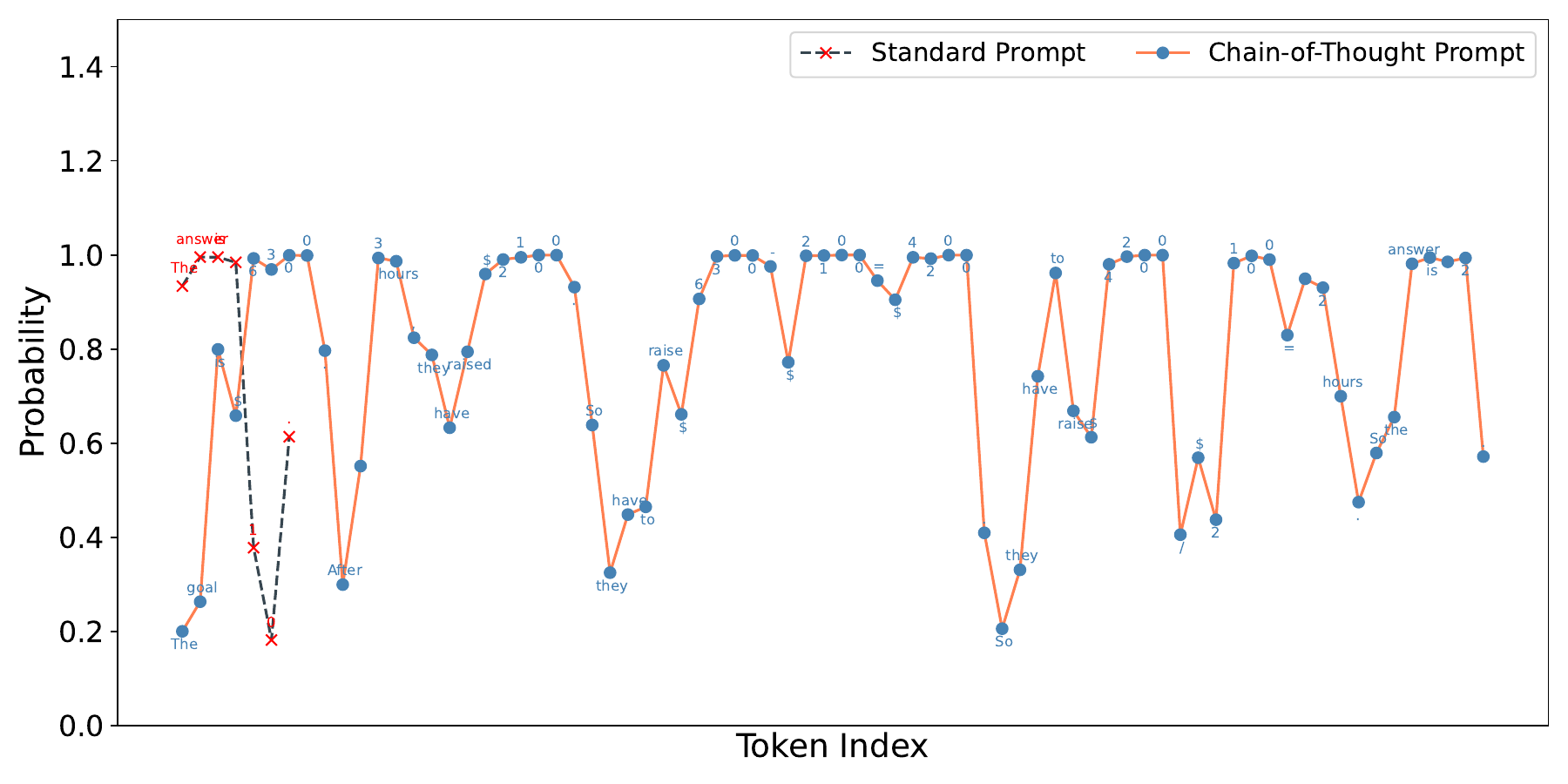}
        \caption{\textit{GSM8K}}
    \end{subfigure}
    \begin{subfigure}{0.45\linewidth}
        \includegraphics[width=\linewidth]{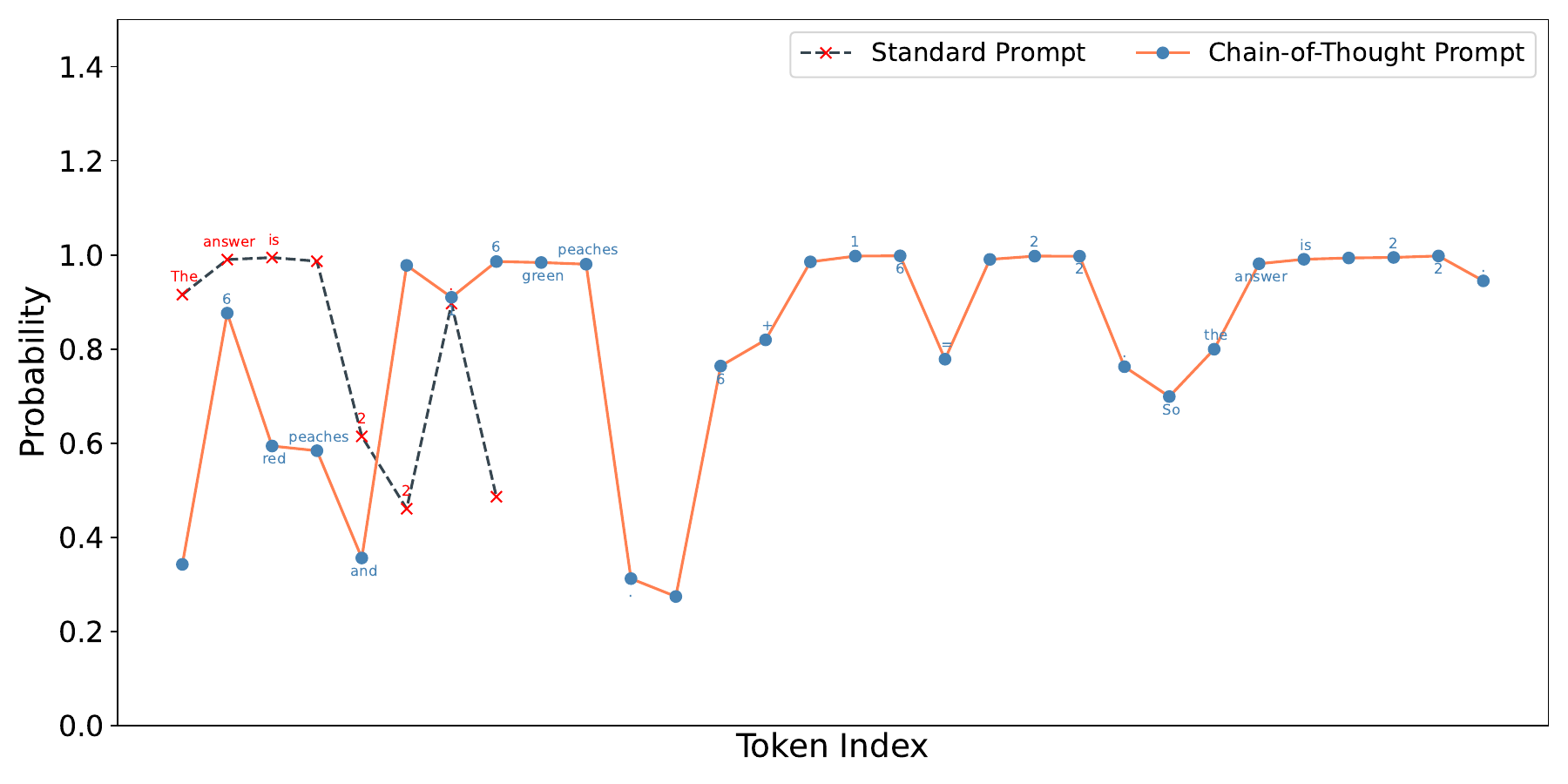}
        \caption{\textit{SVAMP}}
    \end{subfigure}
    \begin{subfigure}{0.45\linewidth}
        \includegraphics[width=\linewidth]
        {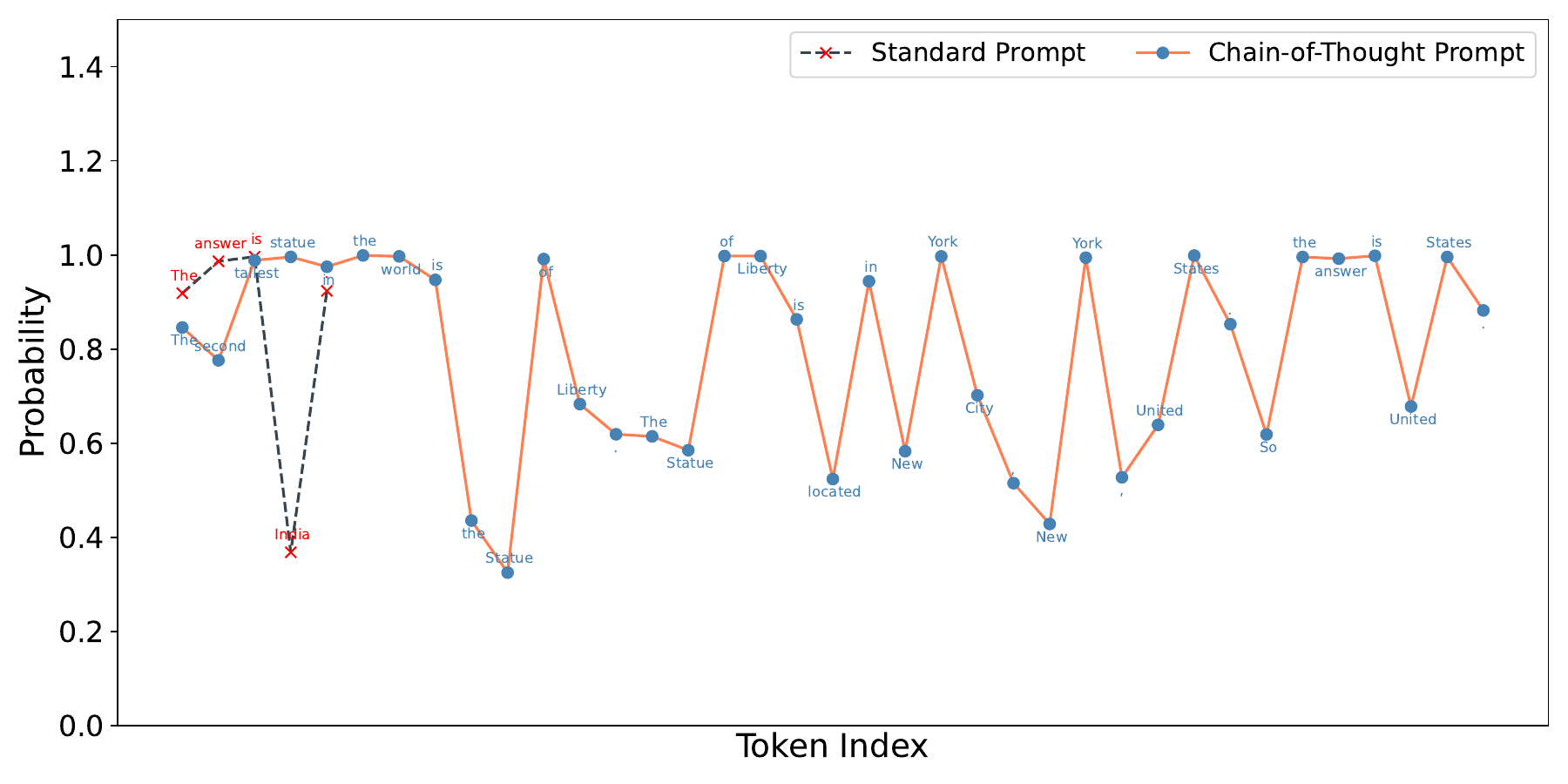}
        \caption{\textit{Bamboogle}}
    \end{subfigure}
    \begin{subfigure}{0.45\linewidth}
        \includegraphics[width=\linewidth]
        {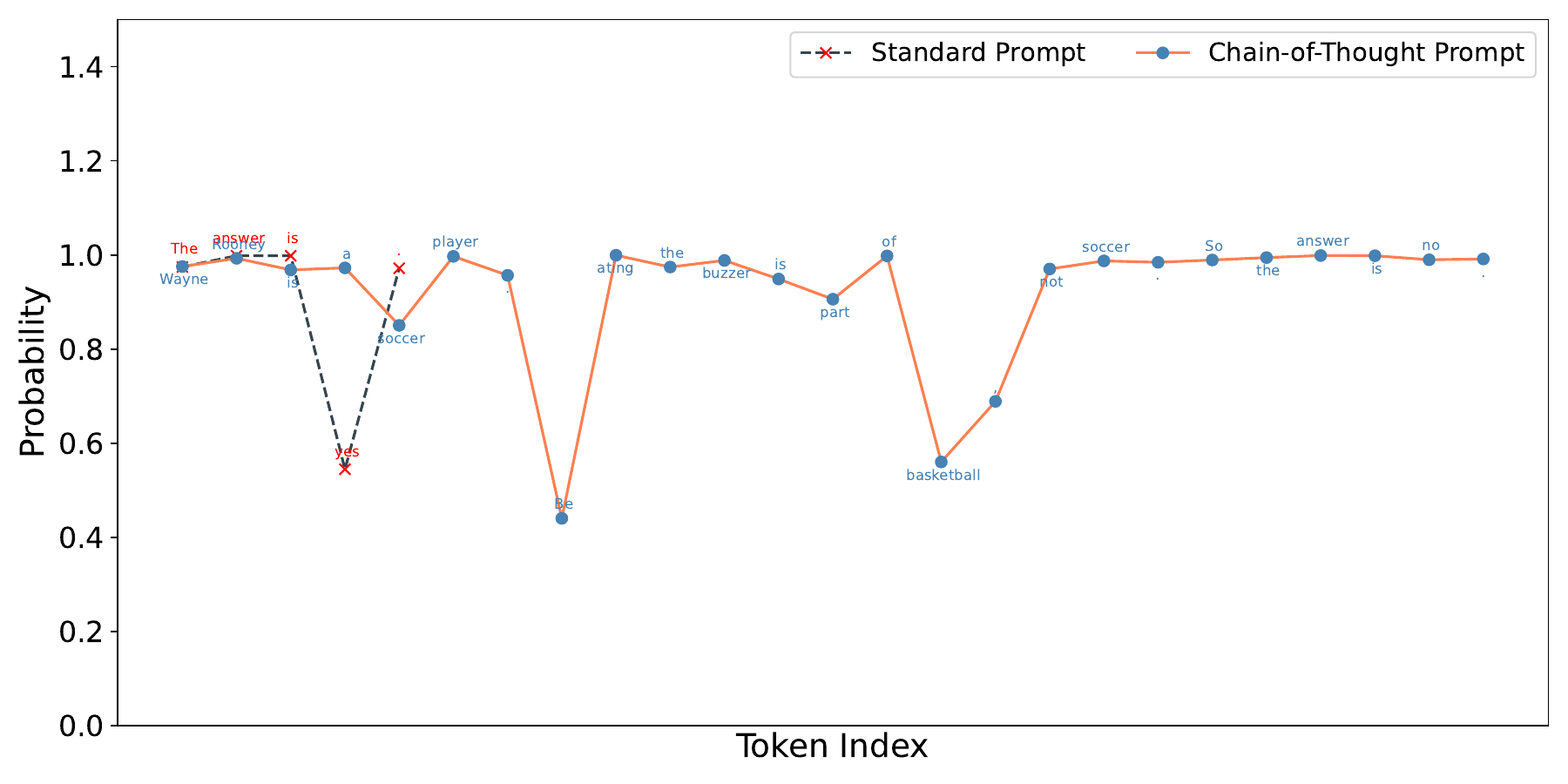}
        \caption{\textit{Sports}}
    \end{subfigure}
    \begin{subfigure}{0.45\linewidth}
        \includegraphics[width=\linewidth]
        {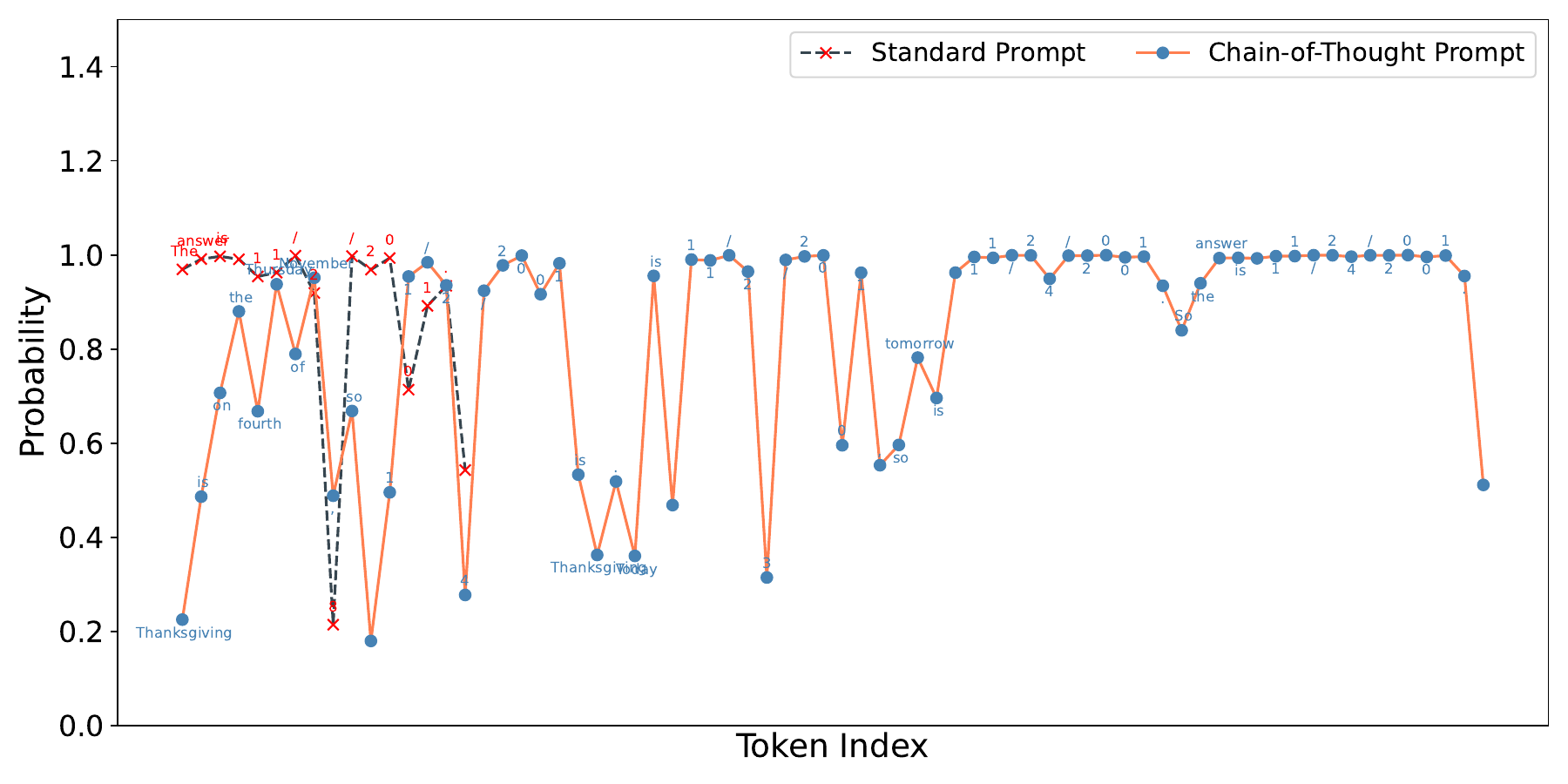}
        \caption{\textit{Date}}
    \end{subfigure}
    \begin{subfigure}{0.45\linewidth}
        \includegraphics[width=\linewidth]
        {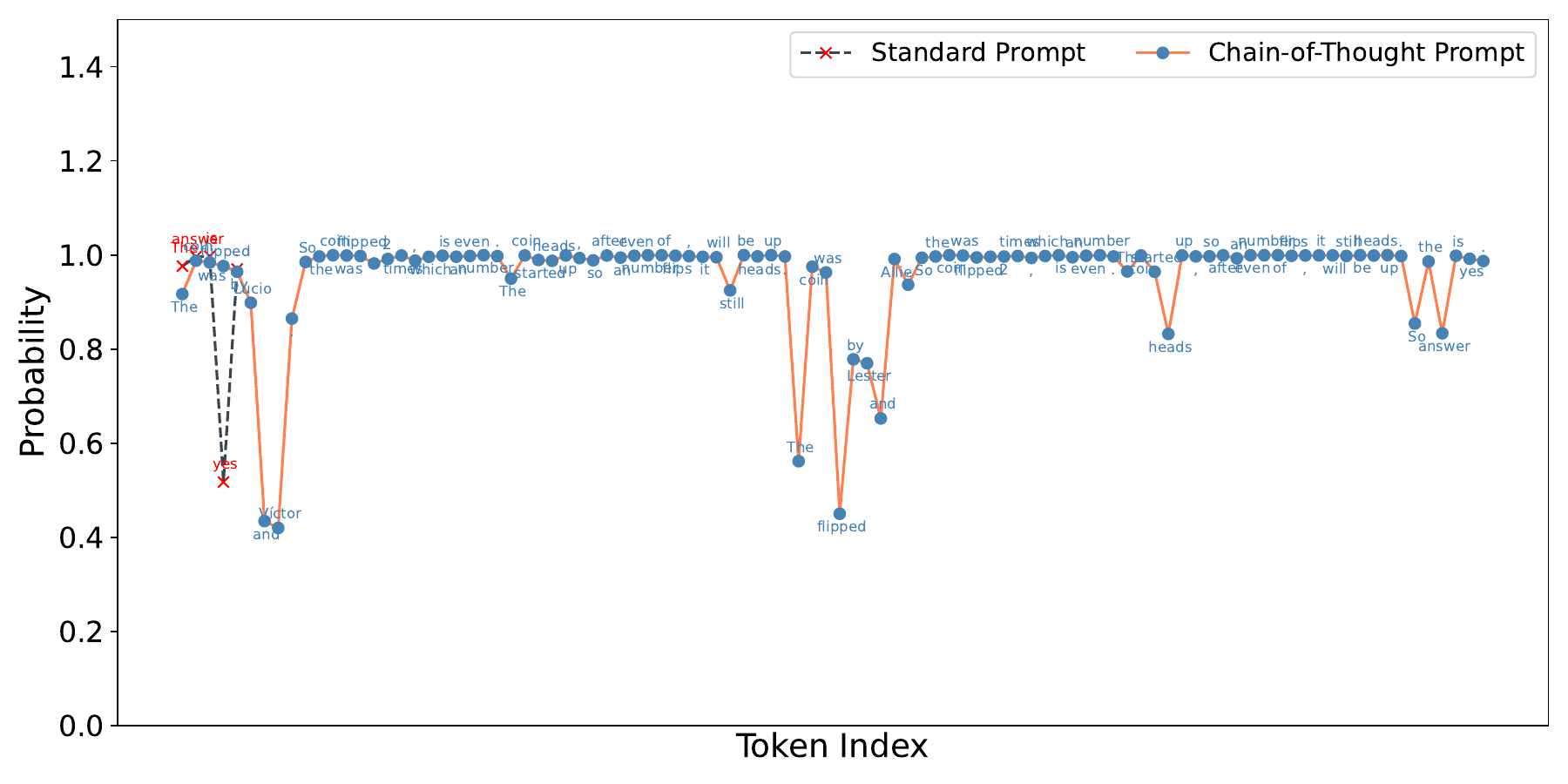}
        \caption{\textit{Coin Flip}}
    \end{subfigure}
     \begin{subfigure}{0.45\linewidth}
        \includegraphics[width=\linewidth]
        {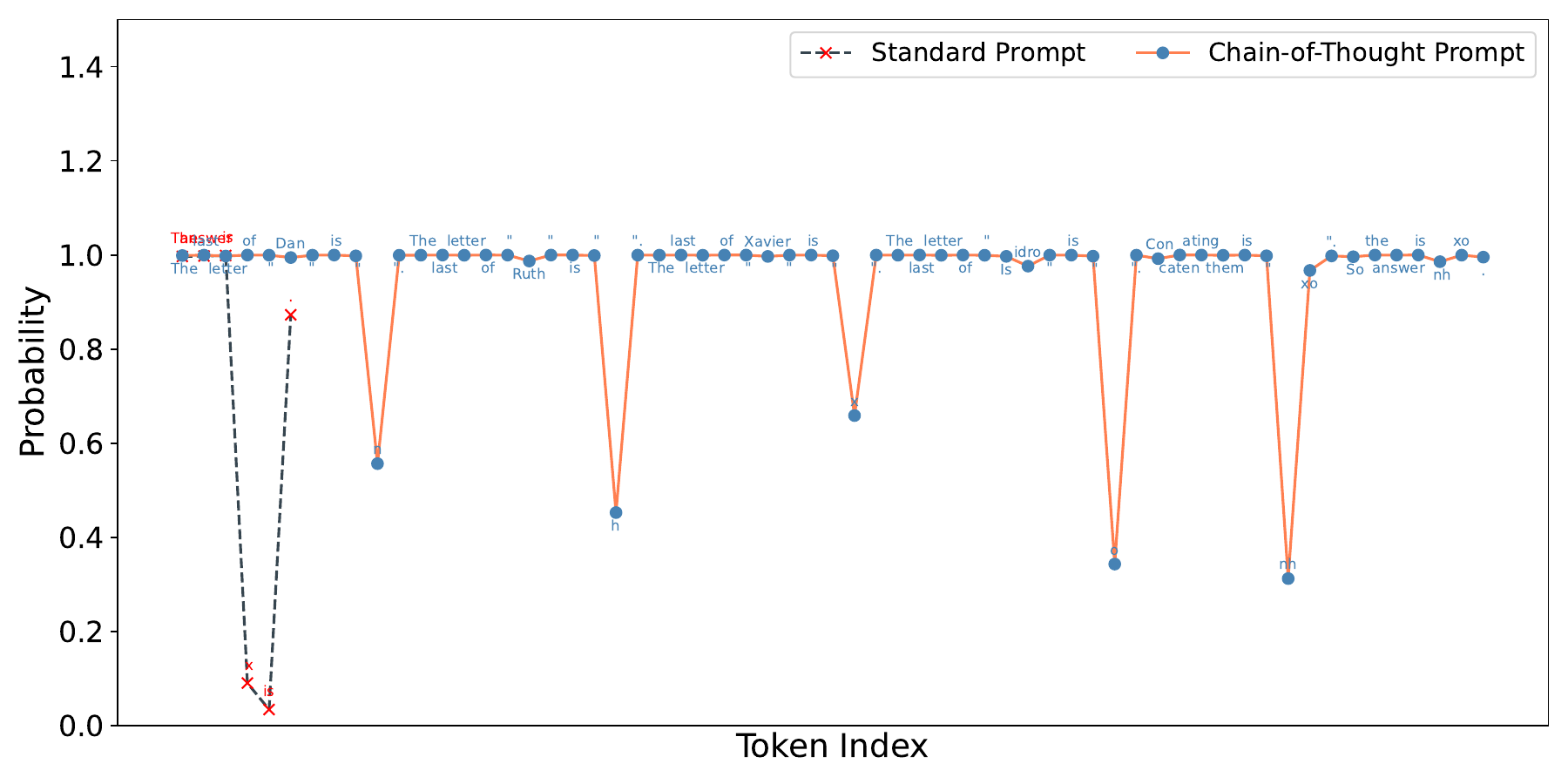}
        \caption{\textit{Last Letter Concatenation}}
    \end{subfigure}
\caption{Probability value of each generated token (results of Gemma2-2b is shown).}
\label{fig:question2_logits_value_2b}
\end{figure*}

\begin{figure*}
\centering
    \begin{subfigure}{0.45\linewidth}
        \includegraphics[width=\linewidth]
        {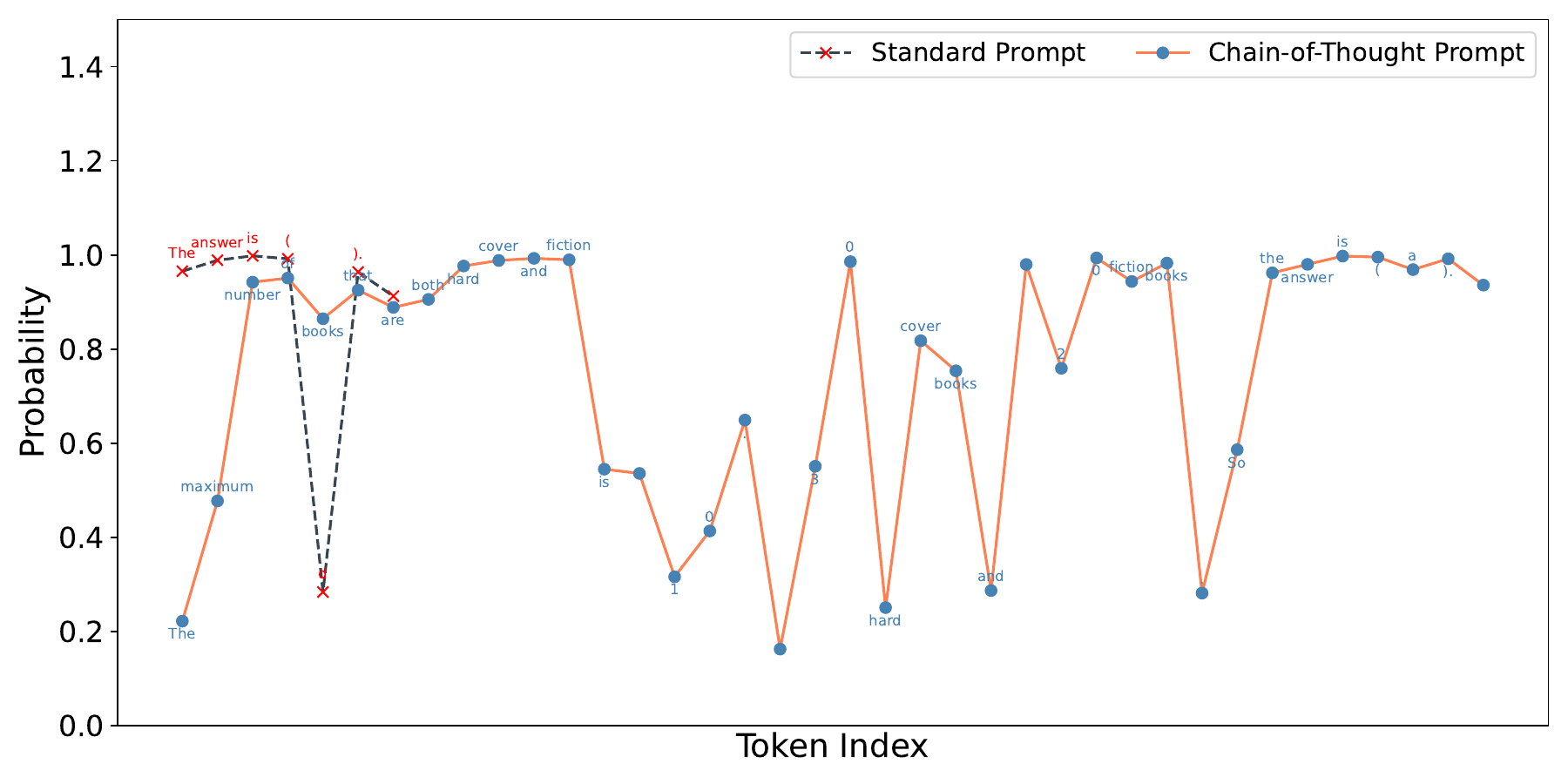}
        \caption{\textit{AQuA}}
    \end{subfigure}
    \begin{subfigure}{0.45\linewidth}
        \includegraphics[width=\linewidth]
        {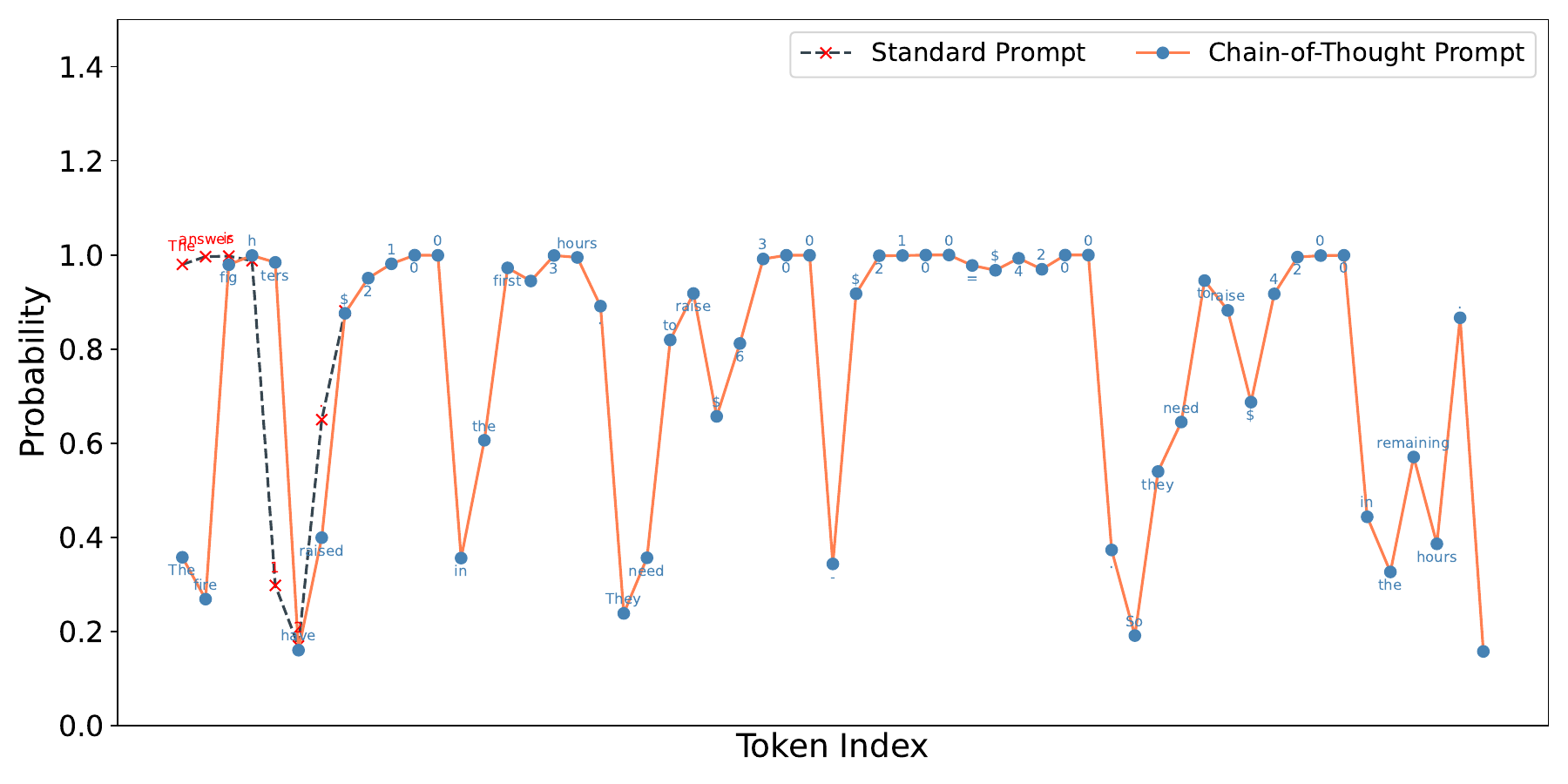}
        \caption{\textit{GSM8K}}
    \end{subfigure}
    \begin{subfigure}{0.45\linewidth}
        \includegraphics[width=\linewidth]{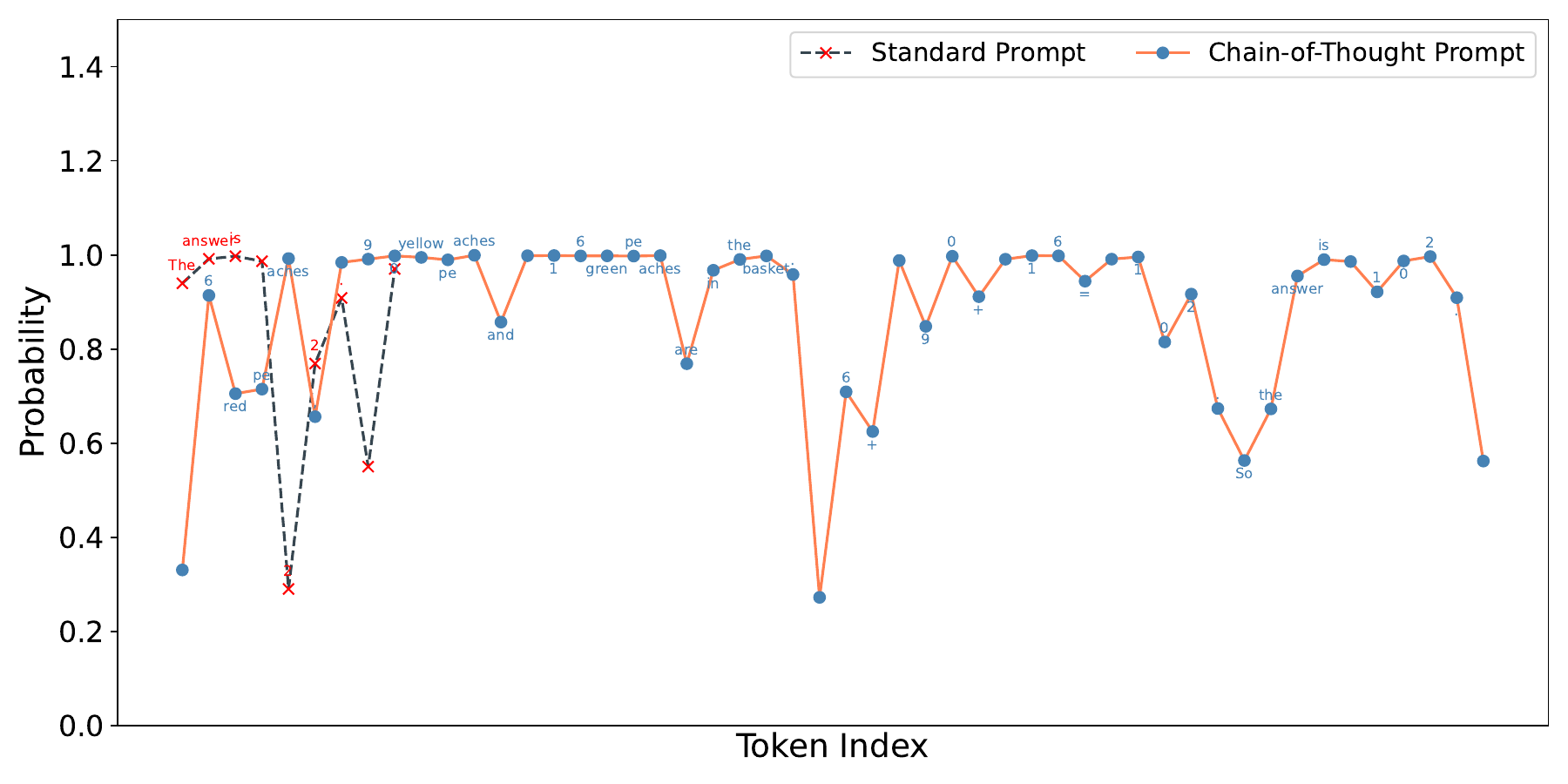}
        \caption{\textit{SVAMP}}
    \end{subfigure}
    \begin{subfigure}{0.45\linewidth}
        \includegraphics[width=\linewidth]
        {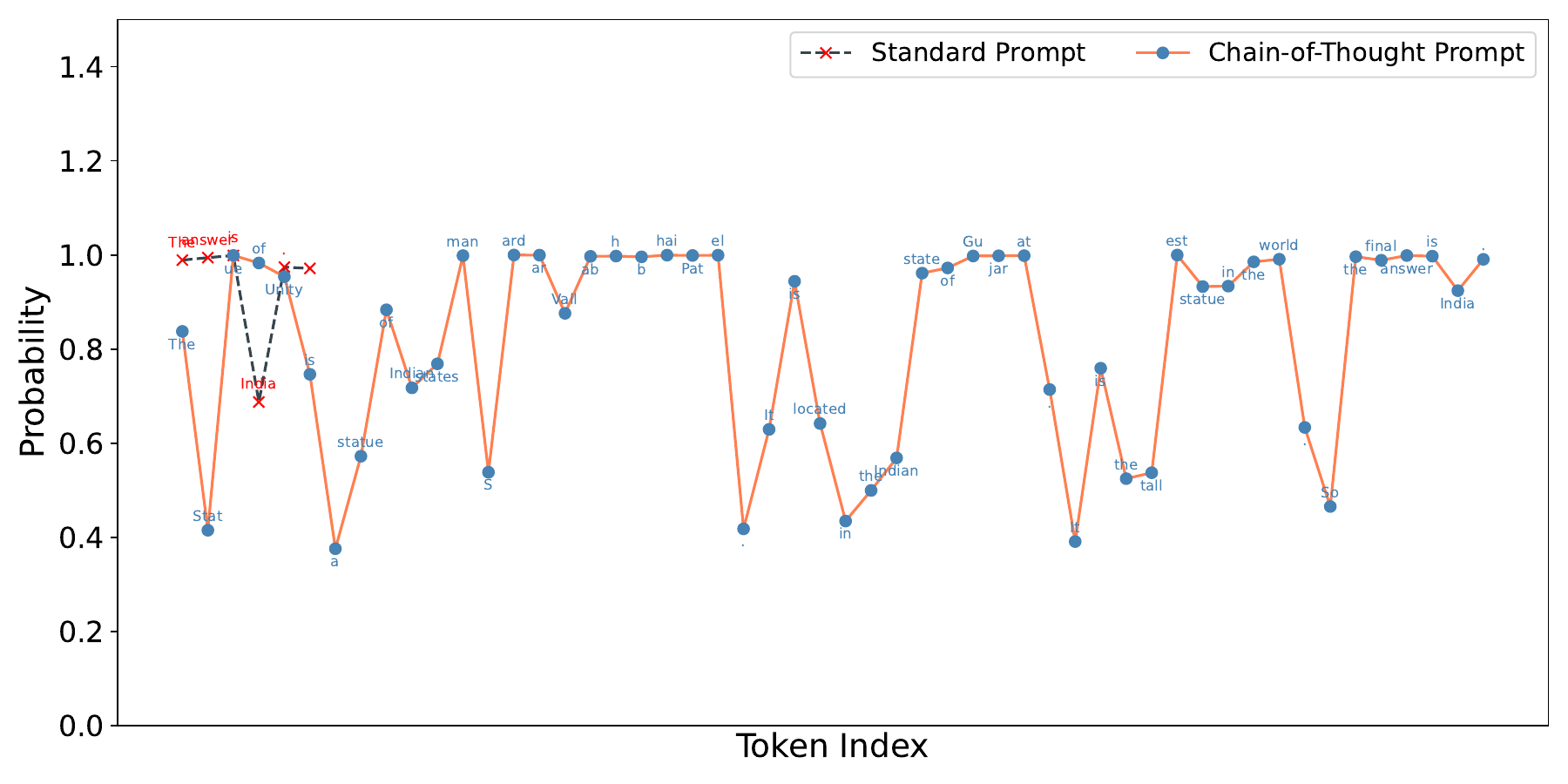}
        \caption{\textit{Bamboogle}}
    \end{subfigure}
    \begin{subfigure}{0.45\linewidth}
        \includegraphics[width=\linewidth]
        {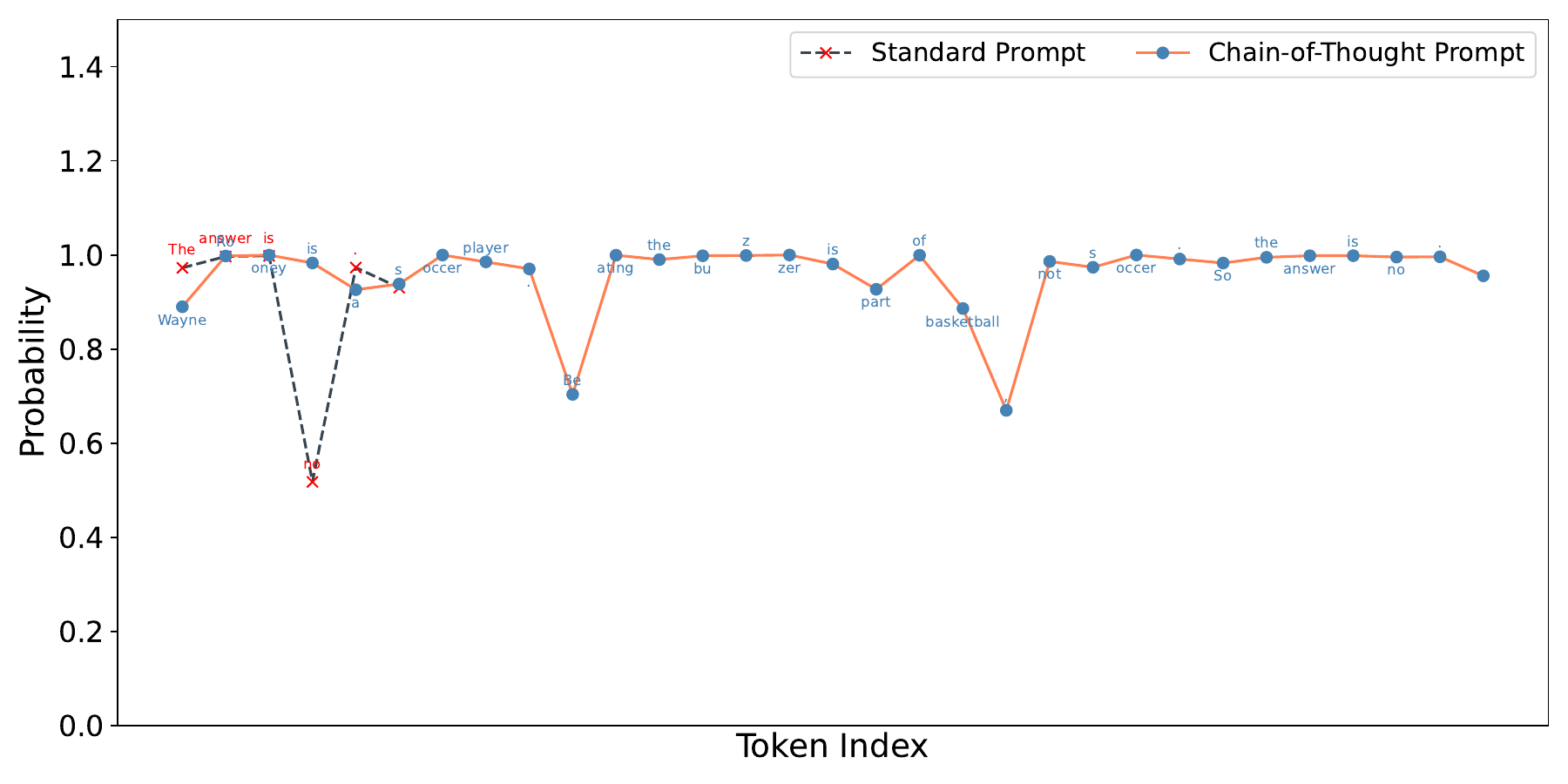}
        \caption{\textit{Sports}}
    \end{subfigure}
    \begin{subfigure}{0.45\linewidth}
        \includegraphics[width=\linewidth]
        {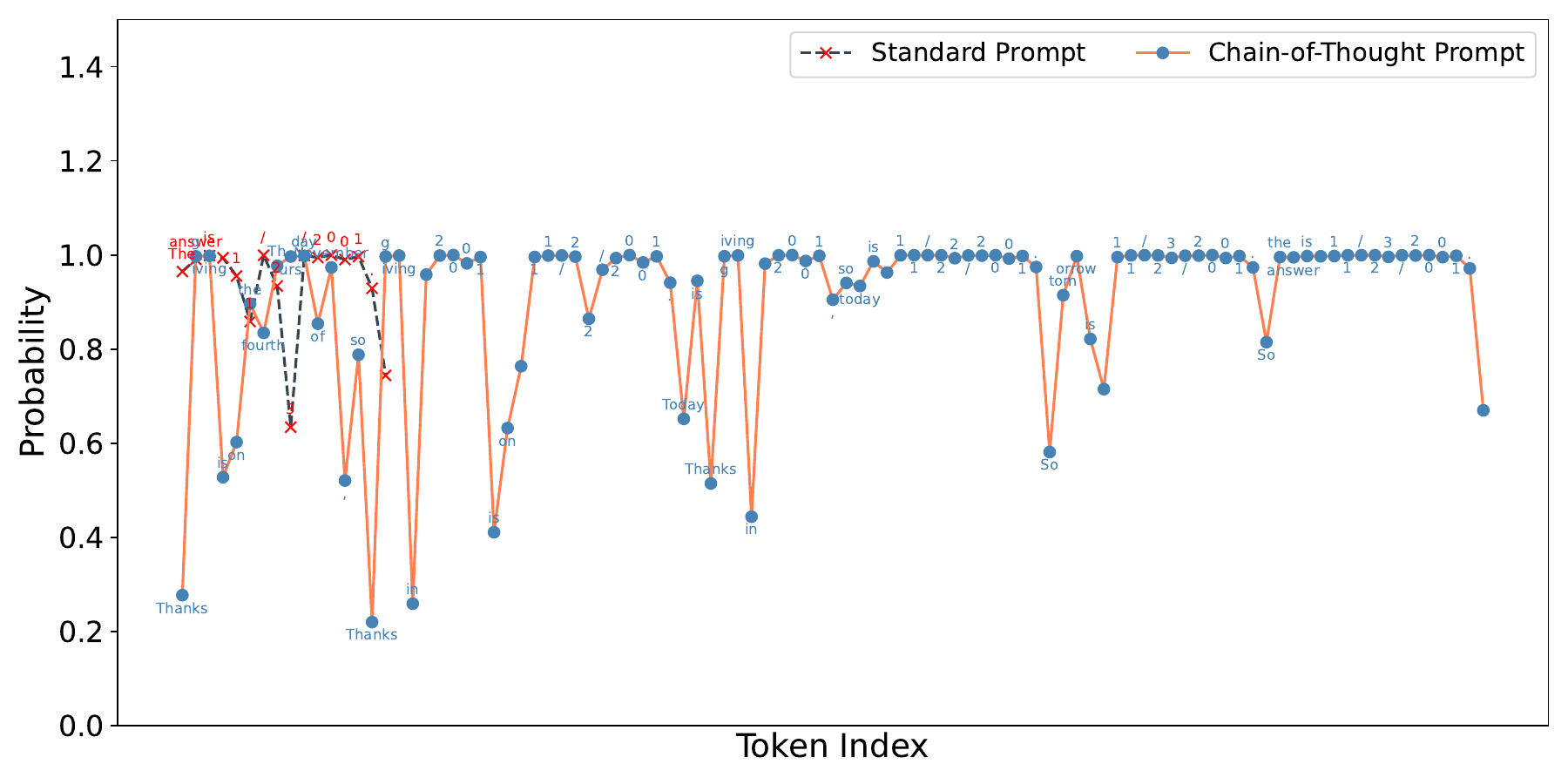}
        \caption{\textit{Date}}
    \end{subfigure}
    \begin{subfigure}{0.45\linewidth}
        \includegraphics[width=\linewidth]
        {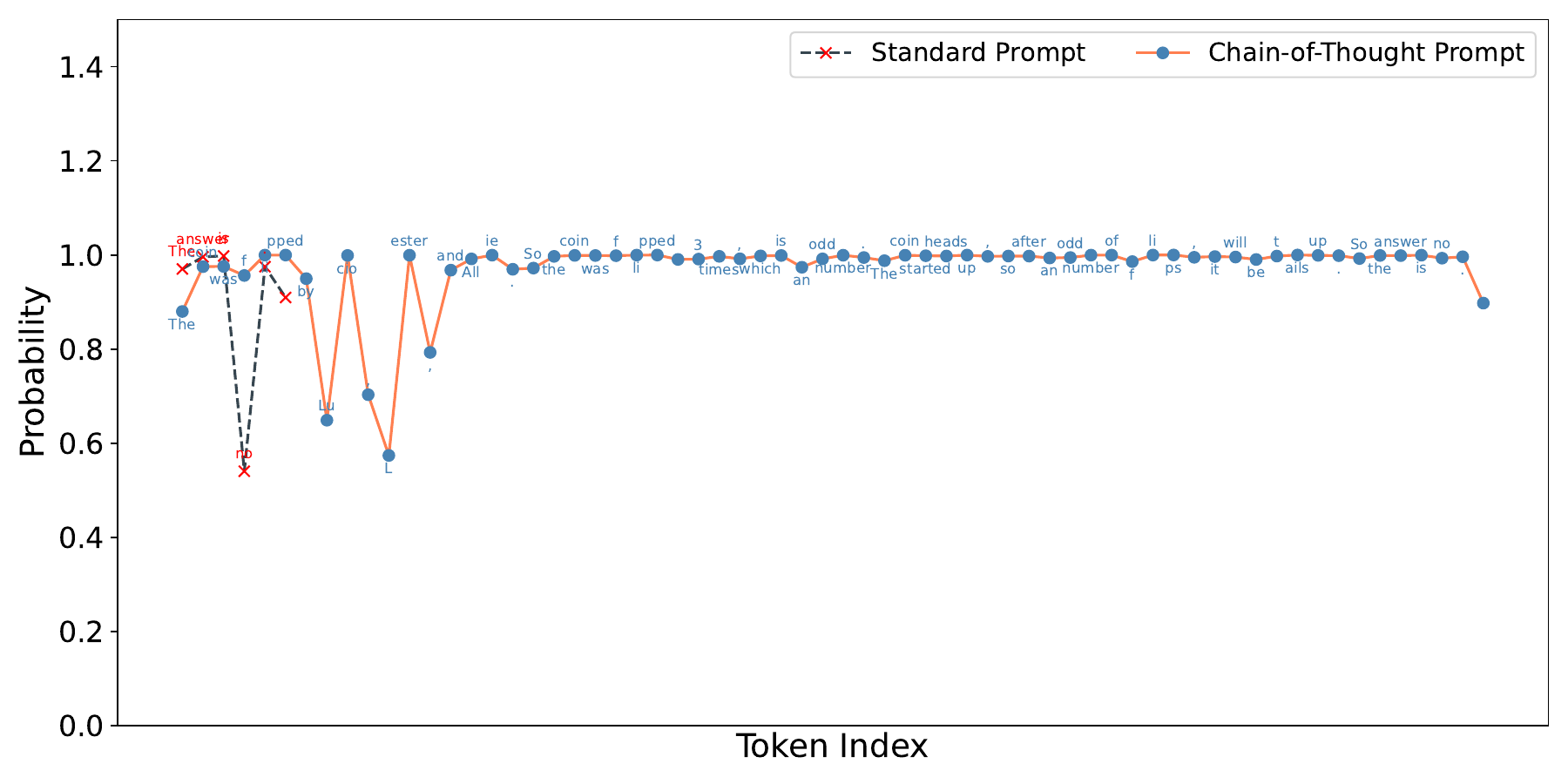}
        \caption{\textit{Coin Flip}}
    \end{subfigure}
     \begin{subfigure}{0.45\linewidth}
        \includegraphics[width=\linewidth]
        {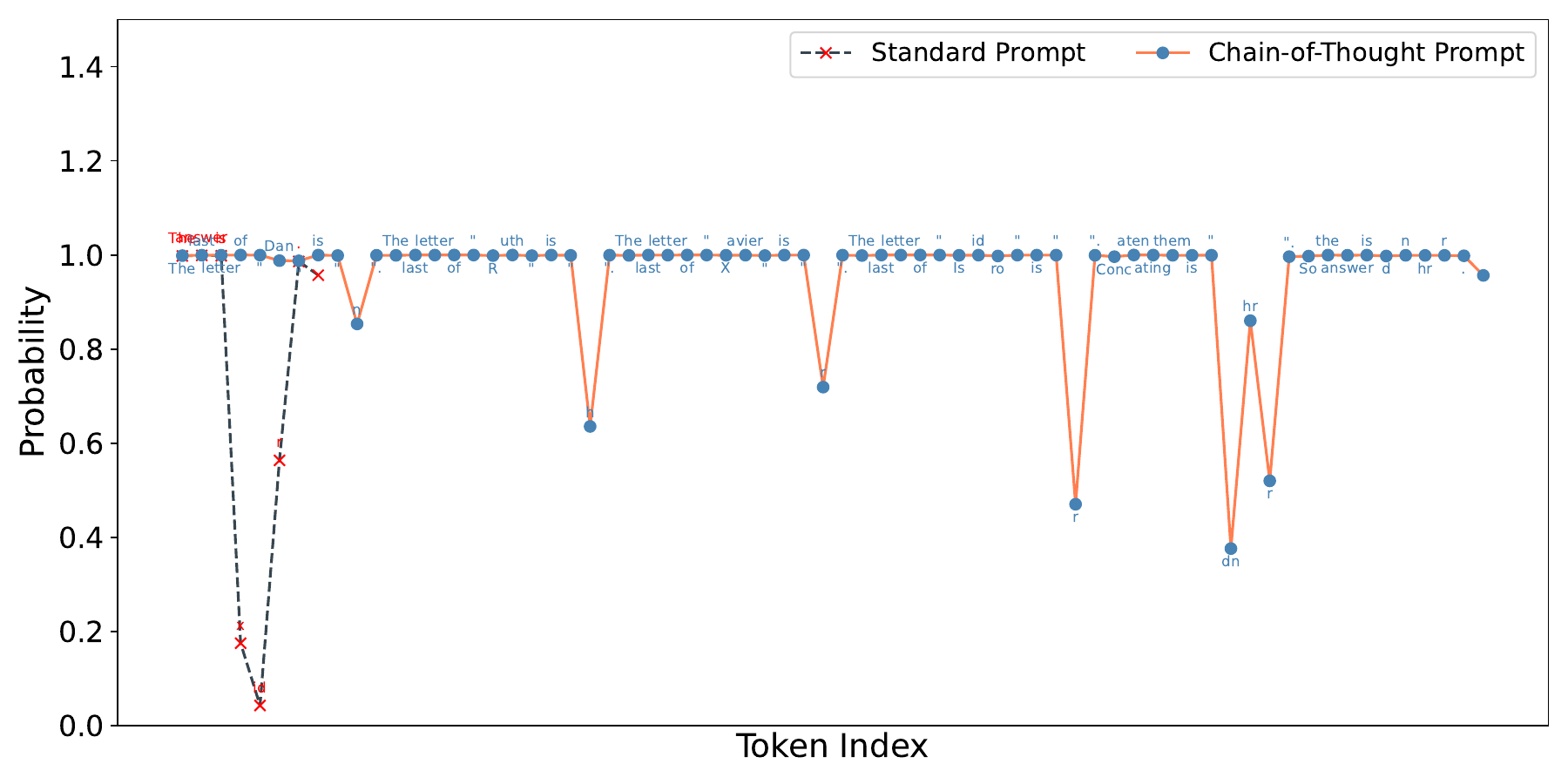}
        \caption{\textit{Last Letter Concatenation}}
    \end{subfigure}
\caption{Probability value of each generated token (the results of LLaMA2-13b is shown).}
\label{fig:question2_logits_value_13b}
\end{figure*}

\begin{figure*}
\centering
    \begin{subfigure}{0.45\linewidth}
        \includegraphics[width=\linewidth]
        {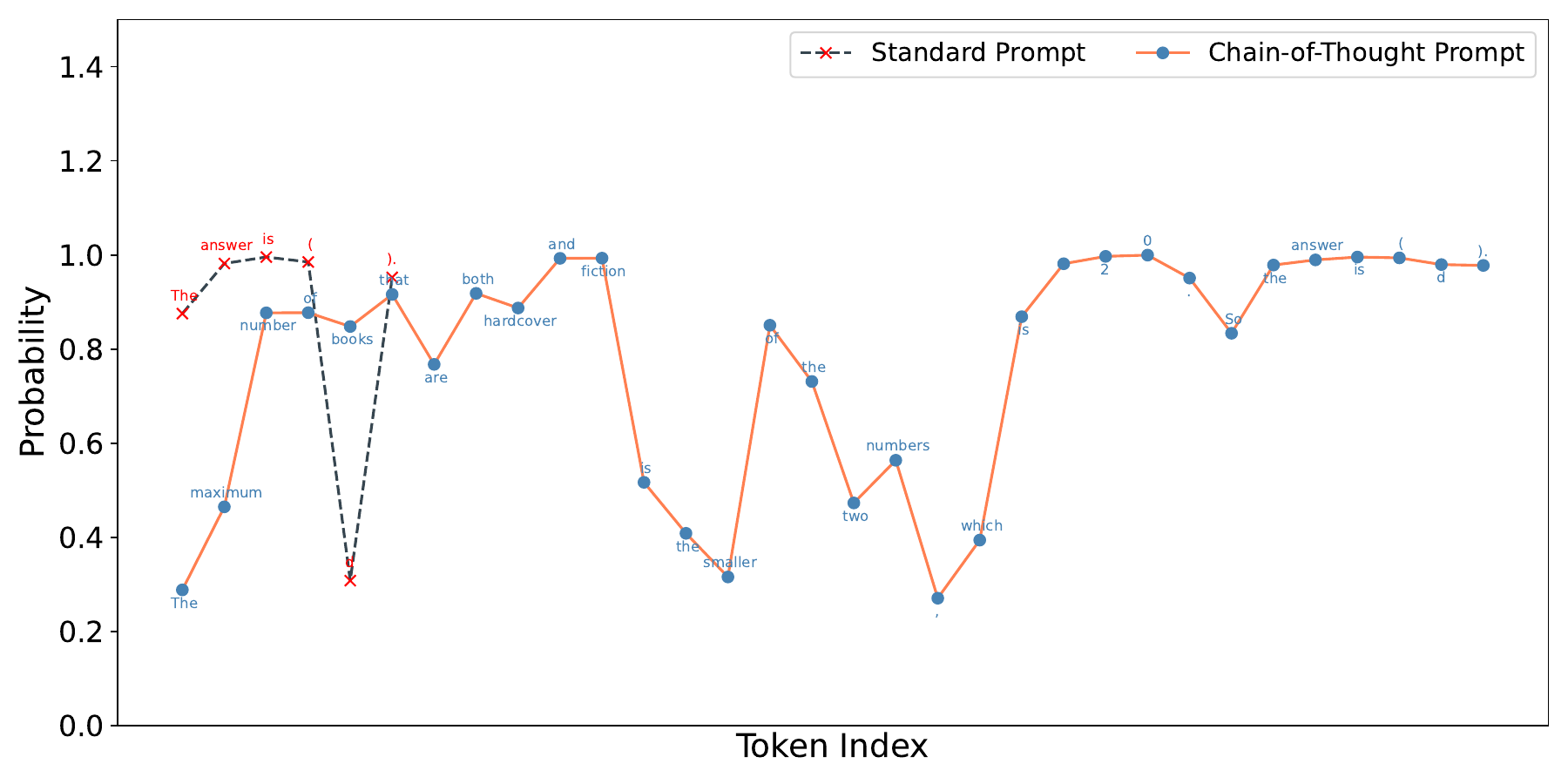}
        \caption{\textit{AQuA}}
    \end{subfigure}
    \begin{subfigure}{0.45\linewidth}
        \includegraphics[width=\linewidth]
        {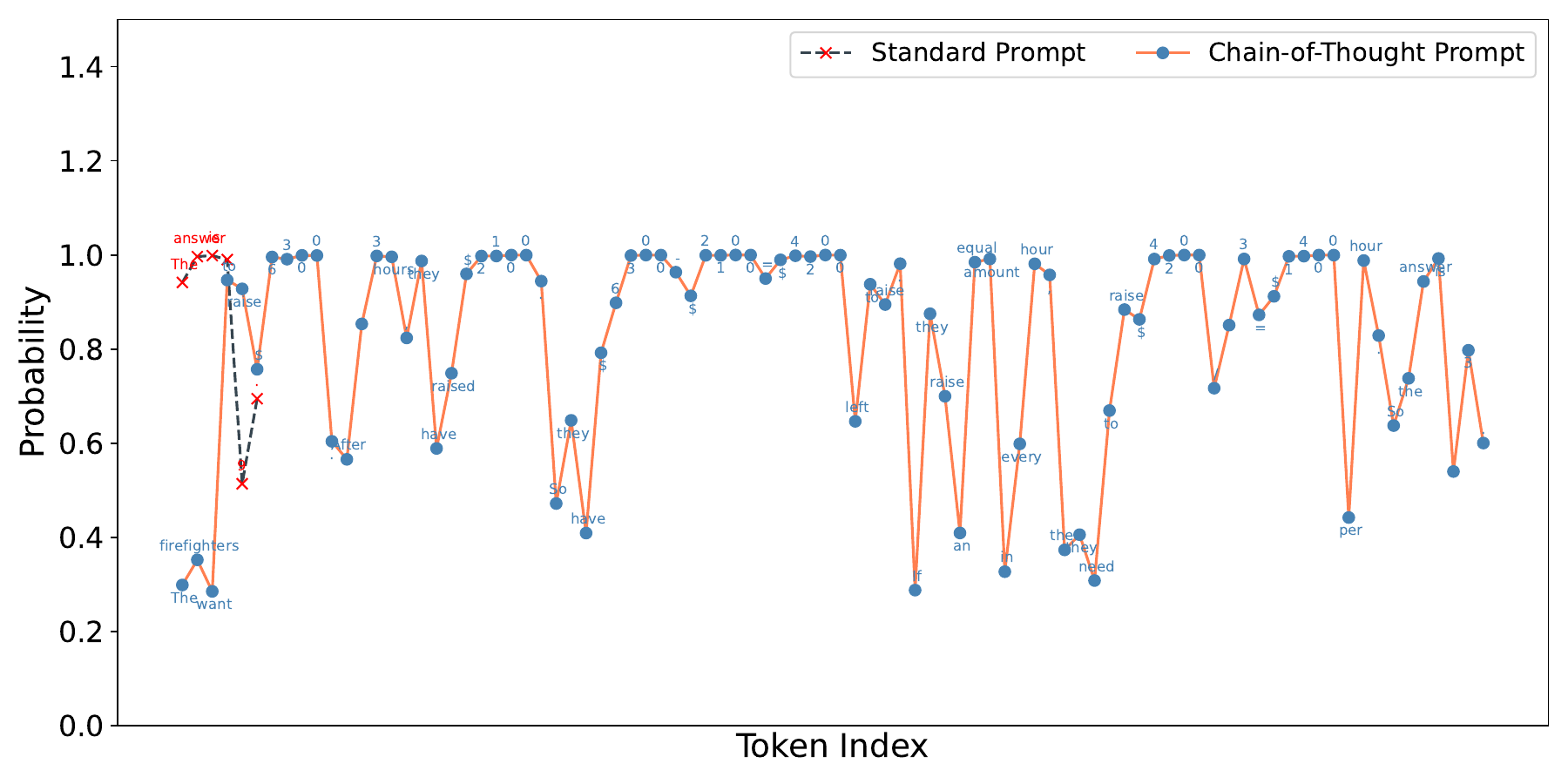}
        \caption{\textit{GSM8K}}
    \end{subfigure}
    \begin{subfigure}{0.45\linewidth}
        \includegraphics[width=\linewidth]{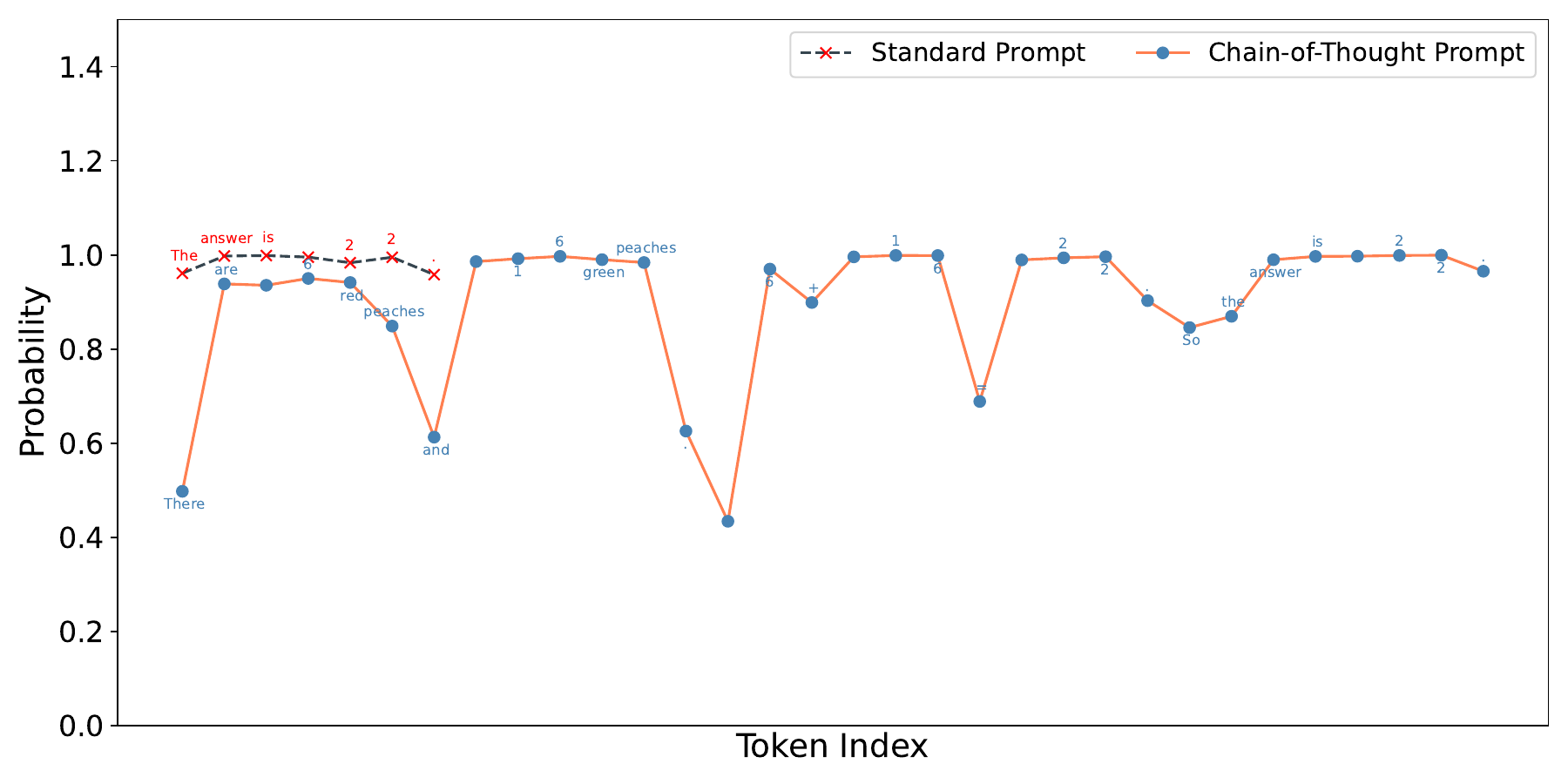}
        \caption{\textit{SVAMP}}
    \end{subfigure}
    \begin{subfigure}{0.45\linewidth}
        \includegraphics[width=\linewidth]
        {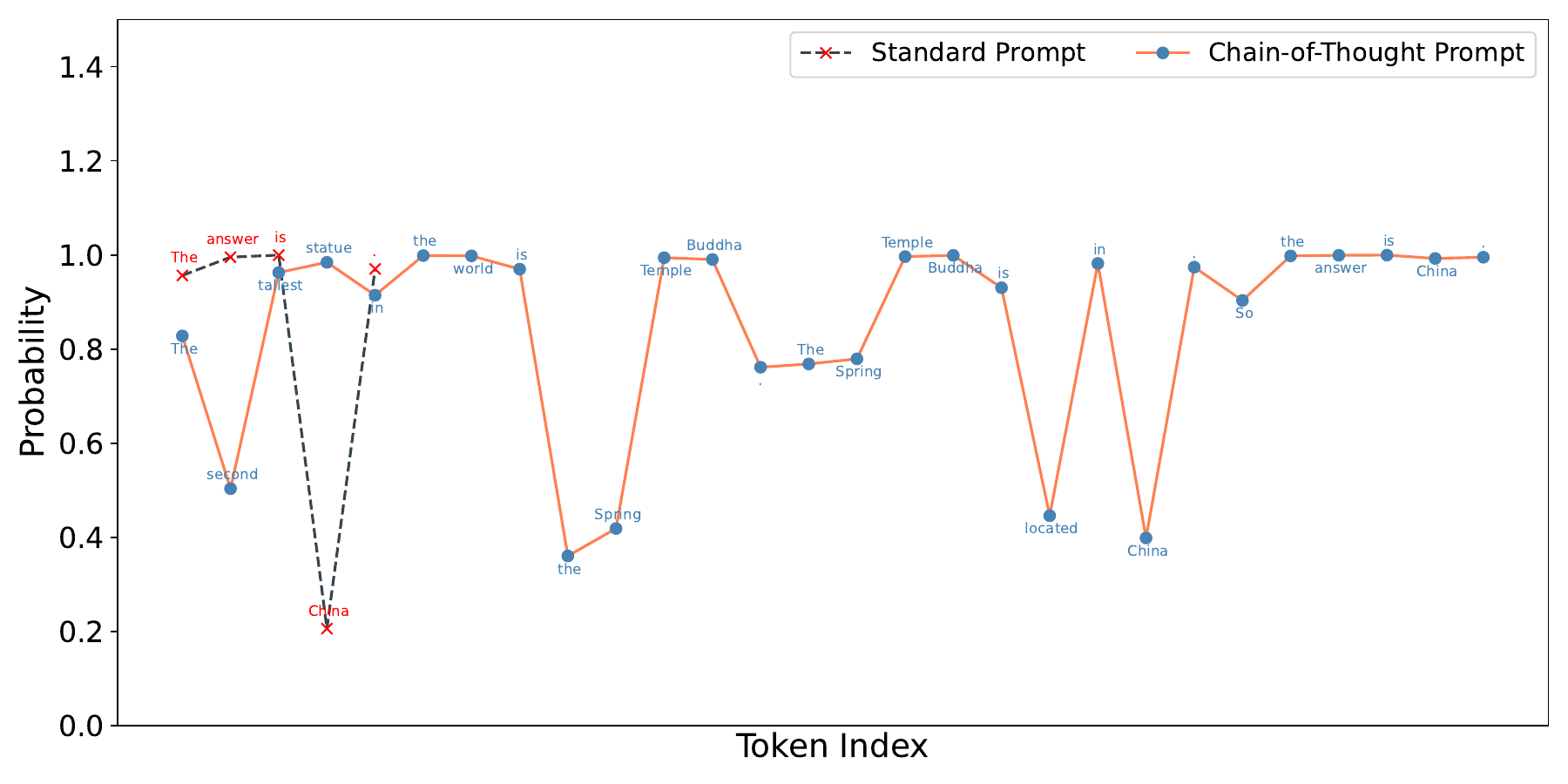}
        \caption{\textit{Bamboogle}}
    \end{subfigure}
    \begin{subfigure}{0.45\linewidth}
        \includegraphics[width=\linewidth]
        {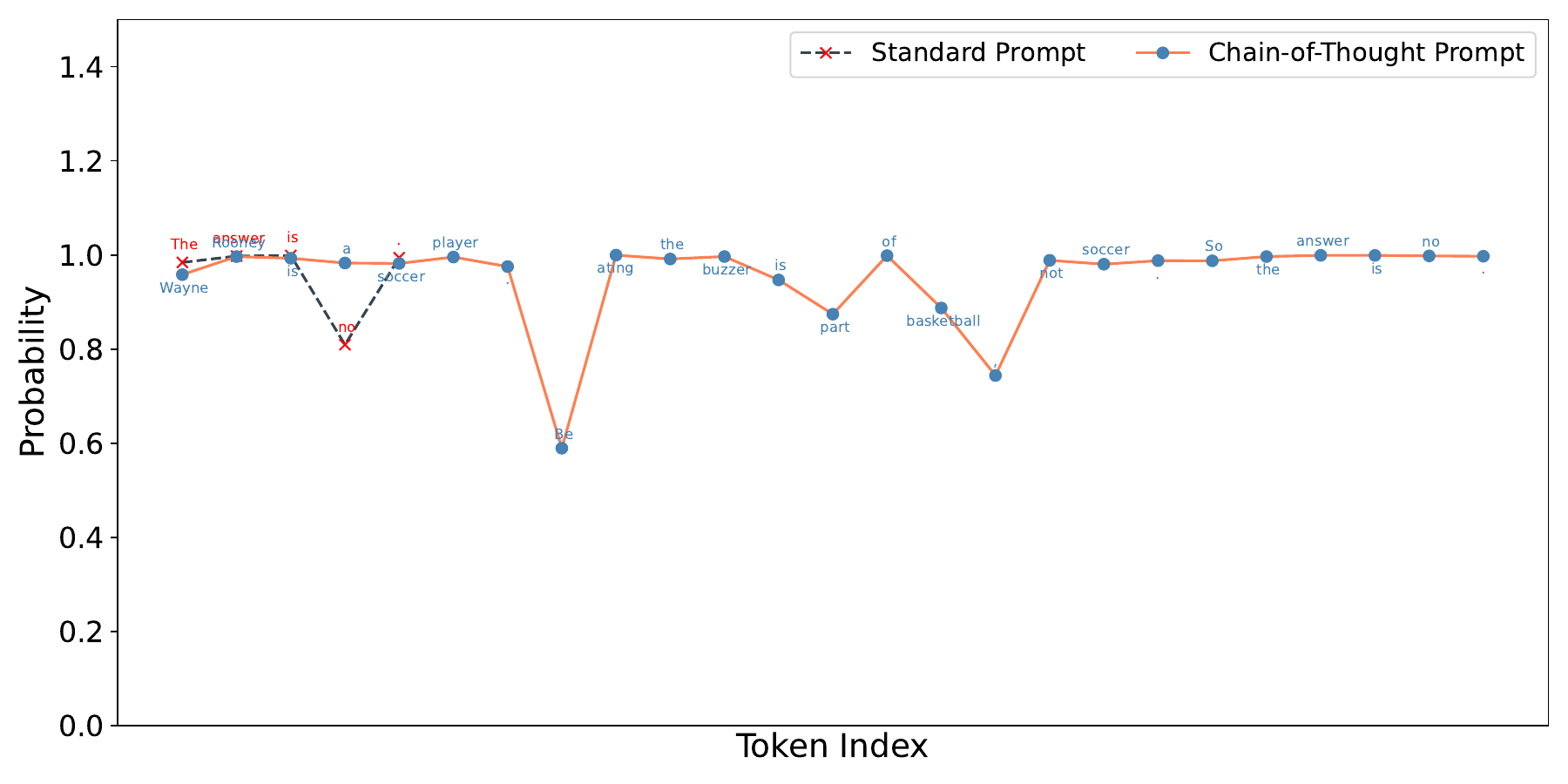}
        \caption{\textit{sports}}
    \end{subfigure}
    \begin{subfigure}{0.45\linewidth}
        \includegraphics[width=\linewidth]
        {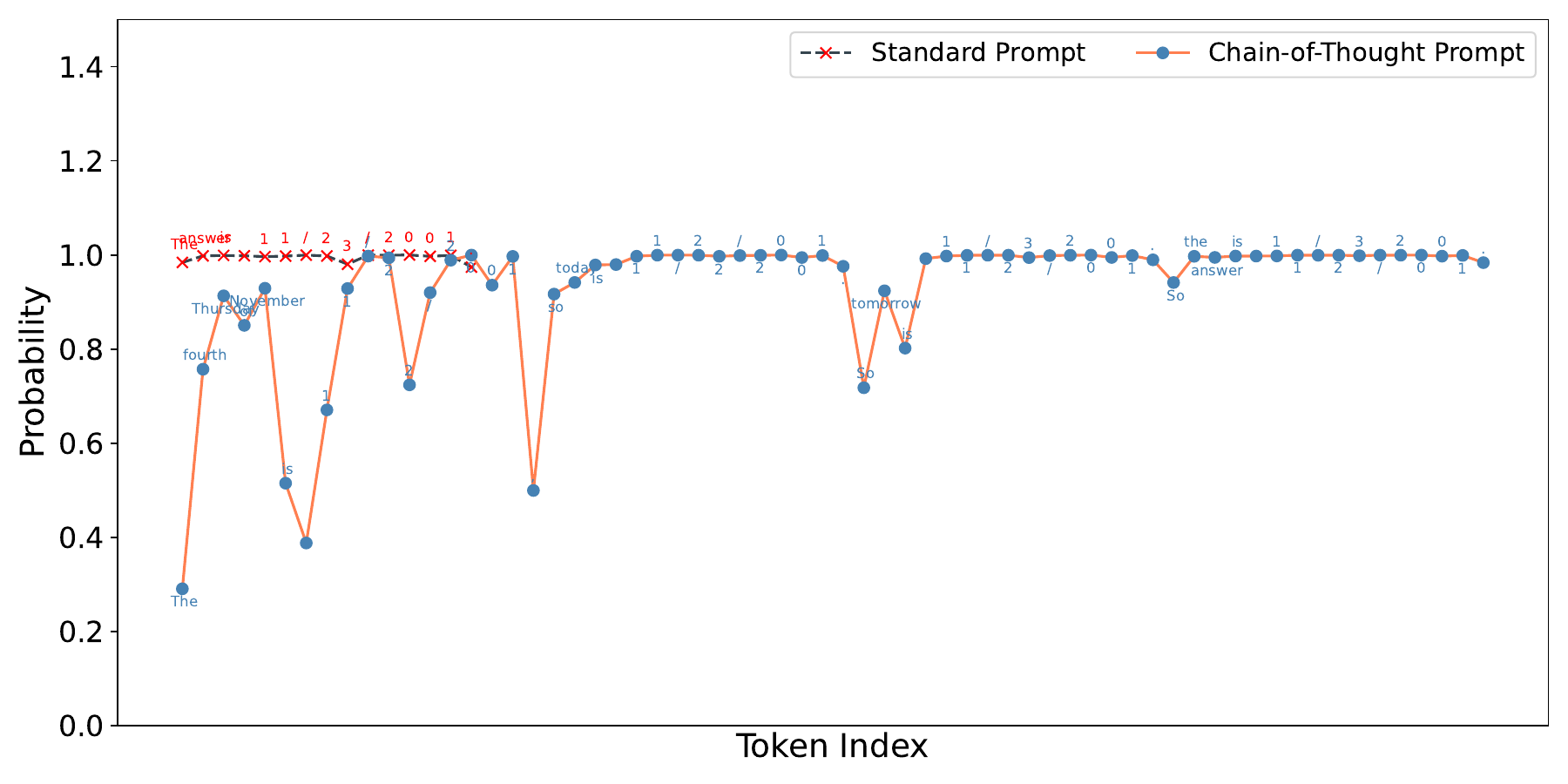}
        \caption{\textit{Date}}
    \end{subfigure}
    \begin{subfigure}{0.45\linewidth}
        \includegraphics[width=\linewidth]
        {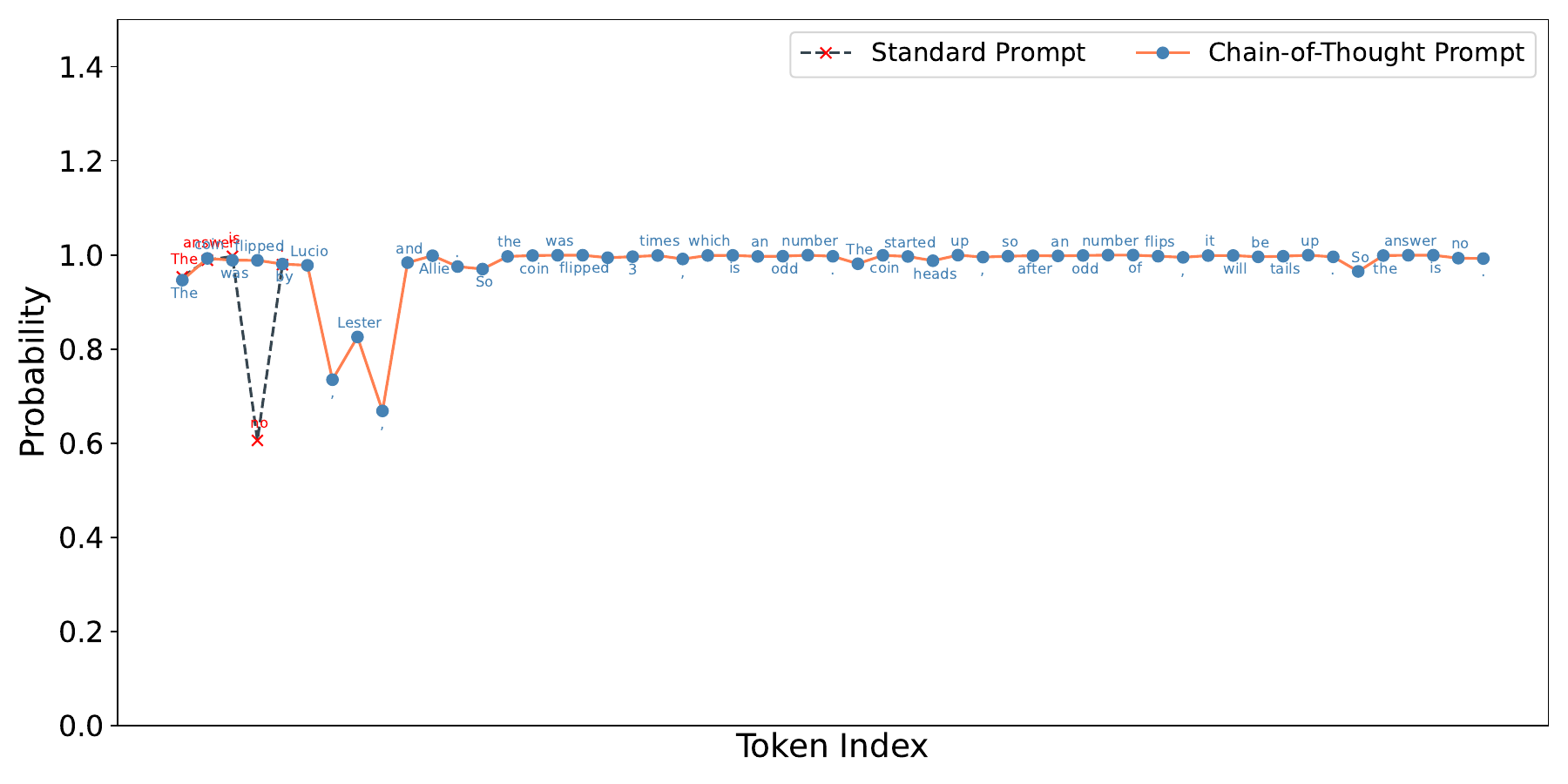}
        \caption{\textit{Coin Flip}}
    \end{subfigure}
     \begin{subfigure}{0.45\linewidth}
        \includegraphics[width=\linewidth]
        {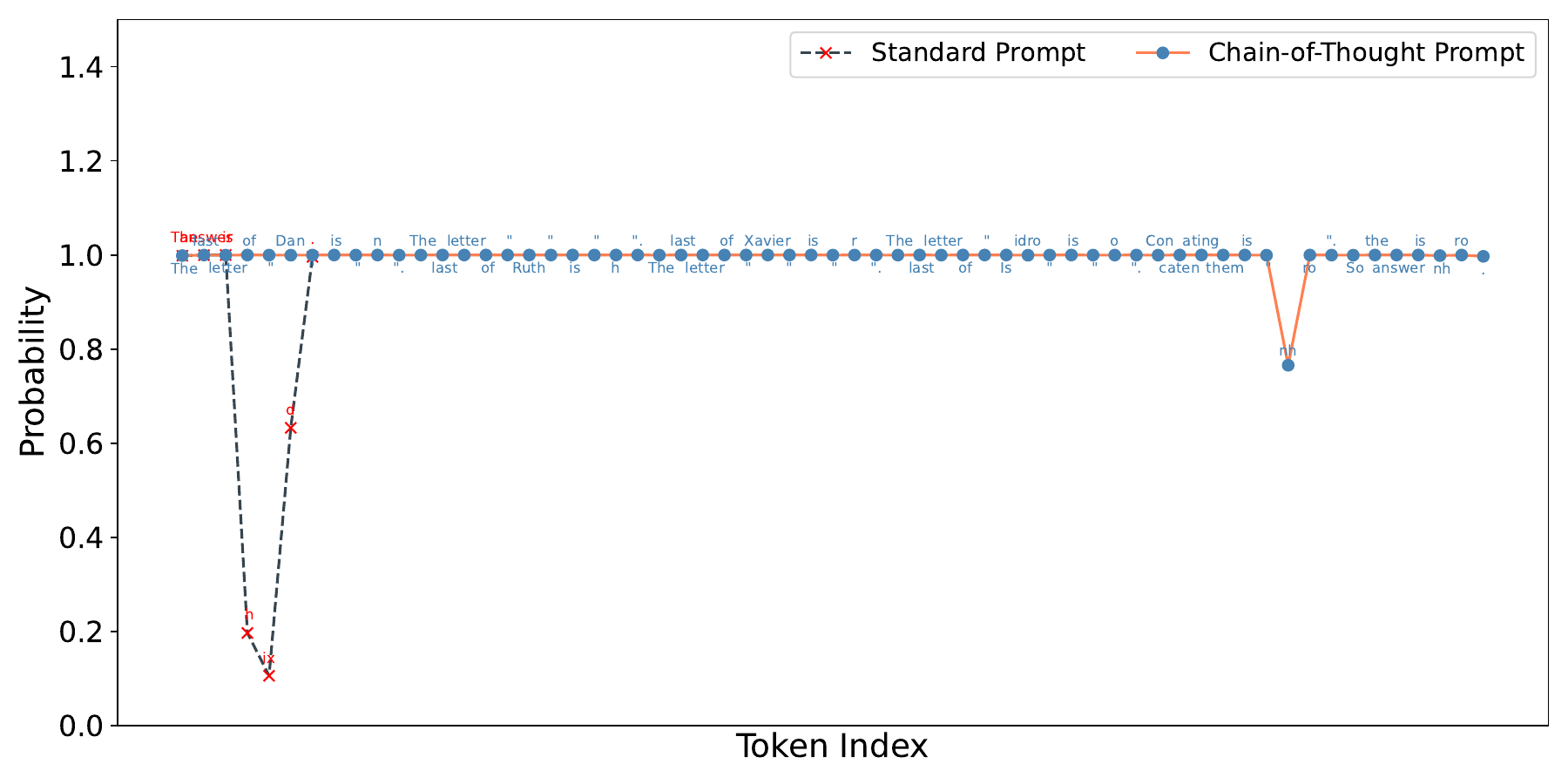}
        \caption{\textit{Last Letter Concatenation}}
    \end{subfigure}
\caption{Probability value of each generated token (the results of Gemma2-27b is shown).}
\label{fig:question2_logits_value_27b}
\end{figure*}

We conducted experiments on all test datasets across various models, obtaining the token sequences and their corresponding generation probability sequences under both CoT and Standard prompting methods. Figure \ref{fig:question2_logits_value_2b} presents the test results for the Gemma2-2b model, Figure \ref{fig:question2_logits_value_13b} shows the results for the LLaMA2-13b model, and Figure \ref{fig:question2_logits_value_27b} illustrates the results for the Gemma2-27b model. The experimental results indicate that the probability trends for token generation under both CoT and Standard prompts are consistent across models of different sizes and series.
\onecolumn
\clearpage

\twocolumn
\section{Kernel Density Estimate}\label{appendix:density}
\begin{figure*}[ht]
\centering
    \includegraphics[width=0.24\linewidth]{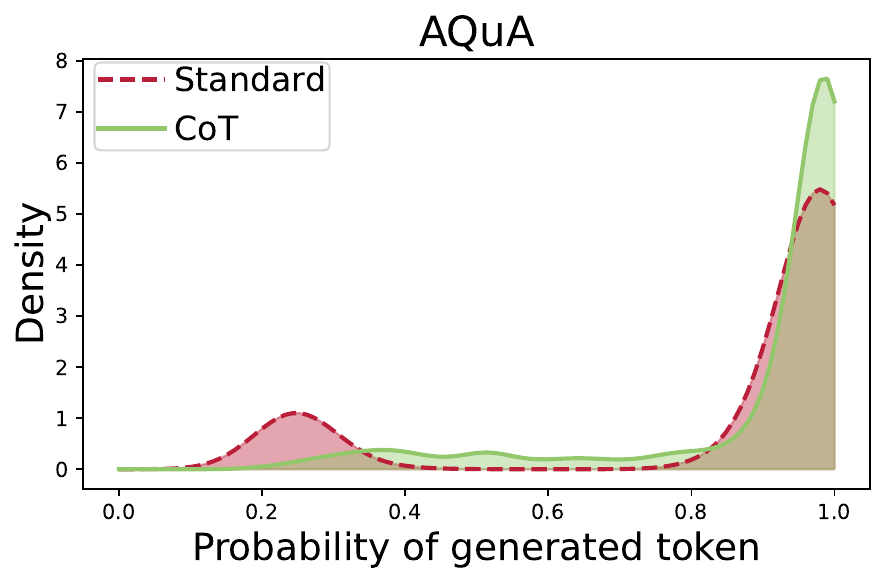}
    \includegraphics[width=0.24\linewidth]{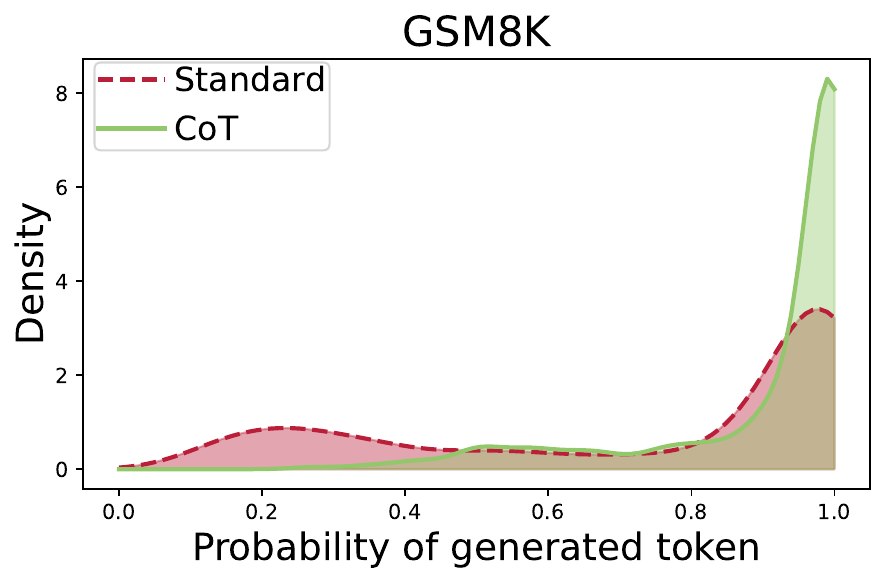}
    \includegraphics[width=0.24\linewidth]{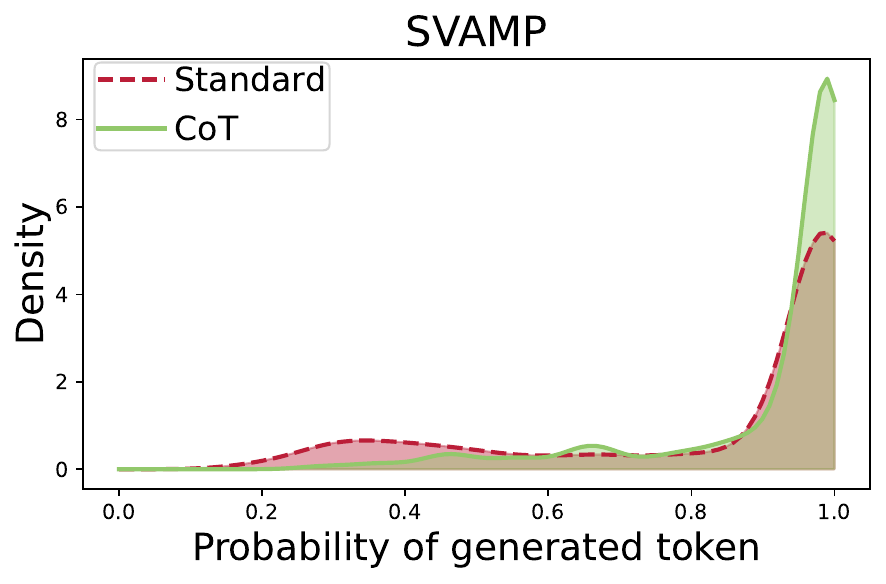}
    \includegraphics[width=0.24\linewidth]{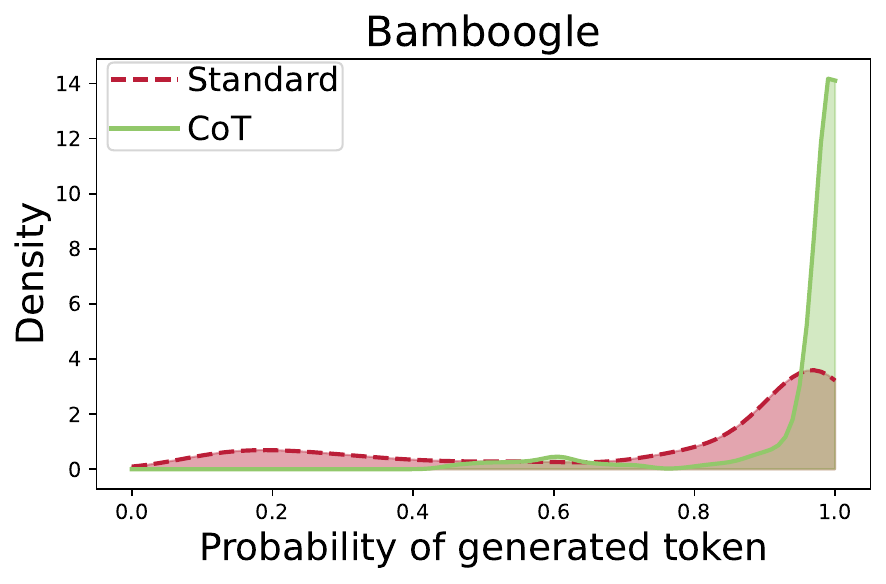}
    \includegraphics[width=0.24\linewidth]{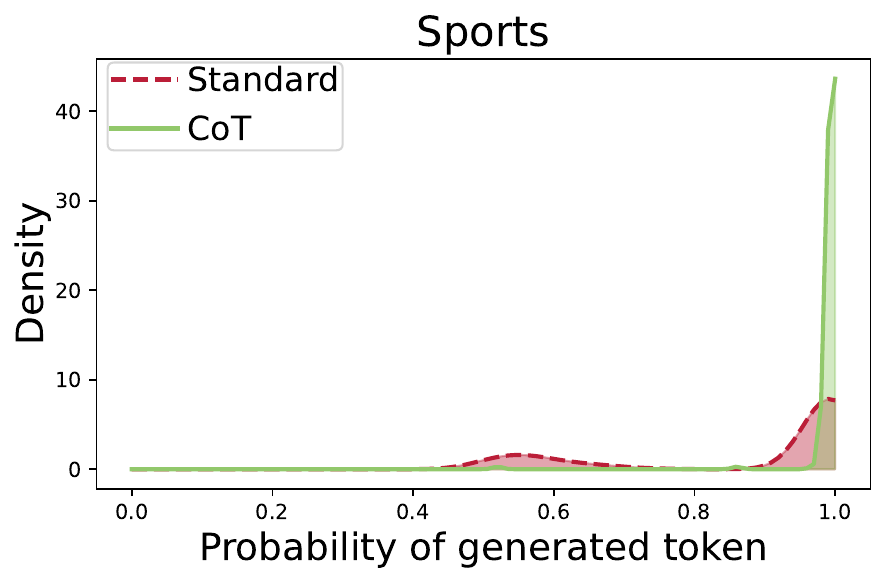}
    \includegraphics[width=0.24\linewidth]{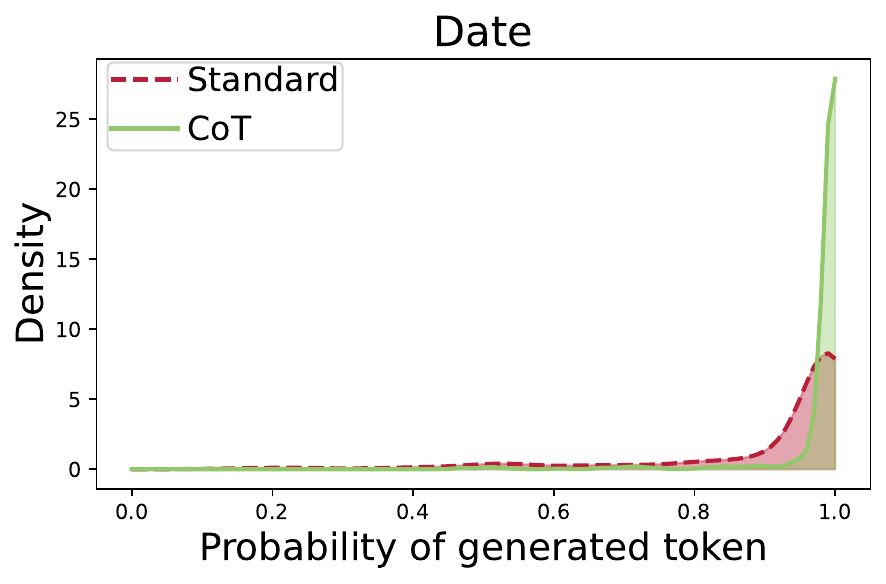}
     \includegraphics[width=0.24\linewidth]{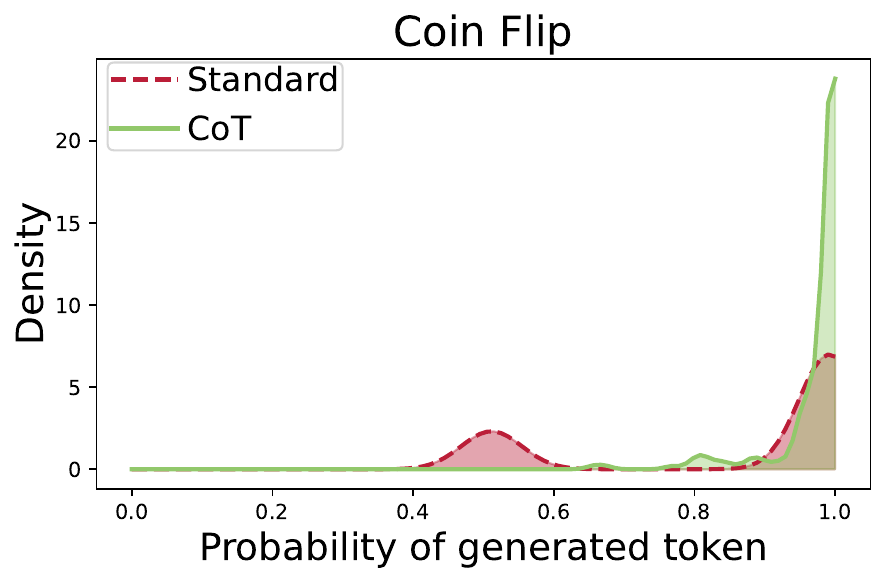}
    \includegraphics[width=0.24\linewidth]{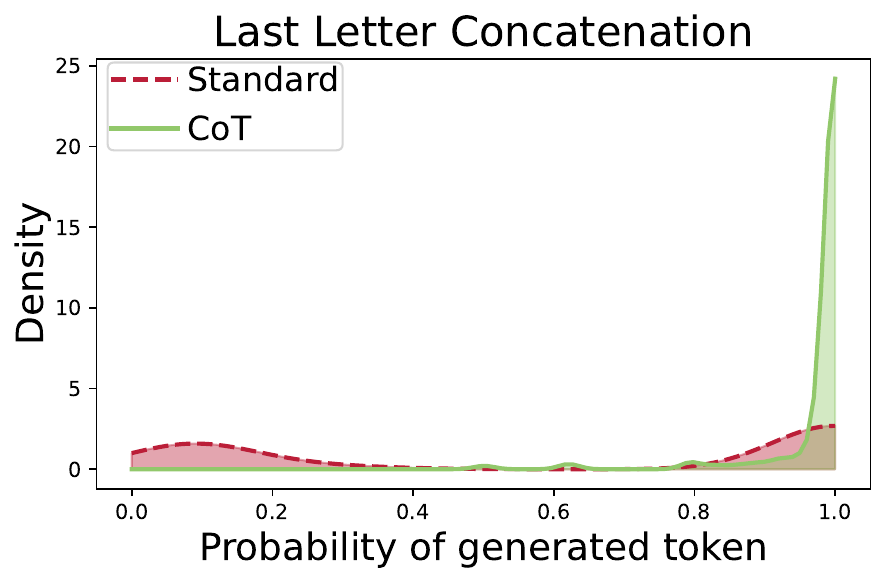}
    \vspace{-4mm}
    \caption{Kernal density of Gemma2-2B.}
    \label{fig:question2_logits_value_kernel_2b}
\end{figure*}

\begin{figure*}[ht]
\centering
    \includegraphics[width=0.24\linewidth]{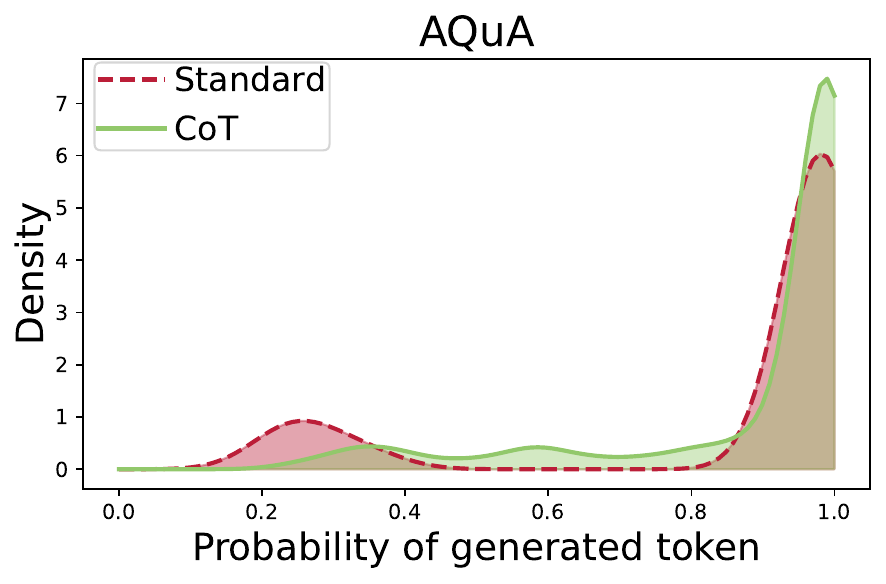}
    \includegraphics[width=0.24\linewidth]{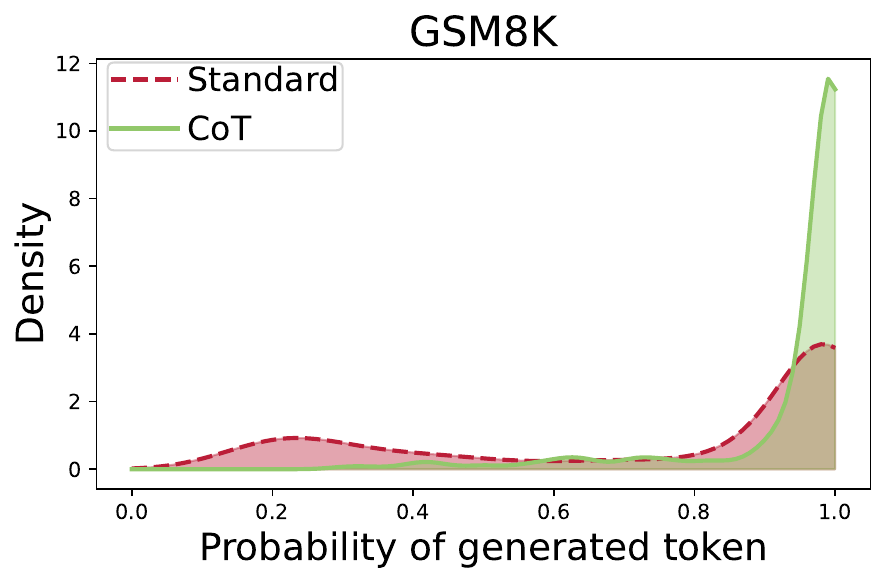}
    \includegraphics[width=0.24\linewidth]{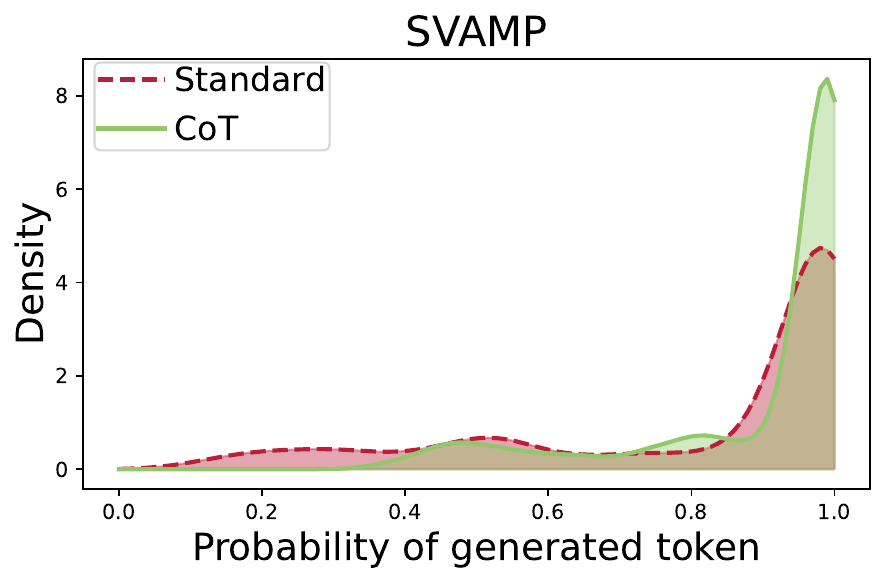}
    \includegraphics[width=0.24\linewidth]{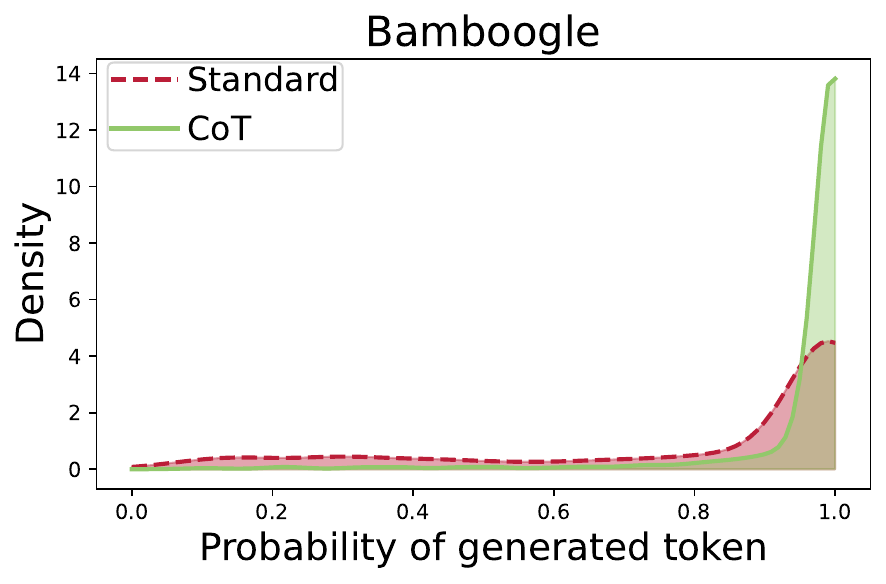}
    \includegraphics[width=0.24\linewidth]{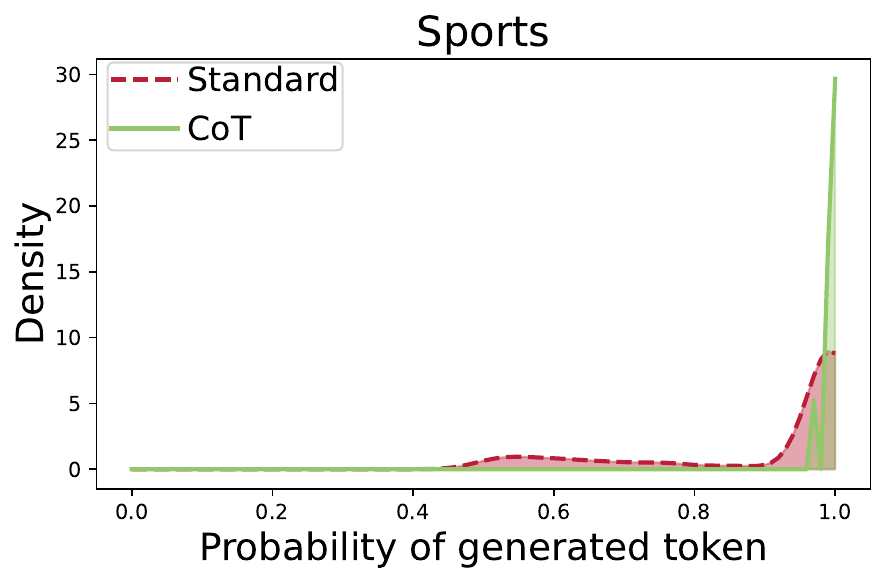}
    \includegraphics[width=0.24\linewidth]{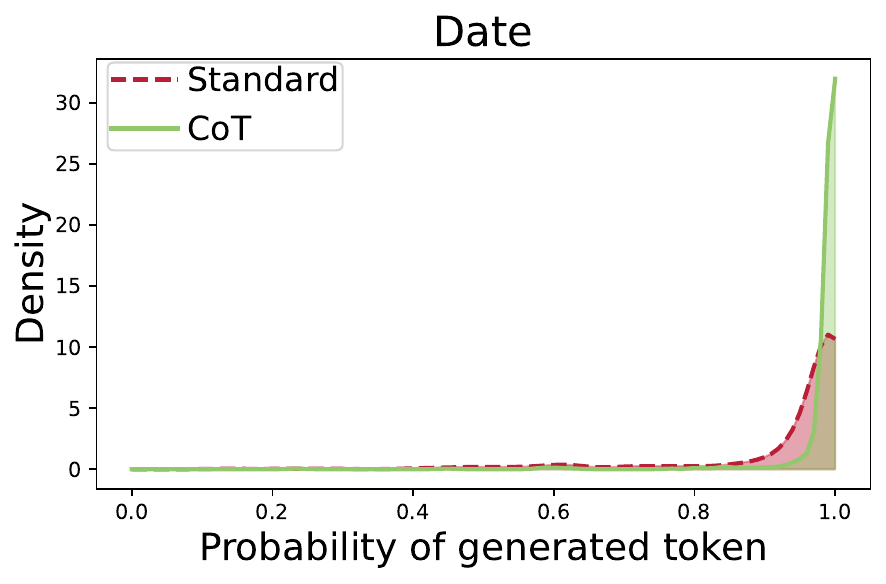}
     \includegraphics[width=0.24\linewidth]{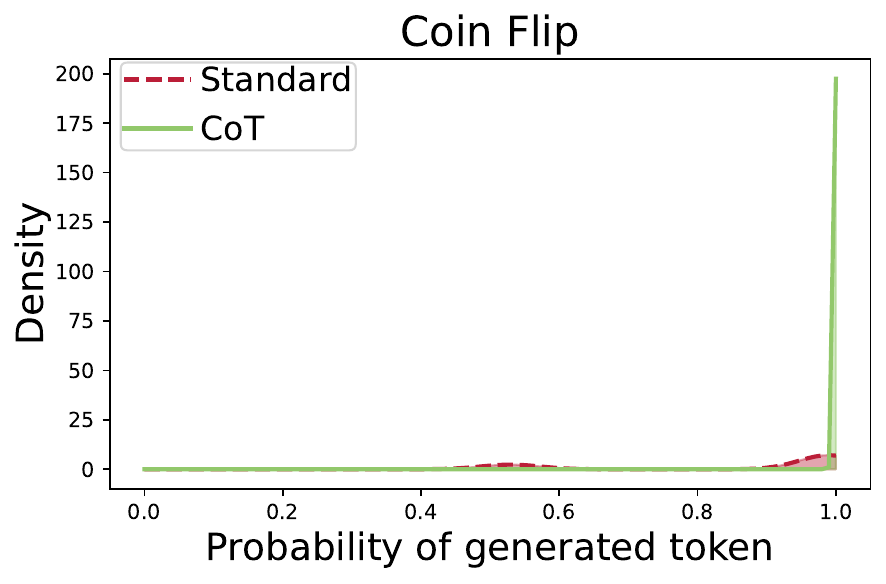}
    \includegraphics[width=0.24\linewidth]{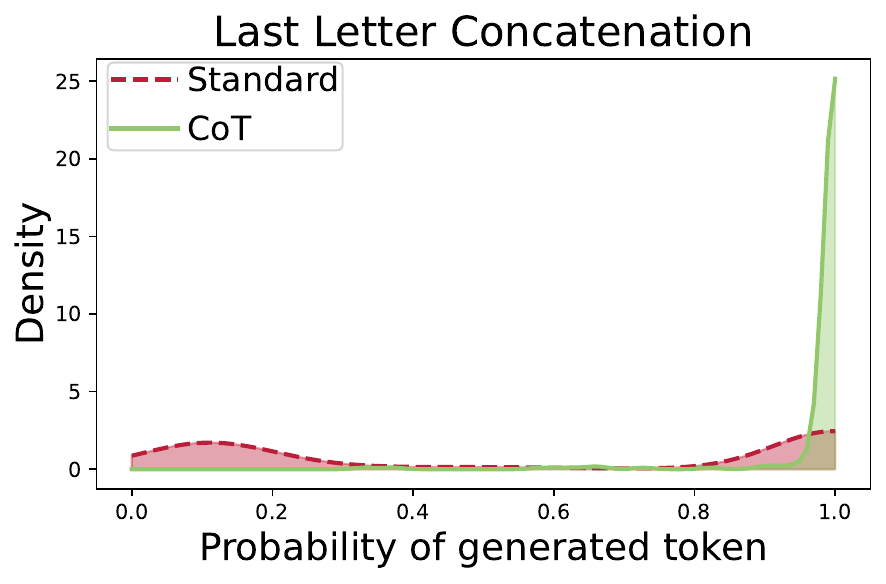}
    \vspace{-4mm}
    \caption{Kernal density of LLaMA2-13B.}
    \label{fig:question2_logits_value_kernel_13b}
\end{figure*}

\begin{figure*}[ht]
\centering
    \includegraphics[width=0.24\linewidth]{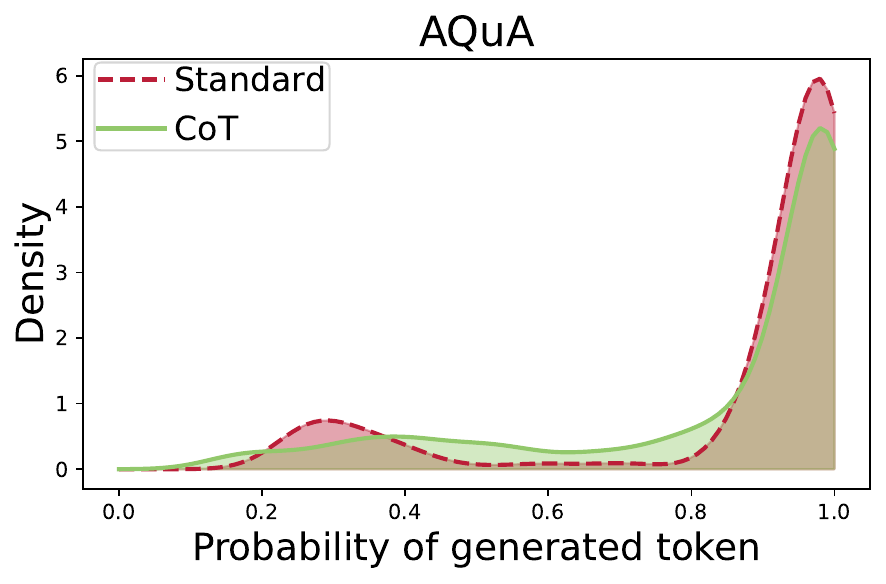}
    \includegraphics[width=0.24\linewidth]{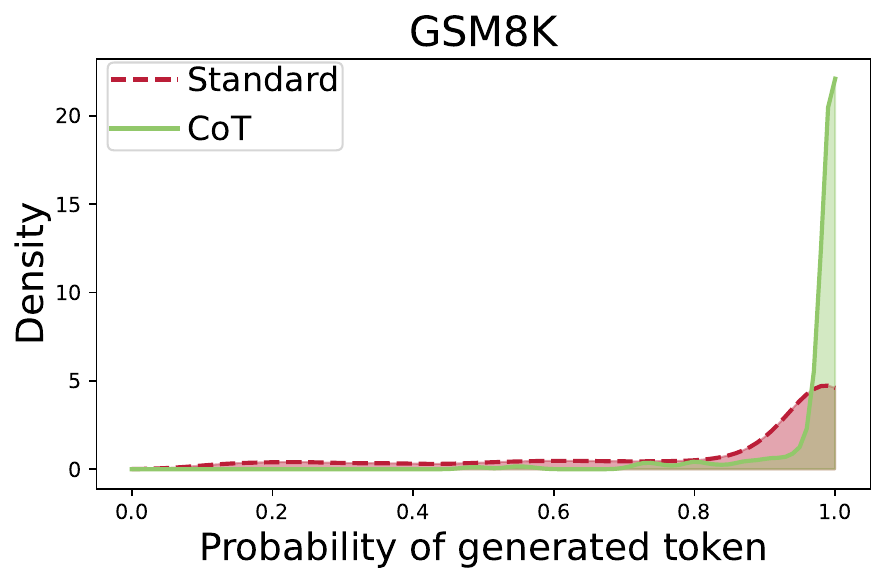}
    \includegraphics[width=0.24\linewidth]{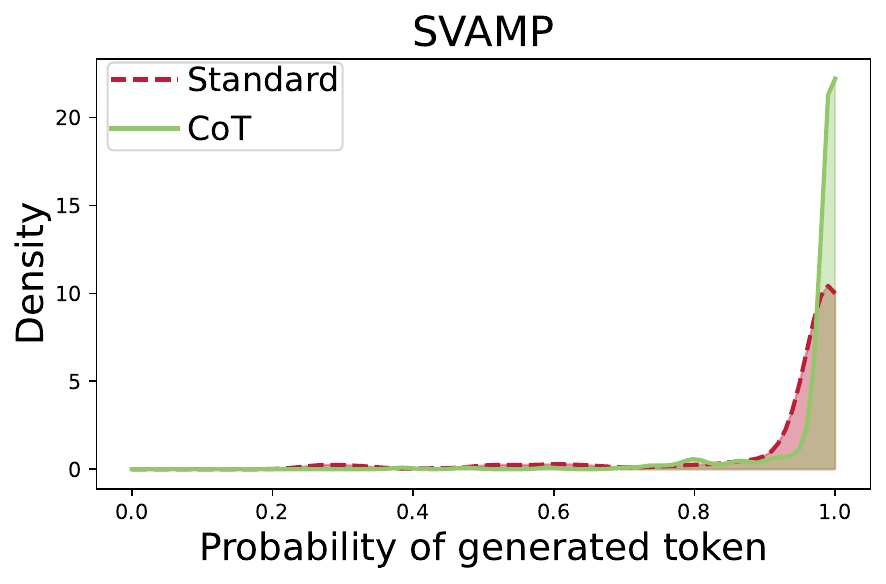}
    \includegraphics[width=0.24\linewidth]{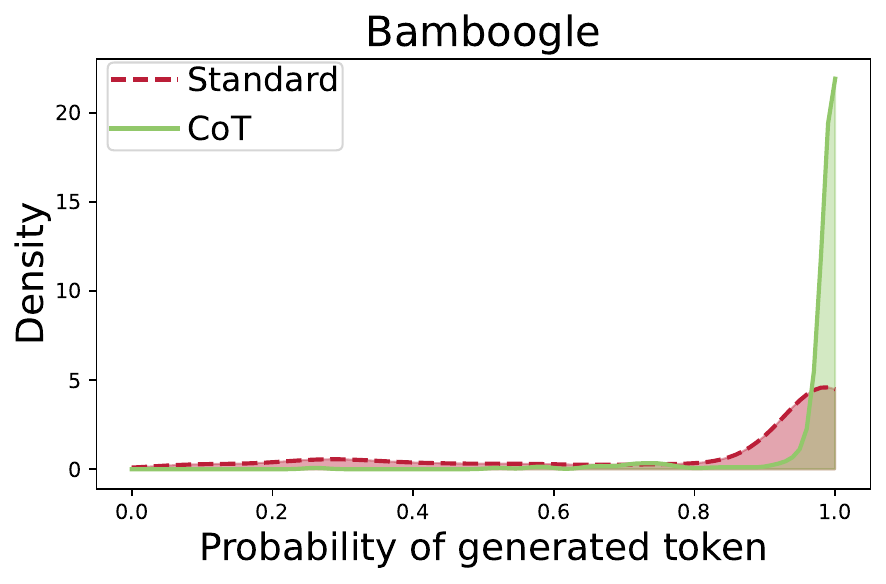}
    \includegraphics[width=0.24\linewidth]{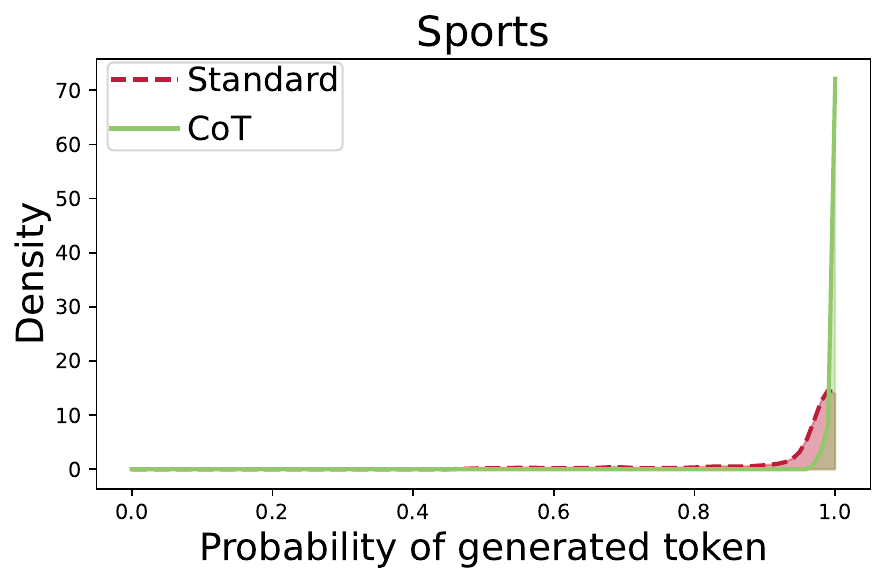}
    \includegraphics[width=0.24\linewidth]{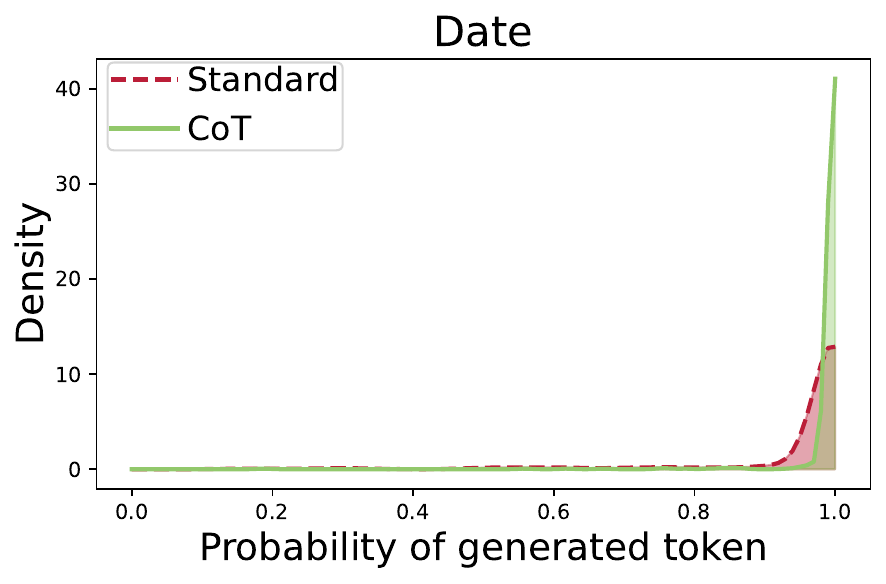}
     \includegraphics[width=0.24\linewidth]{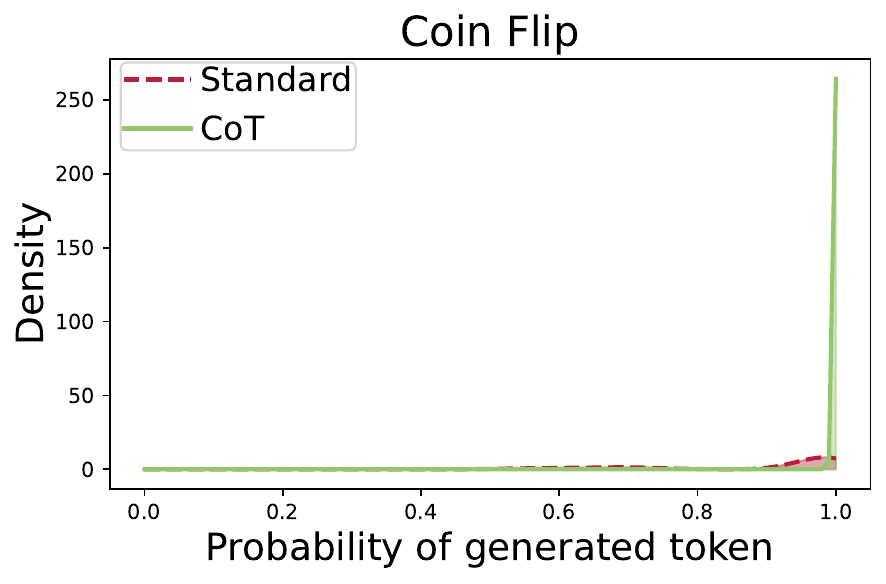}
    \includegraphics[width=0.24\linewidth]{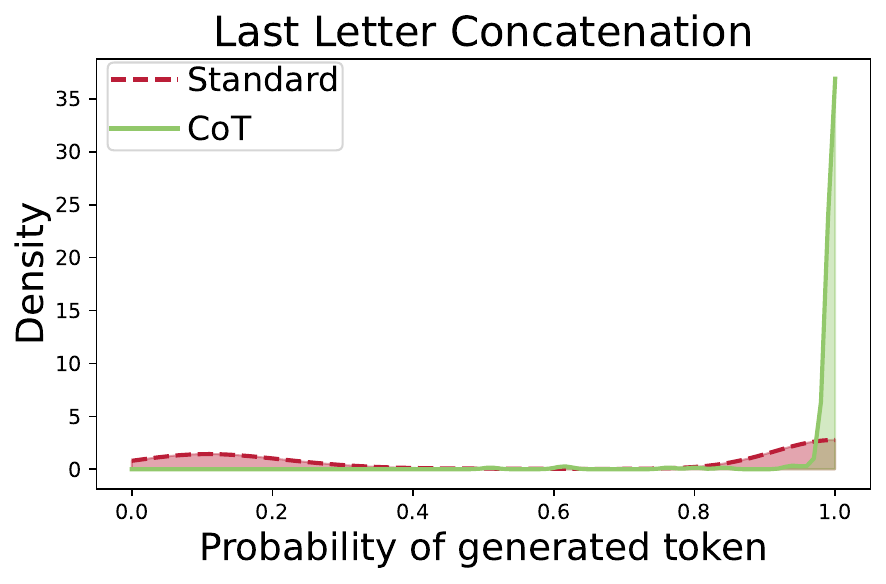}
    \vspace{-4mm}
    \caption{Kernal density  of Gemma2-27B.}
    \label{fig:question2_logits_value_kernel_27b}
\end{figure*}

Kernel Density Estimation (KDE) is a non-parametric method for estimating the probability density function (PDF) of a dataset. It provides a smooth, continuous approximation of the underlying data distribution by summing localized kernel functions centered at each observed data point. The mathematical formulation of KDE is given by Equation \ref{eq:kde}:
\begin{equation}
\label{eq:kde}
\hat{f}(x) = \frac{1}{n h} \sum_{i=1}^{n} K\left( \frac{x - x_i}{h} \right),
\end{equation}
where $\hat{f}(x)$ is the estimated density at point $x$, $n$ is the number of data points, $h$ is the bandwidth (a smoothing parameter), $x_i$ represents the observed data points, and $K(\cdot)$ is a kernel function. In this work, we employed the standard Gaussian kernel, defined as shown in Equation \ref{eq:gaussian}:
\begin{equation}
\label{eq:gaussian}
K(x) = \frac{1}{\sqrt{2\pi}} e^{-\frac{x^2}{2}}.
\end{equation}
Given that our data are confined to a bounded range [0, 1], the standard Gaussian kernel was applied, and the resulting density estimates are plotted within this domain. We conducted experiments using four different models: Gemma2-2b, Gemma2-9b, LLaMA2-13b, and Gemma2-27b. The KDE plots illustrating the experimental results for these models are presented in Figure \ref{fig:question2_logits_value_kernel_2b} (Gemma2-2b), Figure \ref{fig:question2_horizontal_2} (Gemma2-9b), Figure \ref{fig:question2_logits_value_kernel_13b} (LLaMA2-13b), and Figure \ref{fig:question2_logits_value_kernel_27b} (Gemma2-27b).
\onecolumn
\vspace{-8mm}
\vspace{-8mm}
\clearpage

\twocolumn
\section{Entropy of  Probability Distribution}\label{appendix:entropy}
\begin{figure*}[ht]
\centering
    \begin{subfigure}{0.32\linewidth}
        \includegraphics[width=\linewidth]{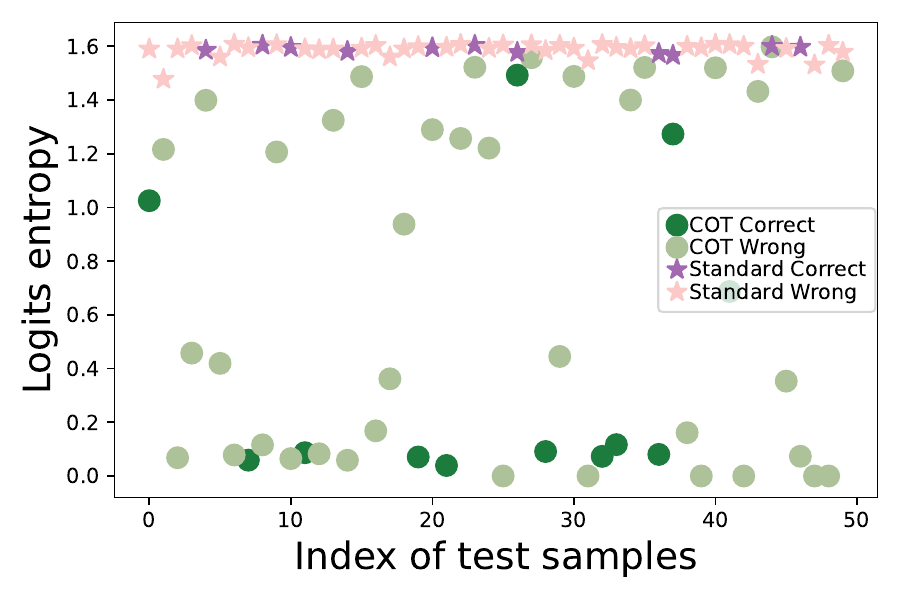}
        \caption{\textit{AQuA}}
    \end{subfigure}
    \begin{subfigure}{0.32\linewidth}
        \includegraphics[width=\linewidth]{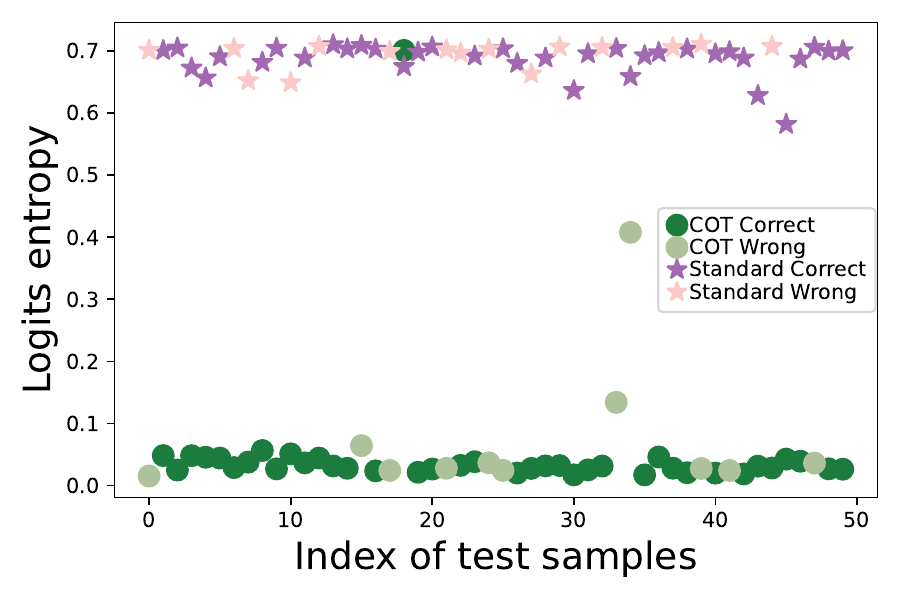}
        \caption{\textit{Sports}}
    \end{subfigure}
    \begin{subfigure}{0.32\linewidth}
        \includegraphics[width=\linewidth]{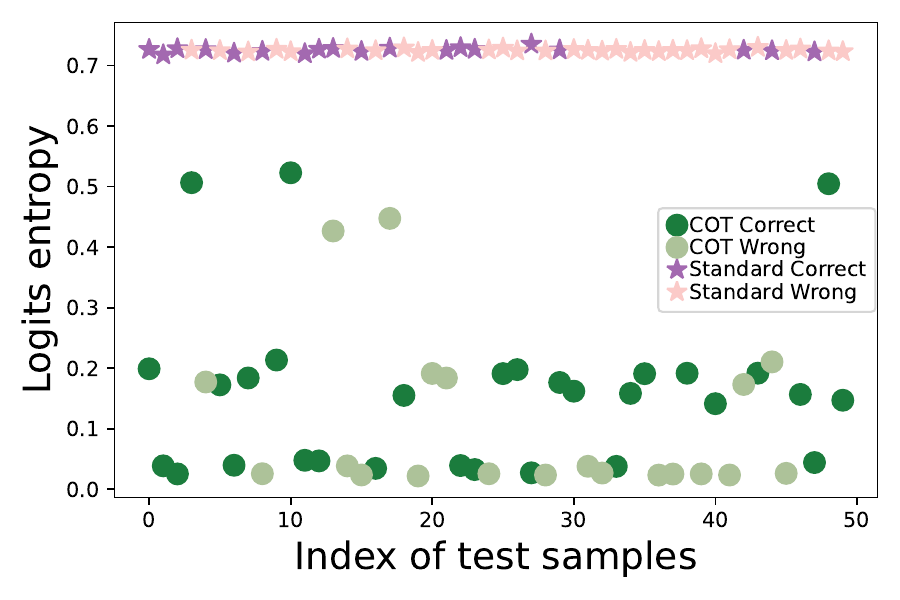}
        \caption{\textit{Coin Flip}}
    \end{subfigure}
\caption{Results of Gemma2-2b.}
\label{fig:question2_vertical_entropy_2b}
\end{figure*}

\begin{figure*}[ht]
\centering
    \begin{subfigure}{0.32\linewidth}
        \includegraphics[width=\linewidth]{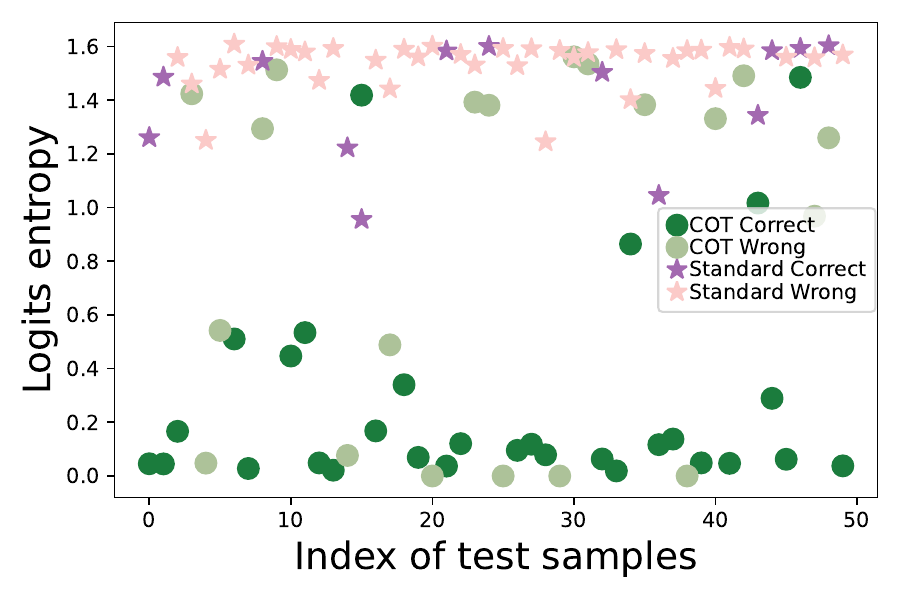}
        \caption{\textit{AQuA}}
    \end{subfigure}
    \begin{subfigure}{0.32\linewidth}
        \includegraphics[width=\linewidth]{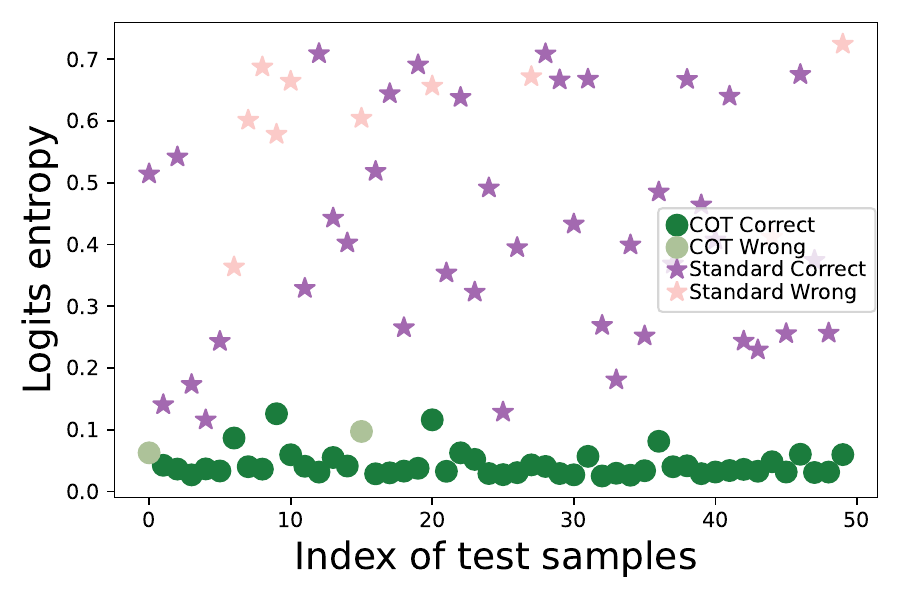}
        \caption{\textit{Sports}}
    \end{subfigure}
    \begin{subfigure}{0.32\linewidth}
        \includegraphics[width=\linewidth]{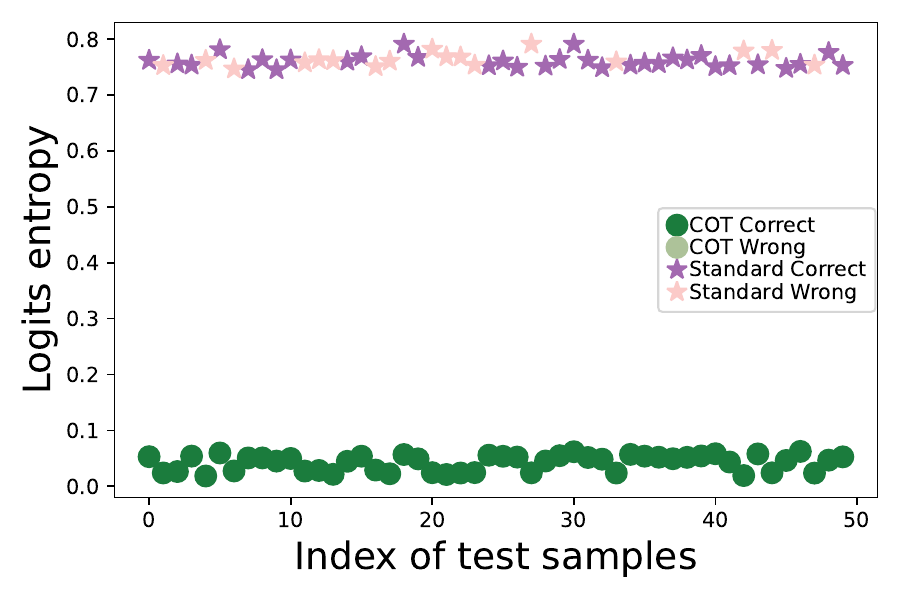}
        \caption{\textit{Coin Flip}}
    \end{subfigure}
\caption{Results of Gemma2-9b.}
\label{question2_vertical_entropy_9b}
\end{figure*}

\begin{figure*}[ht]
\centering
    \begin{subfigure}{0.32\linewidth}
        \includegraphics[width=\linewidth]{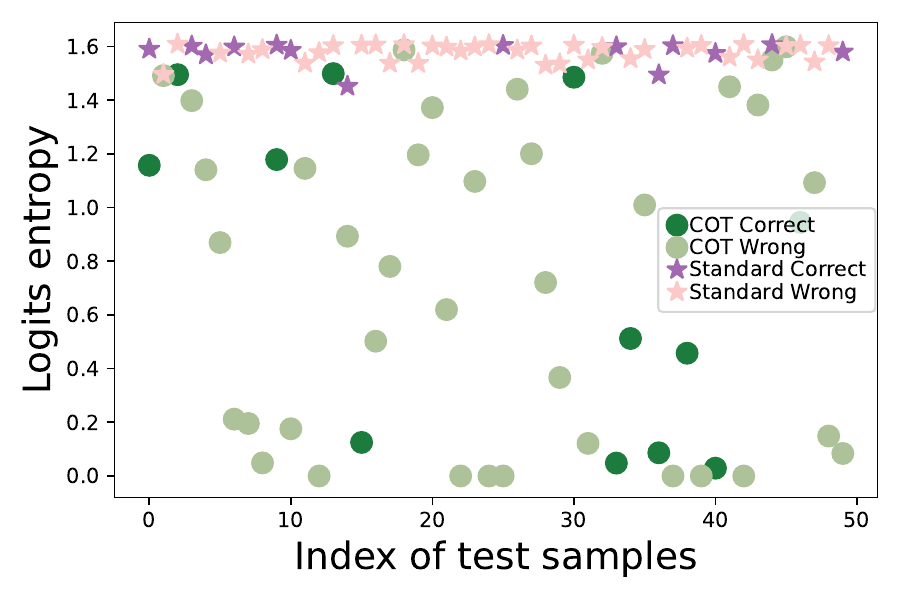}
        \caption{\textit{AQuA}}
    \end{subfigure}
    \begin{subfigure}{0.32\linewidth}
        \includegraphics[width=\linewidth]{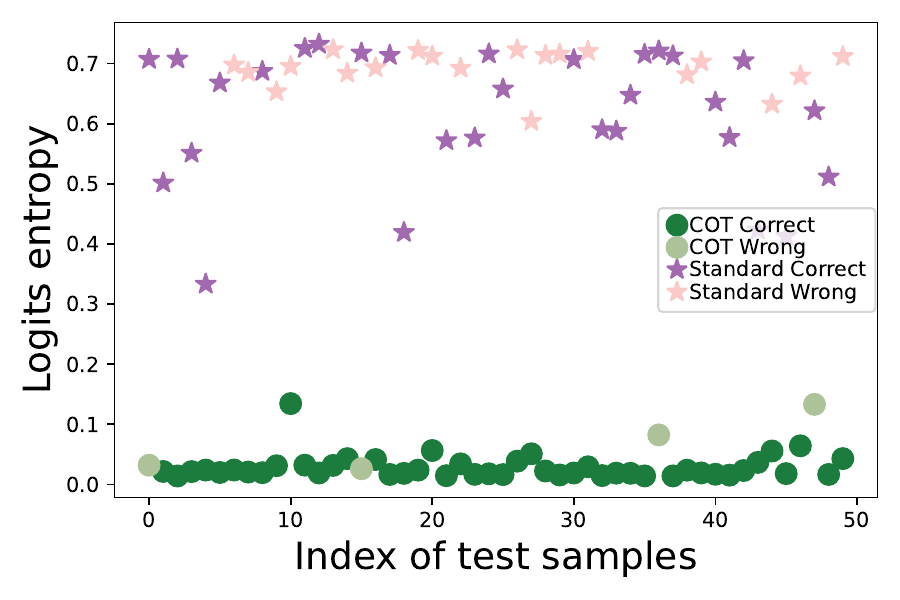}
        \caption{\textit{Sports}}
    \end{subfigure}
    \begin{subfigure}{0.32\linewidth}
        \includegraphics[width=\linewidth]{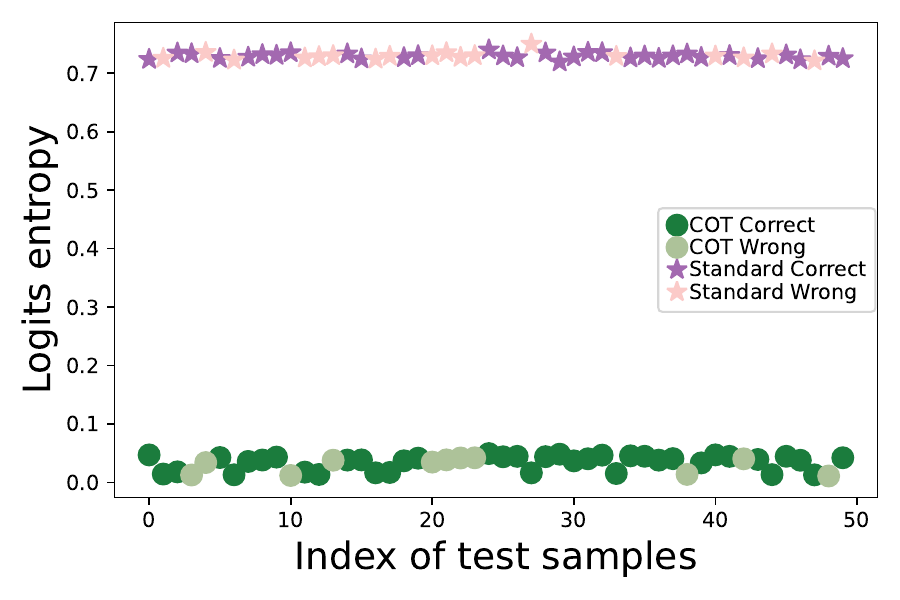}
        \caption{\textit{Coin Flip}}
    \end{subfigure}
\caption{Results of LLaMA2-13b.}
\label{fig:question2_vertical_entropy_13b}
\end{figure*}

To investigate the model's confidence when predicting answers, we examined the probability distribution over candidate answers for the token generated at the answer prediction step. We focused on datasets AQuA, Sports, and Coin Flip, where the number of answer tokens is 1, and the answer space is finite. For datasets with multiple answer tokens and open-domain questions, it is not feasible to retrieve all potential answers and their corresponding probabilities.
Surprisingly, for the AQuA dataset, the top five probabilities at the answer prediction step corresponded exactly to the answer space options: ``a '', ``b '', ``c '', ``d '', and ``e''. Similarly, for the Sports and Coin Flip datasets, the top two probabilities corresponded to their answer spaces, ``yes'' and ``no''.
We selected the top k probabilities at the answer prediction step (where k is the size of the answer space), normalized them, calculated the entropy, and plotted a scatter plot to compare the entropy of answer probabilities generated using the Standard and CoT methods. Figure \ref{fig:question2_vertical_entropy_2b} shows the experimental results for Gemma2-2b, Figure \ref{fig:question2_vertical_entropy_13b} for LLaMA2-13b, and Figure \ref{question2_vertical_entropy_9b} for Gemma2-9b.
\onecolumn

\clearpage
\section{Activation}\label{appendix:activation_difference}

\begin{figure*}[!ht]
\centering
     \includegraphics[width=0.99\linewidth]{figures/activation/GSM8K.pdf}
     \includegraphics[width=0.99\linewidth]{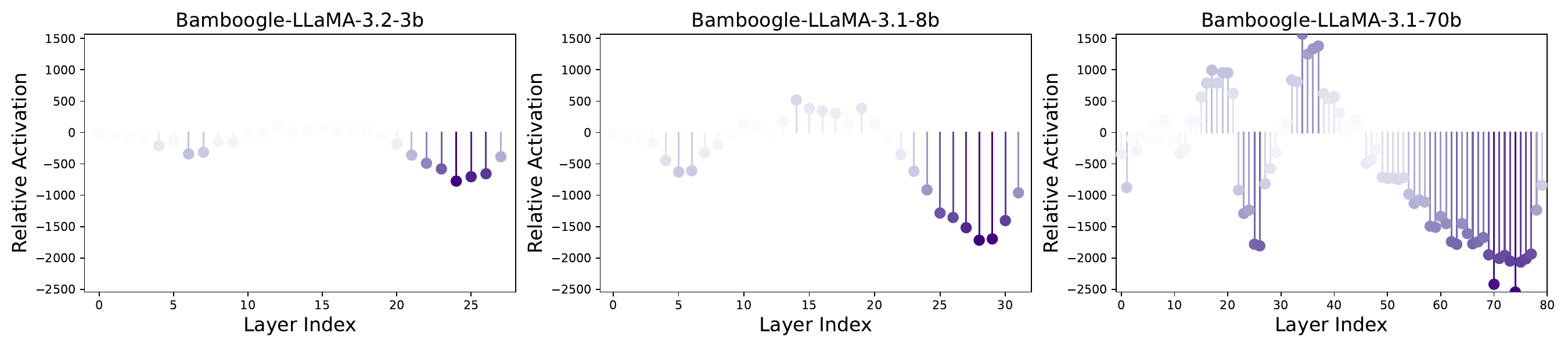}
     \includegraphics[width=0.99\linewidth]{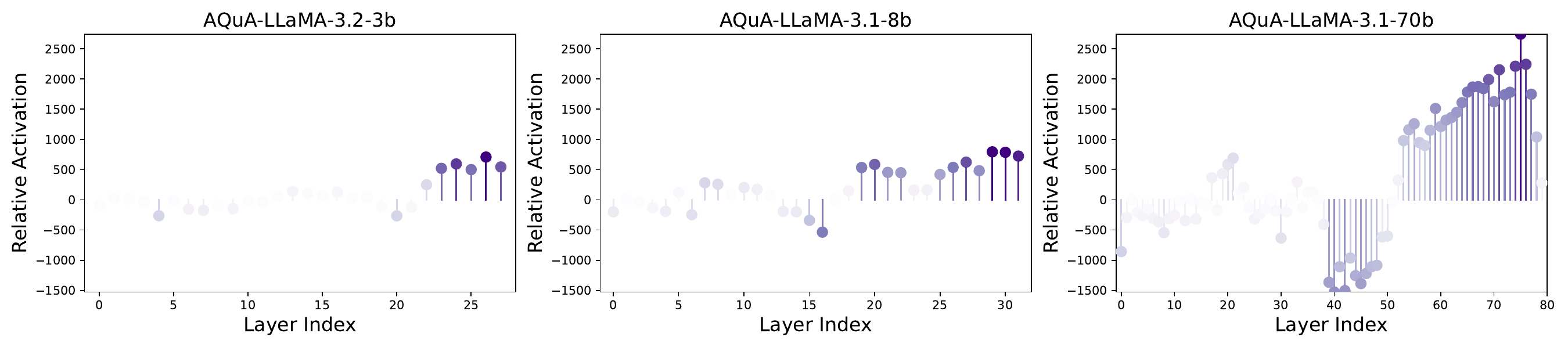}
     \includegraphics[width=0.99\linewidth]{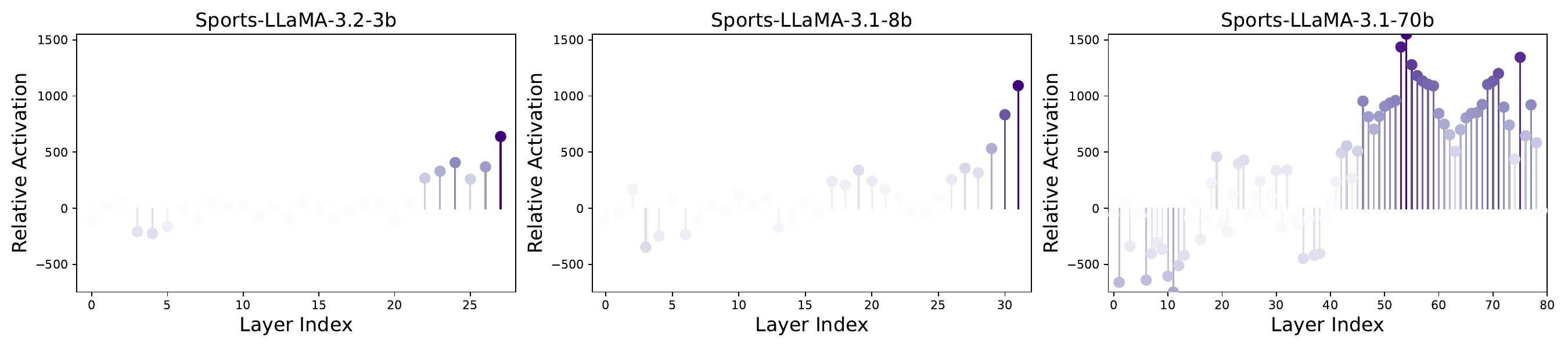}
     \includegraphics[width=0.99\linewidth]{figures/activation/CoinFlip.pdf}
     
\caption{Layer-wise activation differences across layers for models (3B, 8B, 70B) .}
 
\end{figure*}

\clearpage
\section{Activation Count}
\label{nurons}
\begin{figure*}[!ht]
\centering
        \includegraphics[width=0.47\linewidth, height=0.3\linewidth]{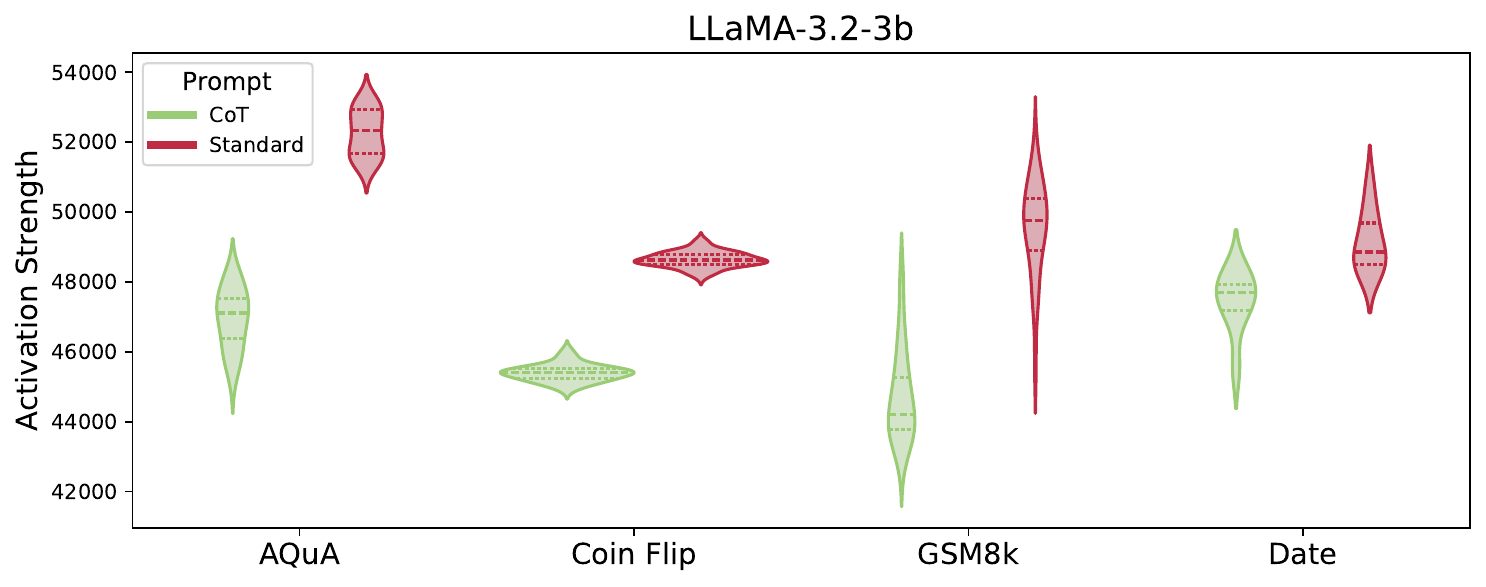}
    \includegraphics[width=0.47\linewidth, height=0.3\linewidth]{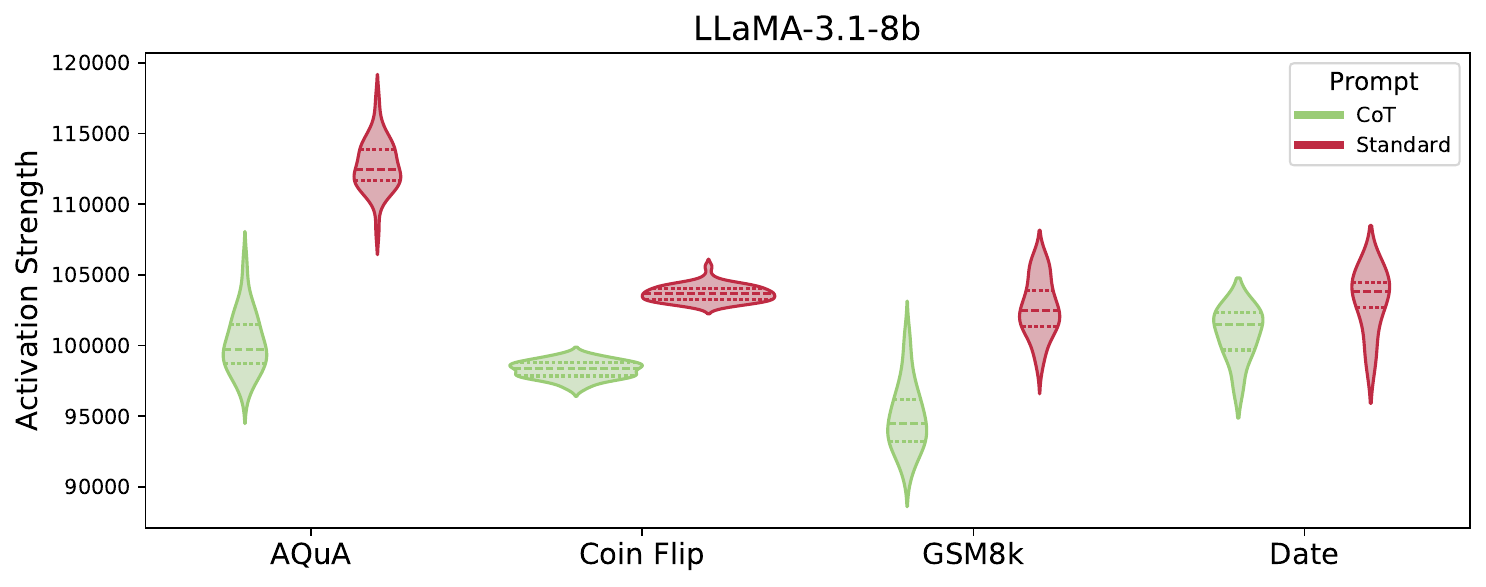}
    \includegraphics[width=0.47\linewidth, height=0.3\linewidth]{figures/question3/bin/LLaMA-3.1-70b_all_datasets_violin.pdf}
    \includegraphics[width=0.47\linewidth, height=0.3\linewidth]{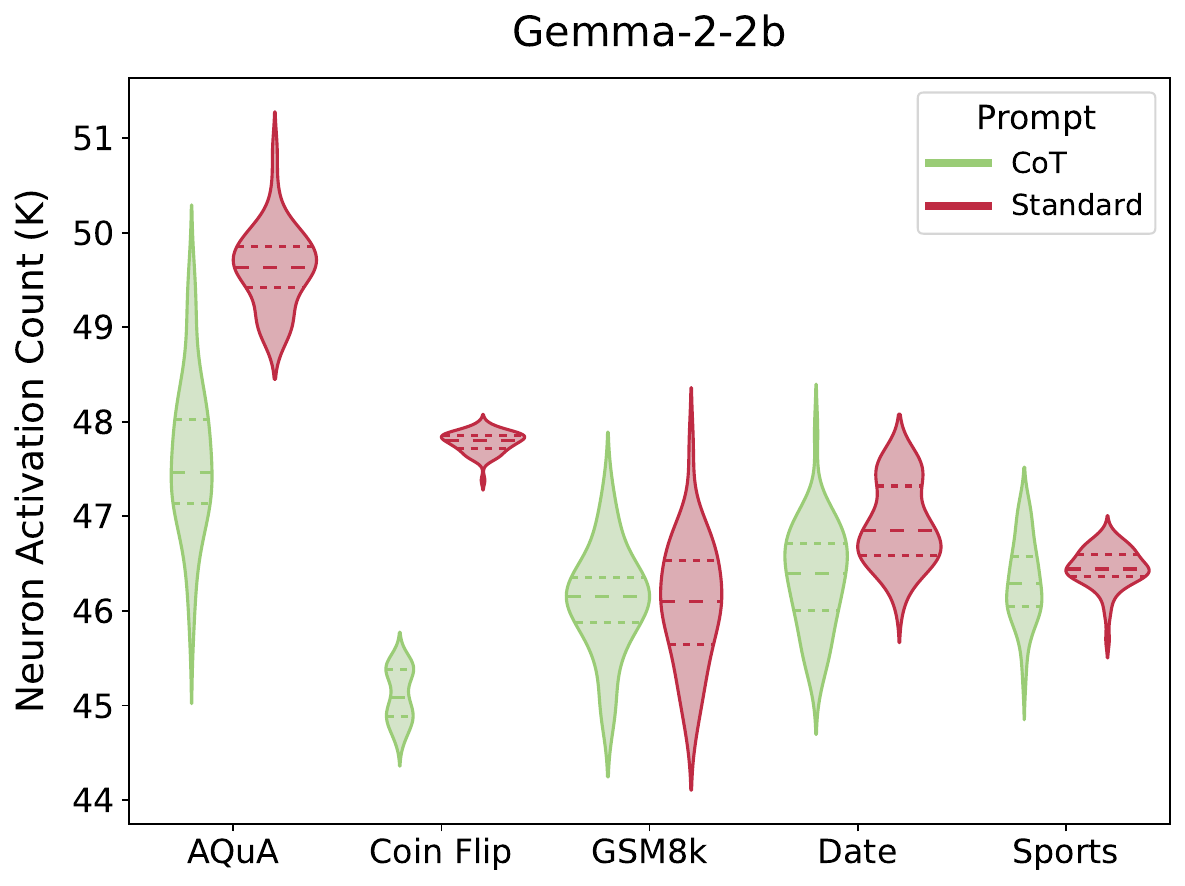}
     \includegraphics[width=0.47\linewidth, height=0.3\linewidth]{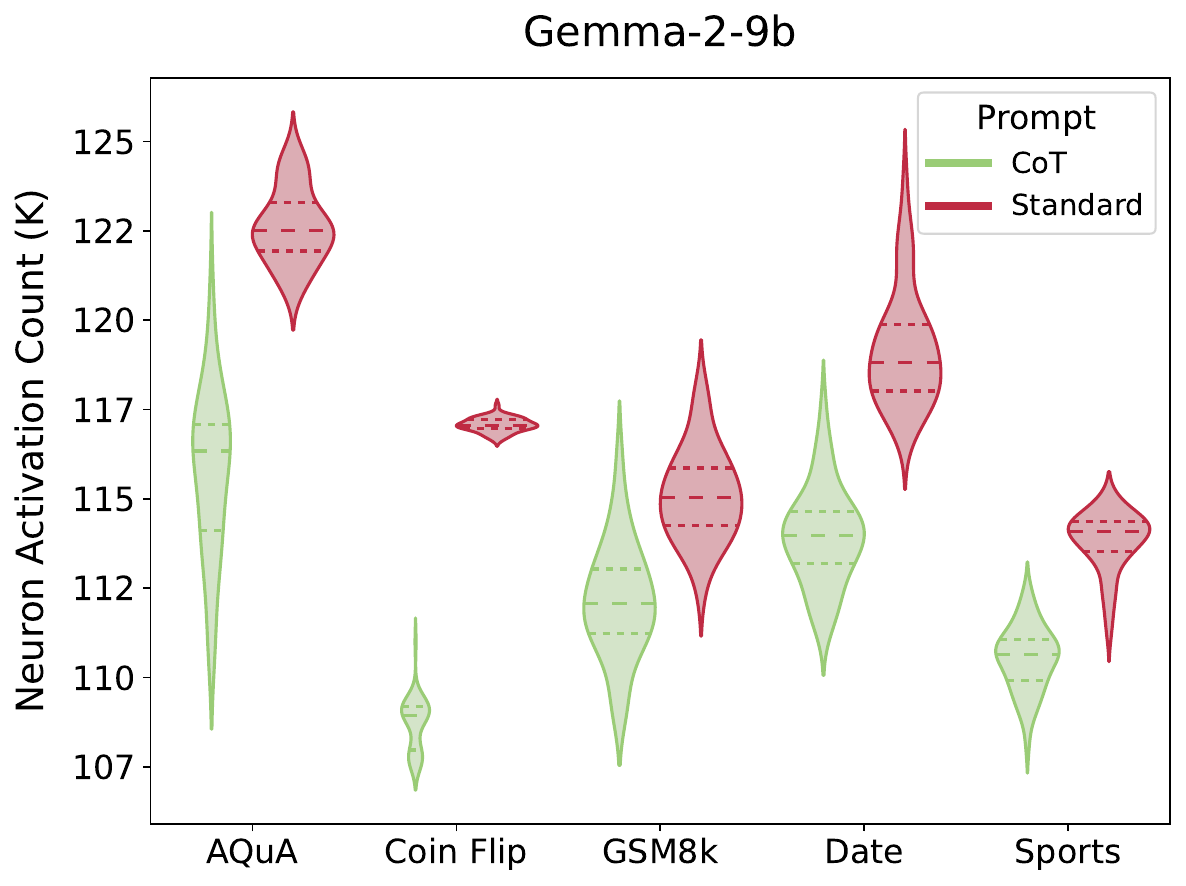}
      \includegraphics[width=0.47\linewidth, height=0.3\linewidth]{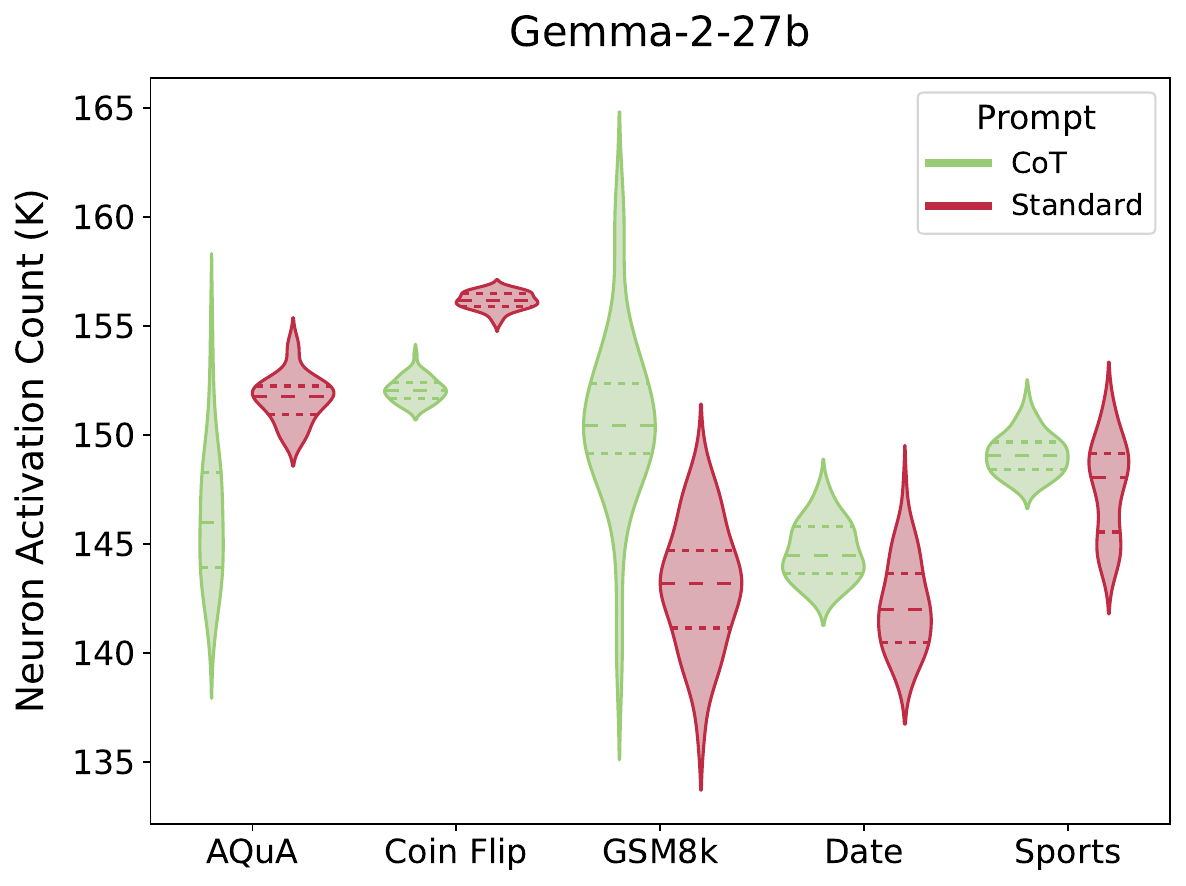}
\end{figure*}

\clearpage
\twocolumn
\section{Prompts Used in This Study}\label{sec:prompts}
This study employs two primary prompt types: Standard prompts and Chain-of-Thought (CoT) prompts, adapted from prior work \citep{wei2022chain, wang2023towards, kojima2022large}. Specifically, the Standard and CoT prompts for the AQuA, GSM8K, SVAMP, Sports, Date, and Coin Flip datasets are derived from \citet{wei2022chain}, those for the Bamboogle dataset are sourced from \citet{wang2023towards}, and those for the Last Letter Concatenation dataset are based on \citet{kojima2022large}. Minor modifications were applied to these prompts to ensure consistency; for instance, CoT prompts conclude with the phrase \textit{``So the answer is...''} while Standard prompts use \textit{``The answer is...''} to present the final response.

For each dataset, we provide four exemplars per prompt type. The following subsections detail the specifications of the Standard and CoT prompts for each dataset.

\subsection{Standard Prompts}
The Standard Prompt supplies the model with multiple question-answer pairs, enabling the model to directly generate the final answer without producing intermediate reasoning steps. The specific Standard Prompts for each dataset are presented as follows: 
Table \ref{standard:gsm8k} for the GSM8K dataset, 
Table \ref{standard:aqua} for the AQuA dataset, 
Table \ref{standard:svamp} for the SVAMP dataset, 
Table \ref{standard:bamboogle} for the Bamboogle dataset, 
Table \ref{standard:date} for the Date dataset, 
Table \ref{standard:sports} for the Sports dataset, 
Table \ref{standard:coin} for the Coin Flip dataset, 
and Table \ref{standard:letter} for the Last Letter Concatenation dataset.
\onecolumn

\begin{table*}[ht]
    \centering
    \footnotesize
    \begin{tabularx}{0.9\textwidth}{X c r}
        \toprule
        \textbf{GSM8K (Standard)} & & \\
        \midrule
        \textbf{Q}: There are 15 trees in the grove. Grove workers will plant trees in the grove today. After they are done, there will be 21 trees. How many trees did the grove workers plant today? & & \\
        \textbf{A}: The answer is 6. & & \\
        \midrule
        \textbf{Q}: Leah had 32 chocolates and her sister had 42. If they ate 35, how many pieces do they have left in total? & & \\
        \textbf{A}: The answer is 39. & & \\
        \midrule
        \textbf{Q}: There were nine computers in the server room. Five more computers were installed each day, from Monday to Thursday. How many computers are now in the server room? & & \\
        \textbf{A}: The answer is 29. & & \\
        \midrule
        \textbf{Q}: Michael had 58 golf balls. On Tuesday, he lost 23 golf balls. On Wednesday, he lost 2 more. How many golf balls did he have at the end of Wednesday? & & \\
        \textbf{A}: The answer is 33. & & \\
        \bottomrule
    \end{tabularx}
    \vspace{-3mm}
    \caption{Standard Prompt exemplars for the GSM8K dataset, adapted from \citep{wei2022chain}.}
    \label{standard:gsm8k}
\end{table*}

\begin{table*}[ht]
    \centering
    \footnotesize
    \begin{tabularx}{0.9\textwidth}{X c r}
        \toprule
        \textbf{Bamboogle (Standard)} & & \\
        \midrule
        \textbf{Q}: Who lived longer, Theodor Haecker or Harry Vaughan Watkins? & & \\
        \textbf{A}: The answer is Harry Vaughan Watkins. & & \\
        \midrule
        \textbf{Q}: Why did the founder of Versus die? & & \\
        \textbf{A}: The answer is shot. & & \\
        \midrule
        \textbf{Q}: Who is the grandchild of Dambar Shah? & & \\
        \textbf{A}: The answer is Rudra Shah. & & \\
        \midrule
        \textbf{Q}: Are both the director of the film FAQ: Frequently Asked Questions and the director of the film The Big Money from the same country? & & \\
        \textbf{A}: The answer is no. & & \\
        \bottomrule
    \end{tabularx}
    \vspace{-3mm}
    \caption{Standard Prompt exemplars for the Bamboogle dataset, adapted from \citep{wang2023towards}.}
    \label{standard:bamboogle}
\end{table*}

\begin{table*}[ht]
    \centering
    \footnotesize
    \begin{tabularx}{0.9\textwidth}{X c r}
        \toprule
        \textbf{AQuA (Standard)} & & \\
        \midrule
        \textbf{Q}: John found that the average of 15 numbers is 40. If 10 is added to each number, then the mean of the numbers is? & & \\
        \textbf{Answer Choices}: (a) 50 (b) 45 (c) 65 (d) 78 (e) 64 & & \\
        \textbf{A}: The answer is (a). & & \\
        \midrule
        \textbf{Q}: If a / b = 3/4 and 8a + 5b = 22, then find the value of a. & & \\
        \textbf{Answer Choices}: (a) 1/2 (b) 3/2 (c) 5/2 (d) 4/2 (e) 7/2 & & \\
        \textbf{A}: The answer is (b). & & \\
        \midrule
        \textbf{Q}: A person is traveling at 20 km/hr and reached his destination in 2.5 hr. Then find the distance? & & \\
        \textbf{Answer Choices}: (a) 53 km (b) 55 km (c) 52 km (d) 60 km (e) 50 km & & \\
        \textbf{A}: The answer is (e). & & \\
        \midrule
        \textbf{Q}: How many keystrokes are needed to type the numbers from 1 to 500? & & \\
        \textbf{Answer Choices}: (a) 1156 (b) 1392 (c) 1480 (d) 1562 (e) 1788 & & \\
        \textbf{A}: The answer is (b). & & \\
        \bottomrule
    \end{tabularx}
    \vspace{-3mm}
    \caption{Standard Prompt exemplars for the AQuA dataset, adapted from \citep{wei2022chain}.}
    \label{standard:aqua}
\end{table*}

\begin{table*}[ht]
    \centering
    \footnotesize
    \begin{tabularx}{0.9\textwidth}{X c r}
        \toprule
        \textbf{SVAMP (Standard)} & & \\
        \midrule
        \textbf{Q}: If there are 3 cars in the parking lot and 2 more cars arrive, how many cars are in the parking lot? & & \\
        \textbf{A}: The answer is 5. & & \\
        \midrule
        \textbf{Q}: Jason had 20 lollipops. He gave Denny some lollipops. Now Jason has 12 lollipops. How many lollipops did Jason give to Denny? & & \\
        \textbf{A}: The answer is 8. & & \\
        \midrule
        \textbf{Q}: Shawn has five toys. For Christmas, he got two toys each from his mom and dad. How many toys does he have now? & & \\
        \textbf{A}: The answer is 9. & & \\
        \midrule
        \textbf{Q}: Olivia has \$23. She bought five bagels for \$3 each. How much money does she have left? & & \\
        \textbf{A}: The answer is 8. & & \\
        \bottomrule
    \end{tabularx}
    \vspace{-3mm}
    \caption{Standard Prompt exemplars for the SVAMP dataset, adapted from \citep{wei2022chain}.}
    \label{standard:svamp}
\end{table*}



\begin{table*}[ht]
    \centering
    \footnotesize
    \begin{tabularx}{0.9\textwidth}{X c r}
        \toprule
        \textbf{Date (Standard)} & & \\
        \midrule
        \textbf{Q}: 2015 is coming in 36 hours. What is the date one week from today in MM/DD/YYYY? & & \\
        \textbf{A}: The answer is 01/05/2015. & & \\
        \midrule
        \textbf{Q}: The first day of 2019 is a Tuesday, and today is the first Monday of 2019. What is the date today in MM/DD/YYYY? & & \\
        \textbf{A}: The answer is 01/07/2019. & & \\
        \midrule
        \textbf{Q}: The concert was scheduled to be on 06/01/1943, but was delayed by one day to today. What is the date 10 days ago in MM/DD/YYYY? & & \\
        \textbf{A}: The answer is 05/23/1943. & & \\
        \midrule
        \textbf{Q}: Jane was born on the last day of February in 2001. Today is her 16th birthday. What is the date yesterday in MM/DD/YYYY? & & \\
        \textbf{A}: The answer is 02/27/2017. & & \\
        \bottomrule
    \end{tabularx}
    \vspace{-3mm}
    \caption{Standard Prompt exemplars for the Date dataset, adapted from \citep{wei2022chain}.}
    \label{standard:date}
\end{table*}

\begin{table*}[ht]
    \centering
    \footnotesize
    \begin{tabularx}{0.9\textwidth}{X c r}
        \toprule
        \textbf{Sports (Standard)} & & \\
        \midrule
        \textbf{Q}: Is the following sentence plausible? ``Kyle Palmieri was called for slashing.'' & & \\
        \textbf{A}: The answer is yes. & & \\
        \midrule
        \textbf{Q}: Is the following sentence plausible? ``Joao Moutinho caught the screen pass in the NFC championship.'' & & \\
        \textbf{A}: The answer is no. & & \\
        \midrule
        \textbf{Q}: Is the following sentence plausible? ``Carson Wentz set the pick and roll.'' & & \\
        \textbf{A}: The answer is no. & & \\
        \midrule
        \textbf{Q}: Is the following sentence plausible? ``Malcolm Brogdon banked the shot in.'' & & \\
        \textbf{A}: The answer is yes. & & \\
        \bottomrule
    \end{tabularx}
    \vspace{-3mm}
    \caption{Standard Prompt exemplars for the Sports dataset, adapted from \citep{wei2022chain}.}
    \label{standard:sports}
\end{table*}

\begin{table*}[ht]
    \centering
    \footnotesize
    \begin{tabularx}{0.9\textwidth}{X c r}
        \toprule
        \textbf{Coin Flip (Standard)} & & \\
        \midrule
        \textbf{Q}: A coin is heads up. Ka flips the coin. Sherrie flips the coin. Is the coin still heads up? & & \\
        \textbf{A}: The answer is yes. & & \\
        \midrule
        \textbf{Q}: A coin is heads up. Maybelle flips the coin. Shalonda does not flip the coin. Is the coin still heads up? & & \\
        \textbf{A}: The answer is no. & & \\
        \midrule
        \textbf{Q}: A coin is heads up. Millicent does not flip the coin. Conception flips the coin. Is the coin still heads up? & & \\
        \textbf{A}: The answer is no. & & \\
        \midrule
        \textbf{Q}: A coin is heads up. Ryan flips the coin. Shaunda flips the coin. Is the coin still heads up? & & \\
        \textbf{A}: The answer is yes. & & \\
        \bottomrule
    \end{tabularx}
    \vspace{-3mm}
    \caption{Standard Prompt exemplars for the Coin Flip dataset, adapted from \citep{wei2022chain}.}
    \label{standard:coin}
\end{table*}


\begin{table*}[ht]
    \centering
    \footnotesize
    \begin{tabularx}{0.9\textwidth}{X c r}
        \toprule
        \textbf{Last Letter Concatenation (Standard)} & & \\
        \midrule
        \textbf{Q}: Take the last letters of each word in ``Tim Candace Cecil Misael'' and concatenate them. & & \\
        \textbf{A}: The answer is mell. & & \\
        \midrule
        \textbf{Q}: Take the last letters of each word in ``Alina Alessandra Amina Bianca'' and concatenate them. & & \\
        \textbf{A}: The answer is aaaa. & & \\
        \midrule
        \textbf{Q}: Take the last letters of each word in ``Felipe Heidi Nino Bradley'' and concatenate them. & & \\
        \textbf{A}: The answer is eioy. & & \\
        \midrule
        \textbf{Q}: Take the last letters of each word in ``Lacey Nora Debra Ashleigh'' and concatenate them. & & \\
        \textbf{A}: The answer is yaah. & & \\
        \bottomrule
    \end{tabularx}
    \vspace{-3mm}
    \caption{Standard Prompt exemplars for the Last Letter Concatenation dataset, adapted from \citep{kojima2022large}.}
    \label{standard:letter}
\end{table*}
\clearpage

\clearpage
\twocolumn
\subsection{Chain-of-Thought Prompts}
The Chain-of-Thought Prompt incorporates intermediate reasoning steps within the exemplars provided to the model, guiding it to derive the final answer through a step-by-step process. The CoT Prompts for each dataset are presented as follows: 
Table \ref{cot:gsm8k} for the GSM8K dataset, 
Table \ref{cot:aqua} for the AQuA dataset, 
Table \ref{cot:svamp} for the SVAMP dataset, 
Table \ref{cot:bamboogle} for the Bamboogle dataset,  
Table \ref{cot:date} for the Date dataset, Table \ref{cot:sports} for the Sports dataset,
Table \ref{cot:coin} for the Coin Flip dataset, and Table \ref{cot:letter} for the Last Letter Concatenation dataset.
\onecolumn

\begin{table*}[ht]
    \centering
    \footnotesize
    \begin{tabularx}{0.9\textwidth}{X c r}
        \toprule
        \textbf{GSM8K (CoT)} & & \\
        \midrule
        \textbf{Q}: There are 15 trees in the grove. Grove workers will plant trees in the grove today. After they are done, there will be 21 trees. How many trees did the grove workers plant today? & & \\
        \textbf{A}: There are 15 trees originally. Then there were 21 trees after some more were planted. So there must have been 21 - 15 = 6. So the answer is 6. & & \\
        \midrule
        \textbf{Q}: Leah had 32 chocolates and her sister had 42. If they ate 35, how many pieces do they have left in total? & & \\
        \textbf{A}: Originally, Leah had 32 chocolates. Her sister had 42. So in total they had 32 + 42 = 74. After eating 35, they had 74 - 35 = 39. So the answer is 39. & & \\
        \midrule
        \textbf{Q}: There were nine computers in the server room. Five more computers were installed each day, from Monday to Thursday. How many computers are now in the server room? & & \\
        \textbf{A}: There were originally 9 computers. For each of 4 days, 5 more computers were added. So 5 * 4 = 20 computers were added. 9 + 20 = 29. So the answer is 29. & & \\
        \midrule
        \textbf{Q}: Michael had 58 golf balls. On Tuesday, he lost 23 golf balls. On Wednesday, he lost 2 more. How many golf balls did he have at the end of Wednesday? & & \\
        \textbf{A}: Michael started with 58 golf balls. After losing 23 on Tuesday, he had 58 - 23 = 35. After losing 2 more, he had 35 - 2 = 33. So the answer is 33. & & \\
        \bottomrule
    \end{tabularx}
    \vspace{-3mm}
    \caption{Chain-of-Thought Prompt exemplars for the GSM8K dataset, adapted from \citet{wei2022chain}.}
    \label{cot:gsm8k}
\end{table*}

\begin{table*}[ht]
    \centering
    \footnotesize
    \begin{tabularx}{0.9\textwidth}{X c r}
        \toprule
        \textbf{SVAMP (CoT)} & & \\
        \midrule
        \textbf{Q}: If there are 3 cars in the parking lot and 2 more cars arrive, how many cars are in the parking lot? & & \\
        \textbf{A}: There are originally 3 cars. 2 more cars arrive. 3 + 2 = 5. So the answer is 5. & & \\
        \midrule
        \textbf{Q}: Jason had 20 lollipops. He gave Denny some lollipops. Now Jason has 12 lollipops. How many lollipops did Jason give to Denny? & & \\
        \textbf{A}: Jason started with 20 lollipops. Then he had 12 after giving some to Denny. So he gave Denny 20 - 12 = 8. So the answer is 8. & & \\
        \midrule
        \textbf{Q}: Shawn has five toys. For Christmas, he got two toys each from his mom and dad. How many toys does he have now? & & \\
        \textbf{A}: Shawn started with 5 toys. If he got 2 toys each from his mom and dad, then that is 4 more toys. 5 + 4 = 9. So the answer is 9. & & \\
        \midrule
        \textbf{Q}: Olivia has \$23. She bought five bagels for \$3 each. How much money does she have left? & & \\
        \textbf{A}: Olivia had 23 dollars. 5 bagels for 3 dollars each will be 5 × 3 = 15 dollars. So she has 23 - 15 dollars left. 23 - 15 = 8. So the answer is 8. & & \\
        \bottomrule
    \end{tabularx}
    \vspace{-3mm}
    \caption{Chain-of-Thought Prompt exemplars for the SVAMP dataset, adapted from \citet{wei2022chain}.}
    \label{cot:svamp}
\end{table*}

\begin{table*}[ht]
    \centering
    \footnotesize
    \begin{tabularx}{0.9\textwidth}{X c r}
        \toprule
        \textbf{AQuA (CoT)} & & \\
        \midrule
        \textbf{Q}: John found that the average of 15 numbers is 40. If 10 is added to each number, then the mean of the numbers is? & & \\
        \textbf{Answer Choices}: (a) 50 (b) 45 (c) 65 (d) 78 (e) 64 & & \\
        \textbf{A}: If 10 is added to each number, then the mean of the numbers also increases by 10. So the new mean would be 40 + 10 = 50. So the answer is (a). & & \\
        \midrule
        \textbf{Q}: If a / b = 3/4 and 8a + 5b = 22, then find the value of a. & & \\
        \textbf{Answer Choices}: (a) 1/2 (b) 3/2 (c) 5/2 (d) 4/2 (e) 7/2 & & \\
        \textbf{A}: If a / b = 3/4, then b = 4a / 3. So 8a + 5(4a / 3) = 22. This simplifies to 8a + 20a / 3 = 22, which means 44a / 3 = 22. So a = (22 × 3) / 44 = 3/2. So the answer is (b). & & \\
        \midrule
        \textbf{Q}: A person is traveling at 20 km/hr and reached his destination in 2.5 hr. Then find the distance? & & \\
        \textbf{Answer Choices}: (a) 53 km (b) 55 km (c) 52 km (d) 60 km (e) 50 km & & \\
        \textbf{A}: The distance that the person traveled would be 20 km/hr × 2.5 hr = 50 km. So the answer is (e). & & \\
        \midrule
        \textbf{Q}: How many keystrokes are needed to type the numbers from 1 to 500? & & \\
        \textbf{Answer Choices}: (a) 1156 (b) 1392 (c) 1480 (d) 1562 (e) 1788 & & \\
        \textbf{A}: There are 9 one-digit numbers from 1 to 9. There are 90 two-digit numbers from 10 to 99. There are 401 three-digit numbers from 100 to 500. 9 × 1 + 90 × 2 + 401 × 3 = 9 + 180 + 1203 = 1392. So the answer is (b). & & \\
        \bottomrule
    \end{tabularx}
    \vspace{-3mm}
    \caption{Chain-of-Thought Prompt exemplars for the AQuA dataset, adapted from \citet{wei2022chain}.}
    \label{cot:aqua}
\end{table*}

\begin{table*}[ht]
    \centering
    \footnotesize
    \begin{tabularx}{0.9\textwidth}{X c r}
        \toprule
        \textbf{Bamboogle (CoT)} & & \\
        \midrule
        \textbf{Q}: Who lived longer, Theodor Haecker or Harry Vaughan Watkins? & & \\
        \textbf{A}: Theodor Haecker was 65 years old when he died. Harry Vaughan Watkins was 69 years old when he died. So the answer is Harry Vaughan Watkins. & & \\
        \midrule
        \textbf{Q}: Why did the founder of Versus die? & & \\
        \textbf{A}: The founder of Versus was Gianni Versace. Gianni Versace was shot and killed on the steps of his Miami Beach mansion on July 15, 1997. So the answer is shot. & & \\
        \midrule
        \textbf{Q}: Who is the grandchild of Dambar Shah? & & \\
        \textbf{A}: Dambar Shah (? - 1645) was the king of the Gorkha Kingdom. He was the father of Krishna Shah. Krishna Shah (? - 1661) was the king of the Gorkha Kingdom. He was the father of Rudra Shah. So the answer is Rudra Shah. & & \\
        \midrule
        \textbf{Q}: Are both the director of the film FAQ: Frequently Asked Questions and the director of the film The Big Money from the same country? & & \\
        \textbf{A}: The director of the film FAQ: Frequently Asked Questions is Carlos Atanes, who is Spanish. The director of the film The Big Money is John Paddy Carstairs, who is British. So the answer is no. & & \\
        \bottomrule
    \end{tabularx}
    \vspace{-3mm}
    \caption{Chain-of-Thought Prompt exemplars for the Bamboogle dataset, adapted from \citet{wang2023towards}.}
    \label{cot:bamboogle}
\end{table*}

\begin{table*}[ht]
    \centering
    \footnotesize
    \begin{tabularx}{0.9\textwidth}{X c r}
        \toprule
        \textbf{Sports (CoT)} & & \\
        \midrule
        \textbf{Q}: Is the following sentence plausible? ``Kyle Palmieri was called for slashing.'' & & \\
        \textbf{A}: Kyle Palmieri is a hockey player. Being called for slashing is a common penalty in hockey. So the answer is yes. & & \\
        \midrule
        \textbf{Q}: Is the following sentence plausible? ``Joao Moutinho caught the screen pass in the NFC championship.'' & & \\
        \textbf{A}: Joao Moutinho is a soccer player. The NFC championship is an American football event, and catching a screen pass is specific to football, not soccer. So the answer is no. & & \\
        \midrule
        \textbf{Q}: Is the following sentence plausible? ``Carson Wentz set the pick and roll.'' & & \\
        \textbf{A}: Carson Wentz is an American football player. The pick and roll is a basketball play, not used in football. So the answer is no. & & \\
        \midrule
        \textbf{Q}: Is the following sentence plausible? ``Malcolm Brogdon banked the shot in.'' & & \\
        \textbf{A}: Malcolm Brogdon is a basketball player. Banking the shot in is a common basketball term. So the answer is yes. & & \\
        \bottomrule
    \end{tabularx}
    \vspace{-3mm}
    \caption{Chain-of-Thought Prompt exemplars for the Sports dataset, adapted from \citet{wei2022chain}.}
    \label{cot:sports}
\end{table*}

\begin{table*}[ht]
    \centering
    \footnotesize
    \begin{tabularx}{0.9\textwidth}{X c r}
        \toprule
        \textbf{Date (CoT)} & & \\
        \midrule
        \textbf{Q}: 2015 is coming in 36 hours. What is the date one week from today in MM/DD/YYYY? & & \\
        \textbf{A}: If 2015 is coming in 36 hours, then it is coming in 2 days. 2 days before 01/01/2015 is 12/30/2014, so today is 12/30/2014. So one week from today will be 01/05/2015. So the answer is 01/05/2015. & & \\
        \midrule
        \textbf{Q}: The first day of 2019 is a Tuesday, and today is the first Monday of 2019. What is the date today in MM/DD/YYYY? & & \\
        \textbf{A}: If the first day of 2019 was Tuesday, then 01/01/2019 was a Tuesday. Today is the first monday, would be six days later. So today is 01/07/2019. So the answer is 01/07/2019. & & \\
        \midrule
        \textbf{Q}: The concert was scheduled to be on 06/01/1943, but was delayed by one day to today. What is the date 10 days ago in MM/DD/YYYY? & & \\
        \textbf{A}: One day after 06/01/1943 is 06/02/1943, so today is 06/02/1943. 10 days before today is 05/23/1943. So the answer is 05/23/1943. & & \\
        \midrule
        \textbf{Q}: Jane was born on the last day of Feburary in 2001. Today is her 16-year-old birthday. What is the date yesterday in MM/DD/YYYY? & & \\
        \textbf{A}: The last day of February is the 28th, so Jane was born on 02/28/2001. Today is her 16-year old birthday, so today is 02/28/2017. So yesterday was 02/27/2017. So the answer is 02/27/2017. & & \\
        \bottomrule
    \end{tabularx}
    \vspace{-3mm}
    \caption{Chain-of-Thought Prompt exemplars for the Date dataset, adapted from \citet{wei2022chain}.}
    \label{cot:date}
\end{table*}

\begin{table*}[ht]
    \centering
    \footnotesize
    \begin{tabularx}{0.9\textwidth}{X c r}
        \toprule
        \textbf{Coin Flip (CoT)} & & \\
        \midrule
        \textbf{Q}: A coin is heads up. Ka flips the coin. Sherrie flips the coin. Is the coin still heads up? & & \\
        \textbf{A}: The coin was flipped by Ka and Sherrie. So the coin was flipped 2 times, which is an even number. The coin started heads up, so after an even number of flips, it will still be heads up. So the answer is yes. & & \\
        \midrule
        \textbf{Q}: A coin is heads up. Maybelle flips the coin. Shalonda does not flip the coin. Is the coin still heads up? & & \\
        \textbf{A}: The coin was flipped by Maybelle. So the coin was flipped 1 time, which is an odd number. The coin started heads up, so after an odd number of flips, it will be tails up. So the answer is no. & & \\
        \midrule
        \textbf{Q}: A coin is heads up. Millicent does not flip the coin. Conception flips the coin. Is the coin still heads up? & & \\
        \textbf{A}: The coin was flipped by Conception. So the coin was flipped 1 time, which is an odd number. The coin started heads up, so after an odd number of flips, it will be tails up. So the answer is no. & & \\
        \midrule
        \textbf{Q}: A coin is heads up. Ryan flips the coin. Shaunda flips the coin. Is the coin still heads up? & & \\
        \textbf{A}: The coin was flipped by Ryan and Shaunda. So the coin was flipped 2 times, which is an even number. The coin started heads up, so after an even number of flips, it will still be heads up. So the answer is yes. & & \\
        \bottomrule
    \end{tabularx}
    \vspace{-3mm}
    \caption{Chain-of-Thought Prompt exemplars for the Coin Flip dataset, adapted from \citet{wei2022chain}.}
    \label{cot:coin}
\end{table*}

\begin{table*}[ht]
    \centering
    \footnotesize
    \begin{tabularx}{0.9\textwidth}{X c r}
        \toprule
        \textbf{Last Letter Concatenation (CoT)} & & \\
        \midrule
        \textbf{Q}: Take the last letters of each words in ``Tim Candace Cecil Misael'' and concatenate them. & & \\
        \textbf{A}: The last letter of ``Tim'' is ``m''. The last letter of ``Candace'' is ``e''. The last letter of ``Cecil'' is ``l''. The last letter of ``Misael'' is ``l''. Concatenating them is ``mell''. So the answer is mell. & & \\
        \midrule
        \textbf{Q}: Take the last letters of each words in ``Alina Alessandra Amina Bianca'' and concatenate them. & & \\
        \textbf{A}: The last letter of ``Alina'' is ``a''. The last letter of ``Alessandra'' is ``a''. The last letter of ``Amina'' is ``a''. The last letter of ``Bianca'' is ``a''. Concatenating them is ``aaaa''. So the answer is aaaa. & & \\
        \midrule
        \textbf{Q}: Take the last letters of each words in ``Felipe Heidi Nino Bradley'' and concatenate them. & & \\
        \textbf{A}: The last letter of ``Felipe'' is ``e''. The last letter of ``Heidi'' is ``i''. The last letter of ``Nino'' is ``o''. The last letter of ``Bradley'' is ``y''. Concatenating them is ``eioy''. So the answer is eioy. & & \\
        \midrule
        \textbf{Q}: Take the last letters of each words in ``Lacey Nora Debra Ashleigh'' and concatenate them. & & \\
        \textbf{A}: The last letter of ``Lacey'' is ``y''. The last letter of ``Nora'' is ``a''. The last letter of ``Debra'' is ``a''. The last letter of ``Ashleigh'' is ``h''. Concatenating them is ``yaah''. So the answer is yaah. & & \\
        \bottomrule
    \end{tabularx}
    \caption{Chain-of-Thought Prompt exemplars for the Last Letter Concatenation dataset, adapted from \citet{kojima2022large}.}
    \label{cot:letter}
\end{table*}
\clearpage

\end{document}